\documentclass{article}
\usepackage{graphicx} 
\usepackage{microtype}
\usepackage{graphicx}
\usepackage{bbm}
\usepackage{amsmath}
\usepackage{amssymb}
\usepackage{mathtools}
\usepackage{amsthm}
\usepackage{xcolor,colortbl}
\usepackage{hyperref}
\hypersetup{colorlinks,allcolors=blue}
\usepackage{booktabs}
\usepackage{caption}
\usepackage{subcaption}
\usepackage{booktabs} 
\usepackage[a4paper, margin=1in]{geometry}
\usepackage{hyperref}
\usepackage{caption} 
\usepackage{lineno}
\usepackage{siunitx}
\usepackage{url}
\providecommand{\keywords}[1]{\textbf{Keywords:} #1}
\usepackage{rotating}

\title{\rule{\linewidth}{1.0 pt} \\ 
       \vspace{0.5em}              
       \textbf{A Deep Learning-based surrogate model for Severe Accidents in nuclear reactors using ASTEC}
       \vspace{0.1em} \\           
       \rule{\linewidth}{1.0pt}}   
\author{
    \textbf{Alessandro Longhi}\thanks{a.longhi@tudelft.nl. The source code for data pre-processing, training and testing of the surrogate model is available at \href{https://github.com/Aleartulon/ASTEC_surrogate_model/tree/main}{\url{https://github.com/Aleartulon/ASTEC_surrogate_model/tree/main}}.
}\\
    \small{TU Delft}\\ \small{Department of Radiation Science and Technology}\\ \small{Delft, The Netherlands} 
    \and
    \textbf{Danny Lathouwers}\\
    \small{TU Delft}\\ \small{Department of Radiation Science and Technology}\\ \small{Delft, The Netherlands} 
    \and
    \textbf{Zoltán Perkó}\\
    \small{TU Delft}\\ \small{Department of Radiation Science and Technology}\\ \small{Delft, The Netherlands}} 
\date{}
\begin{document}
\maketitle
\begin{abstract}
\noindent
Integral codes like the Accident Source Term Evaluation Code (ASTEC) are powerful tools to study the physics of Severe Accidents (SAs) in nuclear reactors. Real time SA simulators can also be helpful in training operators of nuclear plants to react correctly to malfunctions. However, SA simulators can take up to several days per simulation, making their use infeasible for real time applications. In this work we show how to speed up a SA simulator with a fast, Deep Learning based (DL), surrogate model (SM). The SM is built as a combination of a dimensionality reduction stage, via an AutoEncoder, and a time-stepping stage, via a Neural Ordinary Differential Equation. The data on which the SM is trained are obtained from the ASTEC simulator, by sampling a set of operator actions for station blackout (SBO) and loss-of-coolant accidents (LOCA). The objective of the developed SM is to approximate multiple spatio-temporal fields for the thermal-hydraulic physics, core degradation, and fission product release modules in ASTEC's vessel domain. The SM predicts simultaneously around $80$ different physical variables (both scalar and fields), maintaining a stable autoregressive rollout up to $50$ thousand time steps. In addition, the AutoEncoder achieves a dimensionality reduction by a factor of over $300$, which allows the SM to predict up to $40$ hours of simulation in under a minute, both on CPU and GPU. This work is the first study of the capabilities and limits of DL based surrogate modeling in approximating the challenging, highly non-linear physics of ASTEC.
\end{abstract}
\newpage
\section{Introduction}
Surrogate modeling of physical systems by means of \textit{data-driven} methods has emerged in the last years \cite{directions_ML_for_PDEs, aurora} as a new research field to achieve \textbf{fast} simulations of complex physical systems. The main concept is to leverage existing data, obtained from physical simulation and possibly experiments, to approximate the time evolution of physical quantities of interest using (fast) Deep Learning \cite{deep_learning_book} models. In fields of application \textit{where data comes from physical simulations}, the surrogate modeling research is only driven by the need of \textbf{simulation speed up}: we do not expect data-driven surrogate models (SMs) to provide us any new insight into the physics of the system that is simulated; all the known physics is already present in the equations that are numerically solved to get the data, and thus the SM can only approximate it. 

Although all scientific fields would benefit from faster simulations, applications which require \textbf{real-time  control} of physical systems have especially a lot to gain from fast and accurate SMs, as it allows simulating the consequences of specific actions before actually taking them. In this paper, we focus on the field of Severe Accidents (SAs) in Nuclear Reactors. More specifically, we build a SM of the physics of the vessel modeled by the Accident Source Term Evaluation Code (ASTEC) \cite{astec_software, astec}, which simulates the dynamics of a nuclear power plant under severe accident conditions. ASTEC simulates severe accidents given a \textit{set of actions} that the power plants' operators can take in time to mitigate the accidents' effects, however, since it solves the underlying coupled physics partial differential equations for all the phenomena involved in a core meltdown accident in a water-cooled reactor, its computational times are in the order of days. This work is a first step into the construction of fast SMs that can be used in \textit{real time} by power plants operators in training to learn to apply the correct reactions to mitigate SAs.

\subsection{Severe Accidents in nuclear reactors}
According to \cite{severe_accidents_book}: \textit{a severe accident or core melt accident at a Pressurised Water Reactor (PWR) is an accident in which the reactor fuel sustains substantial damage with varying degrees of reactor core melting}. SAs result from multiple sequential failures of equipment and/or personnel, as exemplified by the Three Mile Island and Fukushima-Daiichi accidents. The primary issue in a SA is the failure of core cooling which results in a substantial rise in the temperature of the exposed fuel rods. If core deterioration cannot be halted within a reactor vessel through cooling of the deteriorated core (in-vessel reflooding with coolant), the core melt accident may eventually result in breach of containment integrity and significant releases of radioactivity into the surrounding environment as happened in the Fukushima Daiichi accident \cite{Malizia2021}.

The first step to prevent SAs is to study what scenarios may lead to them. A complete overview a description can be found in \cite{severe_accidents_book}; this work is limited to two types of scenarios: an accident with \textit{station blackout} (SBO) and a \textit{loss-of-coolant} accident (LOCA). 

\textbf{SBOs} occur when the emergency switchboards LHA and LHB fail simultaneously or when loss of offsite and onsite power occurs, where LHA and LHB are two 6.6 kV switchboards whose function is to maintain the continuity of supply
to the relevant loads during normal and abnormal unit operation, and for safe shutdown conditions \cite{eskom2024koeberg}. Response depends on reactor coolant system state. For a closed Reactor Coolant System (RCS), operators act to bring the reactor to a state where injection of water to the Reactor Coolant Pump (RCP) seals is no longer necessary. Seals are the components around the rotating shaft that passes through the wall of the RCP, where a shaft is a rotating rod that transmits mechanical power from a motor to an impeller that circulates the reactor coolant. The objective of seals is to prevent the loss of coolant from the primary circuit. Operators reach this state by using backup systems including the turbine generator, test pump, turbine-driven auxiliary feedwater pump, and steam-dump valves to reduce RCS temperature and pressure to a state characterised by no more than \SI{190}{\celsius} and 45 bar, respectively. At this temperature and pressure the seals are no longer in a demanding environment, since lower temperature and pressure remove the conditions that put the seals at risk, so seal injection is no longer necessary.
For a partially open RCS, seals are not the vulnerability, as there is no high-pressure coolant trying to force its way past them. In this case the focus shifts to making up for coolant losses: operators must bring the reactor to an intermediate state defined by the same temperature and pressure targets (\SI{190}{\celsius} and 45 bar), while the test pump compensates for water lost through the RCS vents. The other systems used in the closed case (the turbine generator, auxiliary feedwater pump, and steam-dump valves) are not used here, as they remove heat through the steam generators, which act as an effective heat sink only when the RCS is closed and pressurised enough to transfer decay heat into them.
For an open RCS, a gravity-fed system is implemented as a short-term measure: water flows into the RCS from an elevated source under its own weight, requiring no power. In the medium term, this supply must be supplemented by pumped injection from the charging pump of an adjacent unit, or from a petrol-powered pump, depending on the reactor design.
In all cases, the SBO generator or combustion turbine (a gas-turbine-driven generator) must be rapidly connected to restore alternating current (AC) power to the plant and return the safety systems to service.
Core melt can result from two situations. When the RCS is closed, it follows from failure of the turbine-driven auxiliary feedwater pump, which removes decay heat through the steam generators, or from failure of seal injection to the reactor coolant pumps, since unprotected seals can degrade and open a leak in the pressure boundary. When the RCS is open, it follows from failure to add water to the RCS, as the water boiling off through the open path is then no longer replaced.

\textbf{LOCAs} occur when the reactor coolant system or connected circuits break, causing coolant leakage and depressurization. Break scenarios vary by reactor state, break size, and location. Depressurization triggers automatic reactor shutdown and activates the safety injection system (SIS), while large breaks also activate the containment spray system (CSS) due to rapid containment pressure increase. Notice that in the SBO the reactor shutdown is initiated by the loss of all AC power (the control rods drop when losing the electromagnetic hold), while for the LOCA it must be initiated by a protection system that detects the depressurisation. Three critical protection functions must operate: 

\begin{enumerate}
    \item reactivity control through automatic shutdown and borated water injection;
    \item water inventory maintenance via passive water accumulators. The SIS operates in two phases: an injection phase using water from the refueling water storage tank (RWST) and a recirculation phase where the water is drawn from the sumps at the bottom of the containment building;
    \item removal of the residual heat generated by the fuel. This happens through vessel water flow, steam generators, and eventually the residual heat removal system (RHRS). Additionally, the containment spray system removes heat when recirculating sump water. These heat-removal operations reject heat through the component cooling water system (CCWS).
\end{enumerate}
 Core melt scenarios for reactors at power involve either safety injection system failure or containment spray system failure during injection or recirculation phases. For shutdown reactors, scenarios depend on whether the reactor coolant system is closed, partially open, or fully open, but always involve failure to maintain adequate core cooling water levels due to human error or equipment failure.
\subsection{The ASTEC code}
The ASTEC software has been initially developed by the former Institut de Radioprotection et de Sûreté Nucléaire (IRSN) in 1995 and by Gesellschaft für Anlagen‑ und Reaktorsicherheit (GRS) \cite{astec, sarnet}. ASTEC is a modular severe accident simulation code that has continuously evolved through several development phases. The V0 series was developed until 2003, followed by the V1 series until 2009, with the first V2 version commissioned in mid-2009. The most significant advancement in V2 involves core damage modeling, where the code can now simulate corium flow in two dimensions using IRSN's ICARE2 mechanistic code \cite{patrick}, representing progression through the core barrel and lower core plates as observed in the TMI-2 accident, whereas V1 modeled flows only in one dimension. Another major improvement concerns the modeling of iodine and ruthenium in reactor cooling systems and containment \cite{severe_accidents_book}.
The V2 versions can simulate various Generation II reactors including French PWRs, German Konvoi reactors, Westinghouse PWRs, and Russian Water-Water Energy Reactors (VVERs), as well as Generation III designs like the European Pressurised Reactor (EPR) and AP1000 \cite{severe_accidents_book, Tarabelli2009}. Research has shown applicability to boiling water reactors, Canada Deuterium Uraniums (CANDUs), and high-temperature reactors with some adaptations. The code can also simulate fuel storage pool accidents and is used extensively for Level 2 Probabilistic Safety Assessments (PSAs) and radioactive release studies. All material properties and chemical reactions are stored in a Material Data Bank (MDB) using the NUCLEA \cite{nuclea} reference base.
\begin{figure}[h]
    \centering
    \includegraphics[width=1.0\textwidth]{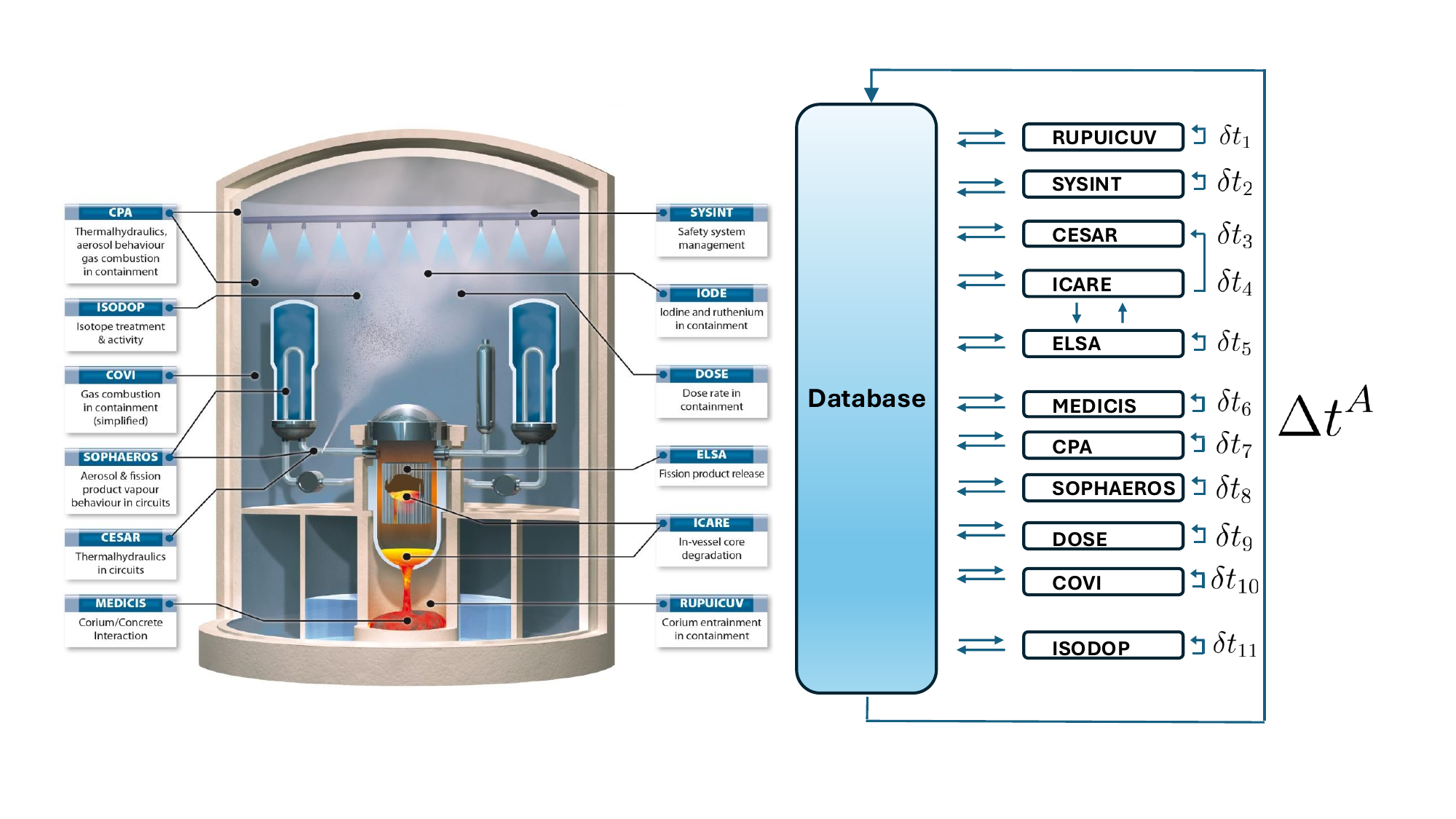}
    \caption{\textbf{Left}: the twelve modules of ASTEC linked to the reactor modules they model. \textbf{Right}: the modules of ASTEC communicate with each other through the ASTEC's dynamic database. $\delta t_i$ is the micro time step of each module, while $\Delta t^A$ is the macro time-step of ASTEC. Image adapted from the ASNR version in \cite{astec2023course}.}
    \label{fig:astec_plant}
\end{figure}
On the left of Figure \ref{fig:astec_plant} we show the physics described by ASTEC. There are a total of twelve modules: \textit{CPA} for the thermalhydraulics modeling in containment; \textit{ISODOP} for the isotope treatment and activity; \textit{COVI} for the gas combustion in containment; \textit{SOPHAEROS} for aerosol and fission product behaviour in circuits and containment; \textit{CESAR} for the thermalhydraulics in the primary and secondary circuits; \textit{MEDICIS} for the interaction between corium and concrete; \textit{SYSINT} for the safety system management; \textit{DOSE} for the dose rate in the containment; \textit{ELSA} for the fission product release in the core part; \textit{ICARE} for the in-vessel core degradation; \textit{RUPUICUV} for the corium entrainment in the containment. On the right of Figure \ref{fig:astec_plant} we show the modular nature of ASTEC: each module interacts with the database at each time step to receive the needed inputs from other modules and write the outputs. $\delta t_i$ is the micro-time step of each module, while $\Delta t^A$ is the macro time-step of ASTEC. Each module is advanced by $\delta t_i$ up to $\Delta t^A$ in the order of interaction from top to bottom as depicted on the right of Figure \ref{fig:astec_plant}. The ASTEC version that has been used to generate the data used in this paper is the 3.1.2.
\subsection{Data-driven surrogate modeling}
\label{subsec:Data-driven surrogate modeling}
Data-driven surrogate modeling has its roots in the field of \textit{Reduced Order Modeling} (ROM) \cite{ROM}, with an application of the \textit{Proper Orthogonal Decomposition} (POD) method to fluid dynamics problems in $1967$ \cite{lumley1967structure}. For dynamical systems, the scope of ROM techniques is to reduce the number of degrees of freedom of the physical system of interest in order to evolve in time (\textbf{quickly}) a reduced representation of the physical state as opposed to evolve in time (\textbf{slowly}) the original high dimensional representation of the physical state. While nowadays the field of surrogate modeling is mainly fully data-driven (thus \textit{non-intrusive}), in the ROM community there have been multiple attempt to reduce the dimensionality of the physical system through \textit{intrusive} approaches \cite{pia,intrusive_rom}, where the known PDEs of the physical system are directly used in the reduction procedure. While successful, intrusive methods can become easily cumbersome and intractable when applied to complex multi-physical systems like most reactor codes, thus most SMs techniques are non-intrusive. Recent works are focusing on extending ROM techniques to non-linear dimensionality reduction methods, like the Spectral SubManifold (SSM) or Operator Inference (OI) theories \cite{ssm, willcox}.

With the rise of Deep Learning (DL), the field of surrogate modeling has been flooded with new methods borrowing techniques from DL; while the goal of such methods is still the increase in the speed of simulations, the fundamental paradigms to achieve it evolved, although the dimensionality reduction concept is still widely adopted by some modern methods. Most methods adopting dimensionality reduction use a non-linear AutoEncoder (AE) to find a reduced (or latent, according to DL terminology) representation of the physical state \cite{romcnn, fresca2021comprehensive, knigge2024space, Longhi2026, Brunton2016}. Since Neural Networks (NNs) are inherently non-linear functions, such AE-based methods are essentially general extensions of SSM and OI methods. The AE is responsible for the reduction of the dimensionality of the physical state, but it does not provide a way to evolve the system in time. The dynamics is thus approximated by sequential NN models, e.g., by LSTM architectures \cite{Wiewel2019}, by Sparse Identification of NonLinear Dynamics (SINDY) \cite{Brunton2016} or by Neural ODEs (NODE) \cite{Longhi2026}. Recently, \textit{Neural Operators} \cite{kovachki2021neural, bartolucci2023representation, li2020fourier, CNO, vcnef-hagnberger:2024} emerged as a novel theoretical framework, born out of the realization that surrogate modeling deals with infinite-dimensional objects (the solution of the system of PDEs), and thus a proper SM should be invariant with respect of the spatial discretization of the grid used to train the model. Parallel to Neural Operator research, new methods have been studied to construct SMs that can work on non-uniform and complex meshes, often times making use of Graph Neural Networks or Transformers based architectures or extending the Neural Operators techniques, as in \cite{Franco2023, UPT, geo_fno}. In line with the intrusive approaches of the ROM community, attempts to embed the known PDEs into the optimization process of finding a SM have also been made in the forms of Physics-Informed-Neural-Networks (PINNs) \cite{pinns} and a framework for optimizing a discrete loss (ODIL) \cite{odil}.

Surrogate modeling of SAs via DL techniques has been studied in \cite{Lee2024, Bae2026} using the simulator Modular Accident Analysis Program (MAAP) \cite{Li2014} and in \cite{Radaideh2020} using the simulator TRACE (TRAC/RELAP5 Advanced Computational Engine) \cite{USNRC2013}. 

\subsection{Contributions}
In this paper we construct the first, to the best of our knowledge, SM of the physics of the vessel of ASTEC. The SM is built on the work carried out in \cite{Longhi2026}, which develops a Deep Learning based SM for general physical phenomena described by time-dependent parametric PDEs and
is based on the coupling of an \textit{AutoEncoder} (AE) and a \textit{Neural Ordinary Differential Equation} (NODE); as such, we will refer to it as \textit{AE-NODE}. The contributions of this work are the following:
\begin{itemize}
    \item We show how \textit{AE-NODE} can be \textbf{adapted} to build a SM of the physics of the \textbf{vessel} of ASTEC by \textbf{decoupling} the vessel from the primary and the secondary circuit. In particular, we modify the AE in order to work with physical variables that belong to different physical domains;
    \item We construct $2$ SMs trained on $2$ different datasets: one for the \textbf{LOCA} accident, one for the \textbf{SBO} accident. In this way we are able to test the capabilities of the SM to approximate different physics. In both cases, \textit{AE-NODE} produces a stable autoregressive rollout within the span of 4 to 40 hours (from 10k to 55k time-steps) across different trajectories in \textbf{less than one minute};
    \item We improve the autoregressive training technique used in \cite{Longhi2026} by using an adaptive window technique;
    \item We demonstrate that the developed SM achieves a \textbf{dimensionality reduction} factor of over $300$ and we study the capabilities and limits of AE-NODE in approximating the full spatio-temporal dynamics of the $80$ vessel domain variables.
\end{itemize}
In section \ref{subsec:Data-driven surrogate modeling} we mentioned that dozens of different methods have been developed in the field of data-driven surrogate modeling, however in this paper we \textbf{restrict} to \textit{AE-NODE} \cite{Longhi2026} for the following reasons:
\begin{itemize}
    \item the SM needs to predict around $80$ physical variables (some of which having spatial dependency as well) per time step which are highly correlated, thus employing a \textbf{dimensionality reduction} technique by means of an AutoEncoder is suitable;
    \item \textit{AE-NODE} handles naturally \textbf{continuous time series prediction} through the use of the (latent) Neural ODE;
    \item \textit{AE-NODE} handles naturally the physical fields \textbf{dependency} on system parameters and boundary conditions; in \cite{Longhi2026} it is shown that \textit{AE-NODE} is noticeable computationally \textbf{faster} at testing time than other SM techniques such as Fourier Neural Operator \cite{li2020fourier} and Vectorized Conditional Neural Fields \cite{vcnef-hagnberger:2024}.
\end{itemize}
Finally, no comparisons with other methods are performed in this work as the scope of the paper is not to find the most optimal surrogate model method for a SA application, but to lay the foundation of surrogate modeling of SAs through the lenses of dimensionality reduction and Neural ODEs.
\section{Severe accident modeling of the vessel}
This work has been developed within the ASSAS (Artificial intelligence for Simulation of Severe AccidentS) project \cite{assas} and the functional specifications of the basic-principles of severe accident simulator are defined in \cite{parrado-rodriguez2024users}. ASTEC is used to generate data modeling a simplified four-loop $1,300$ MWe Western-type Pressurized Water Reactor (PWR) for $2$ different types of accidents \cite{brence2024modelling}:
\begin{enumerate}
\item A Large Break Loss-of-Coolant-Accident (LB-LOCA) with Safety Injection (SI) and
Containment Spray System (CSS) failure;
\item A Station Blackout (SBO) with Auxiliary Feedwater (AFW) failure.
\end{enumerate}

Although ASTEC can simulate the full power plant as depicted in Figure \ref{fig:astec_plant} and described in \cite{chailan2019overview, Drai2025ASTEC}, in this work we focus on the vessel, i.e., we build the SM to substitute the \textit{CESAR} and \textit{ICARE} modules of ASTEC \textbf{within the vessel domain}. The following \textbf{simplifications} have been carried out in order to focus on the most important aspects of severe accident management \cite{brence2024modelling}: 
\begin{itemize}
    \item The Residual Heat Removal System (RHRS), the Chemical and Volume Control System (CVCS), and the balance-of-plant are not considered for the accident management;
 \item In order to limit the amount of generated data the geometric discretization of the input deck and certain physical models have been simplified.
\end{itemize}
Additionally, the Portable Equipment Systems (PES) can be deployed to the reactor in less than 24 hours to mitigate the accident but at least after 20 minutes after the accident events initiator has been activated.
Because of such simplifications there are only $10$ \textit{operator actions} that can be varied in the simulations as described in Section \ref{subsec:Sampling of operator actions}. Both LOCA and SBO accidents begin with the same initial condition, i.e., a reactor operating at its nominal power; thus the only \textbf{source of variation} of the system is given by the vector of \textit{operator actions}. The reactor model is separated into 4 domains: the vessel, the primary circuit, the containment and the secondary circuit. The \textit{vessel} is modeled as a 2D axisymmetric object: 5 concentric rings divided into 15 axial elevations with an additional volume for the lower plenum. The \textit{primary} and \textit{secondary} circuits are described by a 1-D domain with a total of $158$ control volumes. The reactor building is made up of $18$ zones connected by $54$ junctions, while the containment model has $62$ walls.
\begin{figure}[h]
    \centering
    \includegraphics[width=0.8\textwidth]{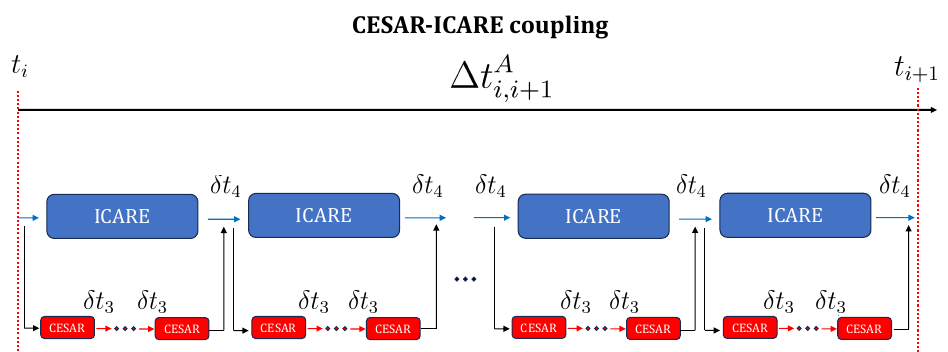}
    \caption{Coupling of the ICARE (core-degradation) and CESAR (thermal-hydraulics) modules. $\Delta t^A_{i,i+1}=t_{i+1}-t_i$ is the macro time-step of ASTEC, $\delta t_3$ is the micro time-step of CESAR while $\delta t_4$ is the micro time-step of ICARE, with $\Delta t^A_{i,i+1} \geq \delta t_4 > \delta t_3$. CESAR is called iteratively within an ICARE time-step to update the initial prediction of ICARE. Figure adapted from \cite{astec2023course}.}
    \label{fig:icare_cesare_coupling}
\end{figure}

\begin{figure}[h]
    \centering
    \begin{subfigure}{0.481\textwidth}
        \centering
        \includegraphics[width=\linewidth]{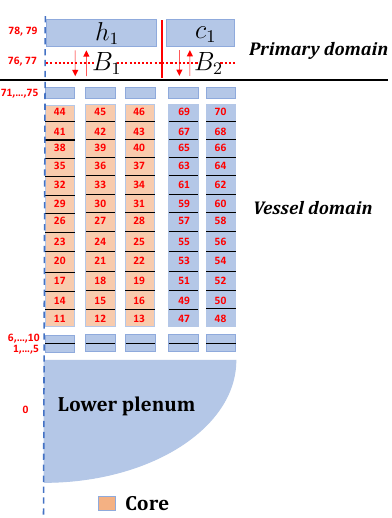}
        \caption{Vessel and core variables}
        \label{fig:vessel_variables}
    \end{subfigure}
    \hspace{0.7cm}
    \begin{subfigure}{0.3\textwidth}
        \centering
        \includegraphics[width=\linewidth]{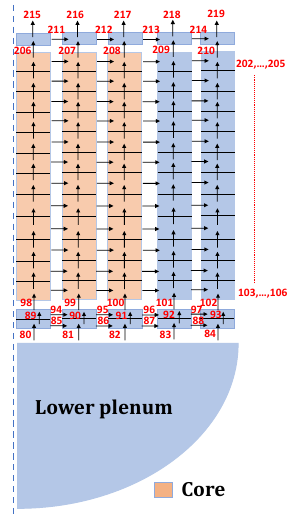}
        \caption{Face variables}
        \label{fig:faces}
    \end{subfigure}
    \caption{(a): blue and orange cells are the volumes considered for the construction of the SM. The vessel domain is made of $5\times 15$ volumes with indices  from $1$ to $75$ plus one additional at the bottom that constitutes the lower plenum with index $0$. The orange volumes make up the core, with indices  from $11$ to $46$. The \textit{vessel domain} is connected to the \textit{primary circuit} of the reactor through the boundary conditions $B_1$ with index $76$ and $B_2$ with index $77$. The volumes $h_1$ with index $78$ and $c_1$ with index $79$ represent the first volumes of the upper plenum and of the cold legs. (b): black arrows represent the \textit{faces} of the vessel, with indices  from $80$ to $219$. Each face contains the physical variables at the interface within volumes. For example, face number $84$ is the interface between the lower plenum and the bottom volume of the fifth column (from the left) of the vessel.}
    \label{fig:vessel_and_faces}
\end{figure}

Importantly, in this work we are going to simulate the physics of the vessel \textbf{up to vessel rupture time}.
\subsection{Sampling of operator actions}
\label{subsec:Sampling of operator actions}
In this work the operator actions are the \textbf{source of variation} of the data, together with the type of accident (LOCA or SBO), since the initial condition of the reactor physical variables is always the same. In Table \ref{tab:op} we show the list of operator actions used in this work. The operator actions are booleans that can happen at a given time step, pressure parameters and percentages. 
\begin{table}[]
\caption{The 10 Operator Actions (OP) which are sampled in order to obtain different simulations to train the surrogate model.}
\label{tab:op}
\begin{tabular}{clp{6cm}}
\toprule

\textbf{OP} & \textbf{Description} & Unit  \\
\midrule 
Opensrv         & Opening percentage pilot-operated relief valve (PORV) during design basis accident (DBA)  & $\%$                                 \\ 
$Pu_5$            & Activating filtered containment venting system when pressure setpoint reached    & Pa                              \\ 
$t_1^{\text{srv}} $          & PORV opening at the value specified in Opensrv    & s                                                                       \\ 
$t_2^{\text{srv}} $          & Fully opening PORV during severe accident (SA) phase     & s                                                       \\ 
$t^{\text{fbseb}}$         & Switching pressurizer valves in feed and bleed mode     & s                                                                 \\ 
$t^{\text{css}}$           & Restoring containment spray system operation           & s                                                        \\ 
$t^{\text{endssg2}}$        & Closing PORV following steam generator tube rupture (SGTR)    & s                                                 \\ 
$t^{\text{pesp}}$           & Water injection in the primary circuit using the PES              & s                                                       \\ 
$t^{\text{pessg}}$          & Water injection in the SG using the PES                      & s                                                       \\ 
$t^{\text{sg2tr}}$         & SGTR occurrence time in affected steam generator          & s                                                   \\ 
\bottomrule
\end{tabular}
\end{table}
What is sampled is a vector $k \in\mathbb{R}_+^{N_{op}}$, where $N_{op} =9$ for the SBO scenario and $N_{op} =7$ for the LOCA scenario as as shown in Tables \ref{tab:parameter_ranges_sbo} and \ref{tab:parameter_ranges_loca}. We add some \textbf{constraints} in time in order to make the sequence of actions meaningful. We can thus define the set $K$ of $N_K$ sampled activation times of the operator actions as $K=\{k_i\in\mathbb{R}_+^{N_{op}}\}_{i=1}^{N_K}$. In Appendix \ref{appendix:appendix_op} more details are presented: in Table \ref{tab:loca_sbo_constraints} the constraints on SBO and LOCA accidents are given. 

In order to sample the Operator Actions, \textit{low-discrepancy} Sobol sequences are used from the OpenTurns library \cite{openturns}; this way a more uniform coverage of the parameter space compared than using random sampling.
\subsection{The vessel}
\label{subsec:the_vessel}
The physics of the vessel \cite{astec2023course} is modeled in ASTEC through the \textbf{coupling} of the ICARE (for the \textit{core degradation} description) and CESAR (for the \textit{thermal-hydraulics} description) modules as shown in Figure \ref{fig:icare_cesare_coupling}: at a time-step $t_i$, ICARE reads the thermo-hydraulic state from CESAR's previous time step and makes a prediction advancing of $\delta t_4$; subsequently CESAR performs a sub-cycle with a time-step $\delta t_3$ from which ICARE corrects its prediction (by taking into account the real thermo-hydraulic state of CESAR in the oxydation models). ICARE modules deal with the following physical phenomena through an \textit{explicit} solver: changes in geometry due to mechanical factors (dilatation, creep, contacts), relocation of materials (decanting/fusion/fragmentation, magma, debris), chemistry, changes in composition and geometry (due to chemistry and relocation), calculation of thermal exchanges coefficients, heat fluxes computation (due to chemistry and relocation) fission products and structural elements movements, calculation of imposed or nuclear power. An \textit{implicit} enthalpy solver is instead called to compute the value of the temperature.

In Figure \ref{fig:vessel_variables} the geometry of the vessel is shown: it has a 2D axisymmetric shape and it comprises $5$ vertical channels. Additionally, a volume is added at the very bottom to model the physics of the \textbf{lower plenum}, with index $0$. Figure \ref{fig:vessel_variables} shows the volumes where the core and vessel variables are computed. While the \textbf{core} variables are computed only in the \textit{orange} grid with indices $11$ to $46$, the \textbf{vessel} variables concern the volumes with indices $1$ to $75$. The vessel is connected to the primary domain through the boundaries in the volumes $B_1$ and $B_2$, with indices  $76$ and $77$. Figure \ref{fig:faces} shows the position of the $140$ \textbf{faces}, with indices  going from $80$ to $219$; a face is a geometric location at the interface within two volumes. For example, face number $84$ contains the values of the variables at the interface between the lower plenum and the bottom volume of the fifth column (from the left) of the vessel. From Figure \ref{fig:vessel_variables} we can see that the vessel is coupled to the primary circuit through the \textbf{boundaries} $B_1$ with index $76$ and $B_2$ with index $77$, which connect the vessel to the first volume of the \textbf{vessel upper plenum} $h_1$ with index $78$ and to the first volume of the \textbf{cold leg} $c_1$ with index $79$.

We now introduce the notation needed to describe the different variables that define the physics of the vessel, given the time domain $\mathcal{D}_t\subseteq\mathbb{R}^+$:
\begin{itemize}
    \item $s_g(t): \mathcal{D}_t\rightarrow \mathbb{R}^{d_g}$ identifies at time $t$ the $d_g$ global variables listed in Table \ref{tab:variables_group1}. $s_g$ both contains \textbf{global variables} that describe the state of the vessel such as the corium mass or the hydrogen mass accumulated (in the core) and $53$ \textbf{fission products} mass flow rates;
    \item given $x_{p}\in\Omega_{p} = \{0\}$, $s_p(x_{p},t): \Omega_{p}\times\mathcal{D}_t\rightarrow \mathbb{R}^{d_p}$ identifies at time $t$ the $d_p$ variables listed in Table \ref{tab:variables_group1}. $s_p$ contains the variables that have as domain the \textbf{lower plenum} (a single volume);
    \item given $x_{cr}\in\Omega_{cr} = \{i|i=11,...,46\}$, $s_{cr}(x_{cr},t):\Omega_{cr}\times\mathcal{D}_t\rightarrow \mathbb{R}^{d_{cr}}$, identifies at time $t$ the $d_{cr}$ 2D fields that have \textit{only} the \textbf{core} as domain; they are listed in Table \ref{tab:variables_group2};
    \item given $x_{v}\in\Omega_{v} = \{i|i=1,...,75\}$, $s_{v}(x_{v},t):\Omega_{v}\times\mathcal{D}_t\rightarrow \mathbb{R}^{d_{v}}$, identifies at time $t$ the $d_{v}$ 2D fields that have \textit{only} the \textbf{vessel} as domain; they are listed in Table \ref{tab:variables_group2};
    \item given $x_{B_1}\in\Omega_{B_1} = \{76\}$ and $x_{B_2}\in\Omega_{B_2} = \{77\}$, $s_{B_1}(x_{B_1}, t):\Omega_{B_1}\times\mathcal{D}_t\rightarrow \mathbb{R}^{d_{B_1}}$ and $s_{B_2}(x_{B_2},t):\Omega_{B_2}\times\mathcal{D}_t\rightarrow \mathbb{R}^{d_{B_2}}$ identify at time $t$ the $d_{B_1}$ and $d_{B_2}$ variables that have as domain the \textbf{boundaries} $B_1$ and $B_2$ respectively, as shown in Figure \ref{fig:vessel_variables}. $s_{B_1}$ and $s_{B_2}$ are described in Table \ref{tab:variables_group1};
     \item given $x_{f}\in\Omega_{f} = \{(i)|i=80,...,219\}$, $s_{f}(x_{f},t):\Omega_{f}\times\mathcal{D}_t\rightarrow \mathbb{R}^{d_{f}}$, identifies at time $t$ the $d_{f}$ variables that have the \textbf{faces} of the vessel as domain, with the indices  of $\Omega_{f}$ following the notation of Figure \ref{fig:faces}; they are listed in the third column of Table \ref{tab:variables_group2};
    \item  given $x_{h_1}\in\Omega_{h_1} = \{78\}$ and $x_{c_1}\in\Omega_{c_1} = \{79\}$, $p_{h_1}(x_{h_1}, t):\Omega_{h_1}\times\mathcal{D}_t\rightarrow \mathbb{R}^{d_{h_1}}$ and $p_{c_1}(x_{c_1},t):\Omega_{c_1}\times\mathcal{D}_t\rightarrow \mathbb{R}^{d_{c_1}}$ identify at time $t$ the $d_{h_1}$ and $d_{c_1}$ variables that have as domain the first volume of the \textbf{vessel upper plenum} and of the \textbf{cold} \textbf{leg} respectively ($h_1$ and $c_1$), as shown in Figure \ref{fig:vessel_variables}. $p_{h_1}$ and $p_{c_1}$ are described in Table \ref{tab:variables_group2};
    \item finally, we define the solution domain $\Omega_{\mathbf{x}} = \Omega_g\times\Omega_p\times\Omega_{cr}\times\Omega_{v}\times\Omega_{f}\times\Omega_{B_1}\times\Omega_{B_2}$, the domain at the intersection of the vessel and the primary circuit $\Omega_{\partial{\mathbf{x}}} = (\Omega_{h_1}\times\Omega_{c_1})\cap(\Omega_{B_1}\times\Omega_{B_2})$ and domain of the primary circuit next to the vessel as $\Omega_{P} = \Omega_{h_1}\times\Omega_{c_1}.$
\end{itemize}
We define $\mathbf{P}_t^k = \{(p_{c_1}(x_{c_1},t^{'}|k),p_{h_1}(x_{h_1},t^{'}|k))|k\in K, t^{'}<t\}$ as the collection of snapshots of the variables in $h_1$ and $c_1$ from the initial state to time $t$ for a given vector of sampled operator actions $k\in K$ as defined in Section \ref{subsec:Sampling of operator actions}. $\mathbf{P}_t^k$ uniquely determines the entire time evolution of the vessel, since the physics equations and the initial condition are always the same. Conversely, $\mathbf{P}_t^k$ in this work is uniquely determined by the sampled vector of \textit{operator actions}, i.e., by $k$.

Thus, given $\mathbf{x}\in\Omega_{\mathbf{x}}$ and $\hat{\mathbf{x}}\in\Omega_{P}$, we define the solution $s$ in the vessel and the solution $p$ in $h_1$ and $c_1$:
\begin{equation}
\label{eq:solution_and_boundary_PDE}
\left\{
\begin{aligned}
s(\mathbf{x},t|k) = & s(\mathbf{x},t|\mathbf{P}_t^k) =(s_g(t|\mathbf{P}_t^k),s_p(x_p,t|\mathbf{P}_t^k),\{s_{cr}(x_{cr},t|\mathbf{P}_t^k)\}_{ x_{cr}\in\Omega_{cr}},\{s_{v}(x_{v},t|\mathbf{P}_t^k)\}_{ x_{v}\in\Omega_{v}},\\&\{s_{f}(x_{f},t|\mathbf{P}_t^k)\}_{ x_{f}\in\Omega_{f}},
s_{B_1}(x_{B_1},t|\mathbf{P}_t^k),s_{B_2}(x_{B_2},t|\mathbf{P}_t^k)):\Omega_{x}\times\mathcal{D}_t\rightarrow\mathcal{S},\\
p(\hat{\mathbf{x}},t|k)=& p(\hat{x},t|\mathbf{P}_t^k) = (p_{c_1}(x_{c_1},t|k),p_{h_1}(x_{h_1},t|k)):\Omega_{P}\times\mathcal{D}_t\rightarrow\mathcal{S}_{P},
\end{aligned}
\right.
\end{equation}
where $\mathbf{x}\in\Omega_{\mathbf{x}}$, $\hat{\mathbf{x}}\in\Omega_{P}$, $\mathcal{S} \subseteq \mathbb{R}^{d_g} \times \mathbb{R}^{d_p} \times \mathbb{R}^{d_{cr}\times 12\times 3} \times \mathbb{R}^{d_v\times 15\times 5 }\times \mathbb{R}^{d_f\times 16\times 9 }  \times \mathbb{R}^{d_{B_1}} \times \mathbb{R}^{d_{B_2}}$ and $\mathcal{S}_{P} \subseteq \mathbb{R}^{d_{c_1}} \times \mathbb{R}^{d_{h_1}}$. In Appendix \ref{appendix:geometry_of_faces_variables} we explain how the geometry of $s_f$ is treated. What we call \textbf{boundaries}, i.e., $s_{B_1}$ and $s_{B_2}$ are part of the \textbf{solution} domain $\Omega_{\mathbf{x}}$, while in this work $p_{c_1}$ and $p_{h_1}$ are the actual boundaries, or \textit{degrees of freedom} of $s(\mathbf{x},t|k)$, that belong to the boundary domain $\Omega_{\partial \mathbf{x}}$; we kept such notation to be coherent with ASTEC terminology, where $s_{B_1}$ and $s_{B_2}$ are the boundaries, while in this work we choose to predict ASTEC boundaries in order to build a SM that can be coupled with the primary domain. In order to formally couple the primary circuit and the vessel, we define the operator $\mathcal{B}:\mathcal{S}_{P}\rightarrow\mathcal{S}_{B}$, where $\mathcal{S}_{B} = \mathbb{R}^{d_{B_1}} \times \mathbb{R}^{d_{B_2}}$,  which takes the variables $p_{h_1}$ and $p_{c_1}$ and computes the variables $s_{B_1}$ and $s_{B_2}$ (and analogously we can define the inverse operator $\mathcal{B}^{-1}:\mathcal{S}_{B}\rightarrow\mathcal{S}_{P}$). We can thus model the physical system represented by the vessel in the following terms:
\begin{equation}
    \left\{
    \begin{aligned}
    & \frac{\partial s(\mathbf{x},t|k)}{\partial t} + \hat{\mathcal{N}}(s(\mathbf{x},t|k),\mathbf{x} )= 0,\\
    &s(\partial{\mathbf{x}},t|k) = \mathcal{B}(p(\hat{\mathbf{x}},t|k)), \\
    &s(\mathbf{x},t=0) = s^0(\mathbf{x}),
    \end{aligned}
    \right.
\label{eq:PDEsystem_vessel}
\end{equation}
where $\hat{\mathcal{N}}$ is a (typically) nonlinear integro-differential operator, $s\in\mathcal{S}$ is the PDE solution, $s^0(\mathbf{x})$ is the initial condition, $\mathbf{x}\in\Omega_{\mathbf{x}}$, $\hat{\mathbf{x}}\in\Omega_{P}$, $\partial{\mathbf{x}}\in\Omega_{\partial{\mathbf{x}}}$, $ t\in\mathcal{D}_t$ and $k\in K$. $s^0(\mathbf{x})$ is \textbf{fixed} and it is given by a reactor operating at its nominal power. By modeling the physics of the vessel through Equation \ref{eq:PDEsystem_vessel}, we have effectively \textbf{replaced the coupling} between ICARE and CESAR depicted in Figure \ref{fig:icare_cesare_coupling} with a single PDE defined by $\hat{\mathcal{N}}$. 
\subsection{Training set construction}
\label{subsec:training_set_construction}
The training dataset is obtained from ASTEC simulations. Following the discussion in Section \ref{subsec:Sampling of operator actions}, in this work we sample for training $N_{tr}$ vectors $k_j\in K_{tr}\subseteq K, j=1,...,N_{tr}$. For each vector $k_j$, the ASTEC code simulates either the LOCA or the SBO accident up to vessel rupture. We thus obtain $N_{tr}$ trajectories in time, and we save snapshots of $s$ and $p$ at each ASTEC \textbf{macro} time-step as defined in Figure \ref{fig:icare_cesare_coupling}. Each trajectory ends at a different time and is defined by different time-steps, thus we need to discretize the time $t$ differently for each trajectory by defining $\mathbf{T}^j = \{(t_1^j,t_2^j,...,t_{F_j}^j)|k_j\in K_{tr},t_i^j\in\mathcal{D}_t,t_{i}^j<t_{i+1}^j, \text{for } i=1,...,F_j-1\}$, i.e., the time discretization associated to the trajectory defined by $k_j$. Based on this, we define the time-discrete version of $\mathbf{P}_t^k$ as $\mathbf{P}_i^j = \{(p_{c_1}(\mathbf{x}_{c_1},t_r^j|k_j),p_{h_1}(\mathbf{x}_{h_1},t_r^j)|k_j) | k_j\in K_{tr}, r<i\}$ for a given vector of sampled operator actions $k_j\in K$, thus $s(\mathbf{x},t_i|k_j) = s(\mathbf{x},t_i^j|k_j) = s(\mathbf{x},t_i|\mathbf{P}_i^j)$ where we omit the superscript $j$ on $t$ for ease of notation (the index $j$ is still present through $k_j$ so no information is lost). We can then build the training set
\begin{align}
\mathcal{M}_{tr}^{l=1} = \{&(s(\mathbf{x},t_1|k_1), p(\hat{\mathbf{x}},t_1|k_1)),...,(s(\mathbf{x},t_{F_1}|k_1), p(\hat{\mathbf{x}},t_{F_1}|k_1)),\\
&...,(s(\mathbf{x},t_1|k_{N_{tr}}), p(\hat{\mathbf{x}},t_1|k_{N_{tr}})),...,(s(\mathbf{x},t_{F_{N_{tr}}}|k_{N_{tr}}), p(\hat{\mathbf{x}},t_{F_{N_{tr}}}|k_{N_{tr}}))\}
\end{align}
as a collection of single snapshots in time each identified by $t_i$ and $k_j$; the superscript $l$ stands for the number of sequential snapshots present in each data-point, in this case $l=1$ as each tuple only contains one snapshot. In general,
\begin{align*}
&\mathcal{M}_{tr}^{l} = \{\big((s(\mathbf{x},t_1|k_1), p(\hat{\mathbf{x}},t_1|k_1)),...,(s(\mathbf{x},t_{1+l}|k_{1}), p(\hat{\mathbf{x}},t_{1+l}|k_1))\big),...,\\&\big((s(\mathbf{x},t_{F_1+Pad_1-l}|k_1), p(\hat{\mathbf{x}},t_{F_1+Pad_1-l}|k_1)),...,(s(\mathbf{x},t_{F_1+Pad_1}|k_{1}), p(\hat{\mathbf{x}},t_{F_1+Pad_1}|k_1))\big),\\
&...,\big((s(\mathbf{x},t_1|k_{N_{tr}}), p(\hat{\mathbf{x}},t_1|k_{N_{tr}})),...,(s(\mathbf{x},t_{1+l}|k_{N_{tr}}), p(\hat{\mathbf{x}},t_{1+l}|k_{N_{tr}}))\big),...,\\&\big((s(\mathbf{x},t_{F_1+Pad_{N_{tr}}-l}|k_{N_{tr}}), p(\hat{\mathbf{x}},t_{F_1+Pad_{N_{tr}}-l}|k_{N_{tr}})),...,(s(\mathbf{x},t_{F_1+Pad_{N_{tr}}}|k_{N_{tr}}), p(\hat{\mathbf{x}},t_{F_1+Pad_{N_{tr}}}|k_{N_{tr}}))\big)\}.
\end{align*}
In words, to make up $\mathcal{M}_{tr}^{l}$, each trajectory is divided into time-windows of size $l$; this means that a datapoint is not a single tuple $(s(\mathbf{x},t_i|k_j), p(\hat{\mathbf{x}},t_i|k_j))$ as in $\mathcal{M}_{tr}^1$, but rather $D^{b,l,j} = \{(s(\mathbf{x},t_i|k_j), p(\hat{\mathbf{x}},t_i|k_j))\}_{i=b}^{b+l-1}$, so a set of tuples going from $t_b$ to $t_{b+l-1}$, with $b=n\,l+1,n\in\mathbb{N}$. Because the length of a trajectory might not be divisible by $l$, we pad each trajectory with zeroes by a quantity $Pad_i = (l-F_i\mod(l))\mod(l)$. The choice of dividing the training dataset in tuples like $D^{b,l,j}$ is driven by the \textit{autoregressive} way the NODE is trained, as explained in Subsection \ref{subsec:NODE}. The size of $l$ can be increased during training to use an adaptive window technique, as explained in Appendix \ref{appendix:appendix_TF_AR}.

\subsection{Coupling of the vessel with the primary circuit}
\begin{figure}[h]
    \centering
    \includegraphics[width=0.5\textwidth]{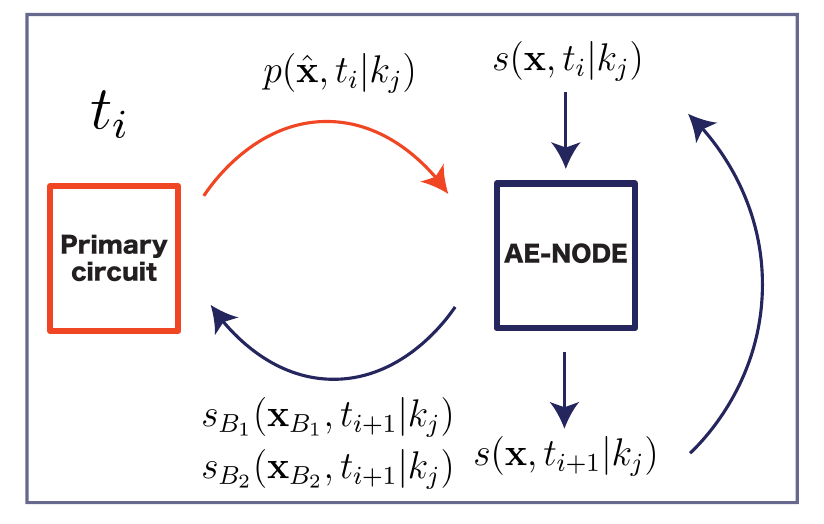}
    \caption{Coupling of the surrogate model (AE-NODE) with the primary circuit. At time $t_i$, the primary circuit computes $p(\hat{\mathbf{x}},t_{i}|k_j)$ which is used by AE-NODE, together with $s(\mathbf{x},t_{i}|k_j)$ as input to compute $s_{B_1}(\mathbf{x}_{B_1},t_{i+1}|k_j)$, $s_{B_2}(\mathbf{x}_{B_2},t_{i+1}|k_j)$ and $s(\mathbf{x},t_{i+1}|k_j)$. At the next iteration $t_{i+1}$, the primary circuit reads $s_{B_1}$ and $s_{B_2}$ and computes $p(\hat{\mathbf{x}},t_{i+1}|k_j)$. Notice that AE-NODE takes autoregressively as input its previous output $s(\mathbf{x},t_{i}|k_j)$.}
    \label{fig:coupling_explained}
\end{figure}

The modules CESAR and ICARE are coupled by models with time-steps $\delta t_1$ and $\delta t_2$ as displayed in Figure \ref{fig:icare_cesare_coupling} and explained in Subsection \ref{subsec:the_vessel}. However, we only consider the values of $s$ at the macro time-steps of ASTEC belonging to $\mathbf{T}^j$, thus ignoring the intermediate steps with micro time-steps $\delta t_1$ and $\delta t_2$ (Figure \ref{fig:icare_cesare_coupling}). This means that the SM has to predict $s(\mathbf{x},t_{i+1}|k_j)$ \textit{given} the state of the system at the previous time-step $s(\mathbf{x},t_{i}|k_j)$ \textbf{in the vessel} and the state of the system $p(\hat{\mathbf{x}},t_{i}|k_j)$ at the previous time-step in the \textbf{first volumes of the primary domain} which is \textbf{given by the primary circuit}. By using $p(\hat{\mathbf{x}},t_{i}|k_j)$ as input to the SM and by having the SM predicting $s_{B_1}$ and $s_{B_2}$, we are completely detaching the vessel from the rest of the reactor and we are posing the foundation for future work, where the prediction of the variables in the primary domain can be done by another SM or by ASTEC itself. In Figure \ref{fig:coupling_explained} we show how the SM developed in this work can be coupled with the primary domain: the primary circuit model computes $p(\hat{\mathbf{x}},t_{i}|k_j)$ at time $t_i$, which is used as input together with $s(\mathbf{x},t_{i}|k_j)$ by the vessel's SM to predict $s(\mathbf{x},t_{i+1}|k_j)$. The primary circuit model takes then the variables $s_{B_1}(\mathbf{x}_{B_1},t_{i+1}|k_j)$ and $s_{B_2}(\mathbf{x}_{B_2},t_{i+1}|k_j)$ and computes $p(\hat{\mathbf{x}},t_{i+1}|k_j)$.

The operator actions considered in this work and described in Subsection \ref{subsec:Sampling of operator actions} do not act directly on the vessel domain; for this reason they \textbf{are not} an \textit{explicit} input to the SM, but rather they are an \textit{implicit} input whose effect is processed by the SM through the information coming from the volumes $h_1$ and $c_1$. 

\section{Surrogate modeling methodology}
\label{sec:SM_methodology}
In order to construct a \textit{data-driven} SM, we adapt AE-NODE from \cite{Longhi2026} to model the physics of the vessel detailed in Subsection \ref{subsec:the_vessel}. AE-NODE is built upon \textbf{two} paradigms: \textit{Dimensionality Reduction} (DR), described in Subsection \ref{subsec:DR}, and \textit{Neural Ordinary Differential Equations} (NODEs), described in Subsection \ref{subsec:NODE}.
\subsection{Dimensionality Reduction}
\label{subsec:DR}
The idea of Dimensionality Reduction (DR) is well consolidated in data-driven models, both in a classic Reduced Order Model perspective and in a modern Deep Learning perspective, where it is known under the name of \textit{manifold hypothesis} \cite{Fefferman2016, higgins2018towards}. The assumption is that \textit{high-dimensional} objects, like the solution function $s$ of Equation \ref{eq:PDEsystem_vessel}, actually live in a \textit{low-dimensional} manifold, i.e., their intrinsic number of degrees of freedom is considerably lower than the dimensionality of their original domain. In our case, this means that although the cardinality of the solution space of $s$ is $N = {d_g} + {d_p} + {d_{cr} \times 3 \times 12} + {d_v \times 5 \times 15} + {d_{B_1}} + {d_{B_2}}=1913$, the \textbf{high-dimensional} physical state of the vessel, i.e., $s(\mathbf{x},t_i|k_j)$, can be described by a \textbf{low-dimensional} vector $\varepsilon(t_i|k_j)\in\mathcal{E}\subseteq\mathbb{R}^\lambda$, with $\lambda<< N$. 
\begin{figure}[h]
    \centering
    \includegraphics[width=1.0\textwidth]{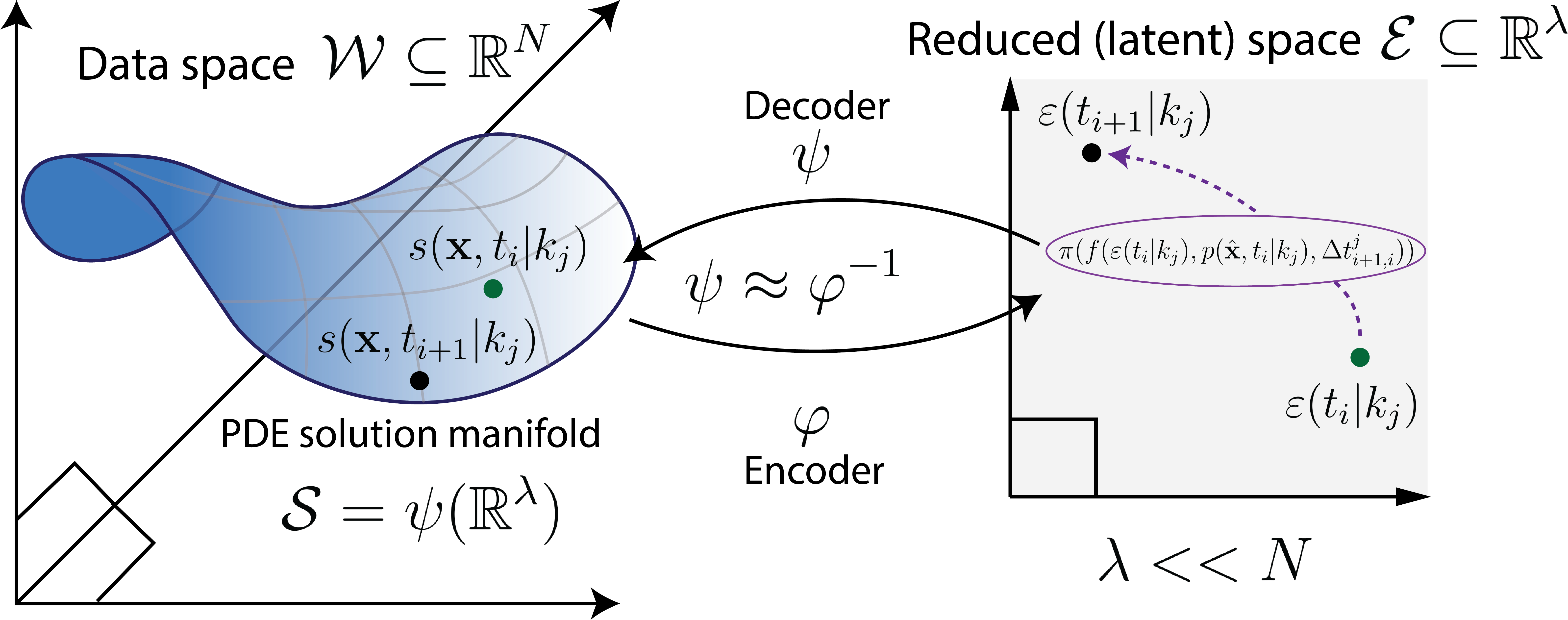}
    \caption{On the left the data space $\mathcal{W}\subset\mathbb{R}^N$ to which the PDE solution $s$ \textbf{potentially} belongs is displayed together with the \textbf{actual} PDE solution manifold $\mathcal{S}$, to which the solution $s$ is constrained by the PDE. Since the dimensionality of $\mathcal{S}$ is $\lambda<<N$, we can describe the solution $s(\mathbf{x},t_i|k_j)$ with a vector $\varepsilon(t_i|k_j)\in\mathcal{E}\subset\mathbb{R}^\lambda$ of only $\lambda$ dimensions displayed on the right. The \textit{Decoder} $\psi:\mathbb{R}^\lambda\rightarrow\mathcal{S}$ maps the low-dimensional $\varepsilon(t_i|k_j)$ into the high-dimensional $s(\mathbf{x},t_i|k_j)$, while the \textit{Encoder} $\varphi:\mathcal{S}\rightarrow\mathbb{R}^\lambda$ performs the opposite mapping (from $s(\mathbf{x},t_i|k_j)$ to $\varepsilon(t_i|k_j)$), with $\psi\approx\varphi^{-1}$.  $\varepsilon(t_{i}|k_j)$ is mapped into $\varepsilon(t_{{i+1}}|k_j)$ by the \textit{Processor} $\pi$, defined in Equation \ref{eq:processor}, given the state of the system at the boundaries $p(\hat{\mathbf{x}},t_i|k_j)$. Image adapted from Keenan Crane's illustrations \cite{crane_autoencoder}.}
    \label{fig:DR}
\end{figure}
Figure \ref{fig:DR} represents the concept of DR graphically. On the left the data space $\mathcal{W}\subseteq\mathbb{R}^N$ is represented, where the solution of the PDE $s$ lives. In principle, $s$ could belong to \textit{any point} of $\mathcal{W}$. However, when we look at the dynamics of $s$, i.e., when we look at the trajectory of $s$ spanned over $\mathcal{W}$, $s$ is \textbf{constrained to a manifold} $\mathcal{S}$ embedded in $\mathcal{W}$. Practically, if we collect multiple data-points we are going to observe some \textbf{structure} when displaying them in $\mathbb{R}^N$, rather than \textbf{uniform randomness}. The dimensionality of the manifold is $\lambda$, and the key \textbf{insight} (and assumption), is that $\lambda<<N$. The complexity, thus non-linearity, of $\mathcal{S}$ is determined by the PDE that constrains $s$. Most standard ROM techniques \cite{ROM} that use techniques like Proper Orthogonal Decomposition (equivalent to Principal Component Analysis), are based on the assumption that $\mathcal{S}$ can be approximated by a linear subspace (assumption that breaks down if $\mathcal{S}$ is highly non-linear). The convenience of a DR framework is that, if we consider the solution points $s(\mathbf{x},t_i|k_j)$ and $s(\mathbf{x},t_{i+1}|k_j)$ on the manifold $\mathcal{S}$, we \textbf{do not need} $N$ degrees of freedom to describe them, but rather only $\lambda$: in other words, we can define a \textit{Decoder} $\psi:\mathbb{R}^\lambda\rightarrow\mathcal{S}$ that constructs the solution $s(\mathbf{x},t_i|k_j)$ from a vector $\varepsilon(t_i|k_j)\in\mathcal{E}$, where $\mathcal{E}$ can be interpreted as the space of \textbf{coordinates of the manifold} $\mathcal{S}$. Conversely, we define the \textit{Encoder} $\varphi:\mathcal{S}\rightarrow\mathbb{R}^\lambda$ as the function that maps the solution $s(\mathbf{x},t_i|k_j)$ into its reduced coordinates $\varepsilon(t_i|k_j)$. Figure \ref{fig:DR} pictures such framework: $s(\mathbf{x},t_i|k_j)$ and $s(\mathbf{x},t_{i+1}|k_j)$ are mapped by the Encoder $\varphi$ into their reduced (or latent) representations $\varepsilon(t_i|k_j)$ and $\varepsilon(t_{i+1}|k_j)$, while the Decoder $\psi$ performs the opposite mapping. Ideally, we would like the Encoder to be exactly the \textit{inverse} of the Decoder; in practice this is not true, and applying $\psi$ after $\varphi$ does not exactly match the original function $s$, thus $\psi\approx\varphi^{-1}$. The right of Figure \ref{fig:DR} anticipates how the \textbf{dynamics} is modeled by the SM, i.e., by the Processor $\pi$, as defined in Equation \ref{eq:processor}.

The application of the Encoder followed by the Decoder, $\psi\circ\varphi(s)$ is known as \textit{AutoEncoding} in Deep Learning, and the couple Encoder-Decoder is called \textit{AutoEncoder} (AE). Following \cite{Longhi2026} we aim at finding $\varphi$ and $\psi$ through a data-driven procedure, thus we parametrize $\varphi$ and $\psi$ via Neural Networks (NNs) with $\varphi_{\theta}\approx\varphi$ and $\psi_{\theta}\approx\psi$, where the subscript $\theta$ are the learned parameters of the NN. Because of the multiple variables with different topologies described in Subsection \ref{subsec:the_vessel} we need to use a more complex AE architecture than the one used in \cite{Longhi2026}. 
\begin{figure}[h]
    \centering
    \includegraphics[width=1.0\textwidth]{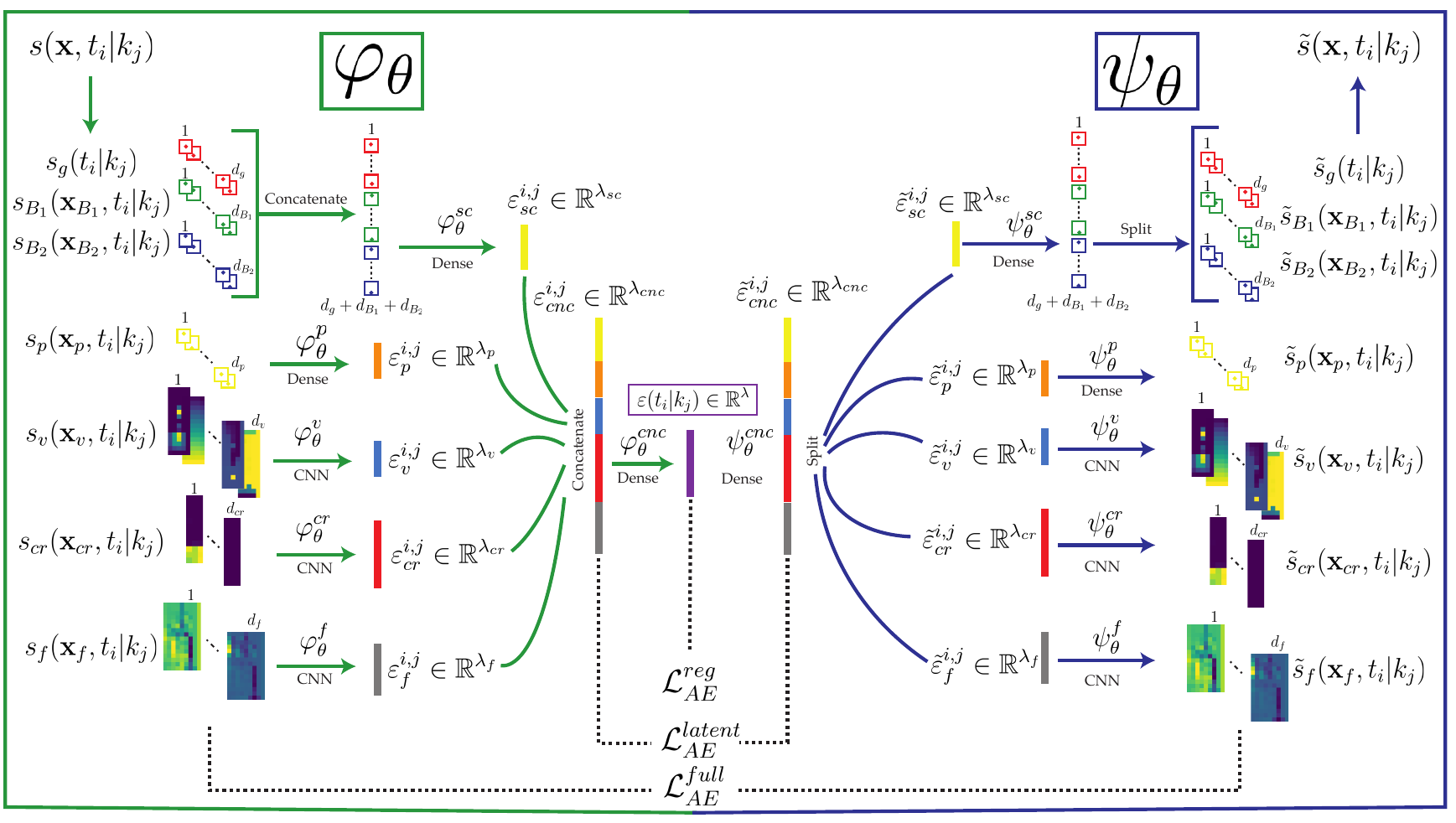}
    \caption{On the \textit{left}, the Encoder $\varphi_\theta=\{\varphi_\theta^{sc}, \varphi_\theta^{p},\varphi_\theta^{v}, \varphi_\theta^{cr}, \varphi_\theta^{f}, \varphi_\theta^{cnc}\}$ is used to encode the solution $s(\mathbf{x},t_i|k_j)$ into the latent representation $\varepsilon(t_i|k_j)\in\mathbb{R}^\lambda$ of dimension $\lambda$. On the \textit{right}, the latent vector $\varepsilon(t_i|k_j)$ is mapped by the Decoder $\psi_\theta=\{\psi_\theta^{sc},\psi_\theta^{sc}, \psi_\theta^{v}, \psi_\theta^{cr}, \psi_\theta^{f}, \psi_\theta^{cnc}\}$ into the reconstructed solution $\tilde{s}(\mathbf{x},t_i|k_j)$. The loss function $\mathcal{L}_{AE}^{reg}$ acts as a regularizer on $\varepsilon(t_i|k_j)$, $\mathcal{L}_{AE}^{latent}$ ensures the symmetry of the AE and $\mathcal{L}_{AE}^{full}$ acts on the reconstruction error of the AE.}
    \label{fig:AE}
\end{figure}
In Figure \ref{fig:AE} we show graphically how the AE is approximated at training time in a data-driven fashion. We define $5$ different sub-Encoders that make up $\varphi_\theta=\{\varphi_\theta^{sc}, \varphi_\theta^{p},\varphi_\theta^{v}, \varphi_\theta^{cr}, \varphi_\theta^{f}, \varphi_\theta^{cnc}\}$ and their corresponding sub-Decoders that make up $\psi_\theta=\{\psi_\theta^{sc},\psi_\theta^{p}, \psi_\theta^{v}, \psi_\theta^{cr}, \psi_\theta^{f}, \psi_\theta^{cnc}\}$ parametrized by NNs: 
\begin{itemize}
    \item $\varphi_\theta^{sc}:\mathcal{S}_{sc}\rightarrow\mathbb{R}^{\lambda_{sc}}$, and $\psi_\theta^{sc}:\mathbb{R}^{\lambda_{sc}}\rightarrow \mathcal{S}_{sc}$, where $\mathcal{S}_{sc}\subseteq\mathbb{R}^{d_g} \times \mathbb{R}^{d_{B_1}} \times \mathbb{R}^{d_{B_2}}$ is the solution space of the \textbf{scalar} variables (global and boundaries);
    \item $\varphi_\theta^{p}:\mathcal{S}_{p}\rightarrow\mathbb{R}^{\lambda_{p}}$, and $\psi_\theta^{p}:\mathbb{R}^{\lambda_{p}}\rightarrow \mathcal{S}_{p}$, where $\mathcal{S}_{p}\subseteq\mathbb{R}^{d_p}$ is the solution space of the variables defined in the \textbf{lower plenum};
    \item $\varphi_\theta^{v}:\mathcal{S}_{v}\rightarrow\mathbb{R}^{\lambda_{v}}$ and $\psi_\theta^{v}:\mathbb{R}^{\lambda_{v}}\rightarrow \mathcal{S}_{v}$, where $\mathcal{S}_v\subseteq \mathbb{R}^{d_v\times 15\times 5}$ is the solution space of the \textbf{vessel} variables;
    \item $\varphi_\theta^{cr}:\mathcal{S}_{cr}\rightarrow\mathbb{R}^{\lambda_{cr}}$ and $\psi_\theta^{cr}:\mathbb{R}^{\lambda_{cr}}\rightarrow \mathcal{S}_{cr}$, where $\mathcal{S}_{cr}\subseteq \mathbb{R}^{d_{cr}\times 12\times 3}$ is the solution space of the \textbf{core} variables;
    \item $\varphi_\theta^{f}:\mathcal{S}_{f}\rightarrow\mathbb{R}^{\lambda_{f}}$ and $\psi_\theta^{f}:\mathbb{R}^{\lambda_{f}}\rightarrow \mathcal{S}_{f}$, where $\mathcal{S}_{f}\subseteq \mathbb{R}^{d_{f}\times 16\times 9}$ is the solution space of the \textbf{face} variables;
    \item $\varphi_\theta^{cnc}:\mathbb{R}^{\lambda_{cnc}}\rightarrow\mathbb{R}^{\lambda}$ and $\psi_\theta^{cnc}:\mathbb{R}^{\lambda}\rightarrow\mathbb{R}^{\lambda_{cnc}}$, where $\lambda_{cnc}=\lambda_{sc}+\lambda_{p}+\lambda_{v}+\lambda_{cr}+\lambda_{f}$, which perform the \textbf{final reduction} and the \textbf{first reconstruction} steps of $\varphi_\theta$ and $\psi_\theta$.
\end{itemize}
On the left of Figure \ref{fig:AE} we can see how the Encoder $\varphi_\theta$ is modeled: at a given time step $t_i$, the variables $s_g(t_i|k_j)$, $s_{B_1}(\mathbf{x}_{B_1},t_i|k_j)$ and $s_{B_2}(\mathbf{x}_{B_2},t_i|k_j)$ are concatenated into a vector of dimension $d_g+d_{B_1}+d_{B_2}$ and mapped by $\varphi_\theta^{sc}$ into $\varepsilon_{sc}^{i,j}\in\mathbb{R}^{\lambda_{sc}}$, where we use the notation $\varepsilon_{a}^{i,j} = \varepsilon_a(t_i|k_j)$. Similarly, $s_p(\mathbf{x}_{p},t_i|k_j)$ is mapped by $\varphi_\theta^p$ into $\varepsilon_{p}^{i,j}\in\mathbb{R}^{\lambda_{p}}$. $s_{v}(\mathbf{x}_{v},t_i|k_j)$, $s_{cr}$ $(\mathbf{x}_{cr},t_i|k_j)$ and $s_f(\mathbf{x}_{f},t_i|k_j)$ are mapped into $\varepsilon_{v}^i\in\mathbb{R}^{\lambda_{v}}$, $\varepsilon_{cr}^i\in\mathbb{R}^{\lambda_{cr}}$ and $\varepsilon_{f}^i\in\mathbb{R}^{\lambda_{f}}$ by $\varphi_\theta^{v}$, $\varphi_\theta^{cr}$ and $\varphi_\theta^{f}$ respectively. The vector $\varepsilon^i_{cnc}$ is then built by concatenation of the vectors $\varepsilon^{i,j}_{sc}$, $\varepsilon^{i,j}_{p}$, $\varepsilon^{i,j}_{v}$, $\varepsilon^{i,j}_{cr}$ and $\varepsilon^{i,j}_{f}$. Finally, the \textbf{actual latent representation} $\varepsilon(t_i|k_j)\in\mathbb{R}^{\lambda}$ of dimension $\lambda$ is obtained by applying $\varphi_\theta^{cnc}$ to $\varepsilon^{i,j}_{cnc}$. 

Conversely, on the right of Figure \ref{fig:AE} we can see how the Decoder $\psi_\theta$ is modeled: the latent vector $\varepsilon(t_i|k_j)$ is given as input to $\psi_\theta^{cnc}$ and mapped into $\tilde{\varepsilon}_{cnc}^{i,j}\in\mathbb{R}^{\lambda_{cnc}}$, where the symbol $\sim$ is used to highlight that $\tilde{\varepsilon}_{cnc}^{i,j}$ is a reconstruction of $\psi_\theta^{cnc}$ which should be equal to $\varepsilon_{cnc}^{i,j}$, i.e., the input of $\varphi_\theta^{cnc}$. $\tilde{\varepsilon}_{cnc}^{i,j}$ is then split into the $5$ reconstructions $\tilde{\varepsilon}_{sc}^{i,j}\in\mathbb{R}^{\lambda_{sc}}$, $\tilde{\varepsilon}_{p}^{i,j}\in\mathbb{R}^{\lambda_{p}}$, $\tilde{\varepsilon}_{v}^{i,j}\in\mathbb{R}^{\lambda_{v}}$, $\tilde{\varepsilon}_{cr}^{i,j}\in\mathbb{R}^{\lambda_{cr}}$ and $\tilde{\varepsilon}_{f}^{i,j}\in\mathbb{R}^{\lambda_{f}}$. The decoder $\psi_\theta^{sc}$ maps $\tilde{\varepsilon}_{sc}^{i,j}\in\mathbb{R}^{\lambda_{sc}}$ into a vector of dimension $d_g+d_{B_1}+d_{B_2}$ which is then split into the final reconstructions $\tilde{s}_g(t_i|k_j)$, $\tilde{s}_{B_1}(\mathbf{x}_{B_1},t_i|k_j)$ and $\tilde{s}_{B_2}(\mathbf{x}_{B_2},t_i|k_j)$ and $\tilde{\varepsilon}_{p}^{i,j}\in\mathbb{R}^{\lambda_{p}}$ is mapped into $\tilde{s}_p(t_i|k_j)$ by $\psi_\theta^p$. Similarly, the vectors $\tilde{\varepsilon}_{v}^{i,j}\in\mathbb{R}^{\lambda_{v}}$, $\tilde{\varepsilon}_{cr}^{i,j}\in\mathbb{R}^{\lambda_{cr}}$ and $\tilde{\varepsilon}_{f}^{i,j}\in\mathbb{R}^{\lambda_{f}}$ are mapped into the reconstructions $\tilde{s}_{v}(\mathbf{x}_{v},t_i|k_j)$, $\tilde{s}_{cr}$ $(\mathbf{x}_{cr},t_i|k_j)$ and $\tilde{s}_f(\mathbf{x}_{f},t_i|k_j)$ by the decoders $\psi_\theta^v$, $\psi_\theta^{cr}$ and $\psi_\theta^f$.

 As specified in Subsection \ref{subsec:training_set_construction}, during training a data-point is not a single point in time, but rather a time-series $D^{b,l,j}$ of length $l$; this is not a problem as $D^{b,l,j}$ can be processed in parallel by $\varphi_\theta$ and $\psi_\theta$. Therefore, we define the loss functions for the AE training as
\begin{align}
\mathcal{L}_{AE}^{reg}\!\left(D^{b,l,j}\right) &= \frac{1}{l\cdot\lambda}\sum_{i=b}^{b+l-1}\varepsilon^{i,j}\cdot\varepsilon^{i,j}, \\
\mathcal{L}_{AE}^{latent}\!\left(D^{b,l,j}\right) &= \frac{1}{l\cdot\lambda_{cnc}}\sum_{i=b}^{b+l-1}\left\|\varepsilon_{cnc}^{i,j}-\tilde{\varepsilon}_{cnc}^{i,j}\right\|_2^2, \\
\mathcal{L}_{AE}^{full}\!\left(D^{b,l,j}\right) &= \frac{1}{l\cdot N}\sum_{i=b}^{b+l-1}\left\|s\!\left(\mathbf{x},t_i\,\middle|\,k_j\right)-\tilde{s}\!\left(\mathbf{x},t_i\,\middle|\,k_j\right)\right\|_2^2,
\end{align}
where $\mathcal{L}_{AE}^{reg}$ acts as a regularizer, $\mathcal{L}_{AE}^{latent}$ ensures that the AutoEncoder is symmetric and $\mathcal{L}_{AE}^{full}$ is the reconstruction error in the solution space.
\subsection{Neural Ordinary Differential Equations (NODE)}
\label{subsec:NODE}
\begin{figure}[h]
    \centering
    \includegraphics[width=0.8\textwidth]{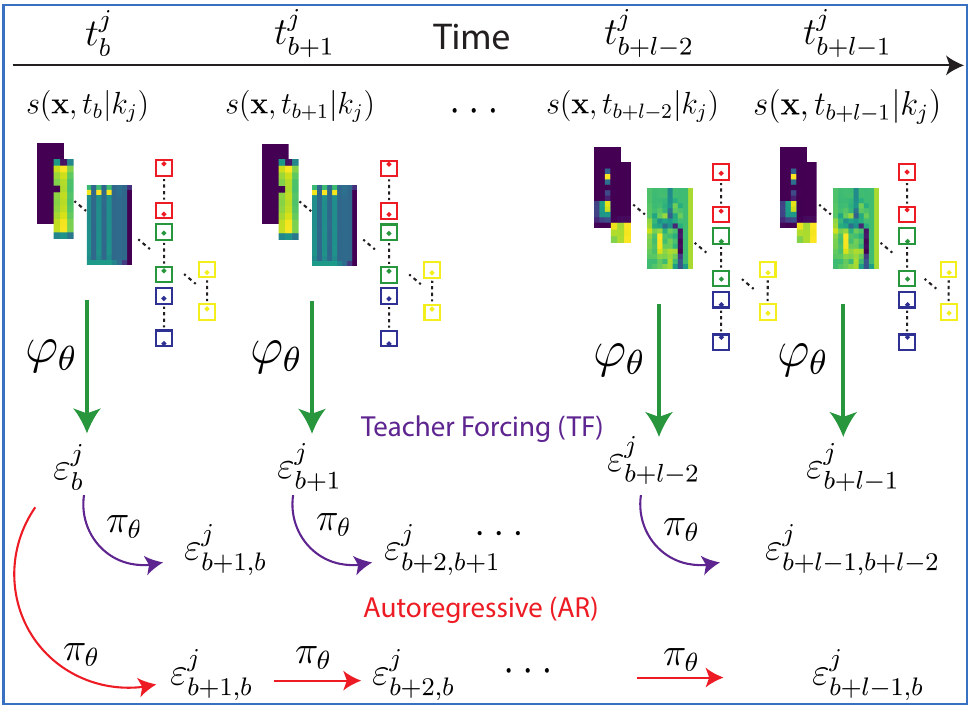}
    \caption{The training procedure to optimize $f_\theta$ during training is split between a Teacher Forcing (TF) and an Autoregressive (AR) approach. At the top the solution fields of a data-point $D^{b,l,j}$, arranged from the left to the right, are encoded into their latent representations going from $\varepsilon_b^{j}$ to $\varepsilon_{b+l}^{j}$. In the TF approach, the Processor $\pi_\theta$ is only applied once to $\varepsilon^j_{b+i}$ to give a latent prediction $\varepsilon^j_{b+i+1,b+i}$ at the next time-step, which then is compared to the encoding of the true state $\varepsilon^j_{b+i+1}$. In the AR approach instead, $\pi_\theta$ is applied autoregressively multiple times starting from $\varepsilon_b^{j}$ up to the prediction $\varepsilon_{b+l-1,b}^{j}$, mimicking what happens at testing time. We remark that $\pi_\theta$ takes as input the boundary conditions $p(\hat{\mathbf{x}},t_i|k_j)$ at timestep $t_i^j$, as defined in Equation \ref{eq:processor}.}
    \label{fig:TF_AR}
\end{figure}
In Subsection \ref{subsec:DR} we described how the AE is modeled and optimized during training. Using $\varphi_\theta$ and $\psi_\theta$ provides us with the latent space $\mathcal{E}$, however it does not describe how the vector $\varepsilon(t_i|k_j)$ evolves in time within $\mathcal{E}$, i.e., we require a model that describes the evolution from $\varepsilon(t_i|k_j)$ to $\varepsilon(t_{i+1}|k_j)$. To do so, we model the evolution of $\varepsilon(t_i|k_j)$ according to the ODE
\begin{equation}
\label{eq:ODE}
    \frac{d}{dt}\varepsilon(t|k_j) =f(\varepsilon(t|k_j),p(\hat{\mathbf{x}},t|k_j)),\quad t\in\mathbb{R}^+,
\end{equation}
from which it follows that
\begin{equation}
\label{eq:processor}
\begin{split}
    \varepsilon(t_{i+1}|k_j) &= \varepsilon(t_{i}|k_j)+\int_{t_i^j}^{t_{i+1}^j} f(\varepsilon(t'|k_j),p(\hat{\mathbf{x}},t_i|k_j))dt' = \\
    &= \pi(f(\varepsilon(t_{i}|k_j),p(\hat{\mathbf{x}},t_i|k_j)), \Delta t_{i+1,i}^j) = \pi(\varepsilon(t_{i}|k_j),\Delta t_{i+1,i}^j)
\end{split}
\end{equation}
where $\Delta t_{i+1,i}^j = t^j_{i+1}-t^j_{i}\in\mathbb{R}^+$, $f:\mathcal{E}\times\Omega_P\rightarrow\mathcal{E}$ is the time derivative of $\varepsilon(t|k)$ and $\pi:f:\mathcal{E}\times\mathbb{R^+}\rightarrow\mathcal{E}$ is the \textit{Processor} that advances in time the latent vector $\varepsilon$. In the final equality we are dropping the dependency on $f$ and $p$ for notational convenience. Note that $p(\hat{\mathbf{x}},t_i|k_j)$ is an \textbf{explicit} input to $f$: in order to go from $\varepsilon(t_{i}|k_j)$ to $\varepsilon(t_{i+1}|k_j)$ $f$ needs the information coming from the primary circuit $\hat{\mathbf{x}}\in\Omega_{P}$ as the vessel can be informed about any variation in the physics of the reactor caused by the activation of an operator action only by the state of the variables $p$ in $\hat{\mathbf{x}}$. Additionally, $p$ is assumed to be constant between $t_i^j$ and $t_{i+1}^j$. Such model is depicted on the right of Figure \ref{fig:DR}, where the Processor $\pi$ acts on $\varepsilon(t_i|k_j)$ and evolves it into $\varepsilon(t_{i+1}|k_j)$. The core of modeling the latent dynamics boils down to finding the derivative $f$. We thus approximate $f$ with a NN $f_\theta\approx f$, and as a consequence we can define $\pi_\theta\approx\pi$ as the approximated Processor. $\pi_\theta$ in practice solves the integral in Equation \ref{eq:processor} by standard numerical techniques such as the explicit Runge-Kutta method \cite{Ascher1998ComputerMF}.

In order to find the optimal weights of $f_\theta$ during training we set up a dual training strategy as described in \cite{Longhi2026}, by combining a \textit{Teacher Forcing} (TF) and an \textit{Autoregressive} (AR) approaches, described in Figure \ref{fig:TF_AR}; at the top the solution fields of a data-point $D^{b,l,j}$ are arranged from the left to the right, starting with $s(\mathbf{x},t_b|k_j)$ and ending with $s(\mathbf{x},t_{b+l-1}|k_j)$. The solution fields are then encoded into their latent representations going from $\varepsilon_b^{j}$ to $\varepsilon_{b+l}^{j}$. We now define $\varepsilon_{r,n}^{j} = \pi_\theta(\varepsilon_{r-1,n}^{j},\Delta t^j_{r,r-1})\circ\cdots\circ\pi_\theta(\varepsilon_n^j,\Delta^j_{n+1,n})\circ\varphi_\theta(s(\mathbf{x},t_n|k_j))$ as the latent vector obtained by encoding the solution at time $t_n^j$ and applying the processor $\pi_\theta$ $r-n$ times  up to the vector $\varepsilon_{r,n}^{j}$. Additionally, we define $\tilde{s}_{r,n}(\mathbf{x},t|k_j) = \psi_\theta(\varepsilon_{r,n}^{j})$. If $r-n=1$, then we call such procedure \textit{Teacher Forcing (TF)}, while if $r-n>1$ then we call it \textit{Autoregressive (AR)}, as shown in Figure \ref{fig:TF_AR}. Ideally we would only want to use the AR approach during training, starting from the very initial condition at $t_1^j$ up to $t_{F_j}^j$, however this leads to large instabilities of the gradients computed by backpropagation during the training and to high training times. This is why the TF approach is important during the initial phase of the training, as it allows the AE to be coupled to the NODE in the most simple setting, while more aggressive AR (large $l$) approaches, with $l$ increased dynamically during the training, are used at later stages of the training procedure. If the AR approach is not used, the training procedure might be subjected to the \textit{distribution shift} \cite{brandstetter2022message}: by using only TF during training, $\pi_\theta$ is always given as input the ground truth, while in practice at inference time $\pi_\theta$ is given as input its previous output, as shown in Figure \ref{fig:inference}. More details about the implementation and motivations of such procedures can be found in Appendix \ref{appendix:appendix_TF_AR}. Based on the TF and AR approaches, we define two loss functions that are used to optimize the weights of $f_\theta$. Given a data-point $D^{b,l,j}$, we define the TF and the AR losses as
\begin{align}
\mathcal{L}_{TF}\!\left(D^{b,l,j}\right) &= \frac{1}{(l-1)\cdot\lambda}\sum_{i=b}^{b+l-2}\left\|\pi_\theta\!\left(\varepsilon_{i}^j,\Delta t^j_{i+1,i}\right)-\varepsilon_{i+1}^j\right\|_2^2, \\
\mathcal{L}_{AR}\!\left(D^{b,l,j}\right) &= \frac{1}{(l-1)\cdot\lambda}\sum_{i=b}^{b+l-2}\left\|\varepsilon^j_{i+1,b}-\varepsilon^j_{i+1}\right\|_2^2.
\end{align}
\begin{figure}[h]
    \centering
    \includegraphics[width=1.0\textwidth]{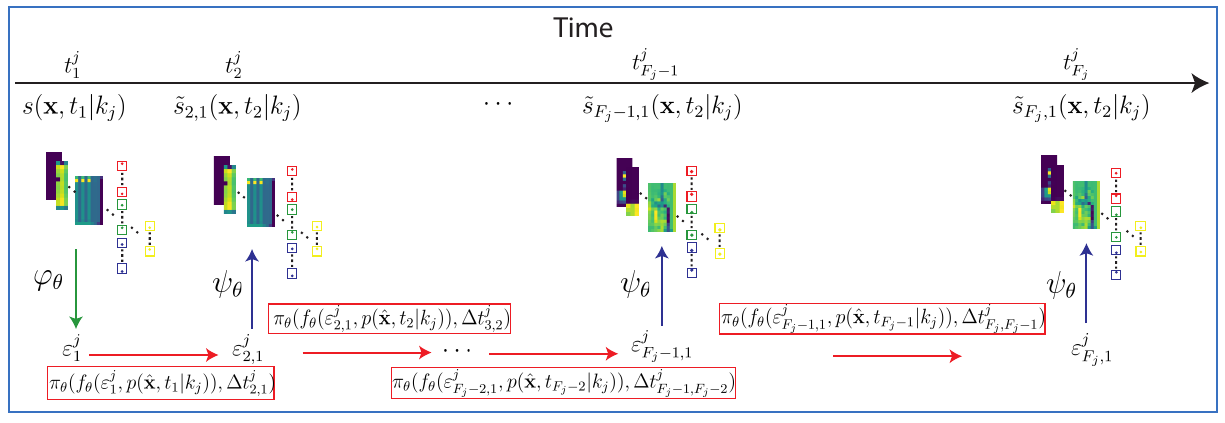}
    \caption{The application of AE-NODE at inference time. From the left: the initial condition $s(\mathbf{x},t_1|k_j)$ is encoded by $\varphi_\theta$ into its latent representation $\varepsilon_1^j$. $\varepsilon_1^j$ is then evolved in time by repeated application of the Processor $\pi_\theta$ to get the predicted latent vectors $\varepsilon_{i,1}^j$ with $i = 2,...,F_j$. Each $\varepsilon_{i,1}^j$ is mapped into the corresponding PDE solution prediction $\tilde{s}_{i,1}(\mathbf{x},t_i|k_j)$ by the Decoder $\psi_\theta$.}
    \label{fig:inference}
\end{figure}
We finally define the total loss function as
\begin{equation}
\begin{aligned}
\mathcal{L}_{tot}\!\left(D^{b,l,j}\right) ={}& \alpha\,\mathcal{L}_{AE}^{reg}\!\left(D^{b,l,j}\right) + \beta\,\mathcal{L}_{AE}^{latent}\!\left(D^{b,l,j}\right) + \gamma\,\mathcal{L}_{AE}^{full}\!\left(D^{b,l,j}\right) \\
&+ \delta\,\mathcal{L}_{TF}\!\left(D^{b,l,j}\right) + \nu\,\mathcal{L}_{AR}\!\left(D^{b,l,j}\right) + \mu\,\mathcal{L}_{t_m}\!\left(D^{b,l,j}\right) \\
&+ \omega\,\mathcal{L}_{AR}^{full}\!\left(D^{b,l,j}\right),
\end{aligned}
\end{equation}
with $\alpha,\beta,\gamma,\delta,\nu,\mu,\omega\in\mathbb{R}^+$ scalar importance weights. $\mathcal{L}_{t_m}$ and $\mathcal{L}_{AR}^{full}$ are defined in Appendix \ref{appendix:training_details_and_speed_of_inference} as they help the training process but do not add any understanding to the main methodology. The AE optimization and the NODE optimization are \textbf{coupled} through $\mathcal{L}_{TF}$ and $\mathcal{L}_{AR}$. Since $\varepsilon_i^j = \varphi_\theta(s(\mathbf{x},t_i|k_j))$, the $\varphi_\theta$ weights are updated also by gradients computed by  $\mathcal{L}_{TF}$ and $\mathcal{L}_{AR}$. During training, we optimize $\mathcal{L}_{tr} = \frac{1}{N_{tr}}\sum_{D^{b,l,j}\in\mathcal{W}_{tr}^l}\mathcal{L}_{tot}(D^{b,l,j})$, with $N_{tr}$ being the total number of data-points in $\mathcal{D}_{tr}^l$. More technical details concerning the training process and the two additional loss functions are given in Appendix \ref{appendix:training_details_and_speed_of_inference}. The specifications concerning the architectural shape of $f_\theta, \psi_\theta$ and $\varphi_\theta$ are described in Appendix \ref{appendix:architectural_details}.

In Figure \ref{fig:inference} we show how $\varphi_\theta$, $\psi_\theta$ and $\pi_\theta$ are jointly used at \textbf{inference time} to give the full spatio-temporal prediction for a given set of operator actions $k_j$. From the left: the initial condition $s(\mathbf{x},t_1|k_j)$ is encoded by $\varphi_\theta$ into its latent representation $\varepsilon_1^j$. $\varepsilon_1^j$ is then evolved in time by repeated application of the Processor $\pi_\theta$ to get the predicted latent vectors $\varepsilon_{i,1}^j$ with $i = 2,...,F_j$. Each $\varepsilon_{i,1}^j$ is mapped into the corresponding PDE solution prediction $\tilde{s}(\mathbf{x},t_i|k_j)$ by the Decoder $\psi_\theta$. At each time point $t_i^j$, $f_\theta$ in $\pi_\theta$ is given as input both the \textbf{state of the vessel} (encoded in $\varepsilon_{i,1}^j$) and the \textbf{boundary conditions}.

The methodology described in Section \ref{sec:SM_methodology} is completely \textbf{general} and can be applied to \textbf{any} physical system modeled by PDEs, even with multiple variables with general spatial domains, as the specific details of the PDEs would only change the mathematical forms of the learnt $f_\theta, \psi_\theta$ and $\varphi_\theta$ approximations.
\section{Results}
\label{sec:results}
In this section we present the results obtained training the SM described in Section \ref{sec:SM_methodology}. In order to better understand the capabilities of the SM, we do two experiments: in Subsection \ref{subsec:LOCA_experiment} we build a SM for the LOCA scenario and in Subsection \ref{subsec:SBO_experiment} for the SBO. We use the metrics $\text{RMSE}_{mean}(vr)$, $\text{RMSE}_{max}(vr)$ and $\text{RMSE}_{std}(vr)$, which are the Root-Mean-Squared-Error divided by the mean, maximum and standard deviations of $s$ as described in Appendix \ref{appendix:metrics}.
\subsection{LOCA scenario}
\label{subsec:LOCA_experiment}
For the LOCA scenario we used $700$ trajectories for training, $100$ for validation and $40$ for testing. 
\begin{figure}[h]
    \centering
    \includegraphics[width=1.0\textwidth]{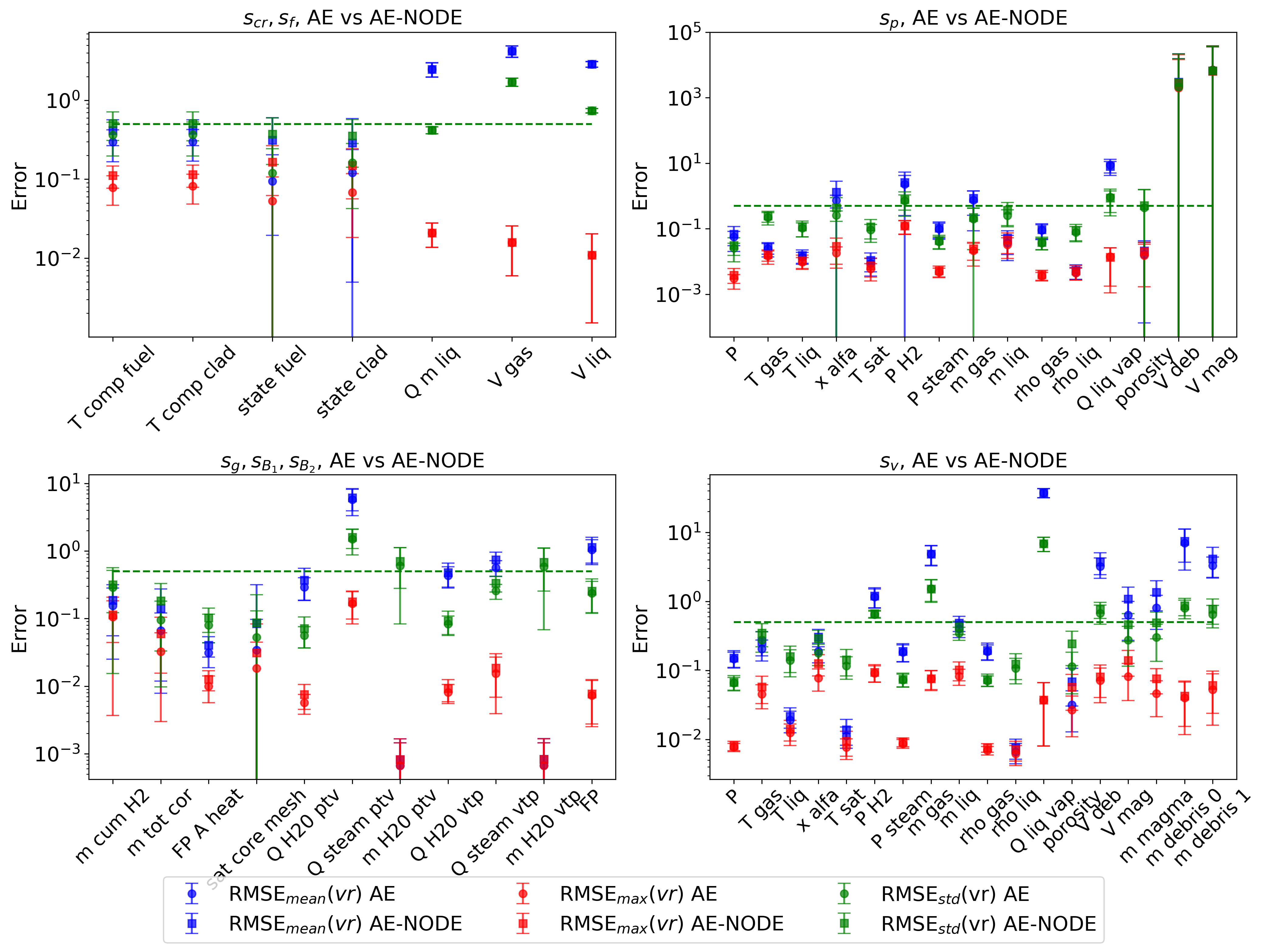}
    \caption{Comparison of $\text{RMSE}_{mean}$, $\text{RMSE}_{max}$ and $\text{RMSE}_{std}$ metrics when doing a simple AutoEncoding (AE) and when running the actual inference (AE-NODE) on the LOCA testing set. The uncertainty bars are the standard deviations of the metrics computed across testing trajectories. The green horizontal line is placed at the value 0.5, which is a common baseline for $\text{RMSE}_{std}$.}
    \label{fig:LOCA_AE_AE_NODE_comparison}
\end{figure}
In Figure \ref{fig:LOCA_AE_AE_NODE_comparison} we show the values of $\text{RMSE}_{mean}(vr)$, $\text{RMSE}_{max}(vr)$ and $\text{RMSE}_{std}(vr)$ (computed on the test set), for all the variables defined in Appendix \ref{appendix:complete_variables} in 2 cases: when doing a simple AutoEncoding of the test set (AE), and when running the full AE-NODE pipeline.

\begin{figure}[htbp]
    \centering
    \begin{subfigure}[b]{0.24\textwidth}
        \includegraphics[width=\textwidth]{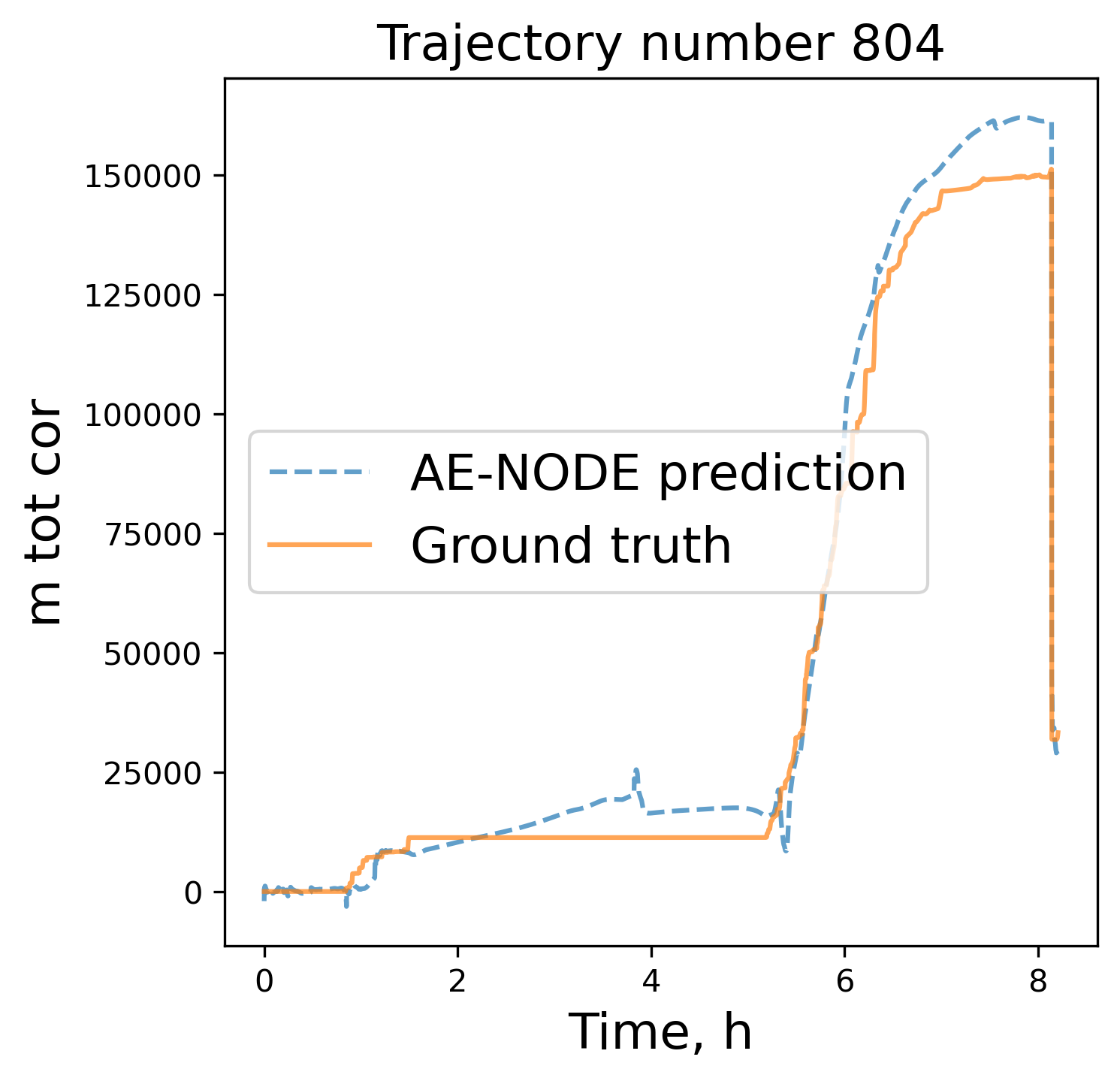}
        \caption{Trajectory 804}
        \label{fig:804_m_tot_cor_LOCA}
    \end{subfigure}
    \hfill
    \begin{subfigure}[b]{0.24\textwidth}
        \includegraphics[width=\textwidth]{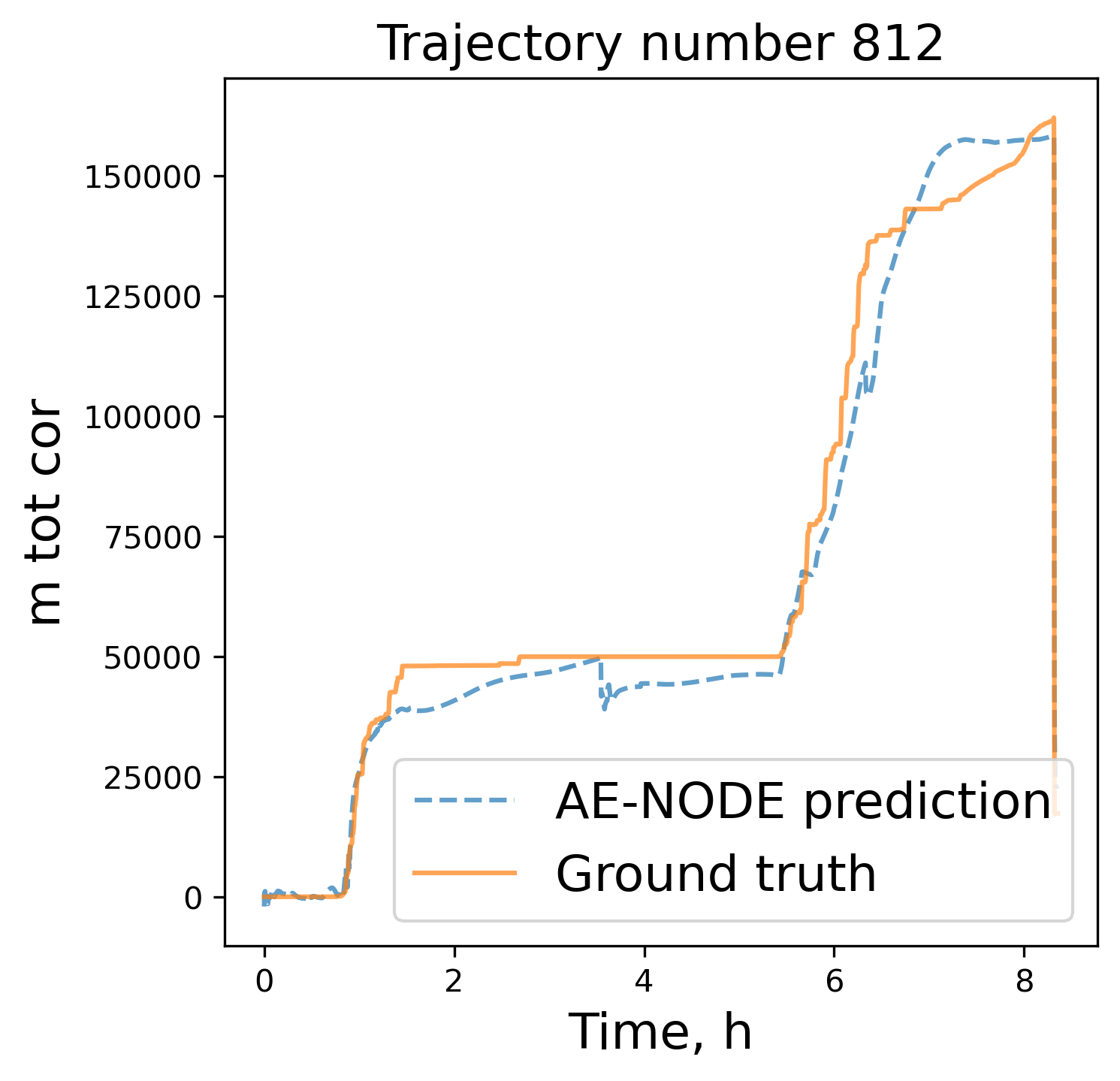}
        \caption{Trajectory 812}
        \label{fig:812_m_tot_cor_LOCA}
    \end{subfigure}
    \hfill
    \begin{subfigure}[b]{0.24\textwidth}
        \includegraphics[width=\textwidth]{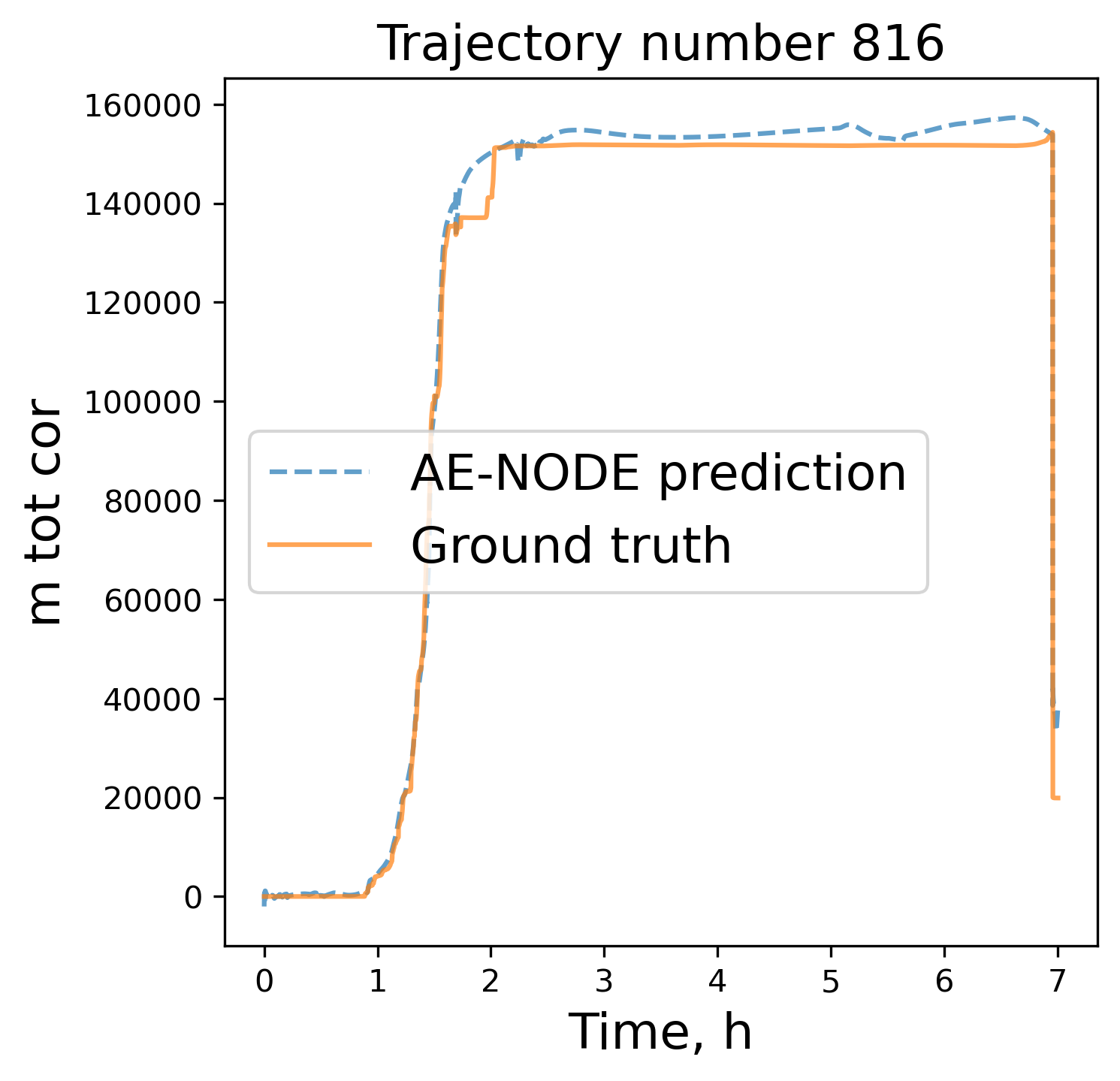}
        \caption{Trajectory 816}
        \label{fig:816_m_tot_cor_LOCA}
    \end{subfigure}
    \hfill
    \begin{subfigure}[b]{0.24\textwidth}
        \includegraphics[width=\textwidth]{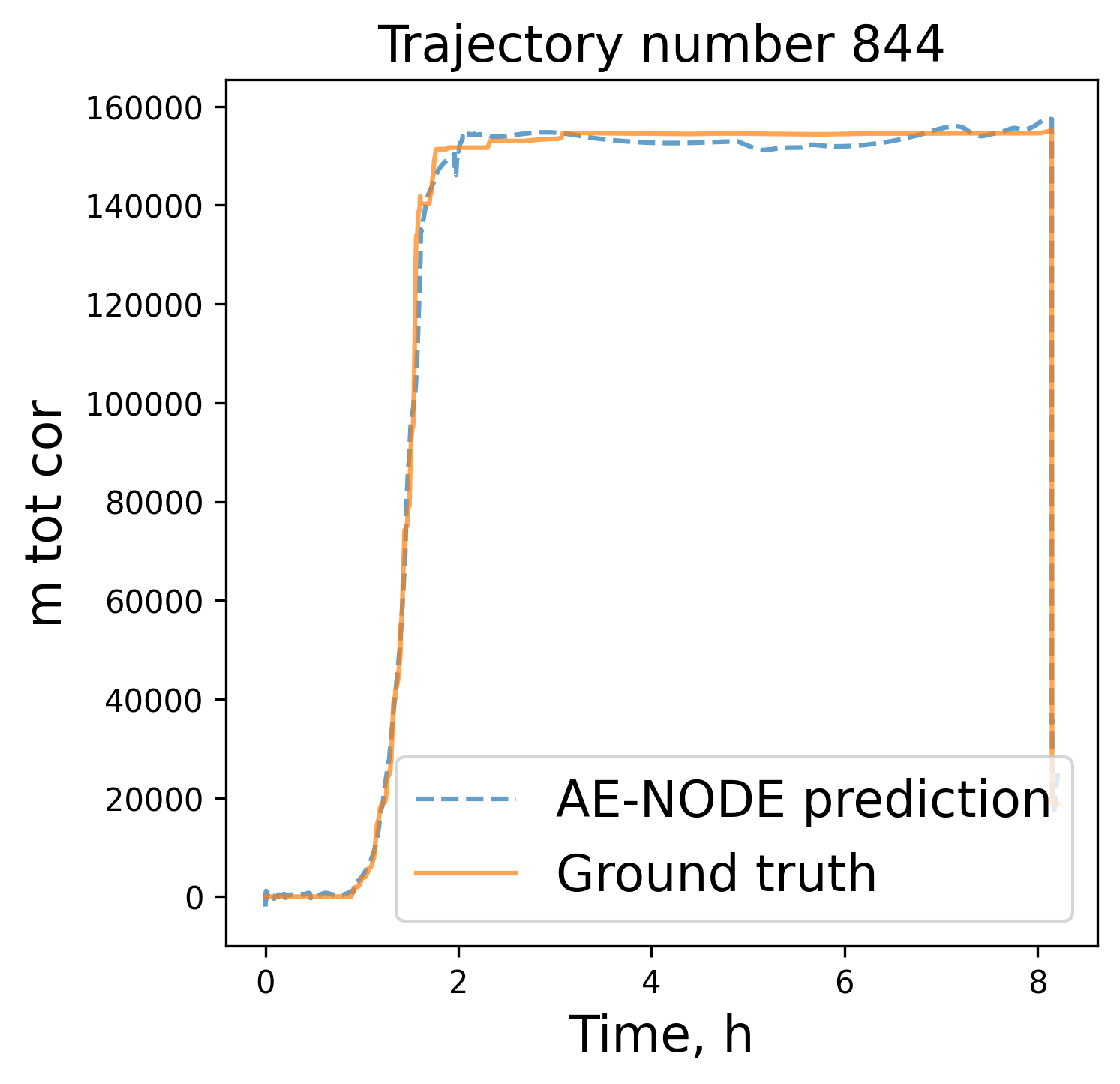}
        \caption{Trajectory 844}
        \label{fig:844_m_tot_cor_LOCA}
    \end{subfigure}
    \caption{AE-NODE predictions vs.\ ground truth for selected (testing) LOCA trajectories of the total corium mass [kg] over time.}
    \label{fig:LOCA_m_tot_cor}
\end{figure}

\begin{figure}[htbp]
    \centering
    \begin{subfigure}[b]{0.24\textwidth}
        \includegraphics[width=\textwidth]{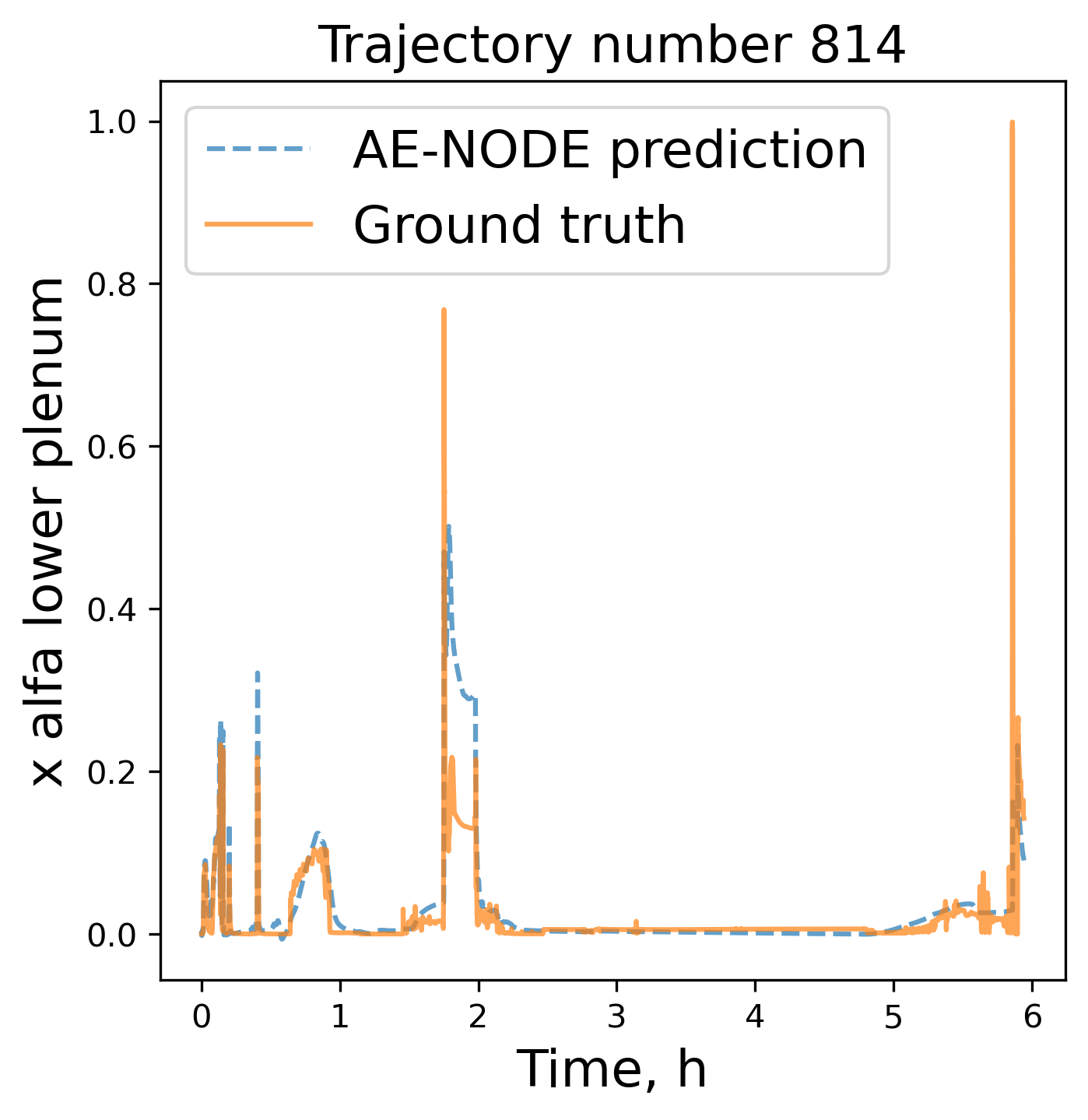}
        \caption{Trajectory 814}
        \label{fig:814_x_alfa_lower_plenum_LOCA}
    \end{subfigure}
    \hfill
    \begin{subfigure}[b]{0.24\textwidth}
        \includegraphics[width=\textwidth]{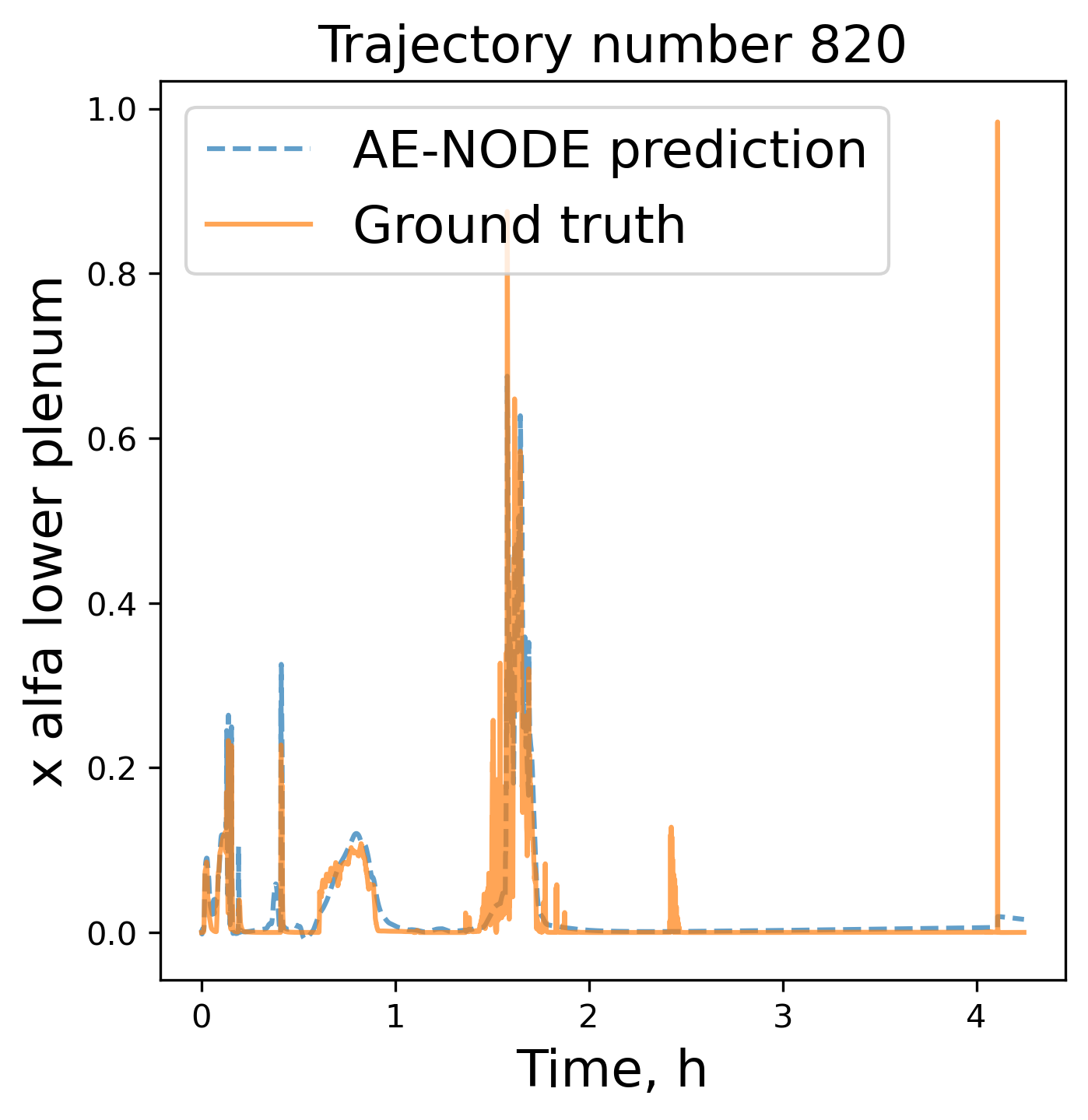}
        \caption{Trajectory 820}
        \label{fig:820_x_alfa_lower_plenum_LOCA}
    \end{subfigure}
    \hfill
    \begin{subfigure}[b]{0.24\textwidth}
        \includegraphics[width=\textwidth]{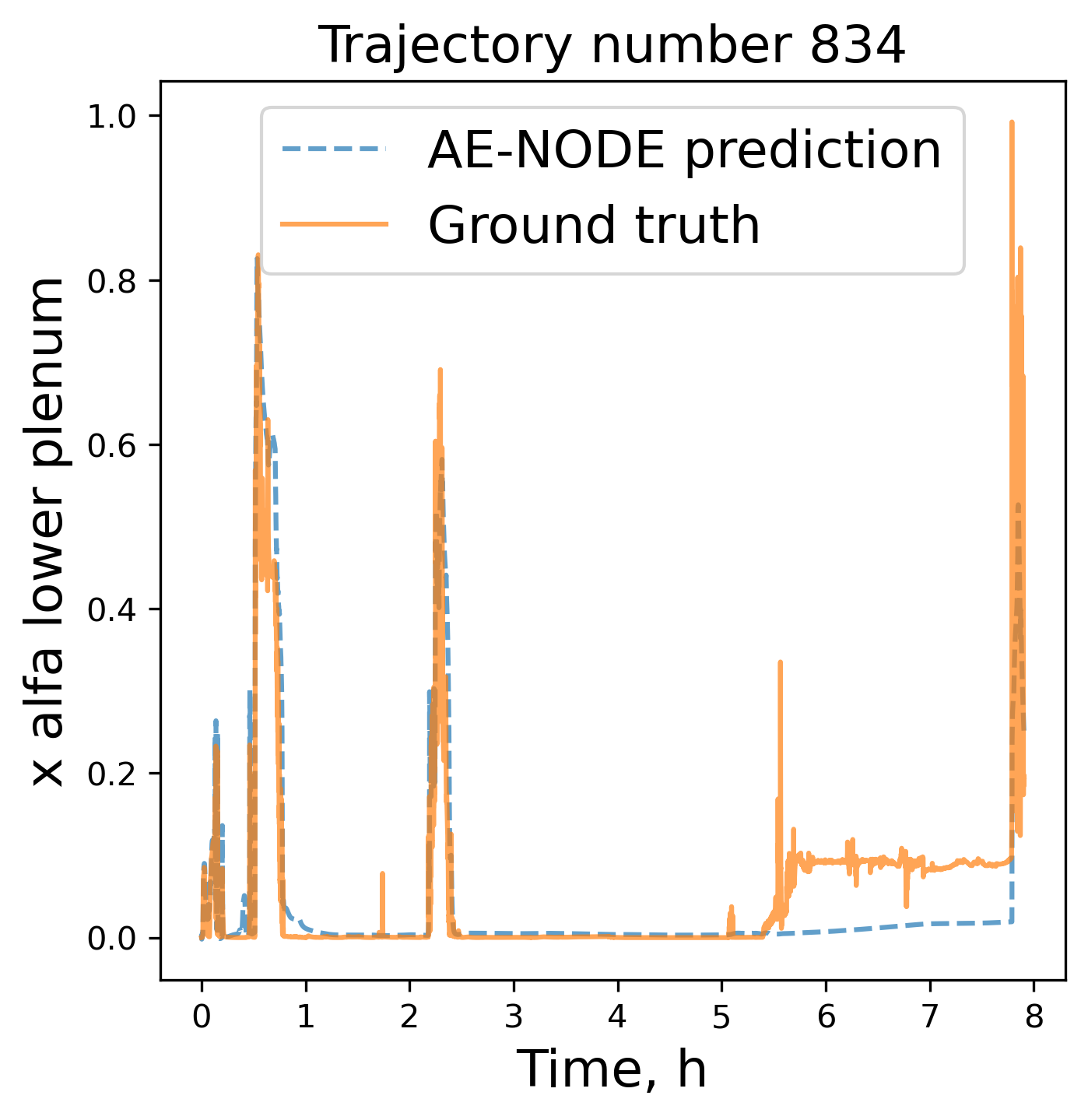}
        \caption{Trajectory 834}
        \label{fig:834_x_alfa_lower_plenum_LOCA}
    \end{subfigure}
    \hfill
    \begin{subfigure}[b]{0.24\textwidth}
        \includegraphics[width=\textwidth]{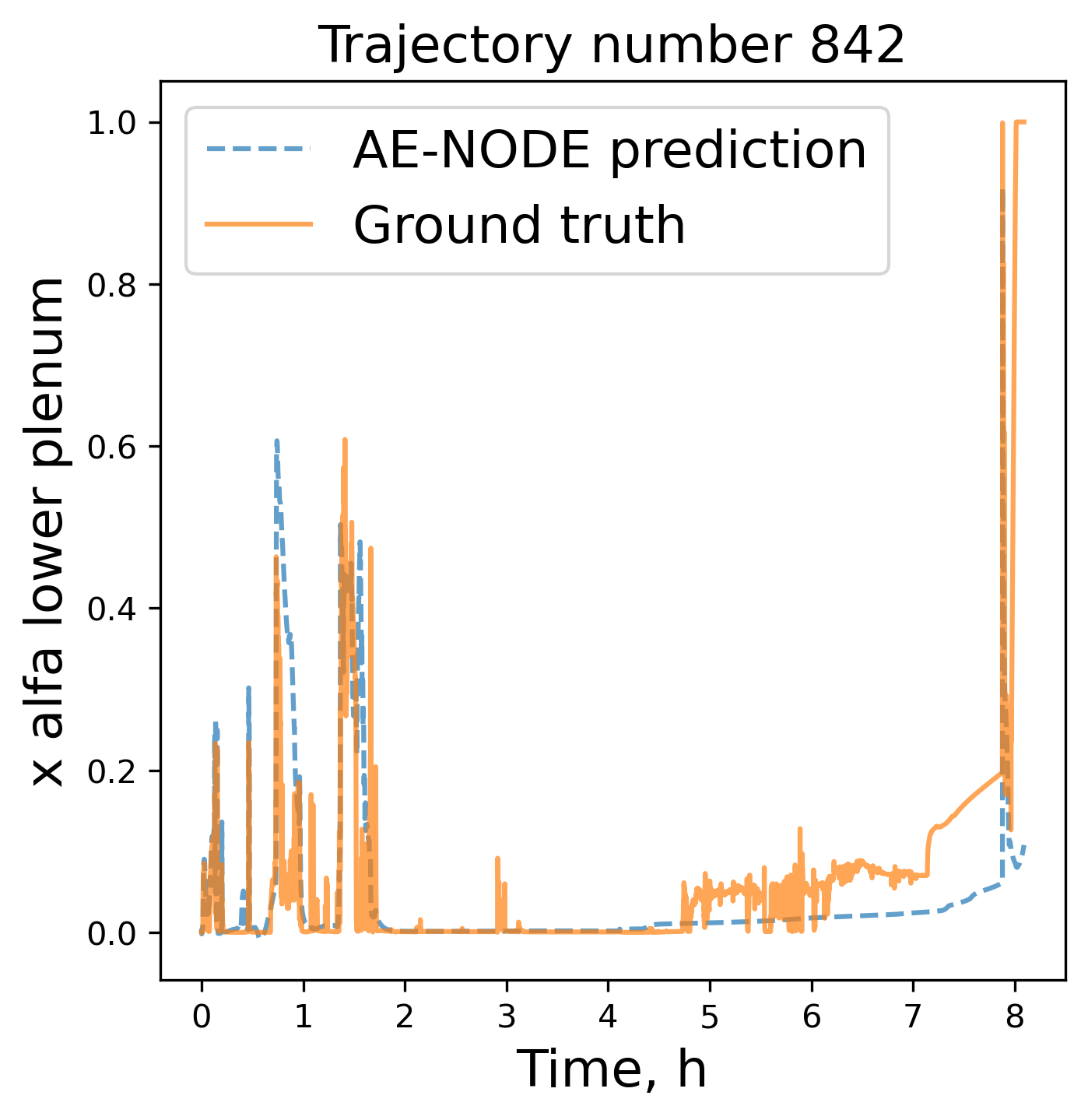}
        \caption{Trajectory 842}
        \label{fig:842_x_alfa_lower_plenum_LOCA}
    \end{subfigure}
    \caption{AE-NODE predictions vs.\ ground truth for selected (testing) LOCA trajectories of the void fraction in the lower plenum over time.}
    \label{fig:LOCA_x_alpha}
\end{figure}
\begin{figure}[htbp]
    \centering
    \begin{subfigure}[b]{0.24\textwidth}
        \includegraphics[width=\textwidth]{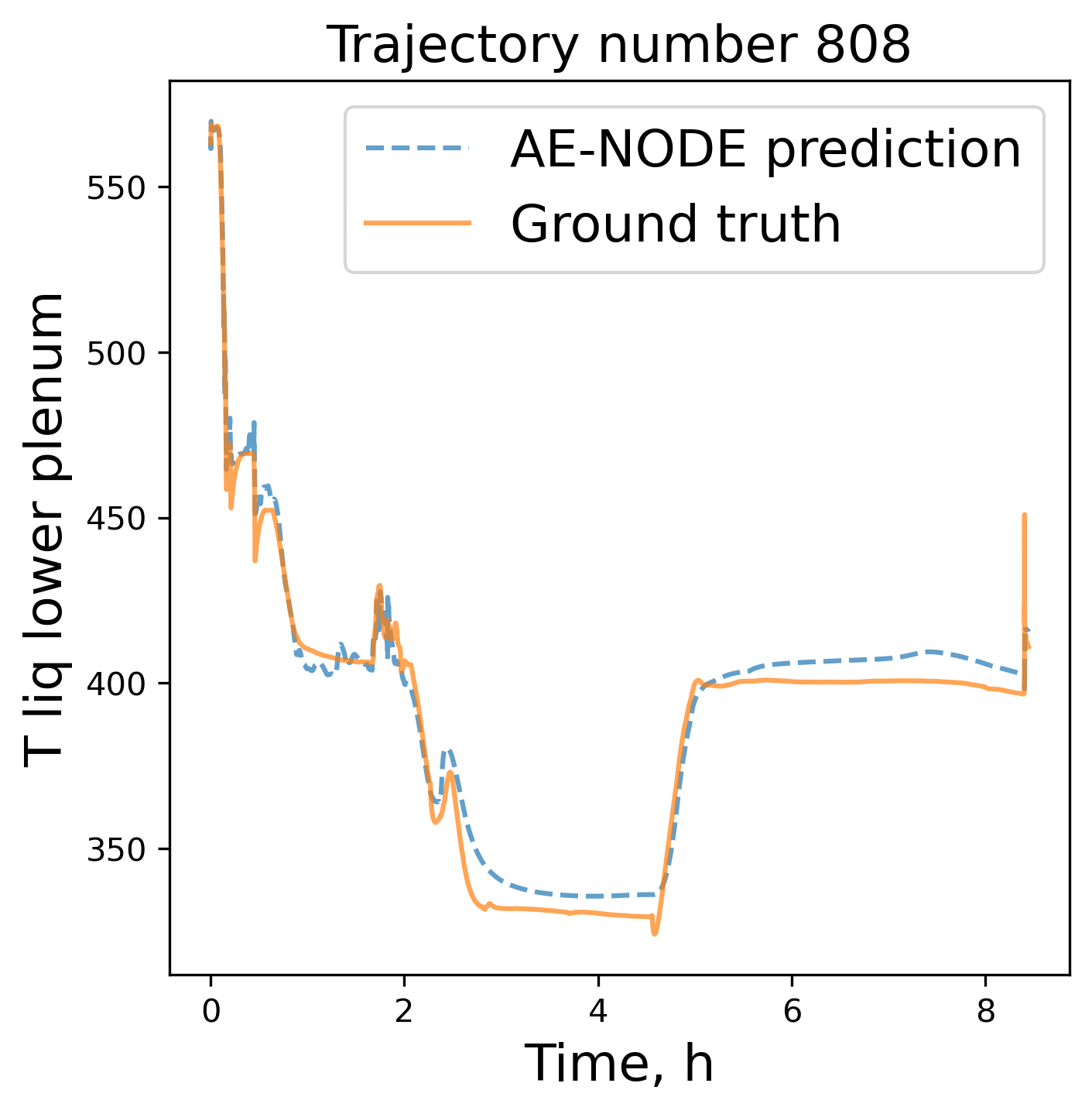}
        \caption{Trajectory 808}
        \label{fig:808_T_liq_lower_plenum_LOCA}
    \end{subfigure}
    \hfill
    \begin{subfigure}[b]{0.24\textwidth}
        \includegraphics[width=\textwidth]{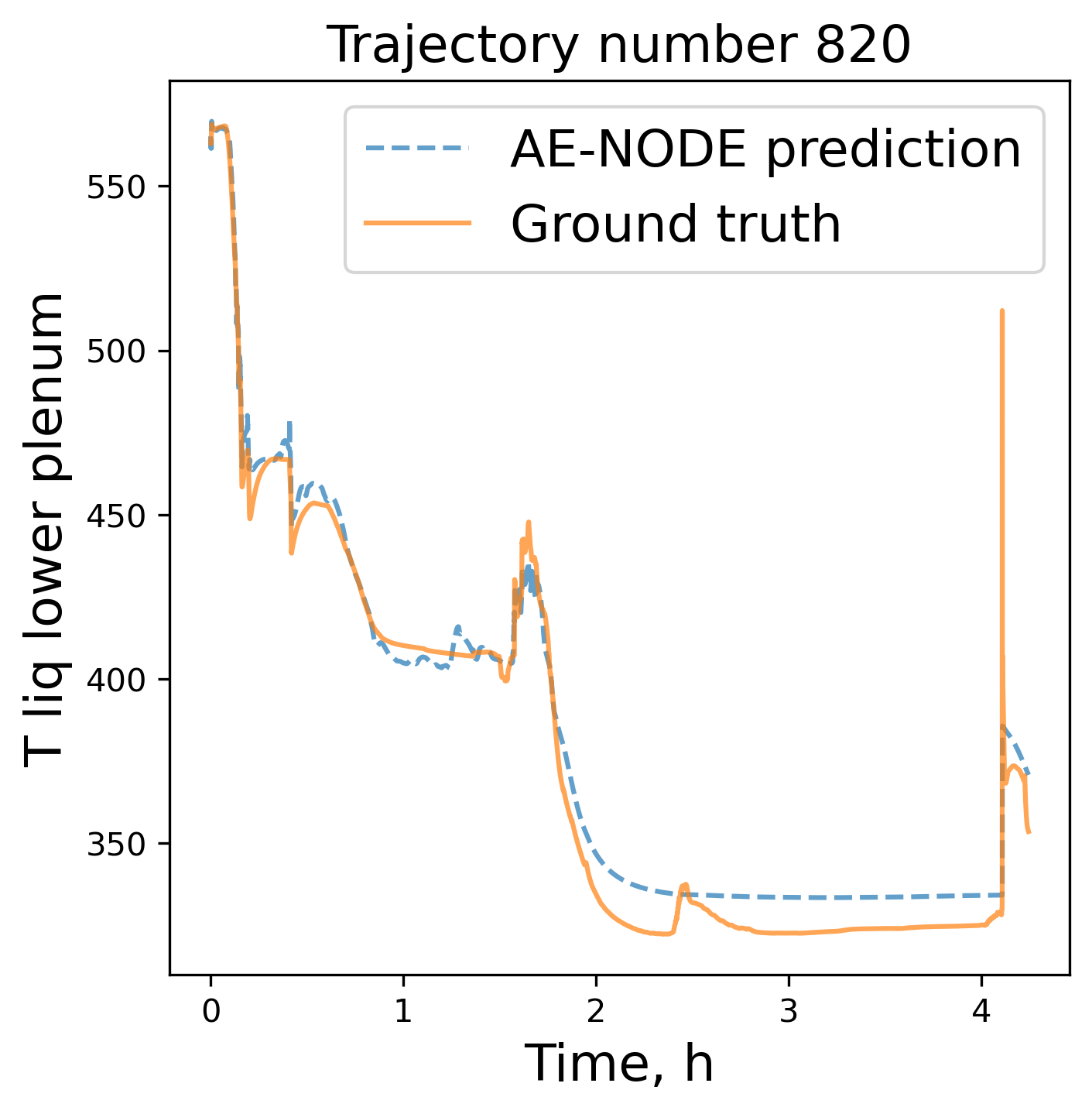}
        \caption{Trajectory 820}
        \label{fig:820_T_liq_lower_plenum_LOCA}
    \end{subfigure}
    \hfill
    \begin{subfigure}[b]{0.24\textwidth}
        \includegraphics[width=\textwidth]{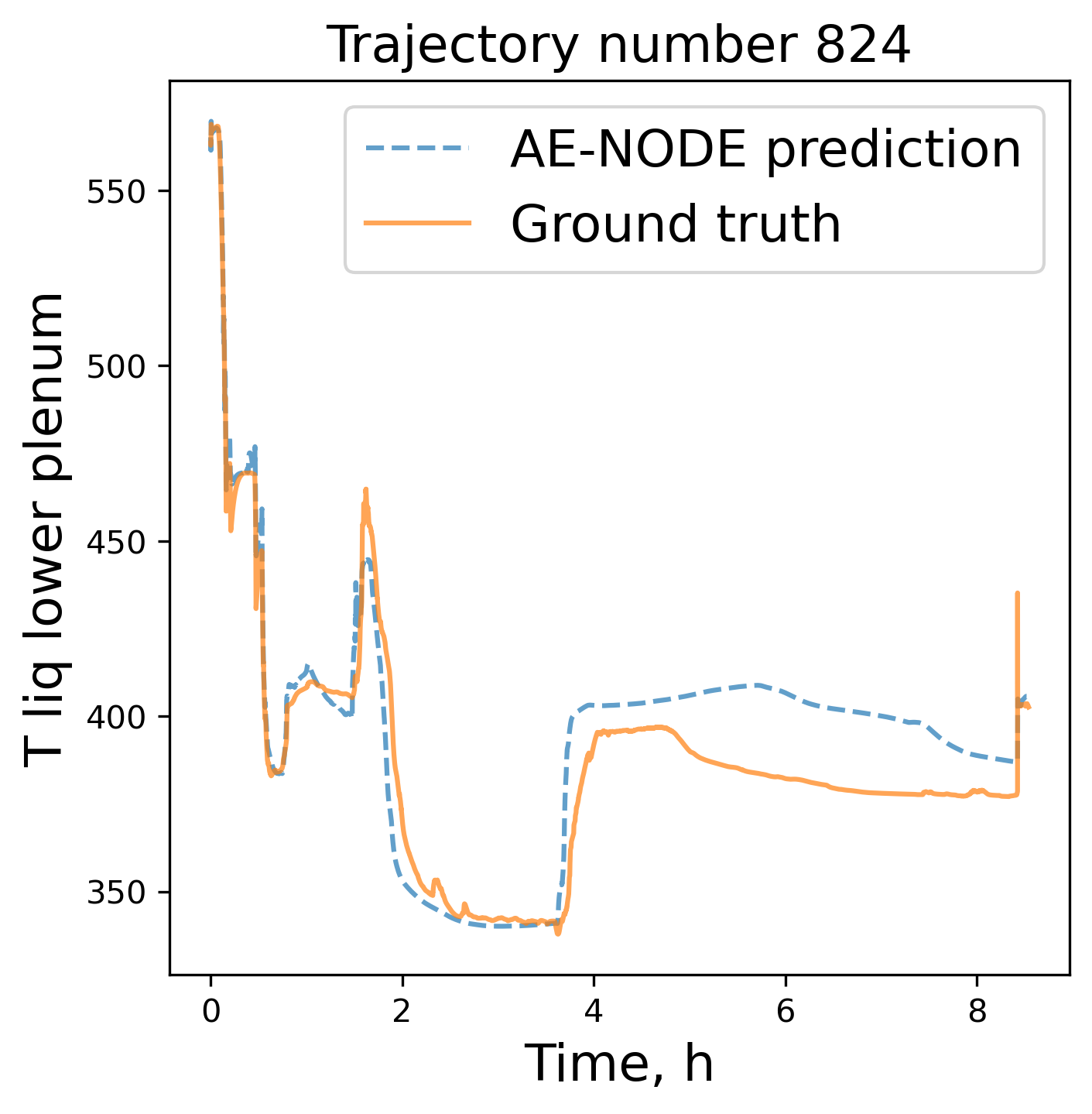}
        \caption{Trajectory 824}
        \label{fig:824_T_liq_lower_plenum_LOCA}
    \end{subfigure}
    \hfill
    \begin{subfigure}[b]{0.24\textwidth}
        \includegraphics[width=\textwidth]{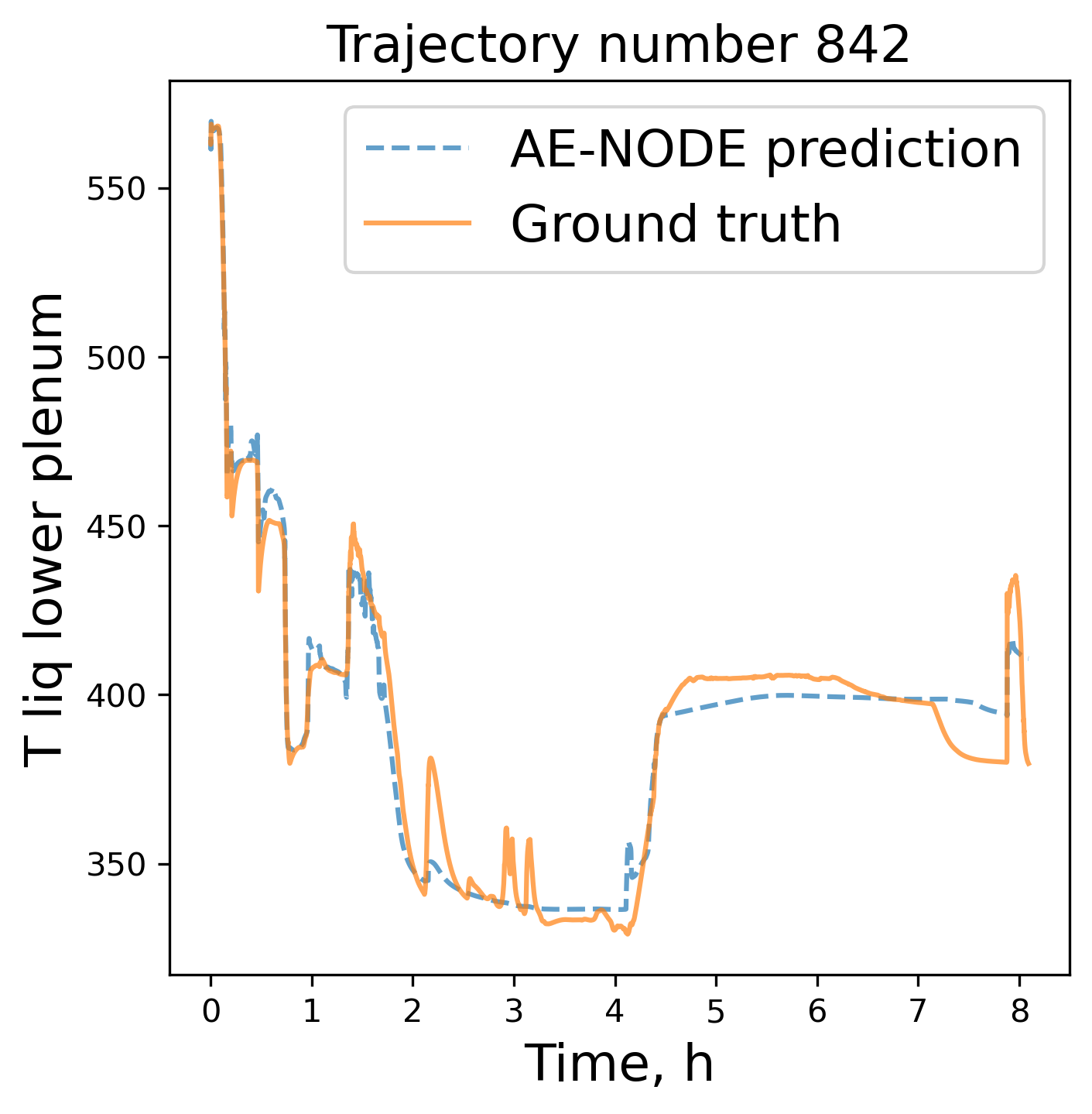}
        \caption{Trajectory 842}
        \label{fig:842_T_liq_lower_plenum_LOCA}
    \end{subfigure}
    \caption{AE-NODE predictions vs.\ ground truth for selected (testing) LOCA trajectories of the liquid temperature in the lower plenum [K] over time.}
    \label{fig:LOCA_T_liq_plenum_predictions}
\end{figure}
\begin{figure}[htbp]
    \centering
    \begin{subfigure}[b]{0.48\textwidth}
        \includegraphics[width=\textwidth]{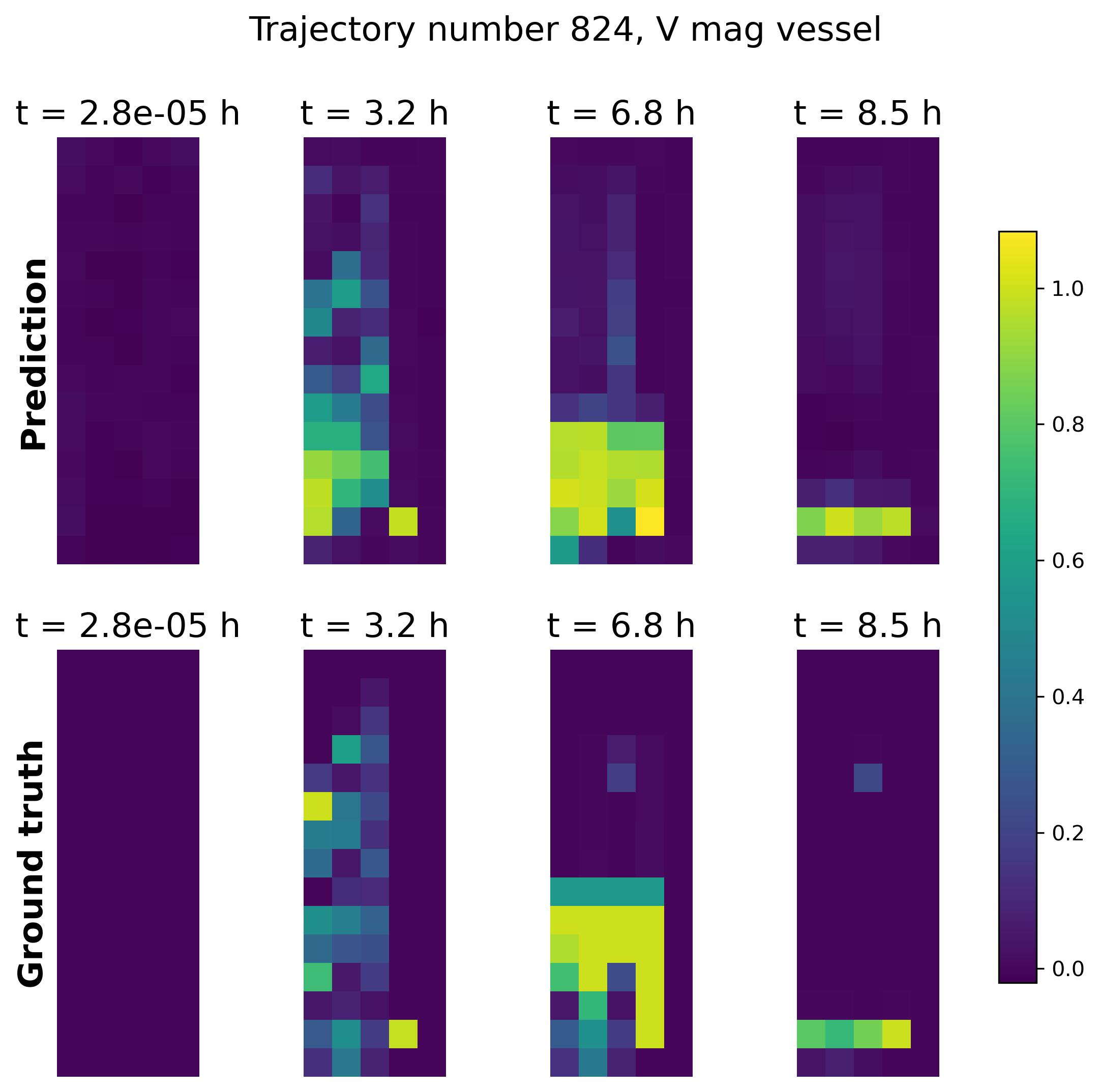}
        \caption{V magma}
        \label{fig:824_V_mag_vessel_LOCA}
    \end{subfigure}
    \hfill
    \begin{subfigure}[b]{0.48\textwidth}
        \includegraphics[width=\textwidth]{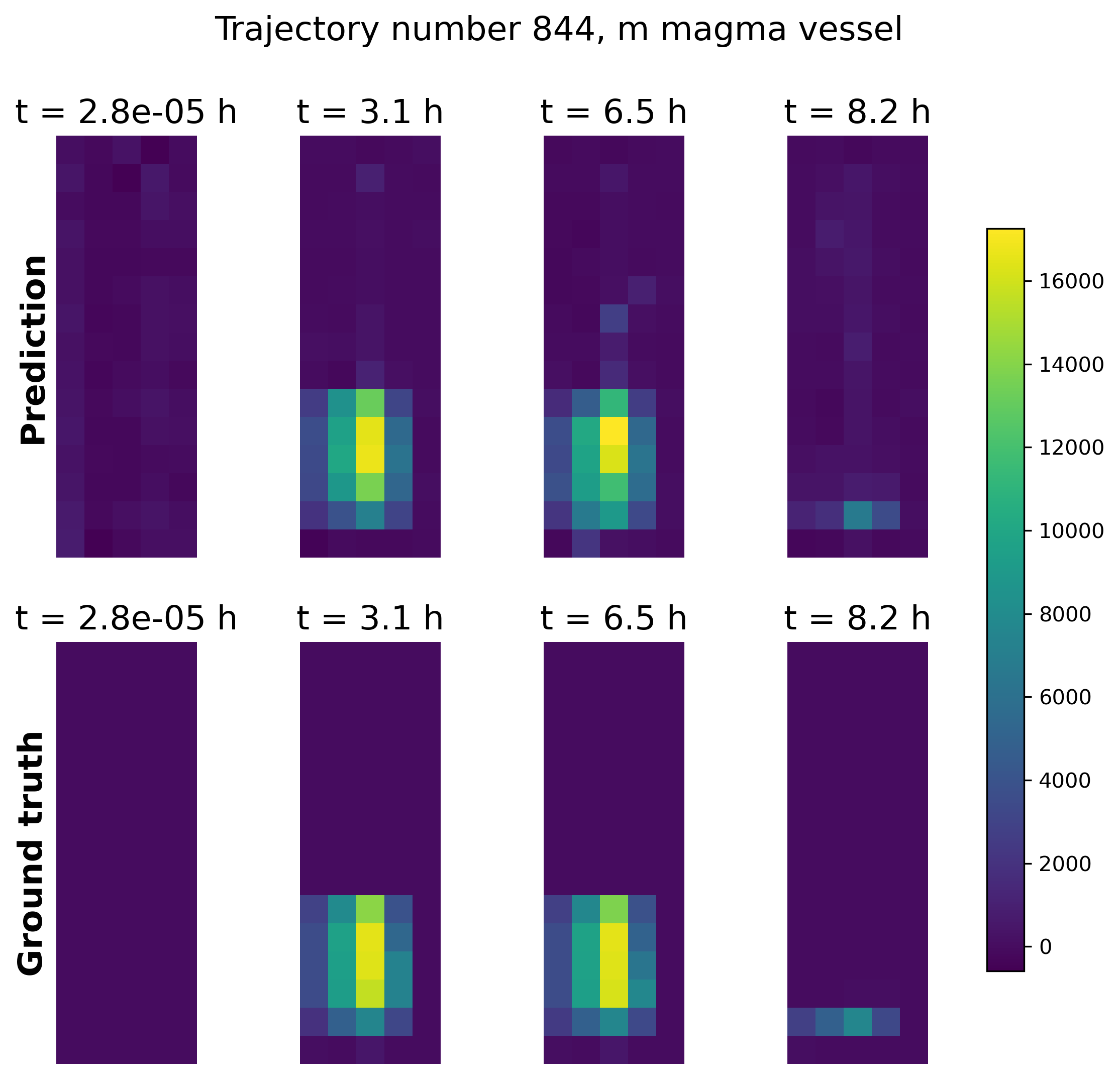}
        \caption{M magma}
        \label{fig:844_m_magma_vessel_LOCA}
    \end{subfigure}
    \caption{AE-NODE predictions vs.\ ground truth at $4$ different time steps of volume and the mass of the magma in the vessel for the LOCA scenario.}
    \label{fig:LOCA_2d_predictions}
\end{figure}
\begin{figure}[htbp]
    \centering
    \begin{subfigure}[b]{0.48\textwidth}
        \includegraphics[width=\textwidth]{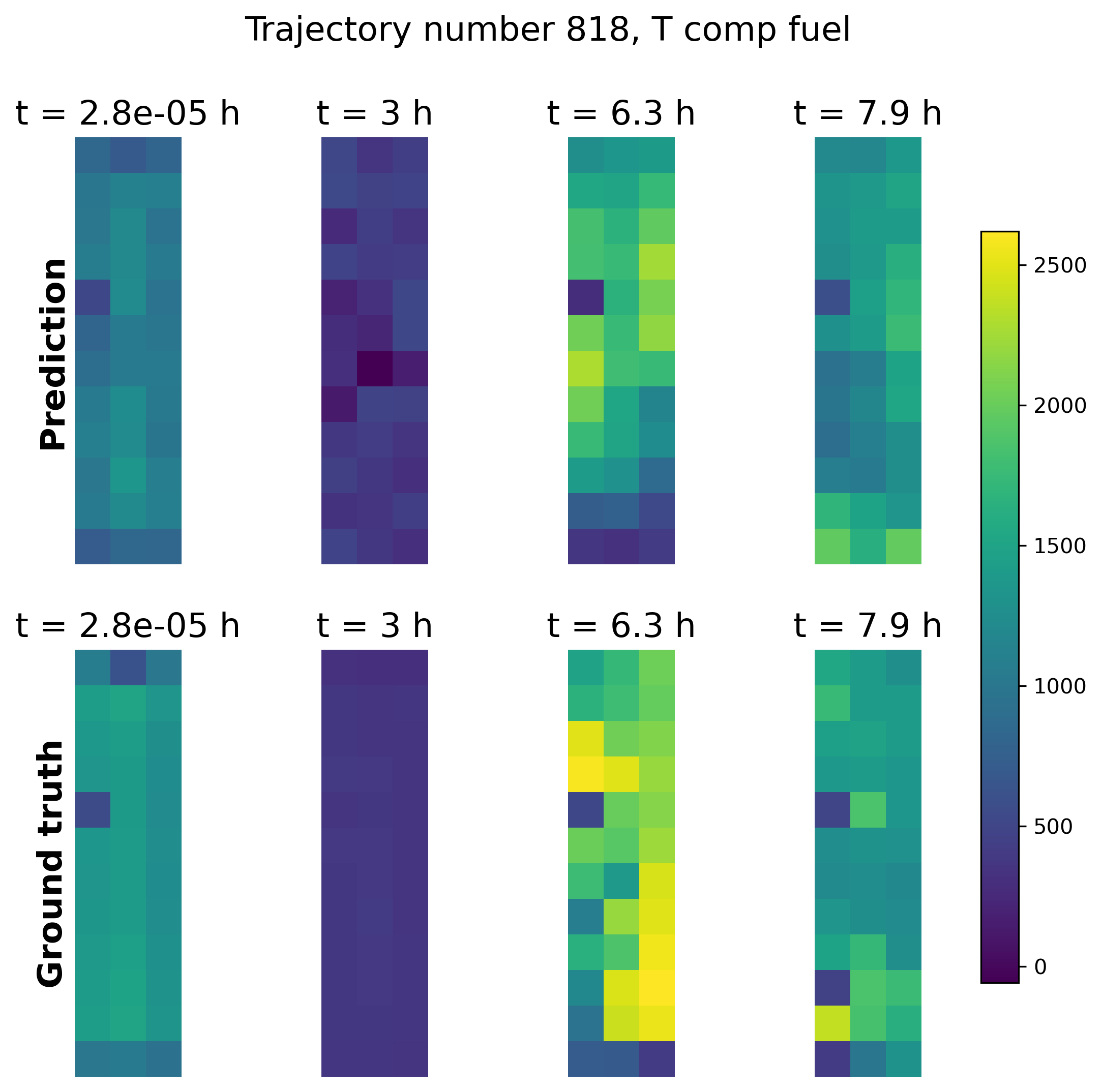}
        \caption{T fuel}
        \label{fig:818_T_comp_fuel_LOCA}
    \end{subfigure}
    \hfill
    \begin{subfigure}[b]{0.48\textwidth}
        \includegraphics[width=\textwidth]{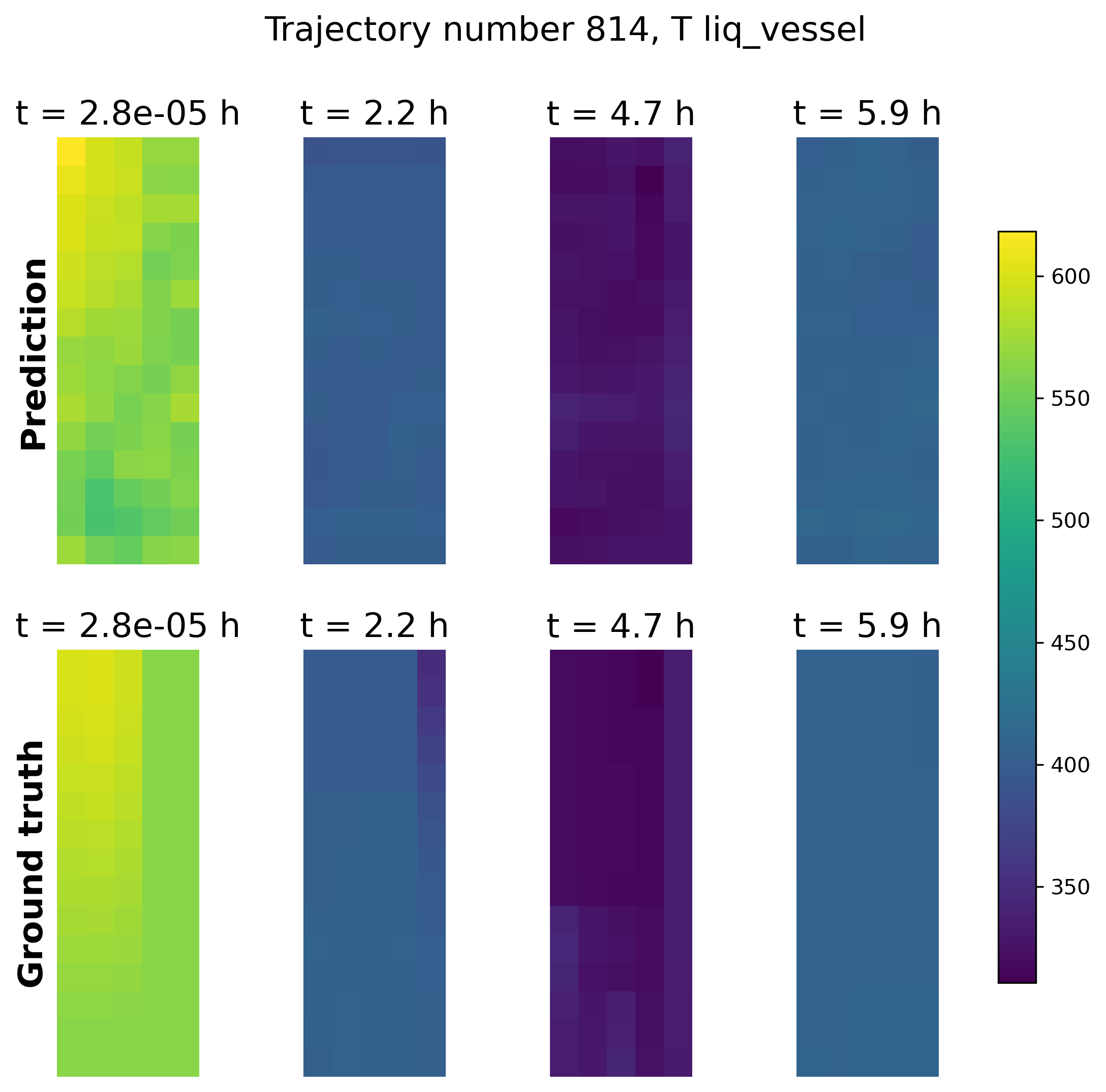}
        \caption{T liquid}
        \label{fig:814_T_liq_vessel_LOCA}
    \end{subfigure}
    \caption{AE-NODE predictions vs.\ ground truth at $4$ different time steps of the temperature of the fuel [K] in the core and of the liquid phase [K] in the vessel for the LOCA scenario.}
    \label{fig:LOCA_2d_temperature}
\end{figure}

In Figure \ref{fig:LOCA_m_tot_cor}, \ref{fig:LOCA_x_alpha} and \ref{fig:LOCA_T_liq_plenum_predictions} we show the predictions of AE-NODE on $4$ different testing LOCA trajectories of the total mass of the corium, of the void fraction in the plenum and of the temperature of the liquid in the plenum. In Figure \ref{fig:LOCA_2d_predictions} and \ref{fig:LOCA_2d_temperature} we display the AE-NODE predictions for the volume and the mass of the magma in the vessel and for the temperature of the fuel and temperature of the liquid phase at 4 snapshots in time. 
\subsection{SBO scenario}
\label{subsec:SBO_experiment}
\begin{figure}[h]
    \centering
    \includegraphics[width=1.0\textwidth]{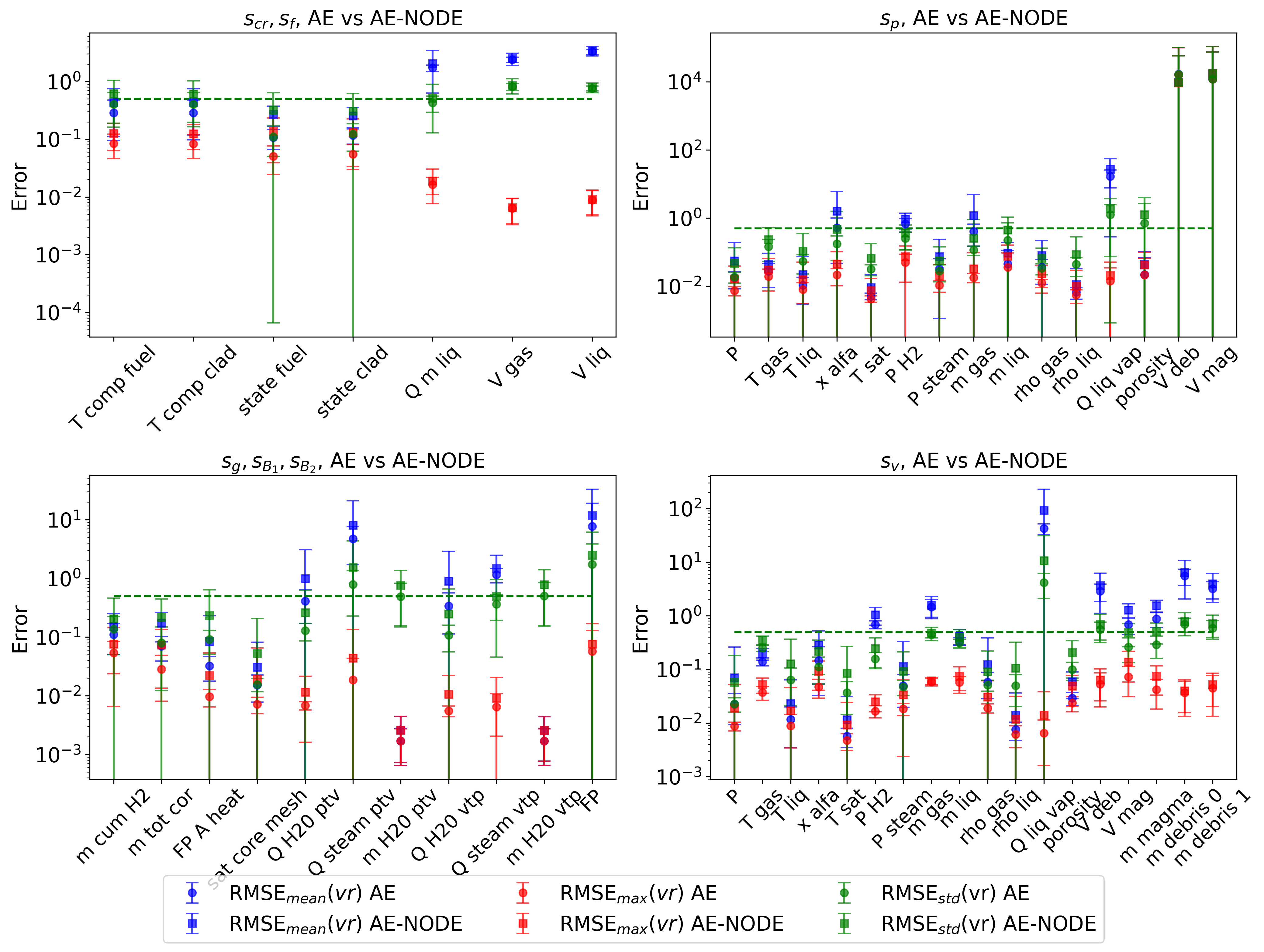}
    \caption{Comparison of $\text{RMSE}_{mean}$, $\text{RMSE}_{max}$ and $\text{RMSE}_{std}$ metrics when doing a simple AutoEncoding (AE) and when running the actual inference (AE-NODE) on the SBO testing set. The uncertainty bars are the standard deviations of the metrics computed across testing trajectories. The green horizontal line is placed at the value 0.5, which is a common baseline for $\text{RMSE}_{std}$.}
    \label{fig:SBO_AE_AE_NODE_comparison}
\end{figure}

\begin{figure}[htbp]
    \centering
    \begin{subfigure}[b]{0.24\textwidth}
        \includegraphics[width=\textwidth]{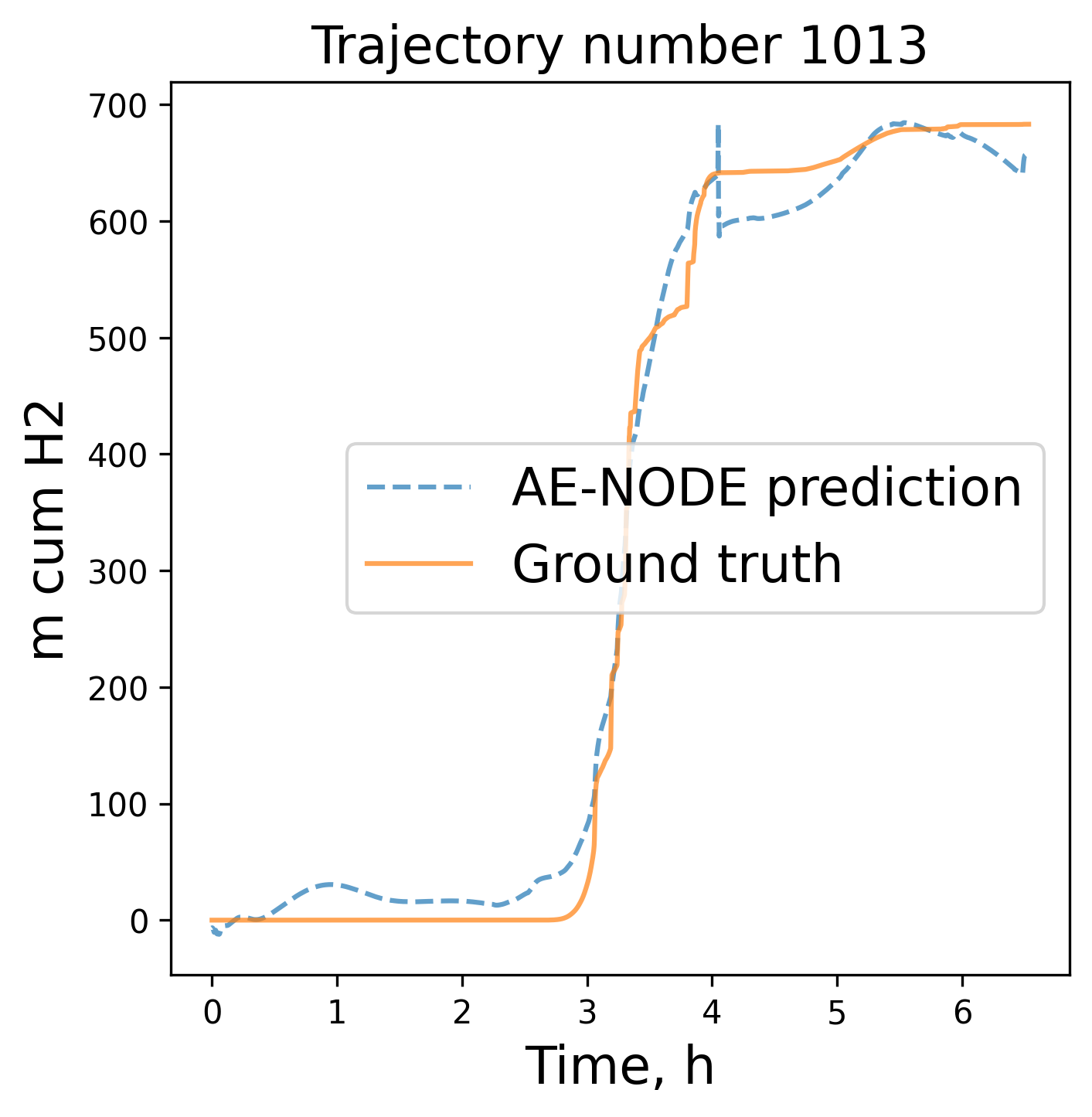}
        \caption{Trajectory 1013}
        \label{fig:1013_m_cum_H2_SBO}
    \end{subfigure}
    \hfill
    \begin{subfigure}[b]{0.24\textwidth}
        \includegraphics[width=\textwidth]{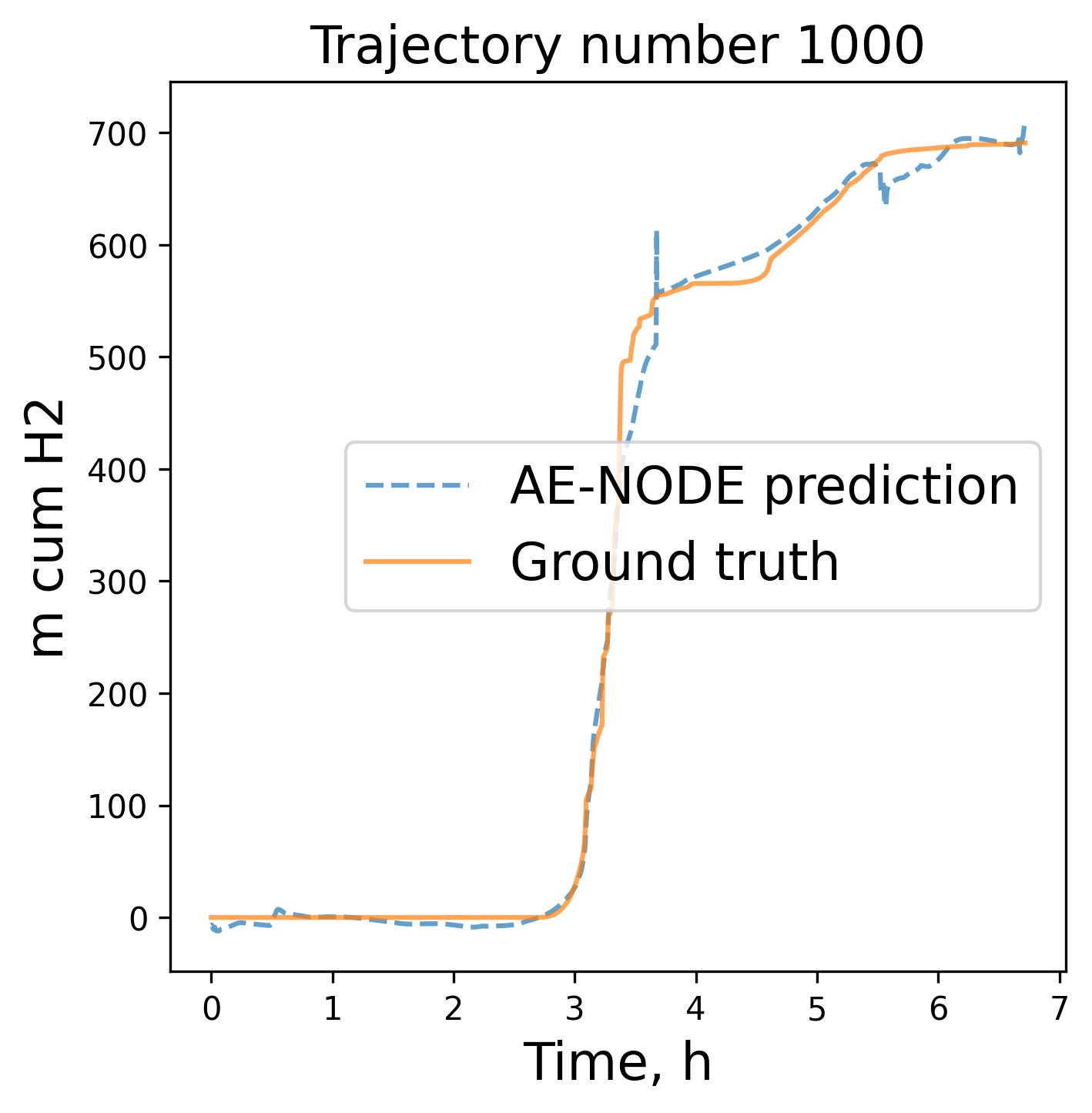}
        \caption{Trajectory 1000}
        \label{fig:1000_m_cum_H2_SBO}
    \end{subfigure}
    \hfill
    \begin{subfigure}[b]{0.24\textwidth}
        \includegraphics[width=\textwidth]{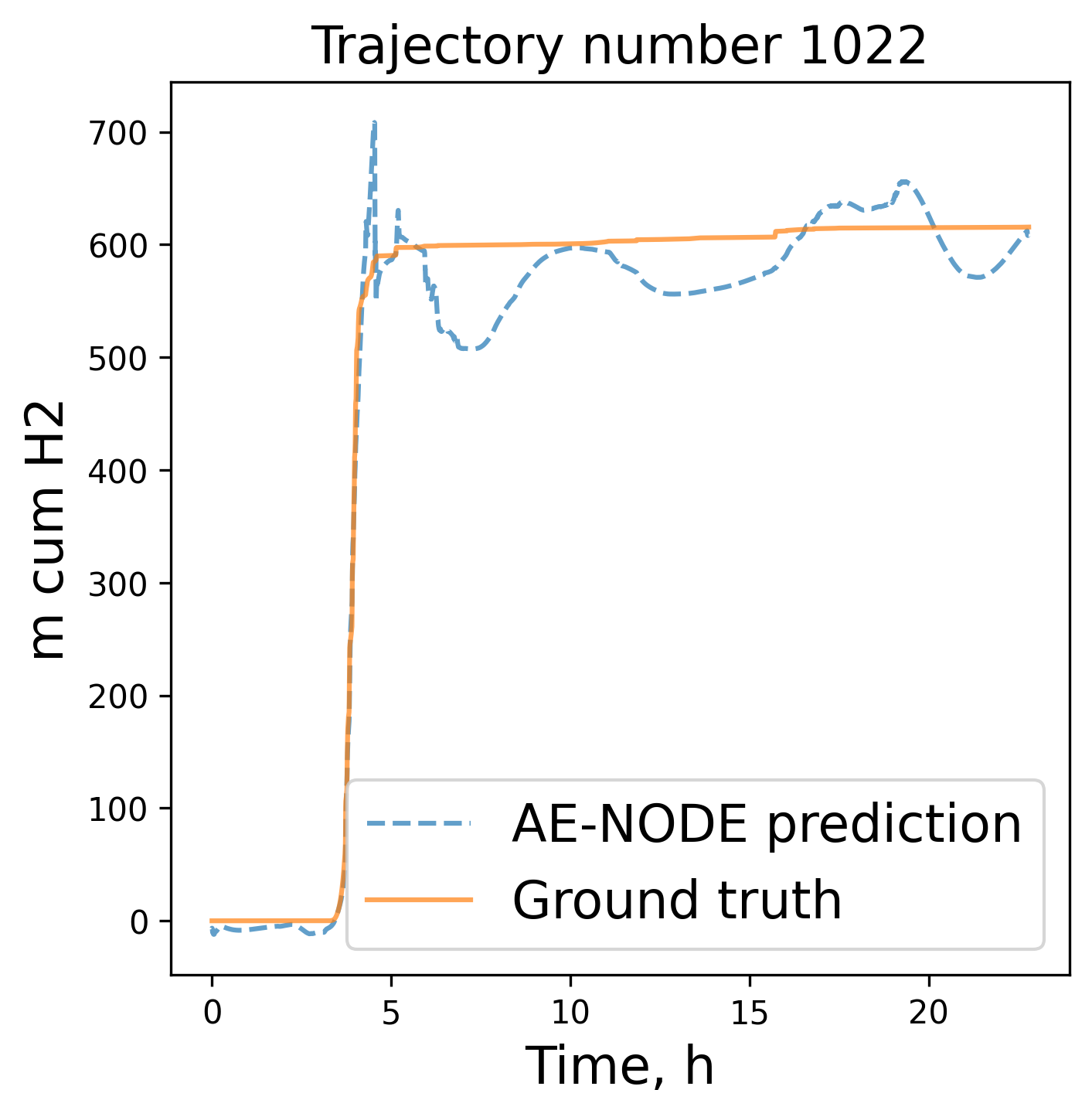}
        \caption{Trajectory 1022}
        \label{fig:1022_m_cum_H2_SBO}
    \end{subfigure}
    \hfill
    \begin{subfigure}[b]{0.24\textwidth}
        \includegraphics[width=\textwidth]{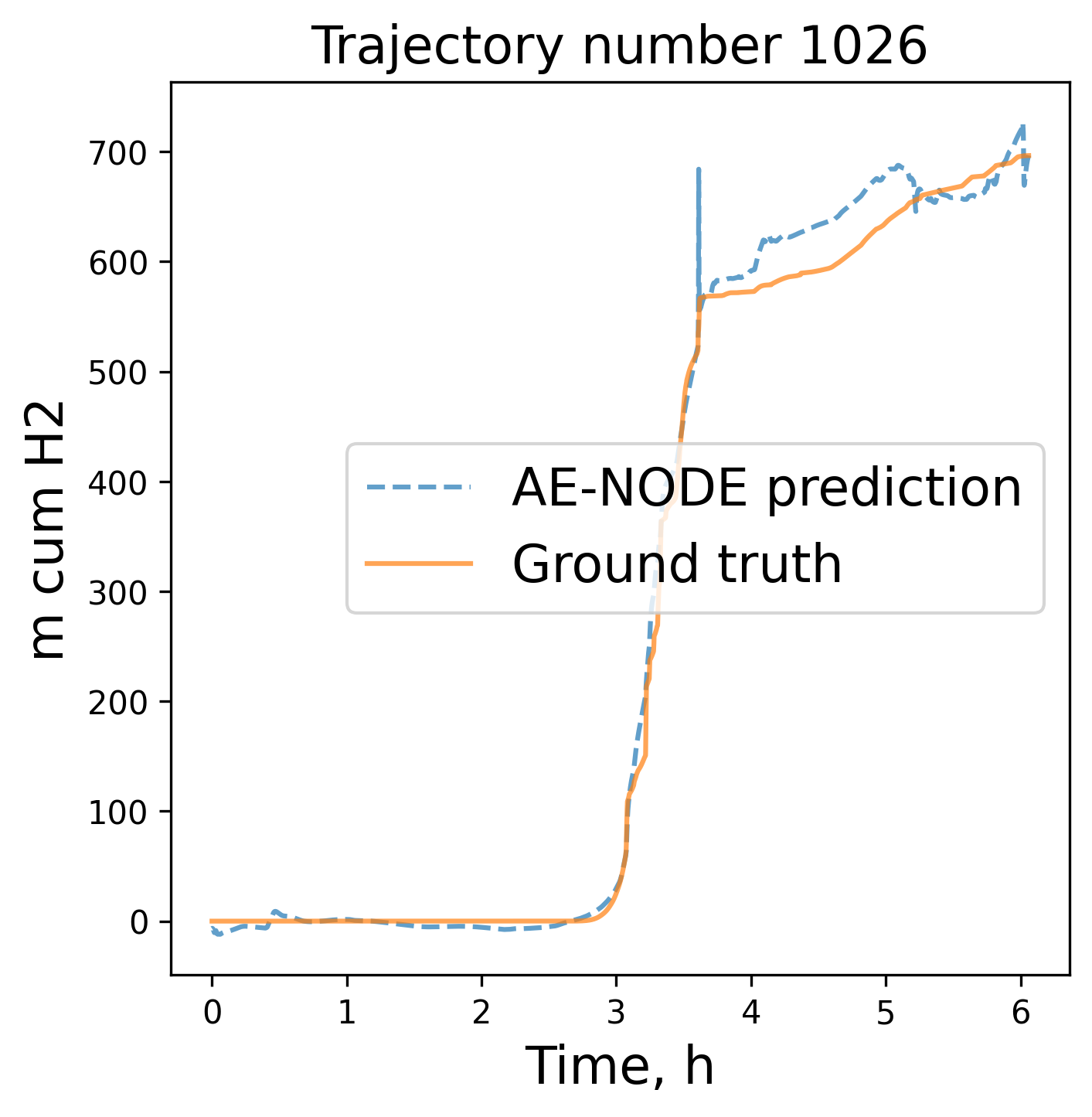}
        \caption{Trajectory 1026}
        \label{fig:1026_m_cum_H2_SBO}
    \end{subfigure}
    \caption{AE-NODE predictions vs.\ ground truth for selected (testing) SBO trajectories of the H2 cumulated mass in the core [kg] over time.}
    \label{fig:SBO_m_cum_H2}
\end{figure}

\begin{figure}[htbp]
    \centering
    \begin{subfigure}[b]{0.24\textwidth}
        \includegraphics[width=\textwidth]{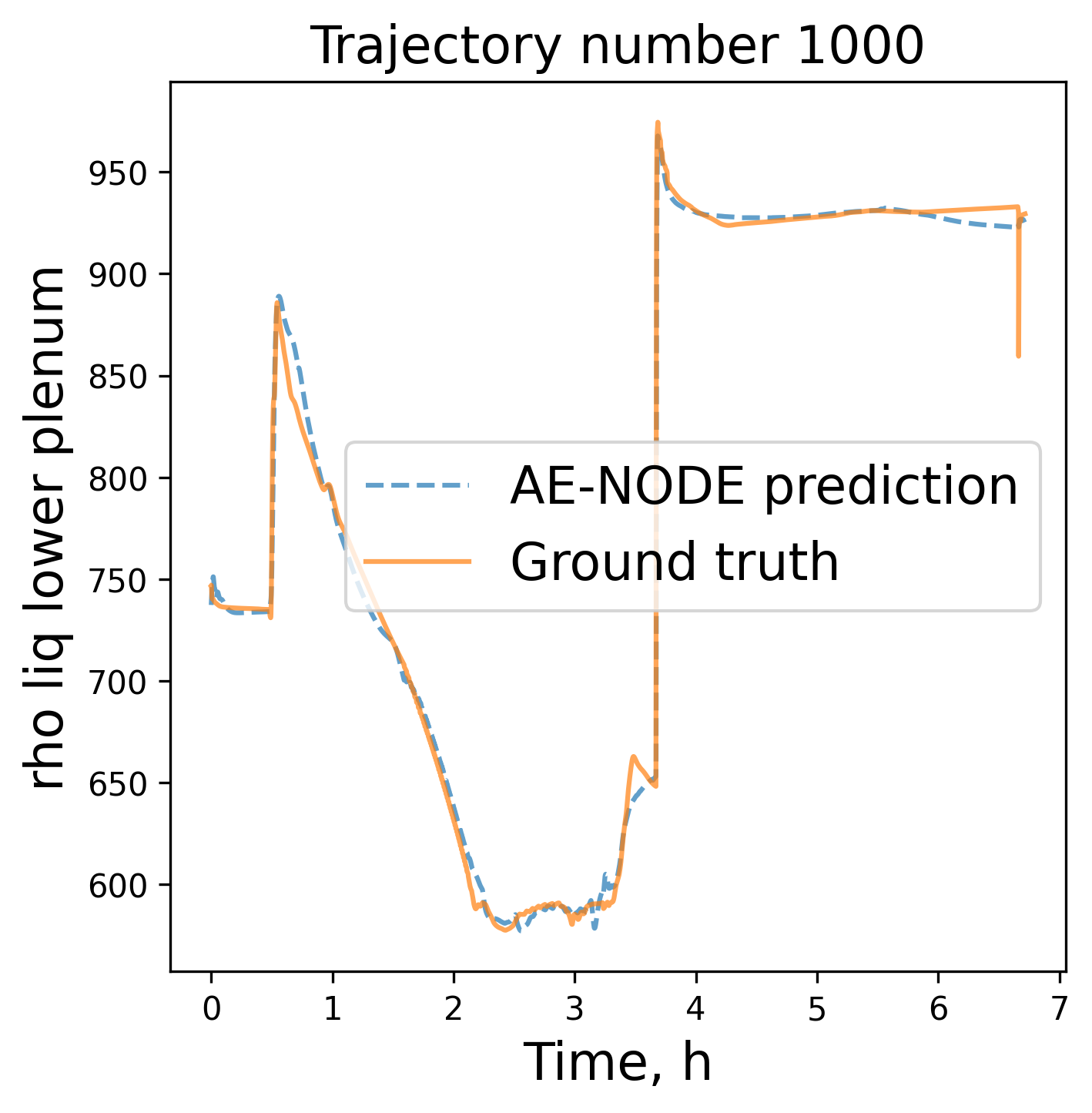}
        \caption{Trajectory 1000}
        \label{fig:1000_rho_liq_lower_plenum_SBO}
    \end{subfigure}
    \hfill
    \begin{subfigure}[b]{0.24\textwidth}
        \includegraphics[width=\textwidth]{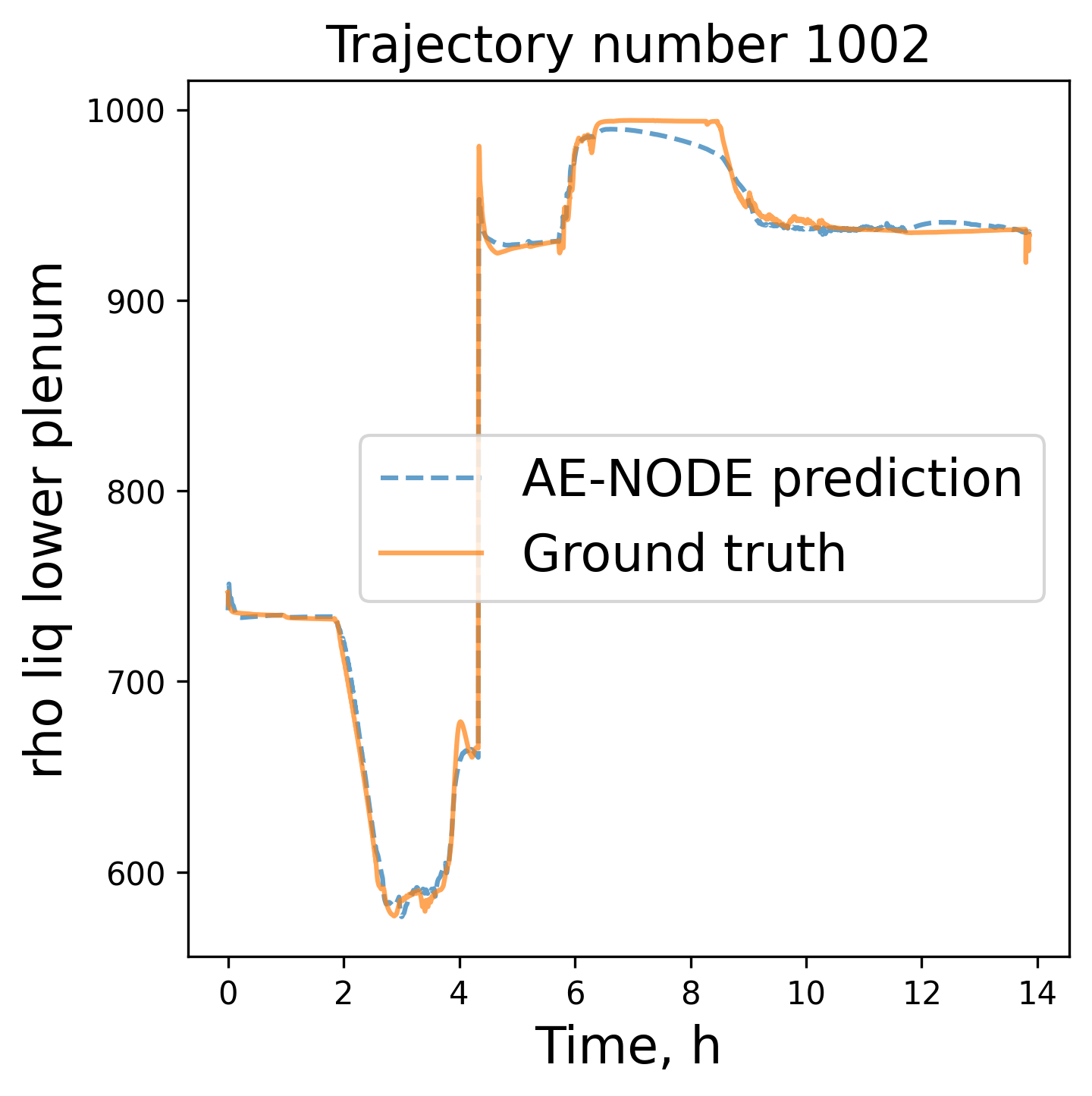}
        \caption{Trajectory 1002}
        \label{fig:1002_rho_liq_lower_plenum_SBO}
    \end{subfigure}
    \hfill
    \begin{subfigure}[b]{0.24\textwidth}
        \includegraphics[width=\textwidth]{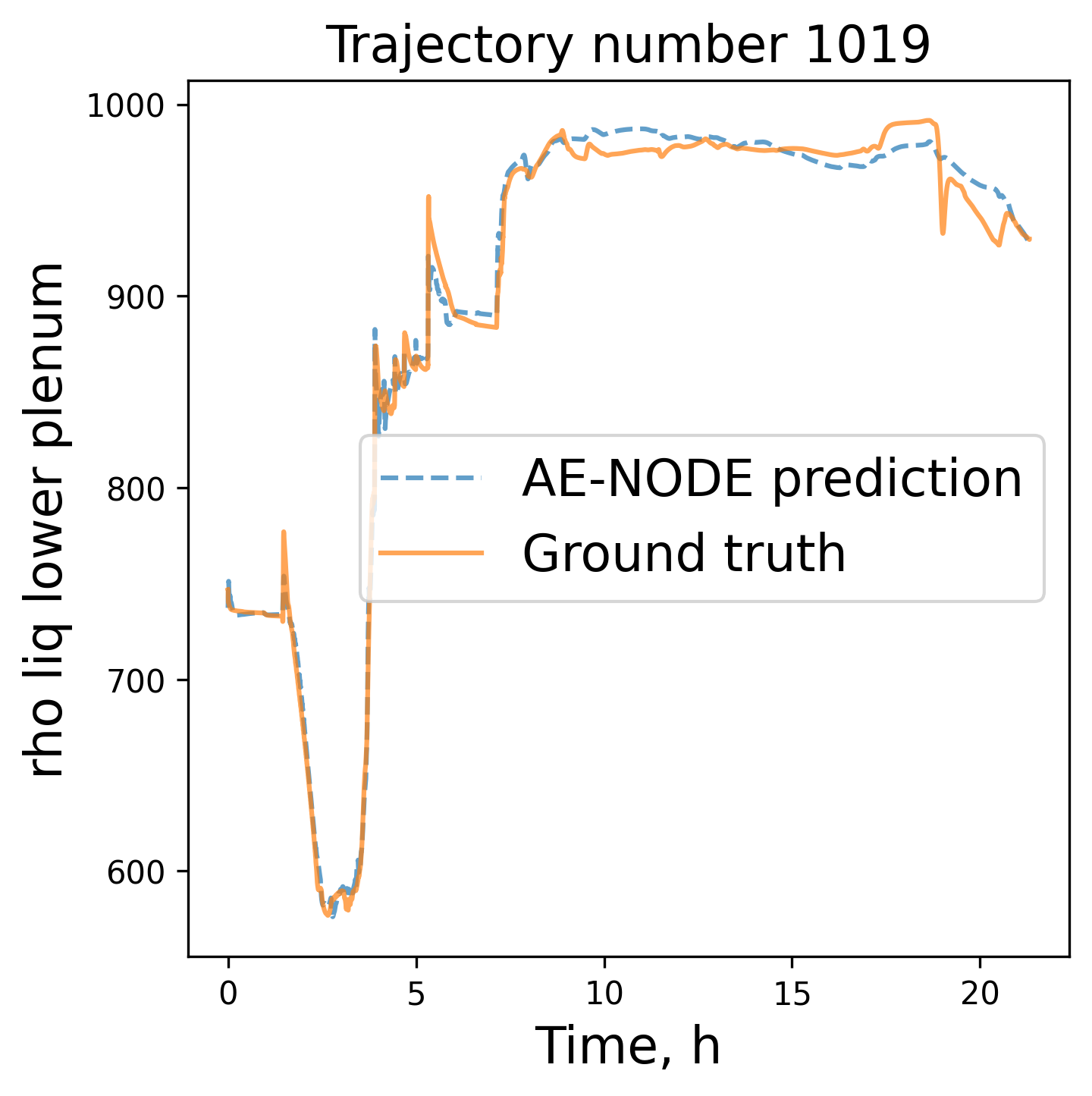}
        \caption{Trajectory 1019}
        \label{fig:1019_rho_liq_lower_plenum_SBO}
    \end{subfigure}
    \hfill
    \begin{subfigure}[b]{0.24\textwidth}
        \includegraphics[width=\textwidth]{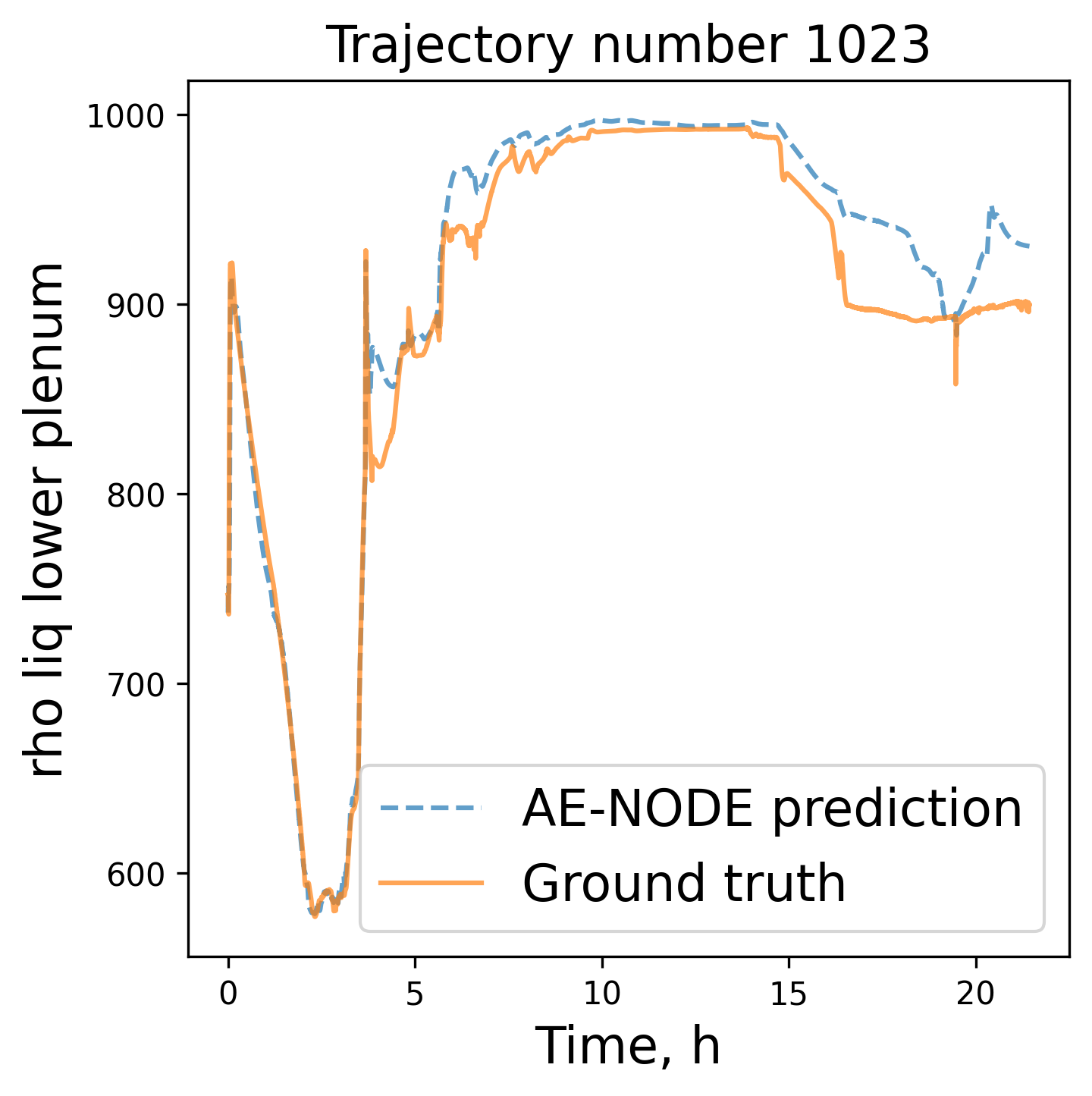}
        \caption{Trajectory 1023}
        \label{fig:1023_rho_liq_lower_plenum_SBO}
    \end{subfigure}
    \caption{AE-NODE predictions vs.\ ground truth for selected (testing) SBO trajectories of the liquid density in the lower plenum [kg/m$^3$] over time.}
    \label{fig:SBO_rho_liq_lower_plenum}
\end{figure}

\begin{figure}[htbp]
    \centering
    \begin{subfigure}[b]{0.24\textwidth}
        \includegraphics[width=\textwidth]{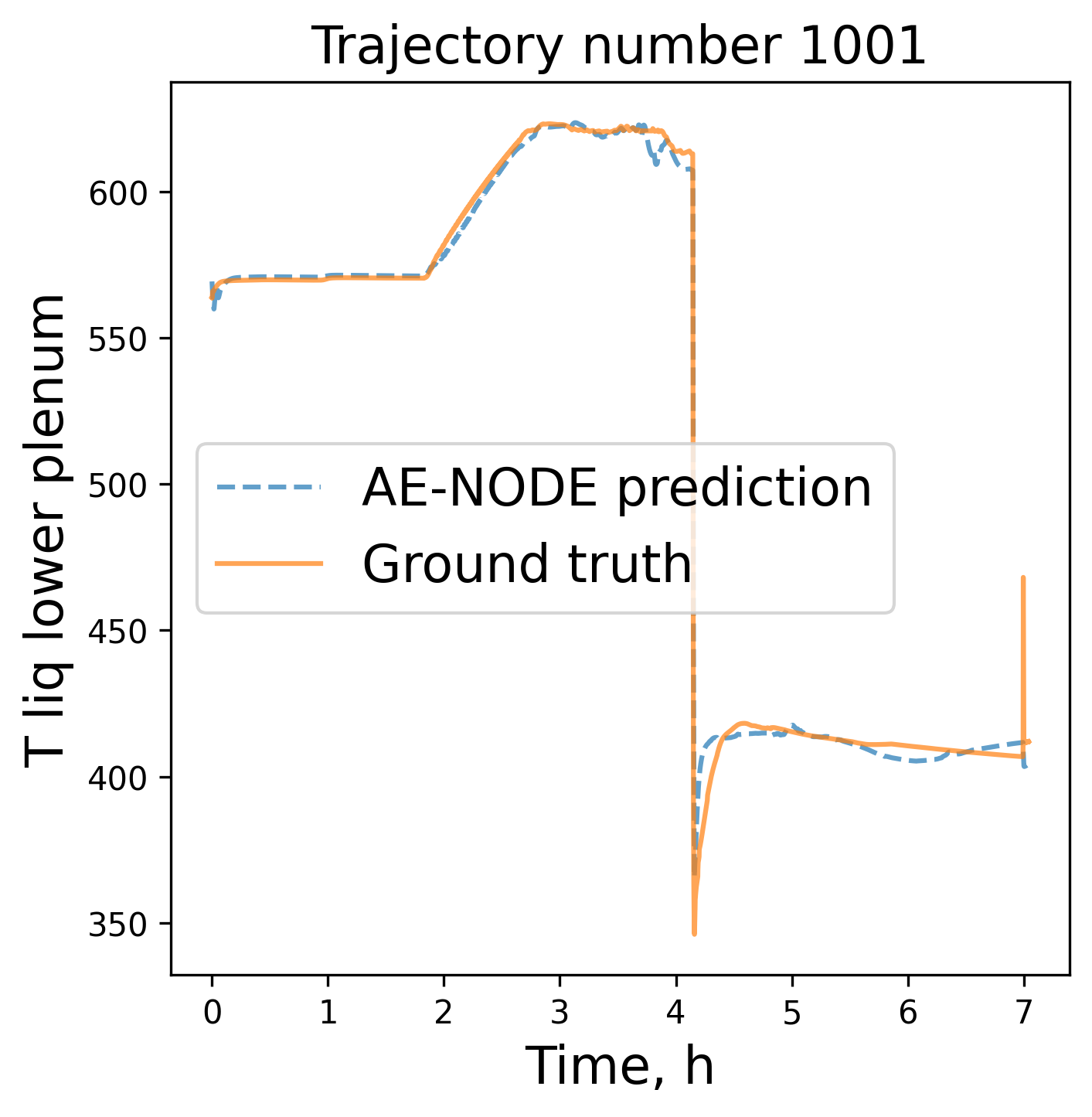}
        \caption{Trajectory 1001}
        \label{fig:1001_T_liq_lower_plenum_SBO}
    \end{subfigure}
    \hfill
    \begin{subfigure}[b]{0.24\textwidth}
        \includegraphics[width=\textwidth]{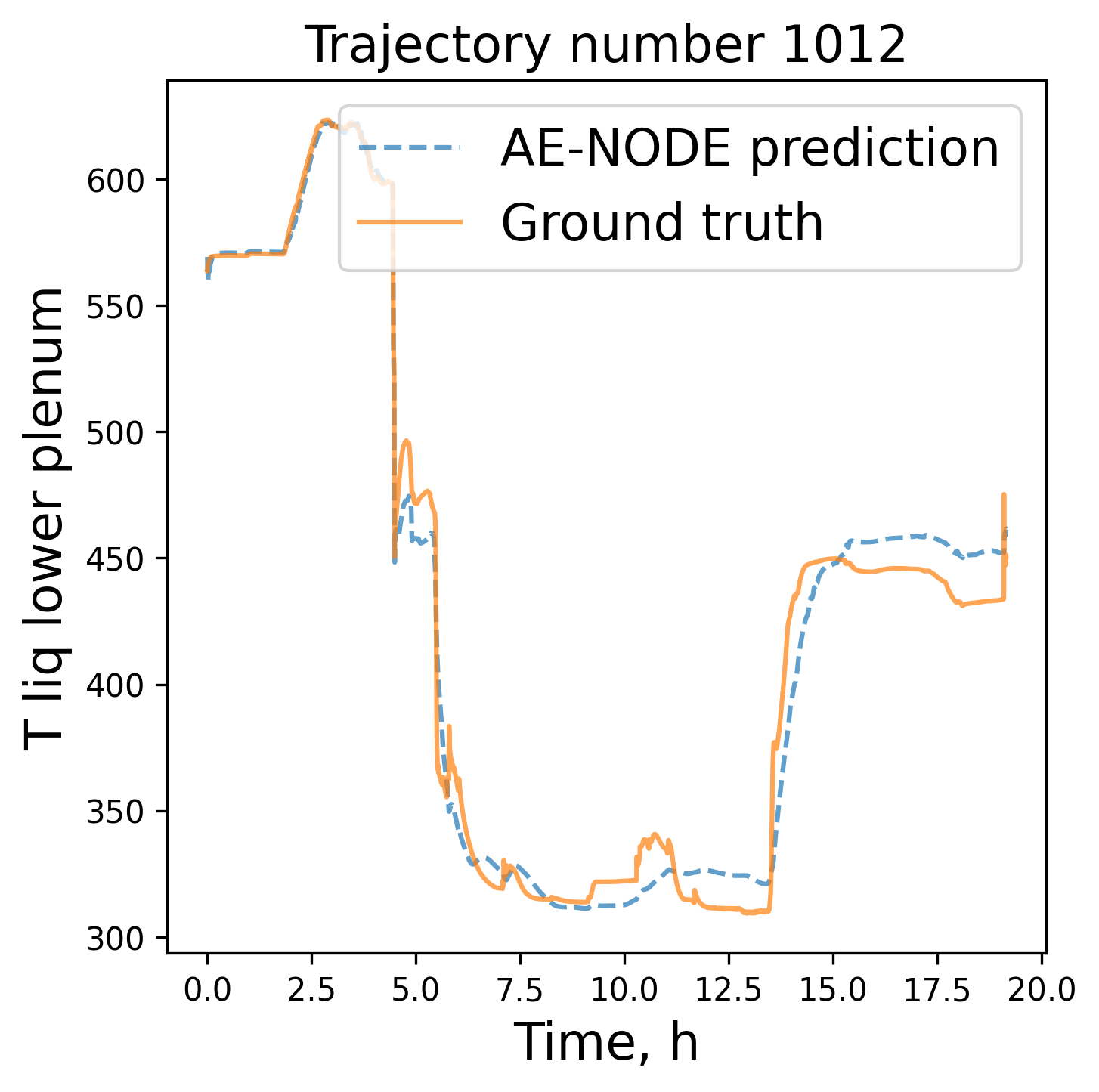}
        \caption{Trajectory 1012}
        \label{fig:1012_T_liq_lower_plenum_SBO}
    \end{subfigure}
    \hfill
    \begin{subfigure}[b]{0.24\textwidth}
        \includegraphics[width=\textwidth]{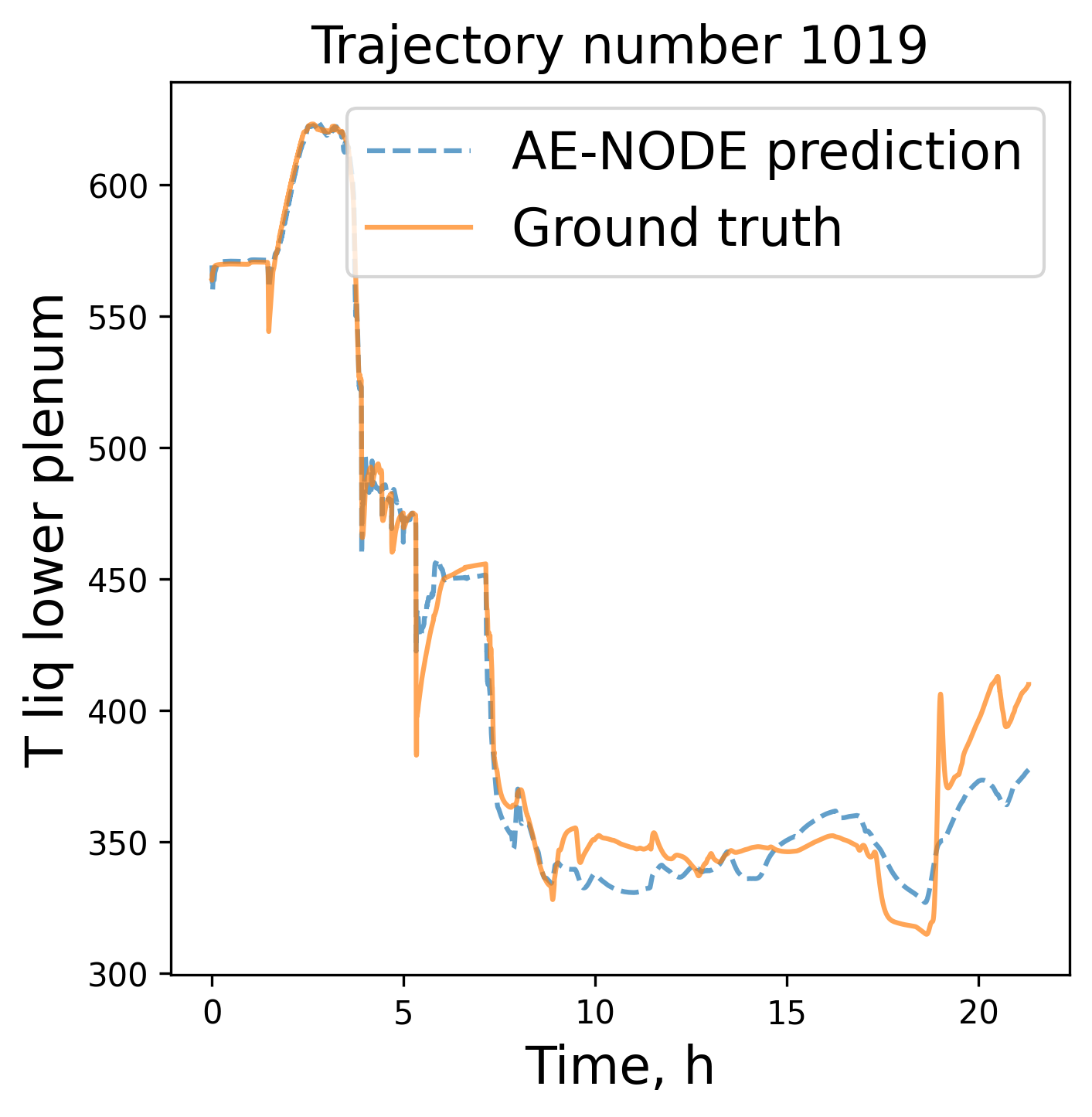}
        \caption{Trajectory 1019}
        \label{fig:1019_T_liq_lower_plenum_SBO}
    \end{subfigure}
    \hfill
    \begin{subfigure}[b]{0.24\textwidth}
        \includegraphics[width=\textwidth]{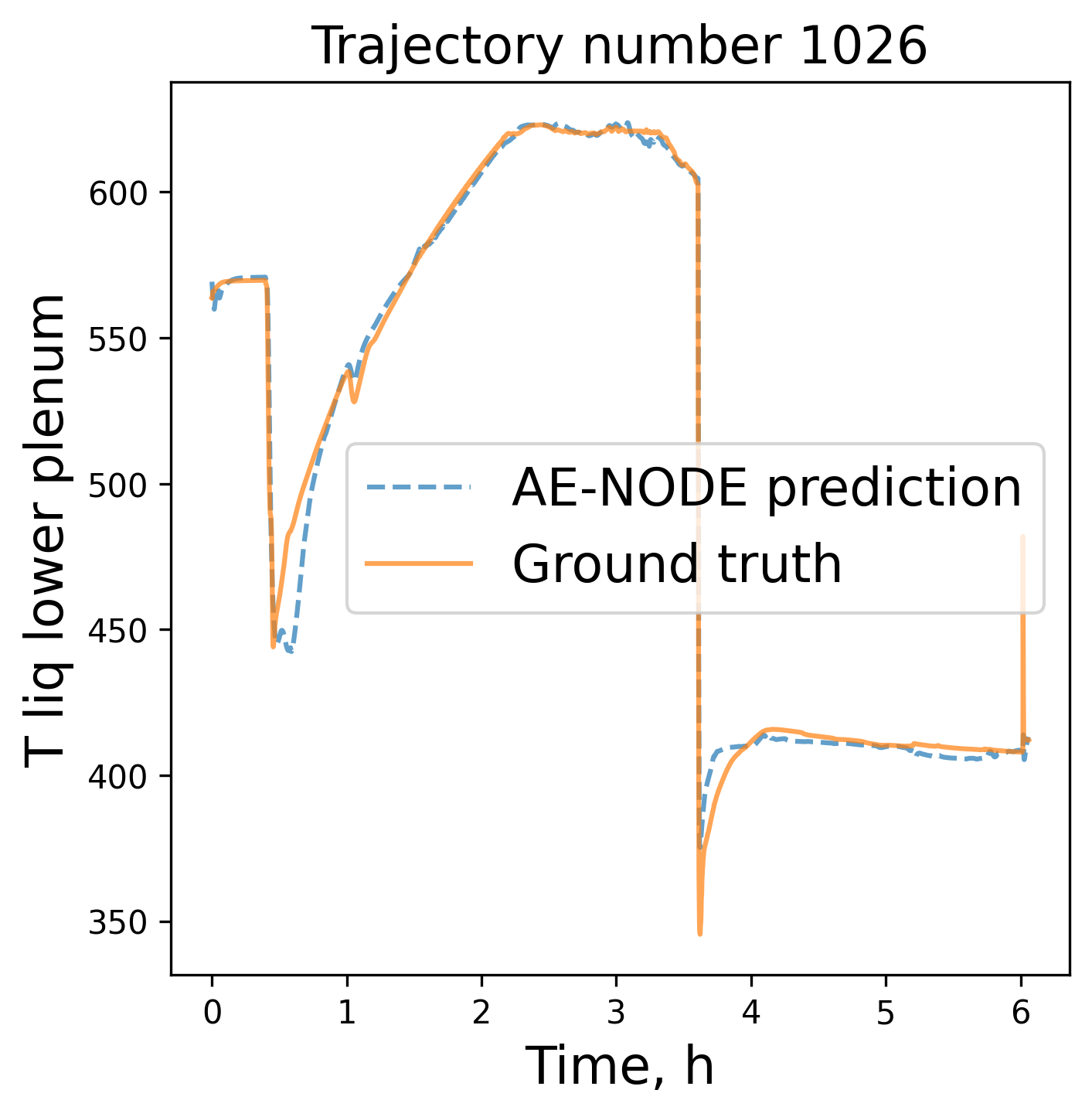}
        \caption{Trajectory 1026}
        \label{fig:1026_T_liq_lower_plenum_SBO}
    \end{subfigure}
    \caption{AE-NODE predictions vs.\ ground truth for selected (testing) SBO trajectories of the temperature of the liquid in the lower plenum [K] over time.}
    \label{fig:SBO_T_liq_lower_plenum}
\end{figure}

\begin{figure}[htbp]
    \centering
    \begin{subfigure}[b]{0.48\textwidth}
        \includegraphics[width=\textwidth]{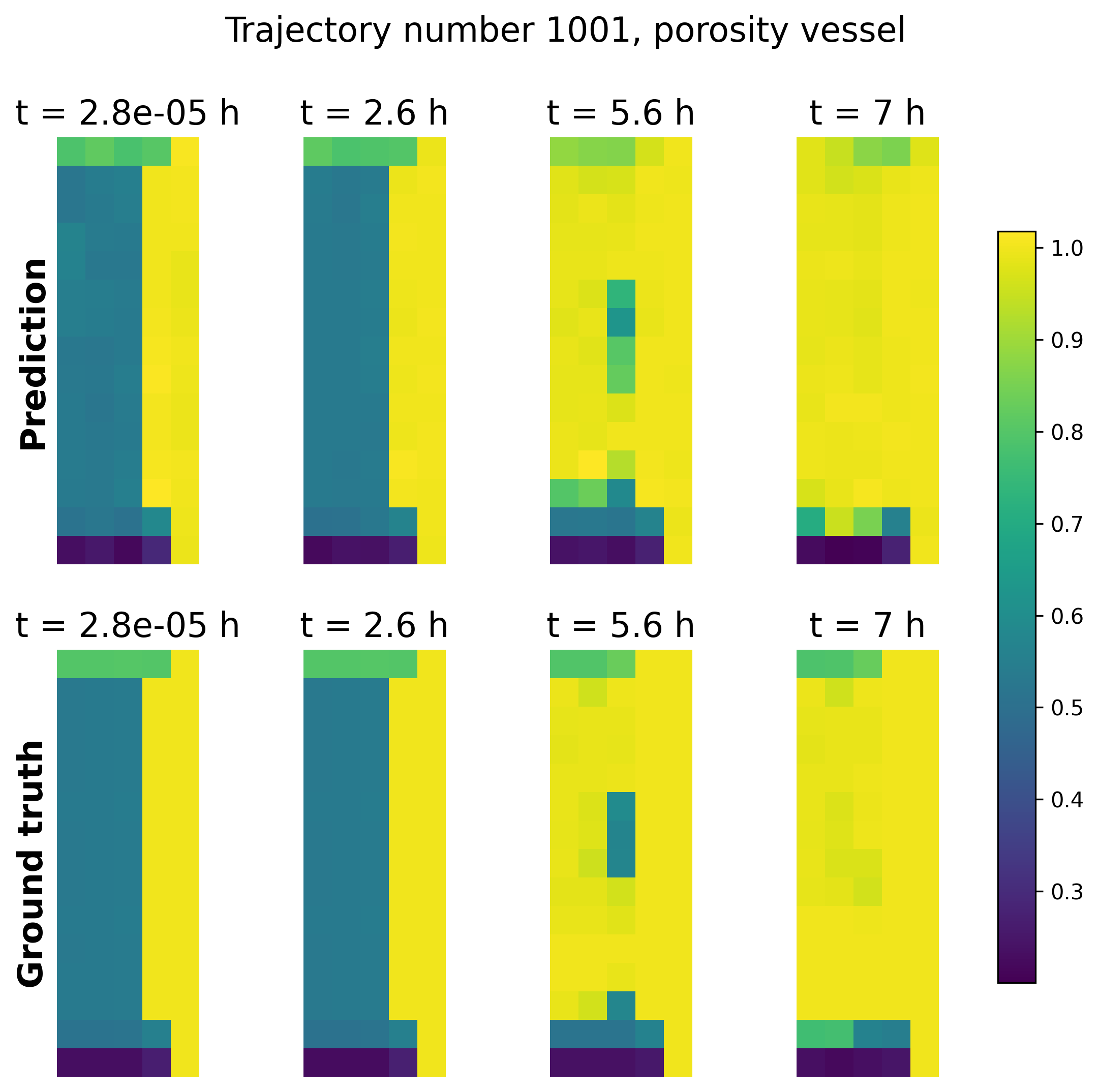}
        \caption{Porosity in the vessel}
        \label{fig:1001_porosity_vessel_SBO}
    \end{subfigure}
    \hfill
    \begin{subfigure}[b]{0.48\textwidth}
        \includegraphics[width=\textwidth]{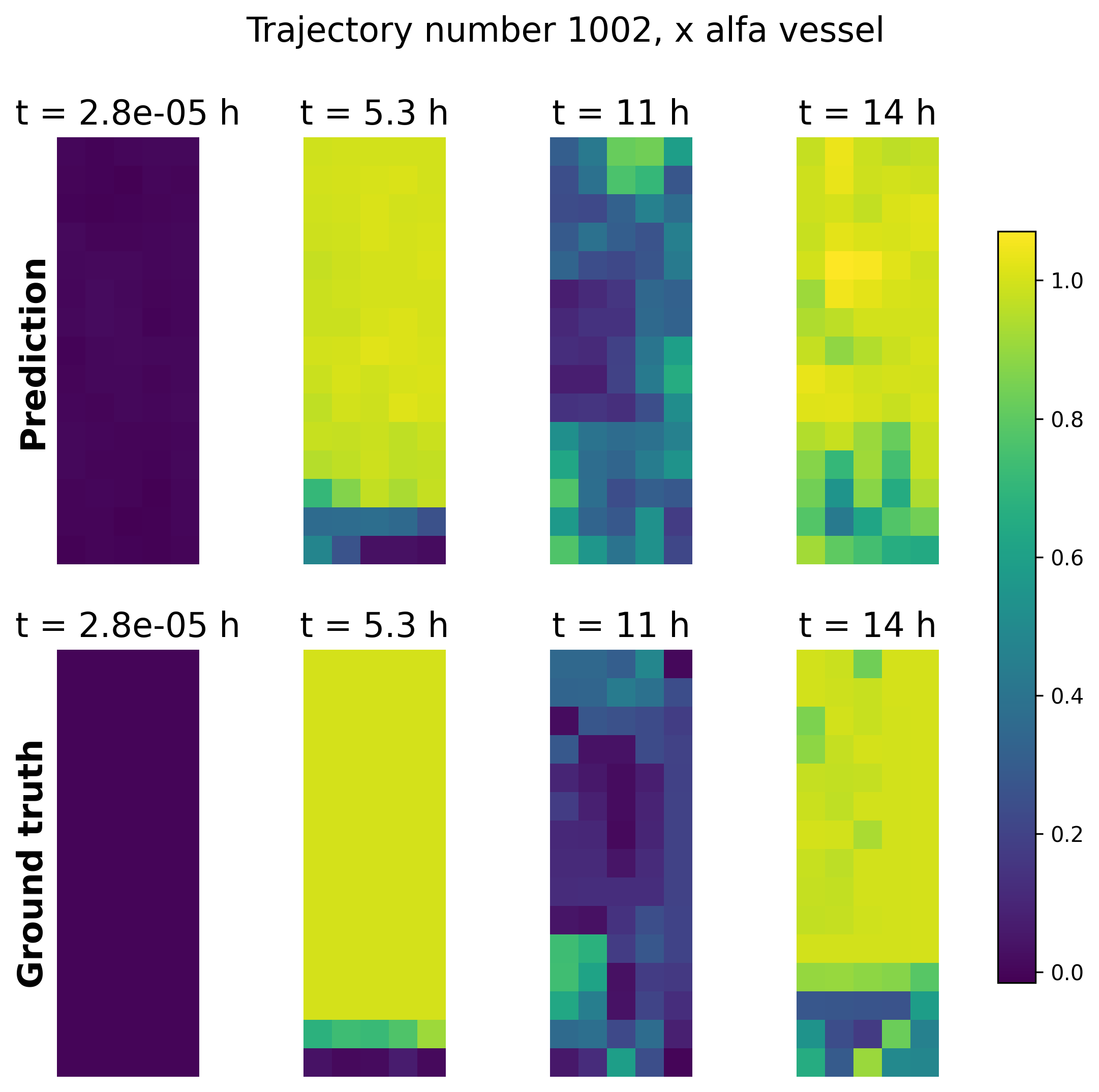}
        \caption{Void fraction in the vessel}
        \label{fig:1002_x_alfa_vessel_SBO}
    \end{subfigure}
    \caption{AE-NODE predictions vs.\ ground truth at $4$ different time steps of porosity and void fraction in the vessel.}
    \label{fig:SBO_2d_predictions}
\end{figure}

\begin{figure}[htbp]
    \centering
    \begin{subfigure}[b]{0.48\textwidth}
        \includegraphics[width=\textwidth]{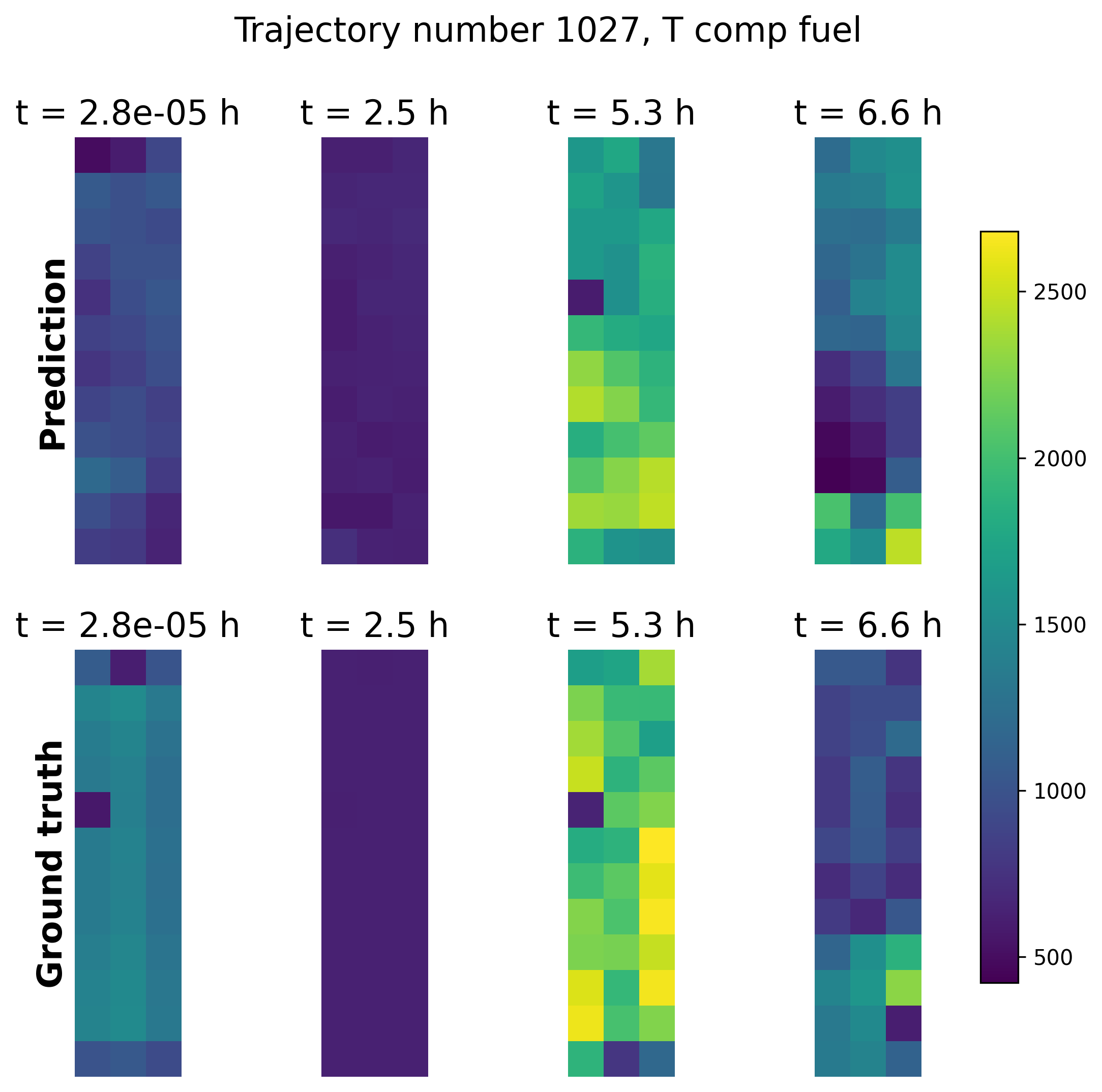}
        \caption{T fuel}
        \label{fig:1027_T_comp_fuel_SBO}
    \end{subfigure}
    \hfill
    \begin{subfigure}[b]{0.48\textwidth}
        \includegraphics[width=\textwidth]{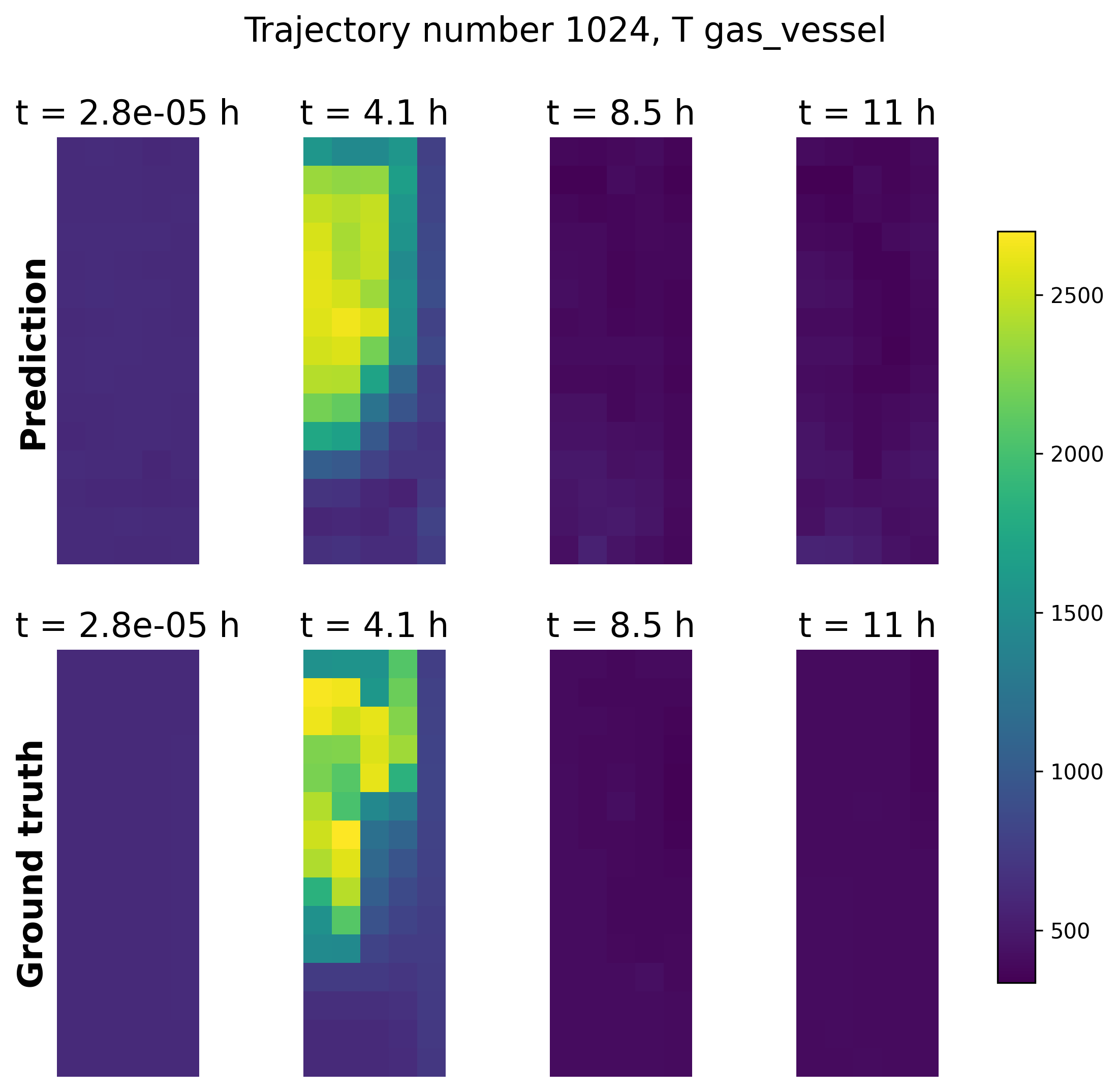}
        \caption{T gas}
        \label{fig:1024_T_gas_vessel_SBO}
    \end{subfigure}
    \caption{AE-NODE predictions vs.\ ground truth at $4$ different time steps of temperature of the fuel in the core [K] and temperature of the gaseous phase in the vessel [K].}
    \label{fig:SBO_2d_temperature}
\end{figure}

For the SBO scenario we train on 286 trajectories, we do validation on 50 and test on 27. In Figure \ref{fig:SBO_AE_AE_NODE_comparison} we show the values $\text{RMSE}_{mean}(vr)$, $\text{RMSE}_{max}(vr)$ and $\text{RMSE}_{std}(vr)$ on the 27 testing trajectories comparing AE and AE-NODE predictions. Figure \ref{fig:SBO_m_cum_H2}, \ref{fig:SBO_rho_liq_lower_plenum} and \ref{fig:SBO_T_liq_lower_plenum} display the AE-NODE predictions over time for four testing trajectories of the H2 cumulated mass in the core, the density of the liquid and of the temperature of the liquid in the lower plenum. Figures \ref{fig:SBO_2d_predictions} and \ref{fig:SBO_2d_temperature} shows the prediction of AE-NODE of the porosity and of the void fraction in the vessel and of the temperature of the fuel (core) and of the gaseous phase (vessel).

\subsection{Discussion}
\label{subsec:discussion}
From the metrics computed on the test set in Figure \ref{fig:LOCA_AE_AE_NODE_comparison} and \ref{fig:SBO_AE_AE_NODE_comparison} we see that there are several scalar variables with large $\text{RMSE}_{mean}$ and $\text{RMSE}_{std}$, while low $\text{RMSE}_{max}$ both for the AE and AE-NODE inference: x alpha, P H2, m gas, Q liq vap, porosity in $s_p$, Q H2O ptv, m H2O ptv, Q steam vtp and FP in $s_g$, $s_{B_{1}}$ and $S_{B_2}$. In Figure \ref{fig:near_zero_variables_SBO} and \ref{fig:near_zero_variables_LOCA} we show examples of such variables from testing simulations from SBO and LOCA testing datasets respectively; while the general trend over time is correctly predicted by the SM, the high non-linearity of the variables over time make it difficult for the SM to make accurate predictions pointwise in time, especially the sharp changes at the end of the simulations. Similarly, variables like Q m liq, V gas and V liq, m magma debris 0 and 1 in $s_f$ and $s_v$ show little variance over time with values close to zero and sudden large increases in value in some volumes over a short time-span as shown in Figure \ref{fig:debris_and_q_liq}. In $s_p$, the variables V deb and V mag showcase large errors for a similar reason, having mostly zero value across time beside a sudden rise at the final time-steps.

\begin{figure}[h]
  \centering
  \includegraphics[width=0.5\textwidth]{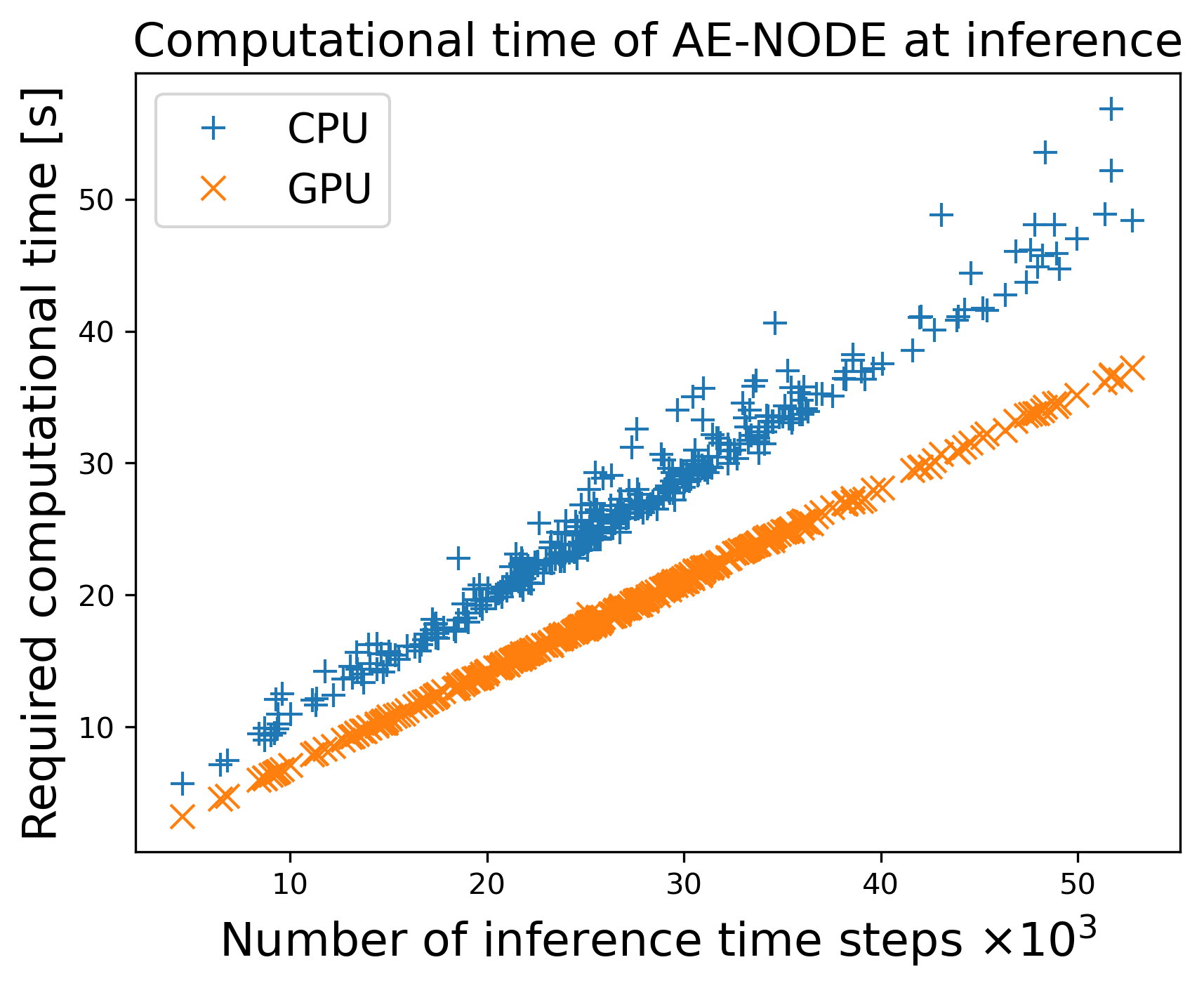}
  \caption{CPU and GPU comparison of time (in seconds) required by AE-NODE to simulate the full spatio-temporal domain of the vessel as a function of the number of time-steps needed to reach vessel rupture. We used an NVIDIA A100 80GB PCIe and an Intel(R) Xeon(R) Gold 6226R CPU @ 2.90GHz.}
  \label{fig:computational_time_at_inference}
\end{figure}

The plots in Figures \ref{fig:LOCA_AE_AE_NODE_comparison} and \ref{fig:SBO_AE_AE_NODE_comparison} quantify how much of the error of the AE-NODE prediction is due to the AutoEncoder and to the NODE: we can use the AE errors as lower bounds, since the NODE in the best case can exactly predict the latent space built by the AE, and thus AE-NODE cannot perform better than the AE. Both in the LOCA and SBO scenario we do not see variables where AE and AE-NODE perform dramatically different, signaling that the NODE well approximates the evolution of the vectors belonging to the reduced space $\mathcal{E}$ and that the AE approximation is the major source of error in AE-NODE. To further analyze the correspondence between the reduced space $\mathcal{E}$ found by the AE and the NODE prediction, we plot in Figures \ref{fig:latent_space_all_LOCA} and \ref{fig:latent_space_all_SBO} some latent trajectories obtained by encoding the ground truth and the NODE predictions obtained autoregressively from the initial condition; we can see that in both cases the NODE is able to achieve a stable autoregressive rollout, with predictions spanning from $4$ to $20$ hours. Looking more closely, it can be seen how the NODE is able to correctly capture different dynamical behaviours across different testing trajectories, hinting at its ability to correctly respond to different inputs coming from the primary circuits. Notably, the AE is able to compress all the physical fields contained in $s$ by a factor $332$, since the $1913$ degrees of freedom in the physical space are mapped into only $6$ latent dimensions. Such reduction results in a fast computation (under a minute), both on CPU and GPU, of the full spatio-temporal domain of the physics of the vessel, as shown in Figure \ref{fig:computational_time_at_inference}. We used an NVIDIA A100 80GB PCIe and an Intel(R) Xeon(R) Gold 6226R CPU @ 2.90GHz. As expected, the time required by AE-NODE to simulate the full spatio-temporal domain grows linearly with the number of time-steps required by each simulation, since the latent dynamics is solved using an explicit (fixed-step) Runge-Kutta solver \cite{Ascher1998ComputerMF}. In Table \ref{tab:icare_time} we show a comparison of the computational time required by ASTEC and by AE-NODE on a set of LOCA simulations: AE-NODE gives a reduction of about 650 times in the mean time and of about 500 in median time (both on CPU and GPU). In Table \ref{tab:icare_time} the ASTEC computational times only consider the ICARE module, hence being a lower bound on the actual time required by ASTEC to solve the physics simulated by AE-NODE (which substitutes the coupling between CESAR and ICARE). The information concerning ASTEC computational times are extracted from the simulations run by the Centre for Energy, Environmental and Technological Research (CIEMAT).

\begin{table}[h]
  \centering
  \caption{LOCA scenario: comparison of mean and median simulation time (in seconds) between ASTEC and AE-NODE (CPU and GPU). For ASTEC we report the simulation time concerning only ICARE, while for AE-NODE the ICARE-CESAR coupling. ASTEC computational time information is limited to the simulations run by the ASSAS partners in the Centre for Energy, Environmental and Technological Research (CIEMAT).}
  \label{tab:icare_time}
  \begin{tabular}{lcc}
    \toprule
    & \textbf{Mean (s)} & \textbf{Median (s)} \\
    \midrule
    ASTEC/ICARE   & 16879.8 & 13082.4 \\
    \midrule
    AE-NODE (CPU) &      26.5  &      25.8  \\
    AE-NODE (GPU) &      18.9  &      18.3  \\
    \bottomrule
  \end{tabular}
\end{table}

The ability of AE-NODE to capture varying physical details across testing trajectories can be observed in Figures \ref{fig:LOCA_m_tot_cor}, \ref{fig:LOCA_x_alpha}, \ref{fig:LOCA_T_liq_plenum_predictions}, \ref{fig:SBO_m_cum_H2}, \ref{fig:SBO_rho_liq_lower_plenum} and \ref{fig:SBO_T_liq_lower_plenum}, where we show the prediction of the AE-NODE vs. the ground truth for some scalar variables across different testing trajectories. In the plotted figures the trend of the variables is correctly predicted and, although some sharp changes are not captured, the differences across trajectories are well predicted by the SM, even for varying time series lengths. In Figures \ref{fig:LOCA_2d_predictions}, \ref{fig:LOCA_2d_temperature}, \ref{fig:SBO_2d_predictions} and \ref{fig:SBO_2d_temperature} we show the AE-NODE predictions for some core ($s_{cr}$) and vessel ($s_{v}$) variables for $4$ snapshots in time. Again, while some finer details are missed, the AE-NODE is able in those cases to reproduce correctly the dynamics of the system. In Figures \ref{fig:LOCA_2d_core_degradation} and \ref{fig:SBO_2d_core_degradation} we plot the 2D variables during the degradation phase (when the mass of the corium is larger than 1 tonne).

We observed that some variables are predicted noticeably better than others, with the AE-NODE performing better on variables that present more meaningful evolution over time and less noisy behaviour, as in Figures \ref{fig:LOCA_T_liq_plenum_predictions}, \ref{fig:SBO_rho_liq_lower_plenum} and \ref{fig:SBO_T_liq_lower_plenum} as opposed to the variables in Figures \ref{fig:near_zero_variables_SBO} and \ref{fig:near_zero_variables_LOCA}. The ground truth of ASTEC is extremely challenging to be approximated, not only because of the sharp changes but also because of the numerous oscillating phenomena often present, as shown in Figure \ref{fig:challenging_examples}: beside the sharp jumps, there are also time intervals where the ground truth exhibits high-frequency and oscillating behaviour. Because this work is the first surrogate modeling attempt done on ASTEC data we initially worked on the raw data without applying any filter, as in the results shown so far. Later, we experimented with the Savitzky–Golay filter as explained in Appendix \ref{appendix:smoothing_of_data}. The applied smoothing kept most of the discontinuities and sharp changes and reduced the high-frequency oscillations of the type seen in Figure \ref{fig:challenging_examples}. Since it is difficult to decide for about $80$ variables across all the trajectories what noise is vs. signal, we decided to not completely smooth out all irregularities. Furthermore, we noticed that the smoothing introduced some artifacts like large negative masses towards discontinuities, and thus in order for it to be properly applied a careful study variable by variable should be conducted. The most noticeable improvement introduced by such smoothing is a more regular and easy to approximate reduced space $\mathcal{E}$, as shown in Figures \ref{fig:latent_space_all_LOCA_smoothed} and \ref{fig:latent_space_all_SBO_smoothed}: the latent dynamics is less compressed compared to Figures \ref{fig:latent_space_all_LOCA} and \ref{fig:latent_space_all_SBO} where smoothing is not applied. In the non-smoothed case the latent space is compressed to make it easier to the NODE to approximate the difficult and noisy latent dynamics, while such compression is no more needed when the smoothing is applied. We observed similar behaviour in the non-smoothing case when observing the transition during training from the AE phase to the autoregressive phase (as explained in Appendix \ref{appendix:appendix_TF_AR}): when the training only comprises the AE we see in the latent dynamics a separation of the latent dimensions as in Figures \ref{fig:latent_space_all_LOCA_smoothed} and \ref{fig:latent_space_all_SBO_smoothed}. When the NODE training is then switched on, the latent dynamics is compressed to make it easier to construct the function $f_\theta$ of the NODE. We do not extend our experiments with the smoothing further as they would require careful study from severe accidents experts in order to establish what is noise and what represents actual physics per each variable. Finally, in Figures \ref{fig:boundaries_over_time_LOCA} and \ref{fig:boundaries_over_time_SBO} we show the RMSE$_{mean}$ per time step for LOCA and SBO testing trajectories for the variables $s_{B_1}$ and $s_{B_2}$, which are important to couple AE-NODE to the primary circuit as shown in Figure \ref{fig:coupling_explained}.

Because this paper is the first attempt to build a surrogate model of the vessel of ASTEC, we cannot compare our results to any other work. However, by looking at Figures [14-18] of \cite{Bae2026}, we see that other DL methods struggle to reproduce the complex non linearities characteristic of SA data.
\section{Conclusions}
We have built the first, purely data-driven (non-intrusive) SM of the physics of the vessel simulated by ASTEC by decoupling the vessel from the primary circuit. By doing so we described a general methodology that can be applied to general (not necessarely nuclear accidents) multi-physics simulations where different coupled modules are used to model a physical system. The SM was modeled building from the method described in \cite{Longhi2026}, which couples an AutoEncoder with a Neural ODE. We slightly modified and improved it, by making the autoregressive training more flexible via the adaptive window technique described in Appendix \ref{appendix:appendix_TF_AR} and by adding the loss function term $\mathcal{L}_{AR}^{full}$. We trained 2 SMs on 2 different datasets, one per accident type: a SBO and a LOCA. In Section \ref{sec:results} we showed the performance of the SMs on some testing trajectories, analyzing the error metrics computed per variables, showing where the SM performs well and where not. Despite the challenging non-linear and discontinuous physics simulated by ASTEC, the SM was able to predict around 80 physical (scalar and field) variables with a stable autoregressive rollout going from 4 to 20 hours of time span (from $10k$ to $50k$ time-steps). Furthermore, the AE compressed the $1913$ degrees of freedom to just $6$, making it possible for the SM to predict the full spatio-temporal domain of the vessel in under a minute (both with CPU and GPU). As shown in Table~\ref{tab:icare_time}, AE-NODE running on CPU is approximately 640× faster than ASTEC in mean simulation time and approximately 510× faster in median time, with GPU inference pushing these speedups to 890× and 720× respectively. This work represents a step toward fast, real-time simulators of severe accidents in nuclear reactors and reinforces the case for data-driven methods such as Deep Learning as a substitute for traditional solvers in such applications. Severe accident simulators can be used to train power plants operators to correctly act in the case of nuclear accidents and are thus an helpful tool in making nuclear plants a safe and reliable source of energy.

\subsection{Limitations and future work}
While the SM proposed in this work has shown some potential in terms of prediction capabilities, we cannot say at this stage that such SM can be used as a reliable substitute of ASTEC in the vessel, because of the failures of the model discussed in Section~\ref{subsec:discussion}. In this work we used the 800 LOCA and 300 SBO simulations that were available within the ASSAS project. The next immediate step would be to generate more training data to understand if the failures of the SM are due to modeling issues or lack of data. Because of the imbalance in LOCA and SBO simulations,  we did not focus on a single SM for LOCA and SBO scenarios, as first trials biased too much the network toward LOCA predictions, because of the imbalanced proportion. 

Perhaps even more importantly, an extensive work on data filtering and smoothing should be done together with severe accidents experts, in order to understand where the physics and where the noise lie in such discontinuous and noisy signals. If it turns out that none of the observed spikes or nonlinearities in the variables are numerical noise, then more data and SMs with larger capacity (more weights), might be able to capture all the difficult transients. If otherwise it is possible to remove most of the noise, while preserving the physical content, then data-driven SMs can become substitutes for traditional solvers even with the data-size and model capacity described in this paper.

Another limitation of this work has to do with the final time of the simulations: in this paper we simply ended the testing simulations based on the end of the ground truth trajectories, while the SM should in principle be able to predict the vessel rupture, either by the use of an additional NN, or by developing a physical model suggested by nuclear physicists that takes as input the predictions of the SM.

Finally, this work did not consider the feedback effect at the volumes of the primary circuit $h_1$ and $c_1$: the predictions of the SM, even when slightly different from the ground truth, would change the predictions of the primary circuit of the variables at $h_1$ and $c_1$, pushing even further from the ground truth the subsequent SM predictions. This issue is connected with the next steps that follow naturally from this work: coupling the SM of the vessel with the primary and secondary circuits, either simulated by ASTEC or by SMs themselves.

\subsection{Acknowledgments}
Funded by the European Union under grant agreement no. 101059682. Views and opinions expressed are however those of the author(s) only and do not necessarily reflect those of the European Union or the European Commission-Euratom. Neither the European Union nor the granting authority can be held responsible for them. This work has been developed within ASSAS (Artificial intelligence for Simulation of Severe AccidentS) \cite{assas}, a Horizon Europe funded project targeting the development of nuclear severe accident simulators. We thank the partners from Nuclear Safety and Radiation Protection Authority (ASNR), CIEMAT, Jožef Stefan Institute (JSI), ENEA, Karlsruhe Institute of Technology (KIT) for data creation and sharing and the partners from CIEMAT, ASNR and Phimeca for helping with the definition of input and output variables from ASTEC. We thank Eric Delaume for his precious comments concerning the ASTEC description and the results section.
\subsection{CRediT authorship contribution statement}
\textbf{Alessandro Longhi}: Conceptualization, Methodology, Software, Validation, Formal analysis, Investigation, Data Curation, Writing - Original Draft, Visualization. \\
\textbf{Danny Lathouwers}: Conceptualization, Methodology, Formal Analysis, Resources, Writing - Review and Editing, Supervision, Project administration, Funding acquisition.\\
\textbf{Zoltán Perkó}: Conceptualization, Methodology, Formal analysis, Resources, Writing - Review and Editing, Supervision, Project administration, Funding acquisition.
\newpage
\bibliographystyle{plainurl}
\bibliography{references}
\newpage
\appendix
\section{Appendix}

\section{Variables}
\label{appendix:complete_variables}
Following the description of the physics of the vessel given in Section \ref{subsec:the_vessel}, we give here a complete description of the physical variables considered in this work. In Table \ref{tab:variables_group1} we list the variables belonging to $s_g$, $s_p$, $s_{B_1}$ and $s_{B_2}$, where $g$ stands for $global$, $p$ for \textit{plenum} and $cr$ for \textit{core} and $B_1$ and $B_2$ indicate the boundaries $B_1$ and $B_2$ of Figure \ref{fig:vessel_variables} respectively. Most global variables are fission product mass flow rates of chemical elements, where 
\begin{align*}
FP = &\{Ac, Ag, Am, As, Ba, Br, Cd, Ce, Cm, Cs, Cu, Dy, Er, Eu, Ga, Gd, Ge, Ho, I, Ln, Kr, La, Mo, \\
&Nb, Nd, Np, Pa, Pd, Pm, Pr, Pu, Ra, 
Rb, Re, Rh, Ru, Sb, Se, Sm, Sn, Sr, Tb, Tc, Te, Th, Tl, \\&Tm, U, Xe, Y, Yb, Zn, Zr\}.
\end{align*} 
In Table \ref{tab:variables_group2} we list the variables belonging to $s_{cr}$, $s_v$, $p_{h_1}$, $p_{c_1}$ and $s_{f}$, where $cr$ stands for \textit{core}, $v$ for \textit{vessel}, $h_1$ and $c_1$ the first volumes of the primary circuit of Figure \ref{fig:vessel_variables} and $f$ stands for \textit{faces}. As explained in Section \ref{subsec:the_vessel}, $p_{h_1}$ and $p_{c_1}$ are an input to the SM, while the other variables are predicted by the SM and given as input to the SM autoregressively.
\begin{table}[]
\caption{Variables belonging to $s_g$, $s_p$, $s_{B_1}$ and $s_{B_2}$}
\label{tab:variables_group1}
\begin{tabular}{cl}
\toprule
\textbf{State space} & \textbf{Variables} \\
\midrule
$s_g$ & H2 cumulated mass in the core (m cum H2) [kg]\\
 & corium mass in the core (m tot cor) [kg]\\
 & total activity in a domain (FP A heat) [Bq]\\
 & maximum saturation in the core meshes (sat core mesh) [-]\\
 & mass flowrate of elements in Fp (FP) [kg/s]\\
\midrule
$s_p$ & pressure (P) [Pa]\\
 & gaseous phase temperature (T gas) [K]\\
 & liquid phase temperature (T liq) [K]\\
 & void fraction (x alpha) [-]\\
 & saturation temperature (T sat) [K]\\
 & hydrogen pressure (P H2) [Pa]\\
 & steam pressure (P steam) [Pa]\\
 & mass of gaseous phase (m gas) [kg]\\
 & mass of liquid phase (m liq) [kg]\\
 & density of the gaseous phase (rho gas) [kg/m$^3$]\\
 & density of liquid (rho liq) [kg/m$^3$]\\
 & liquid to vapor flowrate (Q liq vap) [kg/s]\\
 & porosity of the mesh with rods (porosity) [-]\\
 & volume proportion debris classes (V deb) [-]\\
 & volume proportion magma (V mag) [-]\\
& magma mass in the plenum (m magma) [kg]\\
 & vessel debris mass coming from fuel (m debris 0) [kg]\\
 &vessel debris mass coming from cladding (m debris 1) [kg]\\
\midrule
$s_{B_1}$ & instantaneous value of steam mass flow (Q steam ptv) [kg/s]\\
 & instantaneous value of water flow (Q H2O ptv) [kg/s]\\
 & cumulative total mass of water (m H2O ptv) [kg] \\
\midrule
$s_{B_2}$ & instantaneous value of steam mass flow (Q steam vtp) [kg/s]\\
 & instantaneous value of water flow (vtp) [kg/s]\\
 & cumulative total mass of water (m H2O vtp) [kg]\\
\bottomrule
\end{tabular}
\end{table}

\begin{table}[]
\caption{Physical variables belonging to $s_{cr}$, $s_v$, $p_{h_1}$, $p_{c_1}$ and $s_f$}
\label{tab:variables_group2}
\begin{tabular}{cl}
\toprule
\textbf{State space} & \textbf{Variables} \\
\midrule
$s_{cr}$ & fuel component temperature (T comp fuel) [K] \\
 & clad component temperature (T comp clad) [K] \\
 & categorical integers describing the component state of fuel (state fuel) [-] \\
 & categorical integers describing state of cladding (state clad) [-] \\
\midrule
$s_v$ & pressure (P) [Pa]\\
 & gaseous phase temperature (T gas) [K] \\
 & liquid phase temperature (T liq) [K] \\
 & void fraction (x alpha) [-] \\
 & saturation temperature (T sat) [K] \\
 & hydrogen pressure (P H2) [Pa]\\
 & steam pressure (P steam) [Pa]\\
 & mass of gaseous phase (m gas) [kg]\\
 & mass of liquid phase (m liq) [kg]\\
 & density of the gaseous phase (rho gas) [kg/m$^3$]\\
 & density of liquid (rho liq) [kg/m$^3$]\\
 & liquid to vapor flowrate (Q liq vap) [kg/s]\\
 & porosity of the mesh with rods (porosity) [-]\\
 & volume proportion debris classes (V deb) [-]\\
 & volume proportion magma (V mag) [-]\\
 & magma mass in the vessel (m magma) [kg]\\
& vessel debris mass coming from fuel (m debris 0) [kg]\\
 &vessel debris mass coming from cladding (m debris 1) [kg]\\
\midrule
$p_{h_1}$ & void fraction [-]\\
 & steam partial pressure [Pa]\\
 & gas temperature [K]\\
 & saturation pressure [Pa]\\
 & hydrogen partial pressure [Pa]\\
 & total pressure [Pa]\\
 & steam mass [kg]\\
 & liquid density [kg/m$^3$]\\
 & liquid mass [kg]\\
 & saturation temperature [K]\\
 & void fraction of steam water [-]\\
 & liquid temperature [K]\\
\midrule
$p_{c_1}$ & void fraction [-]\\
 & steam partial pressure [Pa]\\
 & gas temperature [K]\\
 & saturation pressure [Pa]\\
 & hydrogen partial pressure [Pa]\\
 & total pressure [Pa]\\
 & steam mass [kg]\\
 & liquid density [kg/m$^3$]\\
 & liquid mass [kg]\\
 & saturation temperature [K]\\
 & void fraction of steam water [-]\\
 & liquid temperature [K]\\
\midrule
$s_f$ & liquid mass flow rate (Q m liq) [kg/s]\\
 & gas velocity in face of vessel (V gas) [m/s]\\
 & liquid velocity in face of vessel (V liq) [m/s]\\
\bottomrule
\end{tabular}
\end{table}

\section{Sampled Operator Actions}
\label{appendix:appendix_op}
As described in Section \ref{subsec:Sampling of operator actions}, the sampling of the operator action of Table \ref{tab:op} is the source of variation of the data in this work. We use \textit{low-discrepancy} Sobol sequences from the OpenTurns library \cite{openturns} to sample the vectors that contain the time of activation of the operator actions. Because of the nature of the operator actions and because of the nature of the accident (LOCA or SBO), we put some constraints on the sampling, in order to sample scenarios that make sense. In Table \ref{tab:loca_sbo_constraints} we show the constraints chosen for the SBO scenario, namely we want the first Safety Relief Valve (SRV) opening before the second SRV, the primary spray after the feedwater/bleed starts, the steam generator spray after the first SRV operation, the steam generation transition before the feedwater/bleed starts and the Steam Generator (SG) transition before the latest between the SG spray or the first SRV. For the LOCA scenario instead: the first SRV operation must happen before the second SRV operation and the primary spray must be activated after the feedwater/bleed starts. In Table \ref{tab:parameter_ranges_sbo} we list the allowed ranges of the time of activation of the operator actions for the SBO scenario. Finally, in Table \ref{tab:parameter_ranges_loca} we list the allowed ranges for the LOCA scenario.

\begin{table}
\centering
\caption{Constraints for LOCA and SBO scenarios}
\label{tab:loca_sbo_constraints}
\begin{tabular}{clp{7cm}}
\toprule
\textbf{Scenario} & \textbf{Constraint} & \textbf{Description} \\
\midrule
LOCA, SBO & $t_1^{\text{srv}} < t_2^{\text{srv}}$ & First SRV operation before second SRV operation \\
LOCA, SBO & $t^{\text{pesp}} > t^{\text{fbseb}}$ & Primary spray after feedwater/bleed start \\
SBO & $t^{\text{pessg}} > t_1^{\text{srv}}$ & Steam generator spray after first SRV operation \\
SBO & $t^{\text{sg2tr}} < t^{\text{fbseb}}$ & Steam generator transition before feedwater/bleed start \\
SBO & $t^{\text{sg2tr}} < \max(t^{\text{pessg}}, t_1^{\text{srv}})$ & SG transition before the latest between the SG spray and the first SRV \\
SBO & $t^{\text{endssg2}} = t^\text{sg2tr} + 100\, s$ & Closing PORV happens 100 seconds after $t^\text{sg2tr}$ \\
\bottomrule
\end{tabular}
\end{table}

\begin{table}[h!]
\centering
\caption{Ranges of Operator Actions for SBO}
\label{tab:parameter_ranges_sbo}
\begin{tabular}{llll}
\toprule
\textbf{Parameter} & \textbf{Range} & \textbf{Units} & \textbf{Description} \\
\midrule
$t^{\text{sg2tr}}$ & $[10{,}800, 18{,}000]$ & s & Steam generator transition time (3--5 hours) \\
$t^{\text{fbseb}}$ & $[0, 25{,}000]$ & s & Feedwater/bleed start time (0--6.9 hours) \\
$t_1^{\text{srv}}$ & $[0, 35{,}000]$ & s & First SRV operation time (0--9.7 hours) \\
$\text{opensrv}$ & $[0, 100]$ & \% & SRV opening percentage \\
$t_2^{\text{srv}}$ & $[0, 35{,}000]$ & s & Second SRV operation time (0--9.7 hours) \\
$t^{\text{pesp}}$ & $[0, 35{,}000]$ & s & Primary spray time (0--9.7 hours) \\
$t^{\text{pessg}}$ & $[0, 35{,}000]$ & s & Steam generator spray time (0--9.7 hours) \\
$p_{u5}$ & $[5.0 \times 10^5, 6.0 \times 10^5]$ & Pa & Pressure parameter \\
$t^{\text{endssg2}}$ & $[10{,}900, 18{,}100]$ & s & Closing PORV after SGTR \\
\bottomrule
\end{tabular}
\end{table}

\begin{table}[h!]
\centering
\caption{Ranges of Operator Actions for LOCA}
\label{tab:parameter_ranges_loca}
\begin{tabular}{llll}
\toprule
\textbf{Parameter} & \textbf{Range} & \textbf{Units} & \textbf{Description} \\
\midrule
$t^{\text{fbseb}}$ & $[0, 10{,}000]$ & s & Feedwater/bleed start time (0--2.8 hours) \\
$t_1^{\text{srv}}$ & $[0, 10{,}000]$ & s & First SRV operation time (0--2.8 hours) \\
$\text{opensrv}$ & $[0, 100]$ & \% & SRV opening percentage \\
$t_2^{\text{srv}}$ & $[0, 10{,}000]$ & s & Second SRV operation time (0--2.8 hours) \\
$t^{\text{pesp}}$ & $[1{,}200, 8{,}380]$ & s & Primary spray time (0.33--2.3 hours) \\
$t^{\text{css}}$ & $[3{,}600, 86{,}400]$ & s & Containment spray time (1--24 hours) \\
$p_{u5}$ & $[5.0 \times 10^5, 6.0 \times 10^5]$ & Pa & Pressure parameter \\
\bottomrule
\end{tabular}
\end{table}

\section{Geometry of faces variables}
\label{appendix:geometry_of_faces_variables}
\begin{figure}[h]
    \centering
    \includegraphics[width=0.5\textwidth]{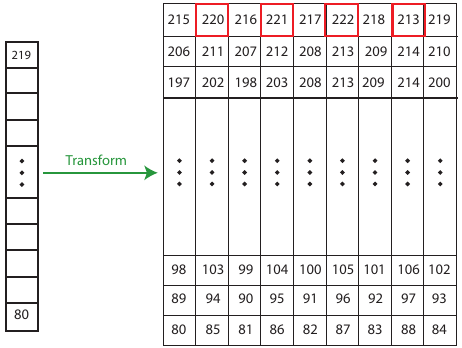}
    \caption{The variables belonging to the faces of the vessel,  given by ASTEC as 1D vectors of dimension $140$, are transformed into a 2D matrix, mimicking the index ordering of Figure \ref{fig:faces}, in order to exploit their local correlations with Convolutional NNs. The values of $s_f$ at the indeces $220,221,222$ and $213$ are obtained as the average of the adiacent boxes row-wise (so for index 220 we compute the average between the value at 215 and the value at 216).}
    \label{fig:faces_transformation}
\end{figure}
In section \ref{subsec:the_vessel} we defined the variables $s_f$ that belong to the \textit{faces} of the vessel as it is modeled by ASTEC. In Figure \ref{fig:faces} we show the indeces of the spatial domains of the faces, going from $80$ to $219$. For each variable of the faces, we get from ASTEC a 1D vector of dimension $140$. However, because of the strong spatial correlations, we want to organize each 1D vector into a 2D matrix so that we can use Convolutional NNs to process $s_f$. In order to do so, we map the 1D vector of dimension $140$ into a 2D matrix of dimensions $16\times 9$ as shown in Figure \ref{fig:faces_transformation}, where on the right it is shown the position of each index coming from the 1D vector on the left. At the very top of the matrix of Figure \ref{fig:faces_transformation} there are $4$ red boxes, with indeces $220$, $221$, $222$ and $213$ which do not belong to the original 1D vector as its maximum index is $219$. To give a value to those indeces, we compute the average of the adiacent boxes row-wise (so for index 220 we compute the average between the value at 215 and the value at 216).
\section{Teacher Forcing and Autoregressive stages via adaptive window training}
\label{appendix:appendix_TF_AR}
In section \ref{subsec:NODE} we described how the reduced dynamics through the NODE defined in Equation \ref{eq:ODE}. We also defined $\mathcal{L}_{TF}$ and $\mathcal{L}_{AR}$ as the 2 loss functions needed to model the TF and AR approaches depicted in Figure \ref{fig:TF_AR}. The effect of AR is bound to the values of $b$ and $l$ of the data-point $D^{b,l,j}$: ideally we would like to have $b=1$ and $b=F_j$, i.e., we would like to train autoregressively on the full time series of a given trajectory identified by $k_j$. However, there are $3$ issues when optimizing $f_\theta$ immediately on the full trajectory autoregressively: the training can be highly \textbf{unstable}, the training time can be very \textbf{large} and, if not enough $k_j$ have been sampled, \textbf{not enough} batches of data would be available, incurring in the large batches phenomena issue \cite{Keskar2016OnLT} which would penalize the training of $f_\theta$ and of the AE. For this reasons we construct the training dataset $\mathcal{M}_{tr}^l$ with a dependency on $l$, so that the time window spanned by each $D^{b,l,j}$ can be increase dynamically during the training. The training process thus consists of the following phases:
\begin{enumerate}
    \item we choose a low value for $l = l_0$, typically $l_0=25$, and we only train the AutoEncoder for a number $N_{AE}$ of epochs on $\mathcal{M}_{tr}^{l_0}$, i.e., we set $\delta=\nu=\mu=\omega=0.0$ from $\mathcal{L}_{tot}$. The idea behind this is that the latent space $\mathcal{E}$ needs to be shaped before the dynamics can be approximated;
    \item we keep $\mathcal{M}_{tr}^{l_0}$ and we switch on the $\mathcal{L}_{TF}$ and $\mathcal{L}_{t_m}$ losses, while keeping $\nu=\omega=0$, i.e., we do not train yet autoregressively. This is done for $N_{TF}$ epochs. The reason why we do so is to start building up $f_\theta$ in the easiest possible setting;
    \item we start with $\mathcal{M}_{tr}^{l_0}$ and we switch on also the AR losses $\mathcal{L}_{AR}$ and $\mathcal{L}_{AR}^{full}$. We wait $N_{l_0}$ epochs and after that we reshape the dataset into $\mathcal{M}_{tr}^{l_1}$, where $l_1>l_0$ and we train on $\mathcal{M}_{tr}^{l_1}$ for $N_{l_1}$ epochs. We repeat this process up to a certain $l_i$ with the waiting times and each $l_i$ pre-defined before the training starts.
\end{enumerate}
Thus, prior to the training, we have to define $N_{AE}$, $N_{TF}$, $L=[l_0,l_1,...,L_F]$ and $N_l = [N_{l_0},N_{l_1},...,N_{l_{F}}]$. We also experimented with a dynamic increase of the coefficient in front of the AR loss term $\nu$. What we do is defining $\nu_0<\nu$ with which we start the training at point $3$. After each epoch, we increase $\nu_0$ by $\nu_0$ until it reaches $\nu$. To address training instabilities, we experimented also with multiplying the $\mathcal{L}_{AR}$ loss with coefficients exponentially decaying from $t_b^j$ to $t_{b+l}^j$, so that the last auto-regressive predictions (likely wrong at the beginning of the training), would matter less. The exponential is gradually flattened over the epochs, so that at a later stage of the training, once the initial time-steps are learned auto-regressively, also the last predictions becomes important in the training procedure. These approach, while still available in the source code, turned out not to be necessary. We found out that ideally one would want to increase $l$ very slowly (increasing by 10), letting each training for a given $l$ to be completed before reshaping the training set. However such approach result in an extremely slow training when GPU resources are scarce, hence we optimized the SM in order to be able to only do a couple of datasets updates.

Finally, we experimented with decoupling the AE optimization from the NODE optimization. This decoupling can be total (the AE is trained first and NODE later), or partial (the AE is trained with the NODE up to a certain accuracy, then the AE's weights are frozen and only the NODE is trained). However, we found better performances when training the two systems coupled, as reported in \cite{Longhi2026} as well. This founding is connected with what we explained in the discussion: when the AR training is switched on, the latent space $\mathcal{E}$ is shaped in such a way that it is more easily approximated by the NODE; however if the AE and NODE trainings are decoupled, such phenomenon does not happen and the NODE struggles to approximate the latent dynamics.
\section{Training details}
\label{appendix:training_details_and_speed_of_inference}
In section \ref{subsec:NODE} we defined the total loss function $\mathcal{L}_{tot}$ which combines the losses needed to train the AE and the losses needed to train the NODE. There are $2$ more losses we have used, which we did not report in the main text as they are additional losses not needed to understand the AE-NODE methodology, but are used to help the training process. We call the first one $\mathcal{L}_{t_M}$. As explained in \cite{Longhi2026}, one of the desiderata of the learned NODE is the ability to do inference at smaller time steps than the one used during training. In order to do so, it can be beneficial to add an intermediate point in time $t_{\tilde{i}}^j$, such that $t_i^j<t_{\tilde{i}}^j<t_{i+1}^j$, with $t_{\tilde{i}}^j$ being extracted \textbf{randomly} from a uniform distribution. We thus define the loss 
\begin{equation}
\label{eq:loss_intermediate_dt}
    \mathcal{L}_{t_m}(D^{b,l,j}) = \frac{1}{(l-1)\cdot\lambda}\sum_{i=b}^{b+l-1}||\pi_\theta(\varepsilon_{\tilde{i},i}^j,\Delta t^j_{i+1,\tilde{i}})\circ\pi_\theta(\varepsilon_{i}^j,\Delta t^j_{\tilde{i},i})-\varepsilon_{i+1}^j||_2^2,
\end{equation}
which is similar to the TF loss, but with a mid-step between $t_{i}^j$ and $t_{i+1}^j$. The second additional loss we define is $\mathcal{L}_{AR}^{full}$, which computes an autoregressive pass with a decoding step, which is missing in $\mathcal{L}_{AR}$:
\begin{equation}
\label{eq:AR_full}
    \mathcal{L}_{AR}^{full}(D^{b,l,j}) = \frac{1}{(l-1)\cdot N}\sum_{i=b}^{b+l-1}||\tilde{s}_{i+1,b}(\mathbf{x},t_{i+1}|k_j)-s(\mathbf{x},t_{i+1}^j|k_j)||_2^2,
\end{equation}
with $\tilde{s}_{i+1,b}(\mathbf{x},t_{i+1}|k_j) = \psi_\theta(\varepsilon^j_{i+1,b})$. For training, we use the ADAM optimizer \cite{adam} and we apply a decaying learning rate scheduler which multiplies by $\gamma_{lr}<1$ the learning rate at each epoch. In Table \ref{tab:AR_hyperparameters} we show the hyper-parameters concerning the AR training for the $2$ experiments described in Section \ref{sec:results}. For training and testing we use either an NVIDIA A40 40 GB or an NVIDIA A100 80GB PCIe depending on availabilities. 

\begin{table}[]
\centering
\caption{Hyper-parameters of AR training}
\label{tab:AR_hyperparameters}
\begin{tabular}{lllll}
         & $N_{AE}$ & $N_{TF}$ & $L$ & $N_l$ \\
\hline
LOCA       &      25      &     10       &     [10,1000]      &    [25,25]           \\
SBO        &     25       &    10        &       [10,1000]    &   [25,25]     \\        
\end{tabular}
\end{table}
\section{Architectural details}
\label{appendix:architectural_details}
\begin{table}[]
\centering
\caption{Training and $f_\theta$ architecture hyper-parameters}
\label{tab:hyper_and_f_details}
\begin{tabular}{lllllllllll}
         & $\alpha$ &$\beta$&$\gamma$&$\delta$&$\nu$ &$\mu$&$\omega$& $\gamma_{lr}$ & $Layers\,f_\theta$ & $Neurons\,f_\theta$\\
\hline
LOCA            & 1.0 &1.0&1.0&1.0 &    1.0  &1.0&1.0    &      0.99          &    4    &  [200,200,200,200]    \\
SBO          &1.0 &1.0&1.0&1.0&    1.0  &1.0&1.0     &      0.99          &     4   &   [200,200,200,200]   \\
\end{tabular}
\end{table}
\begin{table}[]
\centering
\caption{Encoder $\varphi_\theta$ convolutional architecture hyper-parameters. $K_e = [3-3,3-3,3-3]$ and $S_e  = [1-1,2-1,2-1]$.}
\label{tab:cnn_encoder_conv_details}
\begin{tabular}{lllllllllll}
         & $f_e^{v}$ & $K_e^{v}$ & $S_d^{v}$ & $f_e^{cr}$ & $K_e^{cr}$ & $S_d^{cr}$ & $f_e^{f}$ & $K_e^{f}$ & $S_d^{f}$ \\
\hline
LOCA     & [18,32,64] & $K_e$ & $S_e$ & [16,32,64] & $K_e$ & $S_e$ & [16,32,64] & $K_e$ & $S_e$ \\
SBO      & [18,32,64] & $K_e$ & $S_e$ & [16,32,64] & $K_e$ & $S_e$ & [16,32,64] & $K_e$ & $S_e$ \\
\end{tabular}
\end{table}

\begin{table}[]
\centering
\scriptsize 
\caption{Decoder $\psi_\theta$ convolutional architecture hyper-parameters. $K_d = [3-3,4-3,4-3,3-3,3-3,3-3]$ and $S_d  = [1-1,2-1,2-1,1-1,1-1,1-1]$.}
\label{tab:cnn_decoder_conv_details}
\begin{tabular}{llllllllll}
         & $f_d^{v}$ & $K_d^{v}$ & $S_d^{v}$ & $f_d^{cr}$ & $K_d^{cr}$ & $S_d^{cr}$ & $f_d^{f}$ & $K_d^{f}$ & $S_d^{f}$ \\
\hline
LOCA     & [64,32,18,18,18] & [3-3,4-3,3-3,3-3,3-3,3-3] & $S_d$ & [64,32,16,8,4] & $K_d$ & $S_d$ & [64,32,16,8,3] & $K_d$ & $S_d$ \\
SBO      & [64,32,18,18,18] & [3-3,4-3,3-3,3-3,3-3,3-3] & $S_d$ & [64,32,16,8,4] & $K_d$ & $S_d$ & [64,32,16,8,3] & $K_d$ & $S_d$ \\
\end{tabular}
\end{table}

\begin{table}[]
\centering
\caption{Encoder $\varphi_\theta$ layers and neurons hyper-parameters}
\label{tab:dfnn_encoder_layers_neurons_details}
\begin{tabular}{lllllll}
         & $Layers_{e,sc}$ & $Neurons_{e,sc}$ & $Layers_{e,p}$ & $Neurons_{e,p}$ & $Layers_{e,cnc}$ & $Neurons_{e,cnc}$ \\
\hline
LOCA     &        3       &         [70,60,50]        &        3        &     [50,50,50]            &          3        &    [70,50,30]               \\
SBO      &   3             &       [70,60,50]         &       3         &     [50,50,50]            &         3         &    [70,50,30]               \\       
\end{tabular}
\end{table}

\begin{table}[]
\centering
\caption{Decoder $\psi_\theta$ layers and neurons hyper-parameters}
\label{tab:dfnn_decoder_layers_neurons_details}
\begin{tabular}{lllllll}
         & $Layers_{d,sc}$ & $Neurons_{d,sc}$ & $Layers_{d,p}$ & $Neurons_{d,p}$ & $Layers_{d,cnc}$ & $Neurons_{d,cnc}$ \\
\hline
LOCA     &        3      &        [50,60,70]        &        3       &      [50,50,50]           &         3         &     [30,50,70]              \\
SBO      &      3          &        [50,60,70]         &      3          &   [50,50,50]              &       3           &    [30,50,70]               \\         
\end{tabular}
\end{table}
\begin{table}[]
\centering
\caption{Latent dimensions}
\label{tab:latent_dimensions}
\begin{tabular}{lllllll}
         & $\lambda$ & $\lambda_{sc}$ & $\lambda_{p}$ & $\lambda_{v}$ & $\lambda_{cr}$ & $\lambda_{f}$ \\
\hline
LOCA      &       6     &      22     &       5    &    10  & 15 &  10         \\
SBO       &      6      &      11      &      5     &    15   & 5  & 10    \\        
\end{tabular}
\end{table}
We model $\varphi_\theta^{sc}/\psi_\theta^{sc}$, $\varphi_\theta^{p}/\psi_\theta^{p}$ , $\varphi_\theta^{cnc}/\psi_\theta^{cnc}$ as fully connected Autoencoders, while $\varphi_\theta^{v}/\psi_\theta^{v}$, $\varphi_\theta^{cr}/\psi_\theta^{cr}$ and $\varphi_\theta^{f}/\psi_\theta^{f}$ as Convolutional AutoEncoders. We model $f_\theta$ as a fully connected NN. We found beneficial for the stability of the training to multiply $f_\theta$ by a learnable scalar $\gamma_f<1$ in order to prevent $f_\theta$ from reaching NaN values during training due to large gradients. In Tables \ref{tab:hyper_and_f_details}, \ref{tab:cnn_encoder_conv_details}, \ref{tab:cnn_decoder_conv_details}, \ref{tab:dfnn_encoder_layers_neurons_details} and \ref{tab:dfnn_decoder_layers_neurons_details} we list the general hyper parameter of the trainings and the hyperparameters of $f_\theta$, $\varphi_\theta$ and $\psi_\theta$ respectively for the $3$ experiments made in Section \ref{sec:results}. $\alpha$ is the coefficient that multiplies $\mathcal{L}_{AE}^{reg}$, $\lambda$ is the latent dimension, $\gamma_{lr}$ is the coefficient of the scheduling of the learning rate, $Layers\,f_\theta$ is the number of hidden layers of $f_\theta$, $Neurons\,f_\theta$ is the list of neurons of the hidden layers of $f_\theta$. $Layers_{e,j}$ and $Neurons_{e,j}$  are the number of hidden layers and Neurons of $\varphi_\theta^{j}$, while $Layers_{e,j}$ and $Neurons_{e,j}$ refer to $\psi_\theta^j$; $f_e^v$, $f_{e}^{cr}$ and $f_{e}^{f}$ are lists of filters of $\varphi_\theta^{v}$, $\varphi_\theta^{cr}$ and $\varphi_\theta^{f}$ respectively. Similarly, in Tables \ref{tab:cnn_decoder_conv_details}, \ref{tab:dfnn_decoder_layers_neurons_details} there are the same hyperparameter details but for the Decoder $\psi_\theta$, hence the subscript $d$ instead of $e$. We used Group Normalization \cite{group_norm} for the Convolutional Autoencoders and Layer normalization \cite{ba2016layer} for fully connected AutoEncoders (only at the first layer of Encoder and first layer of the Decoder).

$f_\theta$ takes as input the latent vector $\epsilon(...)$ and the boundary conditions $p(\hat{\mathbf{x}},t_i^j|k_j)$ at time $t_i^j$. As done in \cite{Longhi2026}, we experimented with two ways of conveying the combined information from the $2$ inputs:
\begin{enumerate}
    \item simple concatenation. The latent vector is concatenated with the flattened vector $p(\hat{\mathbf{x}},t_i^j|k_j)$, resulting in a concatenated vector of dimension $\lambda+d_{h_1}+d_{c_1}$;
    \item FiLM \cite{perez2018film} application. The latent vector is transformed as $\varepsilon\rightarrow \beta_\theta^{film}(p)\odot \varepsilon + \delta_\theta^{film}(p)$, where $\beta_\theta^{film}:\mathcal{S}_p\rightarrow\mathbb{R}^\lambda$ and $\delta_\theta^{film}:\mathcal{S}_p\rightarrow\mathbb{R}^\lambda$ are two linear transformations parametrized by weights found during training. $\odot$ is the point-wise (Hadamard product).
\end{enumerate}
We did not find noticeable difference in the application of the two, so for simplicity we use concatenation (it results in faster training times). Before using either FiLM or concatenation, we map $p(\hat{\mathbf{x}},t_i^j|k_j)$ into a vector of dimension $12$ through a learnable linear multiplication.
\section{Smoothing of data}
\label{appendix:smoothing_of_data}
ASTEC data can be quite non-linear and noisy; hence we experiment with the Savitzky–Golay filter from the SciPy library \cite{scipy} on the training and testing datasets.
\begin{figure}[h]
    \centering
    \includegraphics[width=1.0\textwidth]{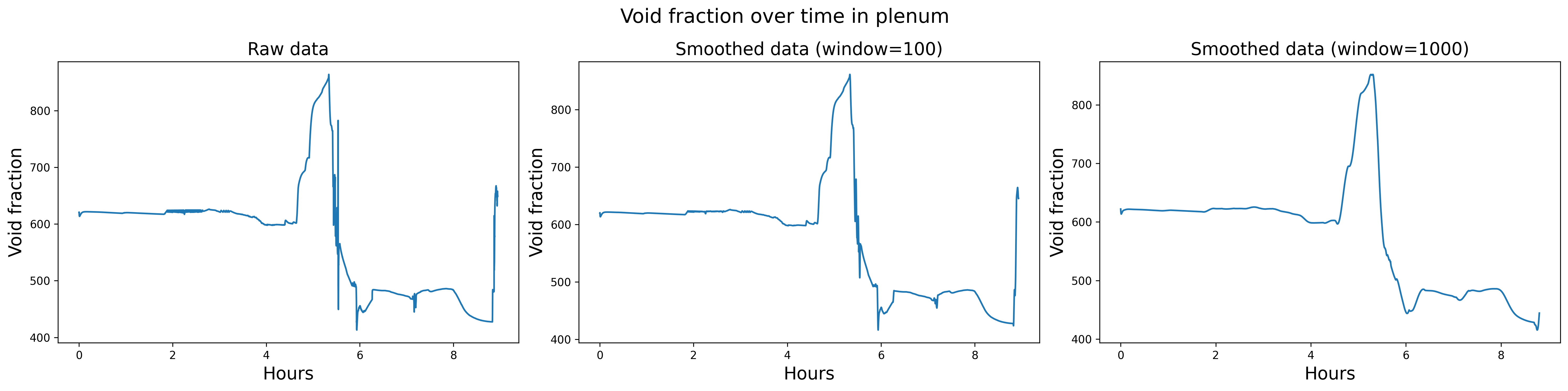}
    \caption{Application of the Savitzky–Golay filter on the void fraction in the plenum over time.}
    \label{fig:smoothing}
\end{figure}
In Figure \ref{fig:smoothing} we show the effect of such smoothing for the void fraction in the plenum over time when the window length is $100$ and when $1000$. In both cases the general shape is kept, however when the window is set to $1000$ most granular details are lost. Hence, we use a polynomial order of $3$ and a window length of $100$ time steps.
\section{Metrics}
\label{appendix:metrics}
Finding a unique working metric for this work is challenging, as the different fields represented in $s$ have different characteristics that can result in a biased error. For example, normalizing the error by the expected value results in an infinite error for scalar predictions that are $0$ over some time periods, while at the same time not normalizing the error can result in high values for variables that have large magnitude. For such reasons, we use the following diverse metrics, defined per trajectory $j$ and for a given prediction $\tilde{s}(\mathbf{x}_m,t_{i+1}|k_j)$:
\begin{enumerate}
    \item $\text{RMSE}^j_m = \frac{1}{F_j-1}\sum_{i=1}^{F_j-1}\left(\frac{||\tilde{s}(\mathbf{x}_m,t_{i+1}|k_j)-s_m(\mathbf{x}_m,t_{i+1}|k_j)||_2^2}{|\Omega_m|}\right)^{\frac{1}{2}}\in\mathbb{R}^{d_m}$;
    \item $\text{RMSE}^j_{m,mean}=  \frac{\text{RMSE}^j_m}{\text{Mean}\{{s_m(\mathbf{x}_m,t_{i}|k_j)\}_{i=1}^{F_j}}}\in\mathbb{R}^{d_m}$;
    \item $\text{RMSE}^j_{m,max}=  \frac{\text{RMSE}^j_m}{\text{Max}\{{s_m(\mathbf{x}_m,t_{i}|k_j)\}_{i=1}^{F_j}}}\in\mathbb{R}^{d_m}$
    \item $\text{RMSE}^j_{m,std}=  \frac{\text{RMSE}^j_m}{\text{Std}\{{s_m(\mathbf{x}_m,t_{i}|k_j)\}_{i=1}^{F_j}}}\in\mathbb{R}^{d_m}$
\end{enumerate}
where the index $m\in\{g,p,cr,v,B_1,B_2\}$ as defined in Subsection \ref{subsec:the_vessel} and $\tilde{s}(\mathbf{x}_m,t_{i+1}|k_j)$ can be either $\tilde{s}^m_{i+1,1}(\mathbf{x}_m,t_{i+1}|k_j)$, i.e., the actual prediction of AE-NODE, or $\psi\circ\varphi(s_m(\mathbf{x}_m,t_{i+1}|k_j))$, i.e., a simple autoencoding, as done in Figures \ref{fig:SBO_AE_AE_NODE_comparison} and \ref{fig:LOCA_AE_AE_NODE_comparison}. So each metric is a vector containing $d_m$ errors, each associated to a variable belonging to $s_m$. We are interested in looking at the error to each variable within $s_m$, so we define $\text{RMSE}^j_{m,a}(vr)=\text{RMSE}^j_{a}(vr)\in\mathbb{R}$ as the error at one of the variables defined in Appendix \ref{appendix:complete_variables}, where $a\in\{mean, max, std\}$ and $vr\in V$, where $V$ is the set of the names of all the variables of Appendix \ref{appendix:complete_variables}. Finally, given a testing set $M_{test}$ built from $k_j\in K_{test}\subset K, j=1,...,N_{test}$, we define $\text{RMSE}_{a}(vr) = \frac{1}{N_{test}}\sum_{i=1}^{N_{test}}\text{RMSE}^i_{a}(vr)$.

We experimented with $2$ types of normalization, min-max and mean-std. The four operations are computed across time and space \textit{per variable} within each $s_m$ (so not per $s_m$, but per variable within every $s_m$). So:
\begin{align}
    s_m(\mathbf{x},t_i|k^j)&\rightarrow \frac{s_m(\mathbf{x},t_i|k^j)-\text{min}\{s_m(\mathbf{x},t_n|k^j)\}_{n=1}^{F_j}}{\text{max}\{s_m(\mathbf{x},t_n|k^j)\}_{n=1}^{F_j}-\text{min}\{s_m(\mathbf{x},t_n|k^j)\}_{n=1}^{F_j}},\\  
    s_m(\mathbf{x},t_i|k^j)&\rightarrow \frac{s_m(\mathbf{x},t_i|k^j)-\text{mean}\{s_m(\mathbf{x},t_n|k^j)\}_{n=1}^{F_j}}{\text{std}\{s_m(\mathbf{x},t_n|k^j)\}_{n=1}^{F_j}},
\end{align}
where \text{min}, \text{max}, \text{mean}, \text{std} compute the minimum, maximum mean and standard deviation respectively. Note that $\text{min}/\text{max}/\text{mean}/\text{std}\{s_m(\mathbf{x},t_n|k^j)\}_{n=1}^{F_j}\in \mathbb{R}^{d_m}$, i.e., they have dimension $d_m$. Each statistic \textit{per variable} within each $s_m$ computed before training is saved and used during testing to compute the inverse transformations. We found min-max normalization to perform better than the mean-std one.
\section{Additional images from results}
\begin{figure}[htbp]
    \centering
    \begin{subfigure}[b]{0.48\textwidth}
        \includegraphics[width=\textwidth]{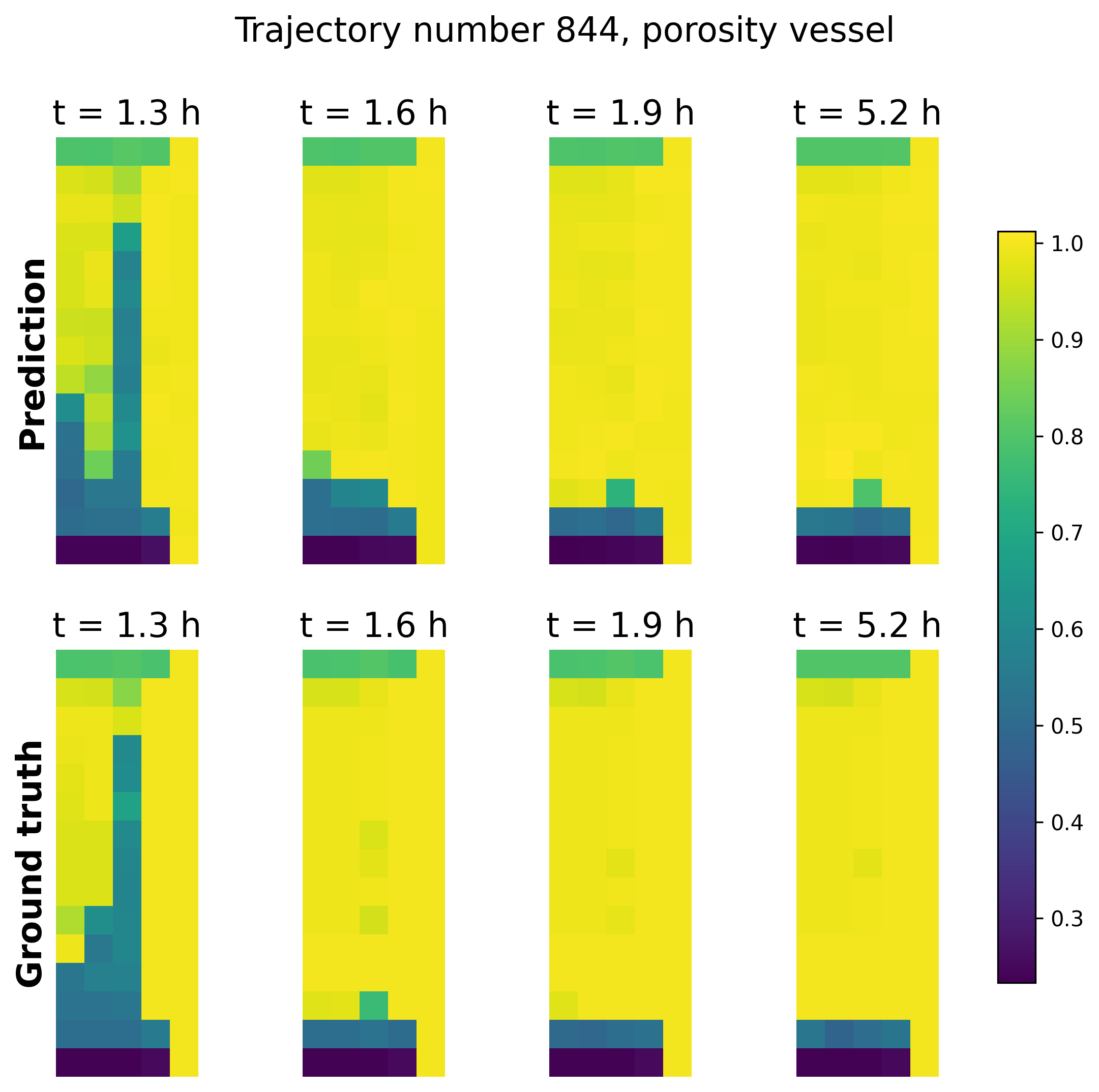}
        \caption{Porosity in the vessel}
        \label{fig:844_porosity_vessel_degradation_LOCA_cd}
    \end{subfigure}
    \hfill
    \begin{subfigure}[b]{0.48\textwidth}
        \includegraphics[width=\textwidth]{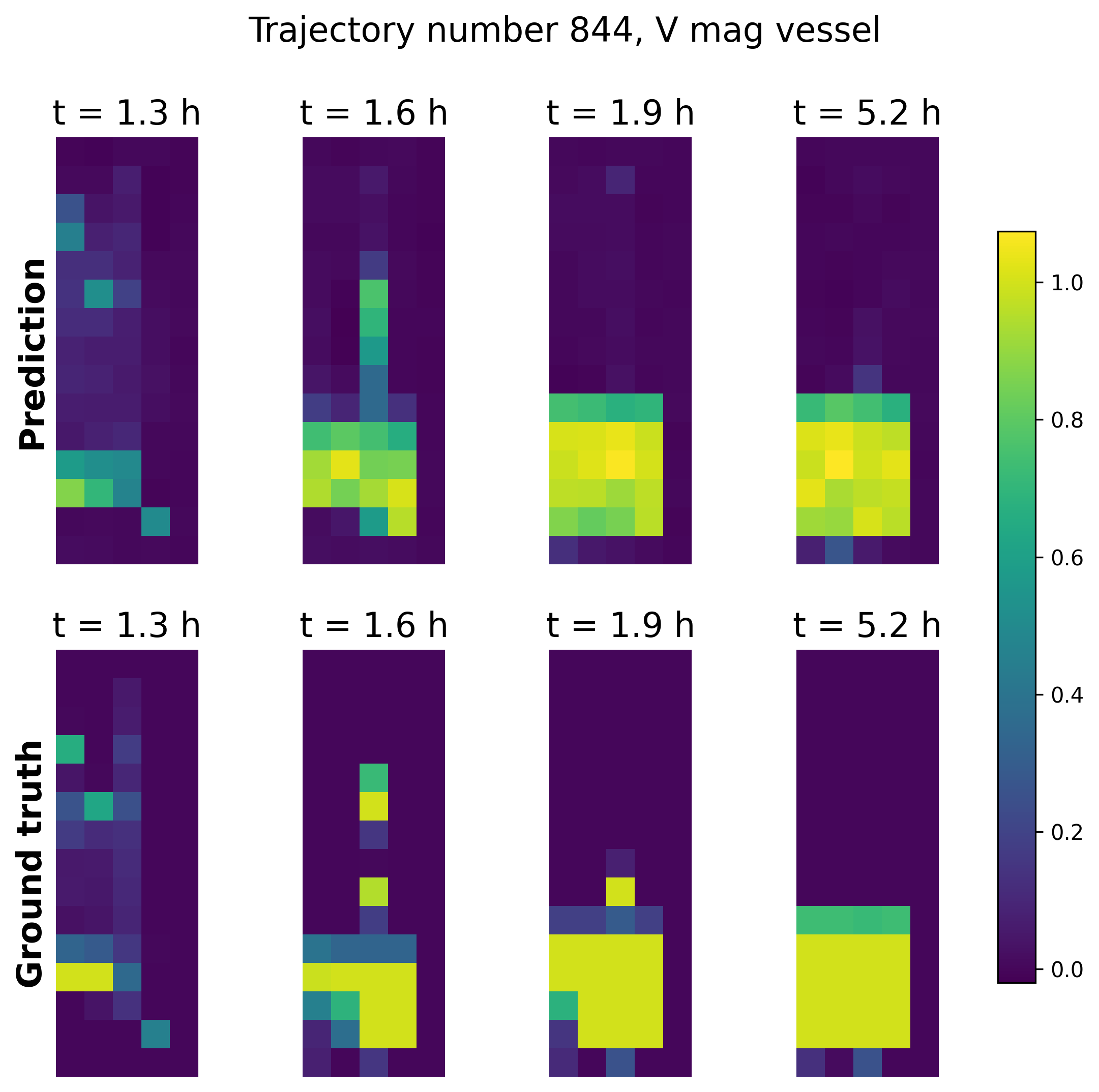}
        \caption{V magma}
        \label{fig:844_V_mag_vessel_degradation_LOCA}
    \end{subfigure}
    \caption{AE-NODE predictions vs.\ ground truth at $4$ different (during the degradation phase) time steps of the porosity in the vessel and the volume of the magma for the LOCA scenario.}
    \label{fig:LOCA_2d_core_degradation}
\end{figure}

\begin{figure}[htbp]
    \centering
    \begin{subfigure}[b]{0.48\textwidth}
        \includegraphics[width=\textwidth]{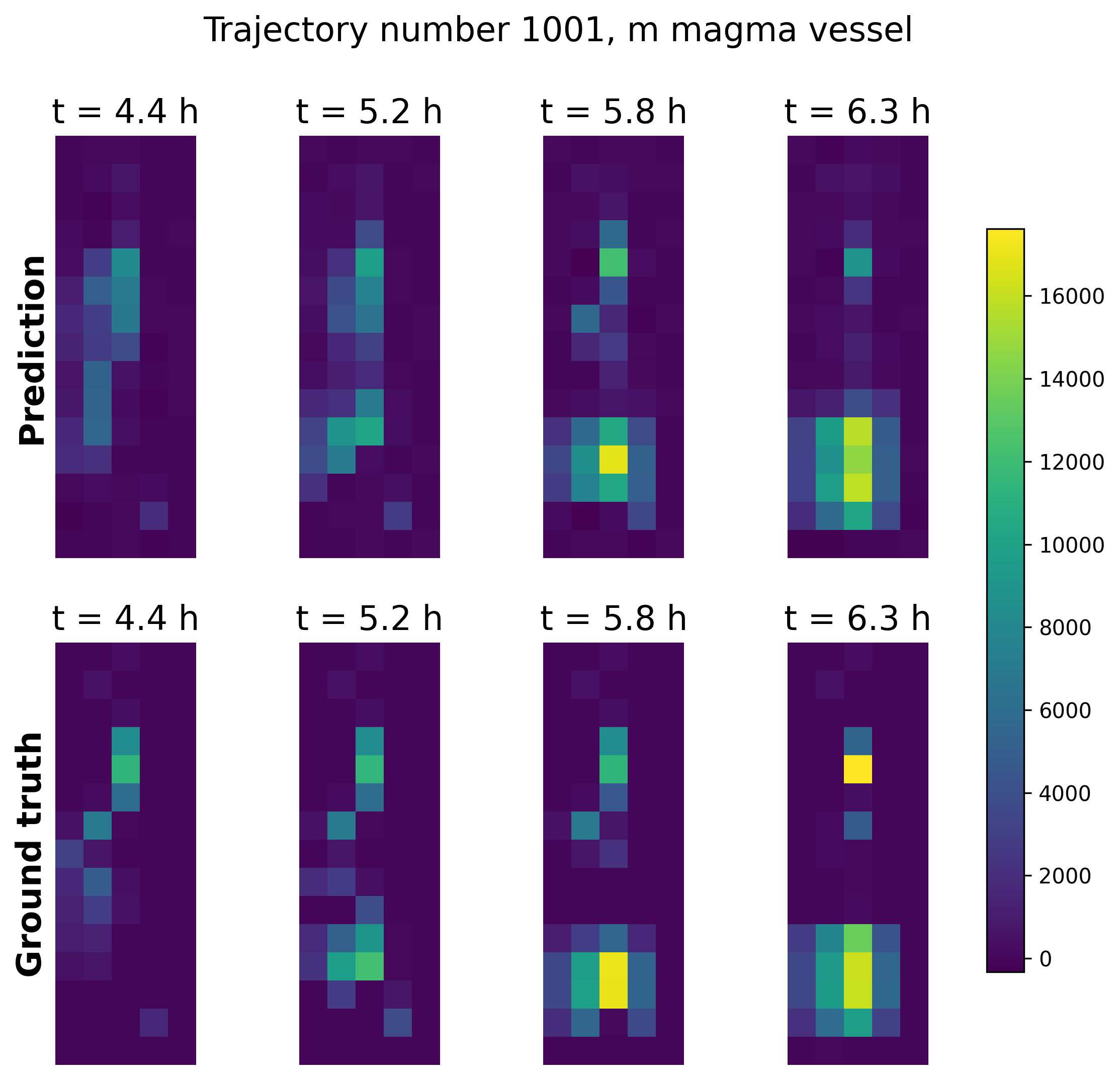}
        \caption{M magma}
        \label{fig:1001_m_magma_vessel_degradation_SBO_cd}
    \end{subfigure}
    \hfill
    \begin{subfigure}[b]{0.48\textwidth}
        \includegraphics[width=\textwidth]{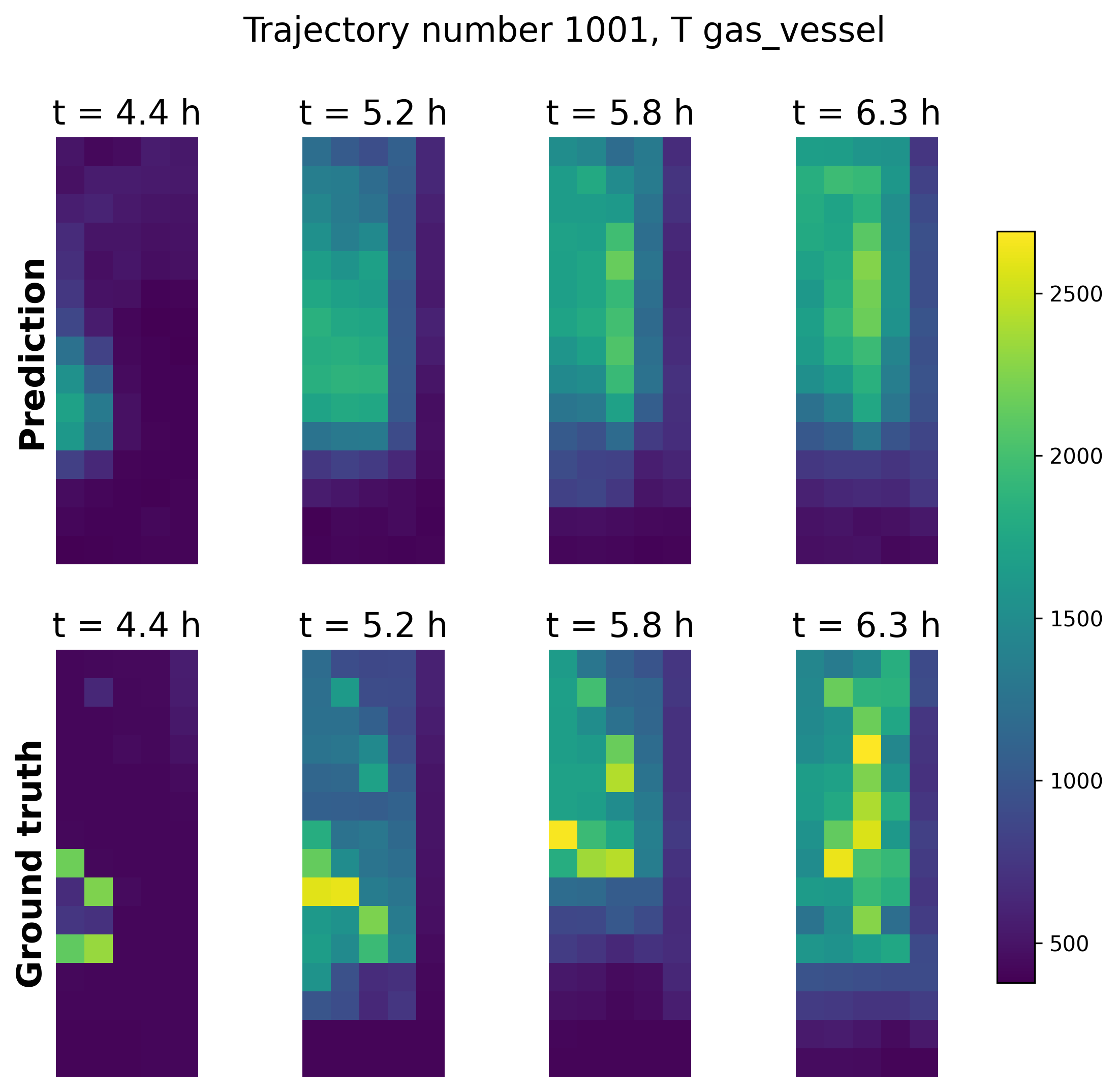}
        \caption{T gas}
        \label{fig:1001_T_gas_vessel_degradation_SBO_cd}
    \end{subfigure}
    \caption{AE-NODE predictions vs.\ ground truth at $4$ different (during the degradation phase) time steps of mass of the magma and temperature of the gaseous phase for the SBO scenario.}
    \label{fig:SBO_2d_core_degradation}
\end{figure}

In Figures \ref{fig:LOCA_2d_core_degradation} and \ref{fig:SBO_2d_core_degradation} we plot the 2D variables during the degradation phase (when the mass of the corium is larger than 1 tonne).

\begin{figure}[htbp]
    \centering
    \begin{subfigure}[b]{0.19\textwidth}
        \includegraphics[width=\textwidth]{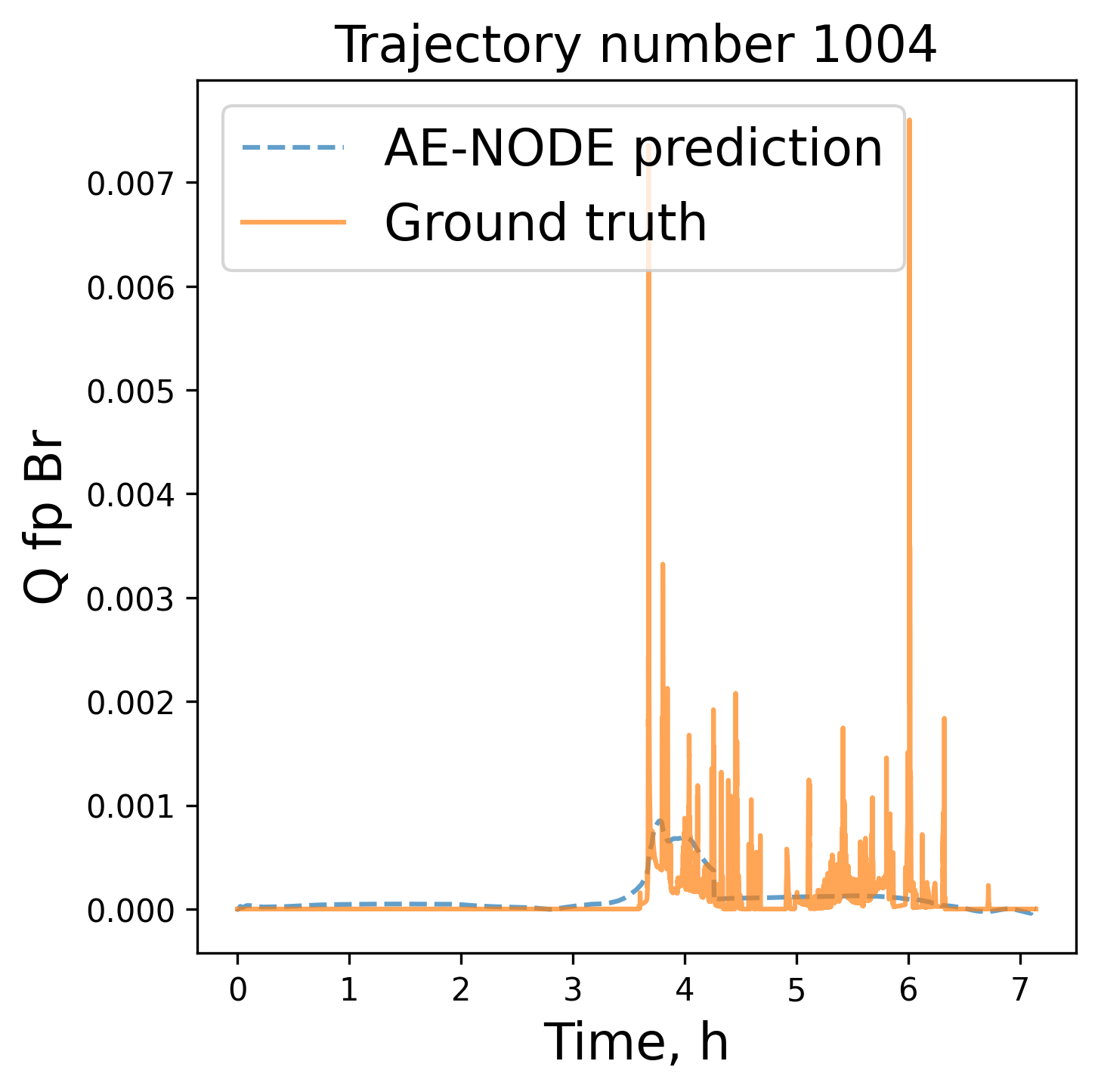}
    \end{subfigure}
    \hfill
    \begin{subfigure}[b]{0.19\textwidth}
        \includegraphics[width=\textwidth]{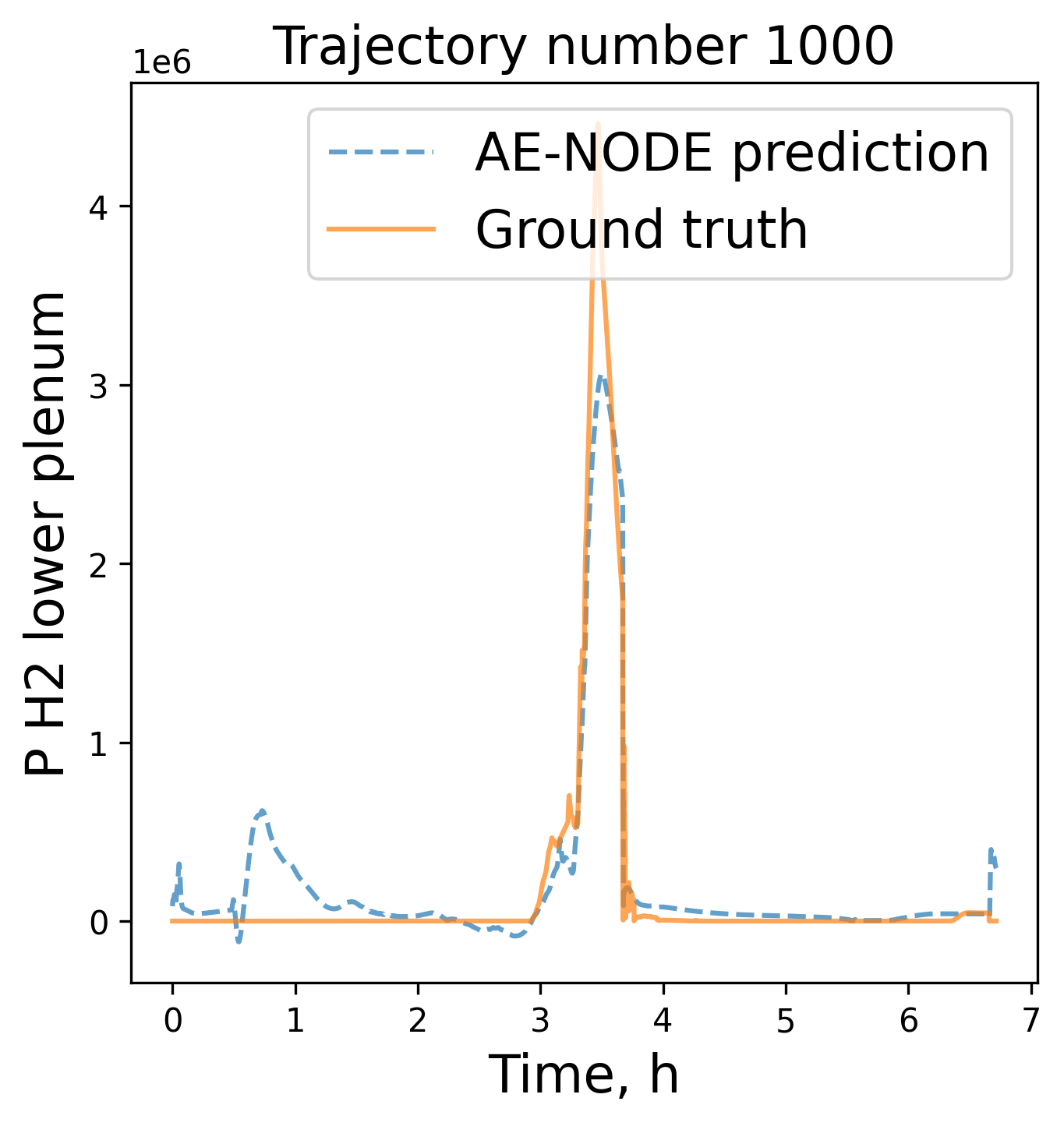}
    \end{subfigure}
    \hfill
    \begin{subfigure}[b]{0.19\textwidth}
        \includegraphics[width=\textwidth]{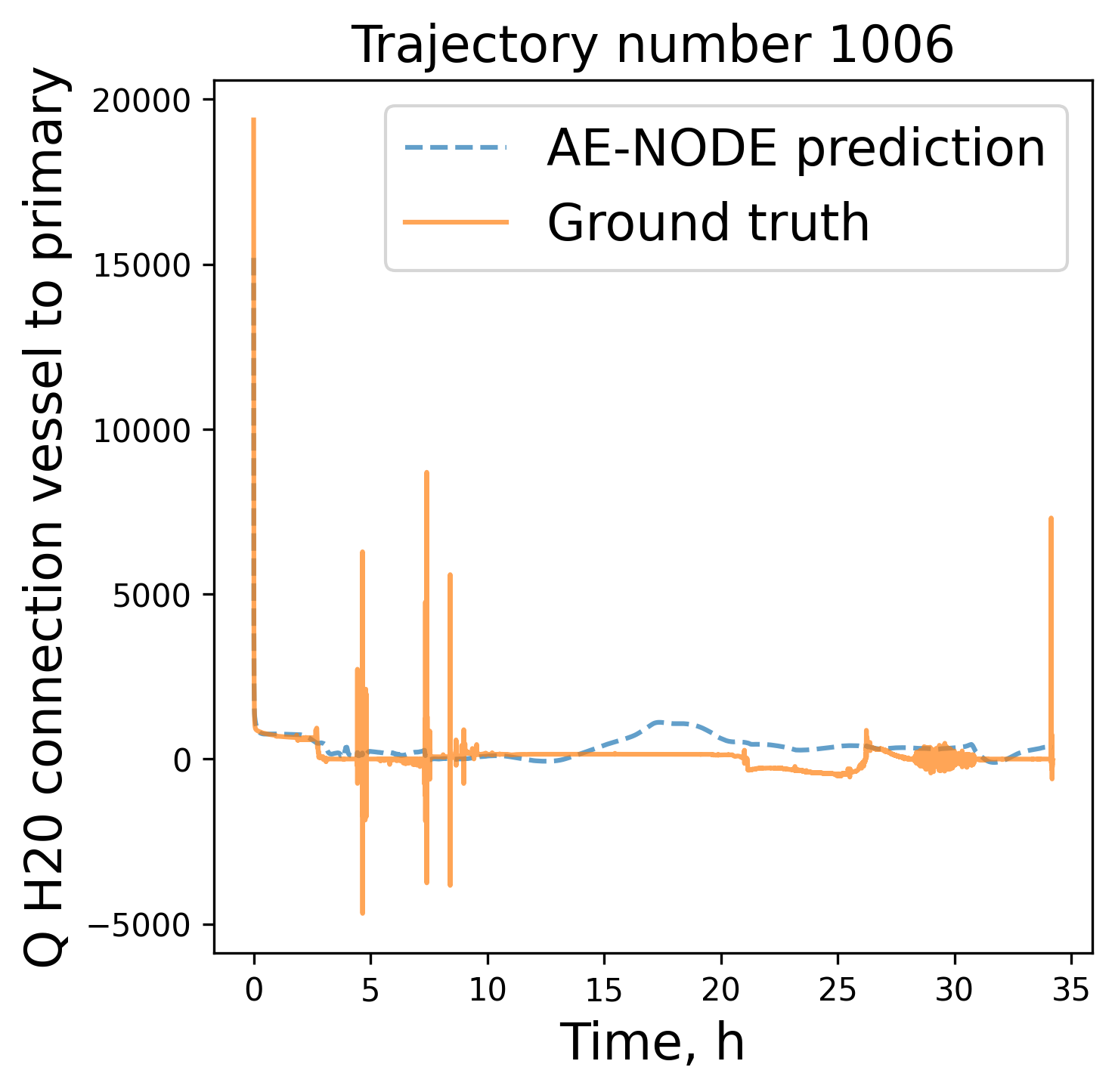}
    \end{subfigure}
    \hfill
    \begin{subfigure}[b]{0.19\textwidth}
        \includegraphics[width=\textwidth]{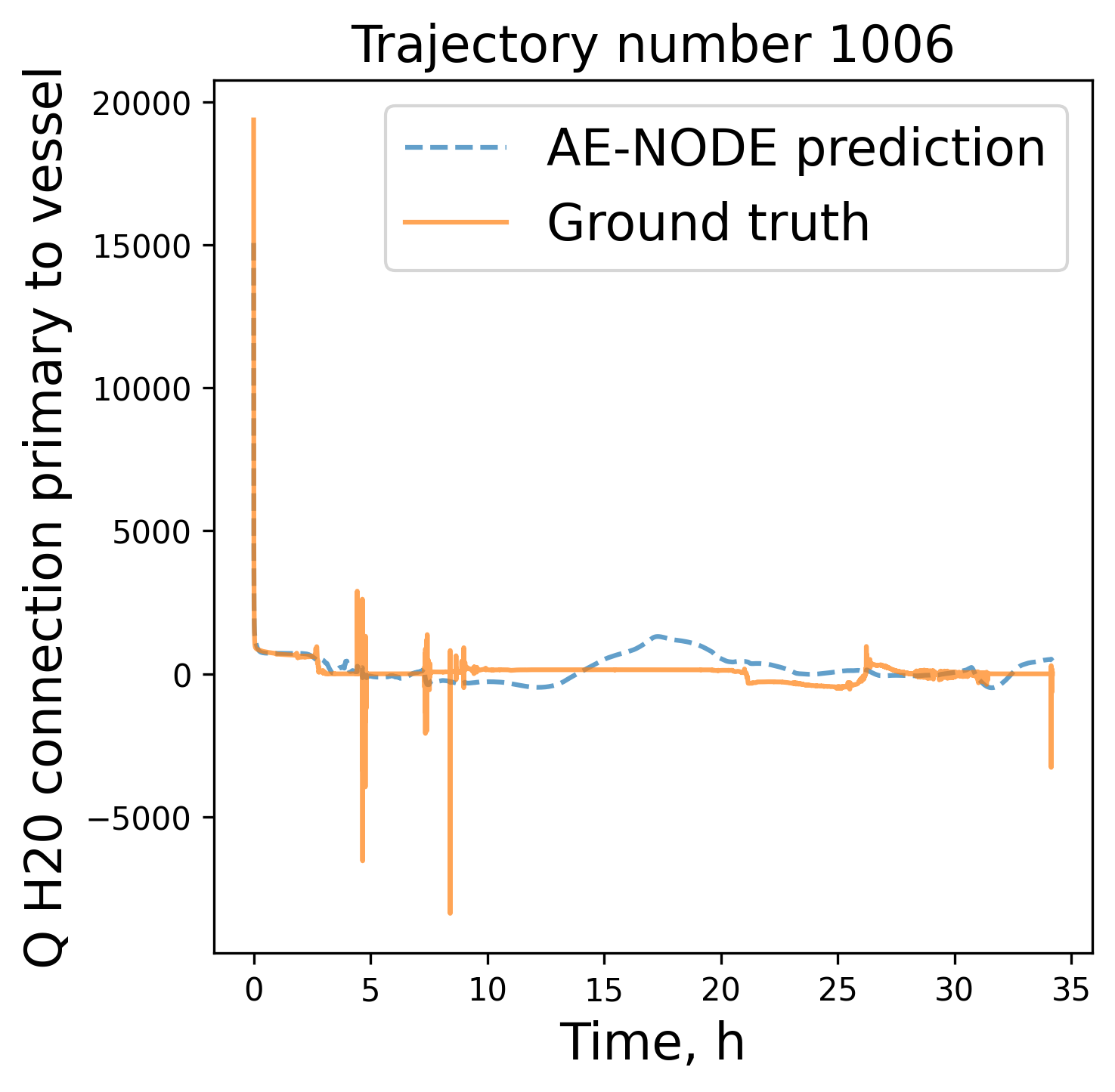}
    \end{subfigure}
    \hfill
    \begin{subfigure}[b]{0.19\textwidth}
        \includegraphics[width=\textwidth]{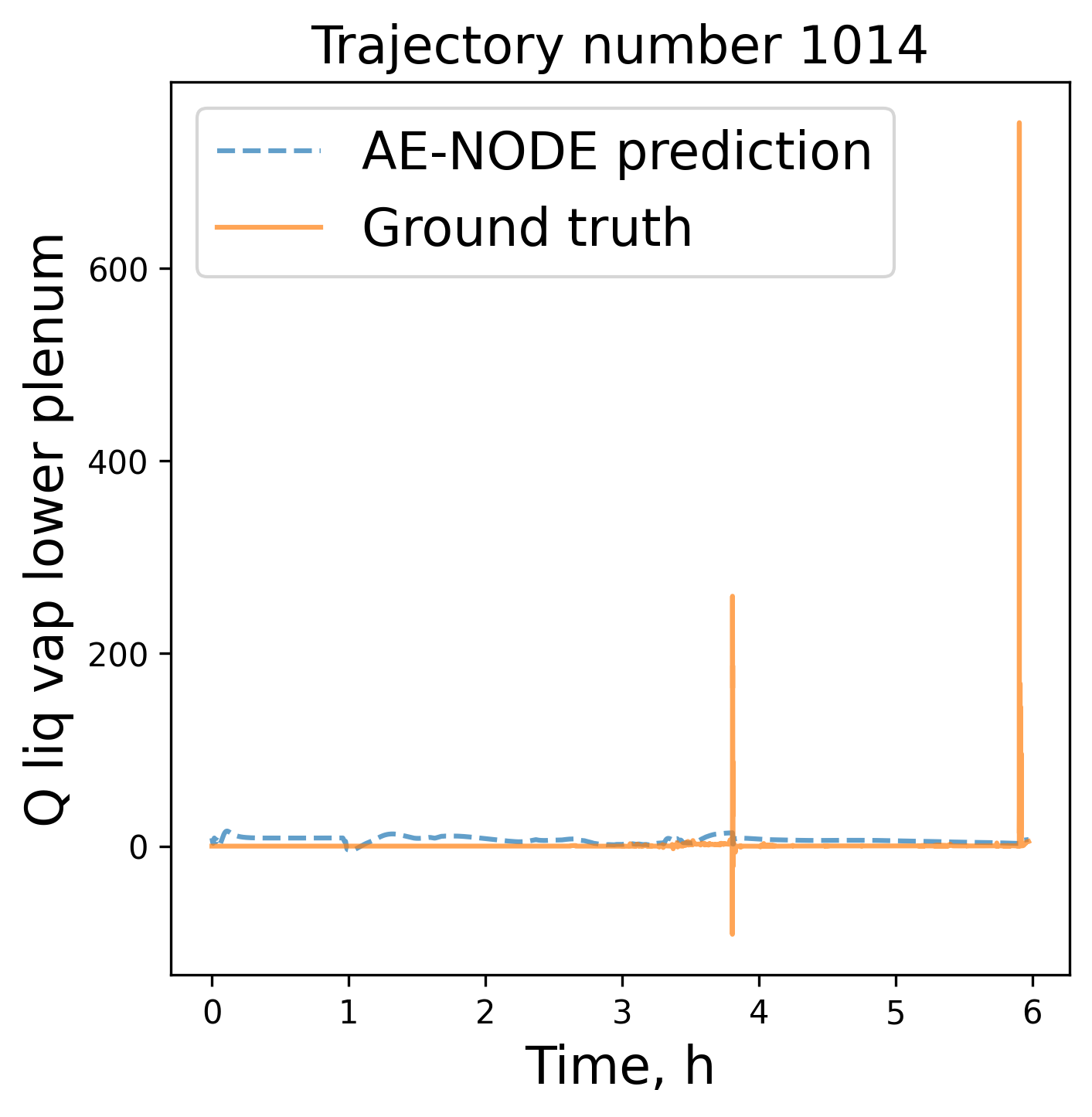}
    \end{subfigure}

    \vspace{0.5em}

    \begin{subfigure}[b]{0.19\textwidth}
        \includegraphics[width=\textwidth]{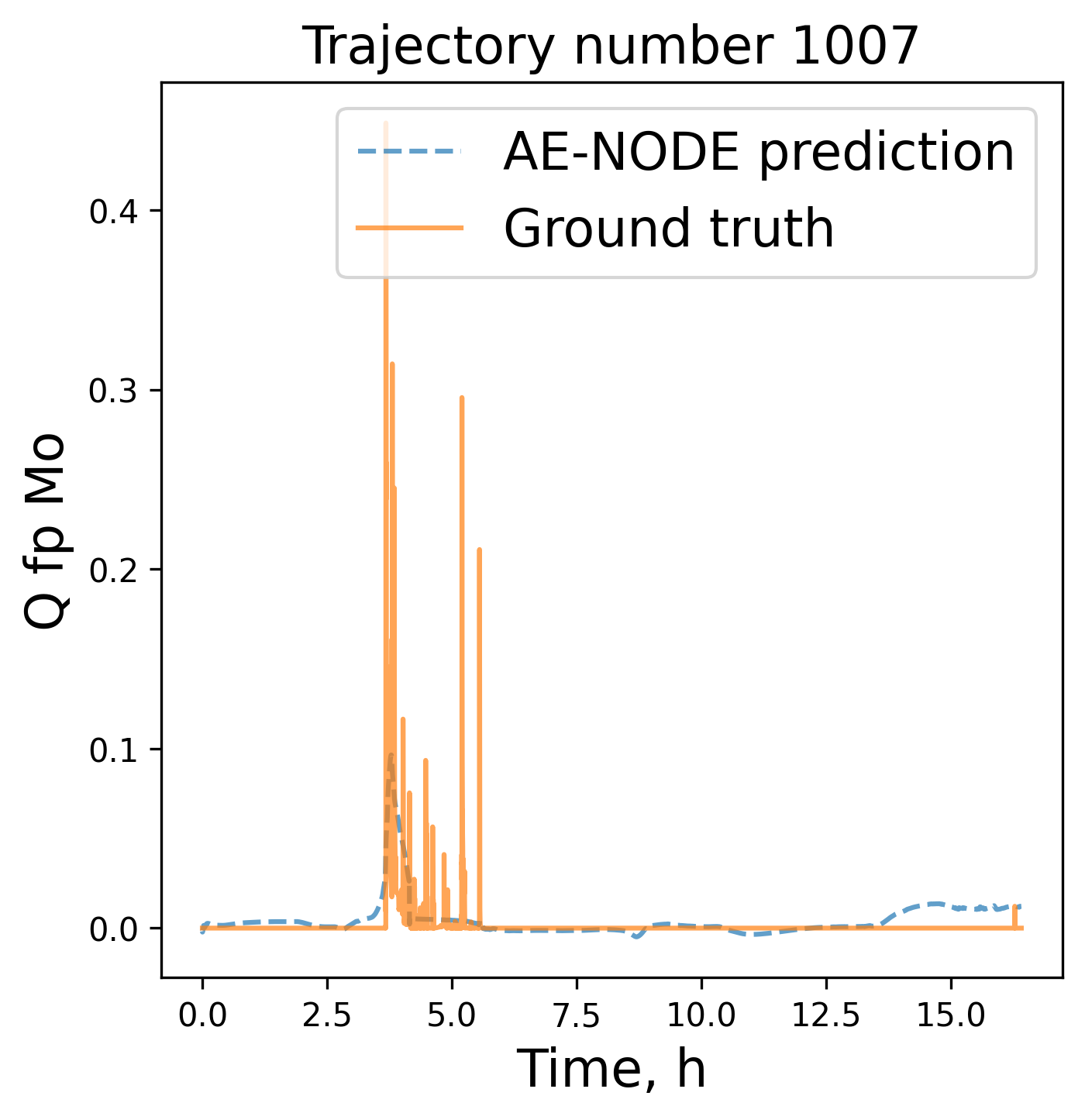}
    \end{subfigure}
    \hfill
    \begin{subfigure}[b]{0.19\textwidth}
        \includegraphics[width=\textwidth]{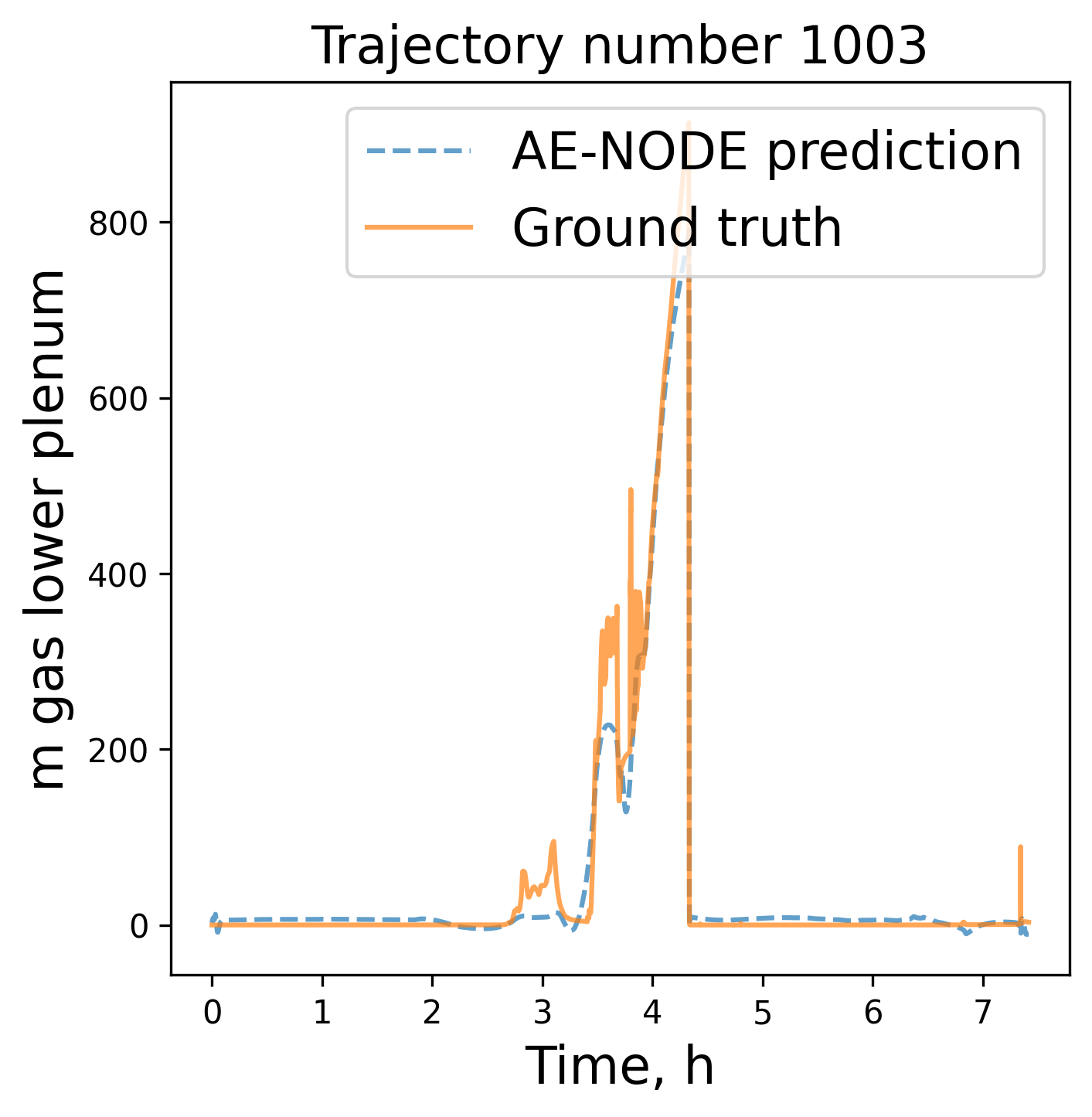}
    \end{subfigure}
    \hfill
    \begin{subfigure}[b]{0.19\textwidth}
        \includegraphics[width=\textwidth]{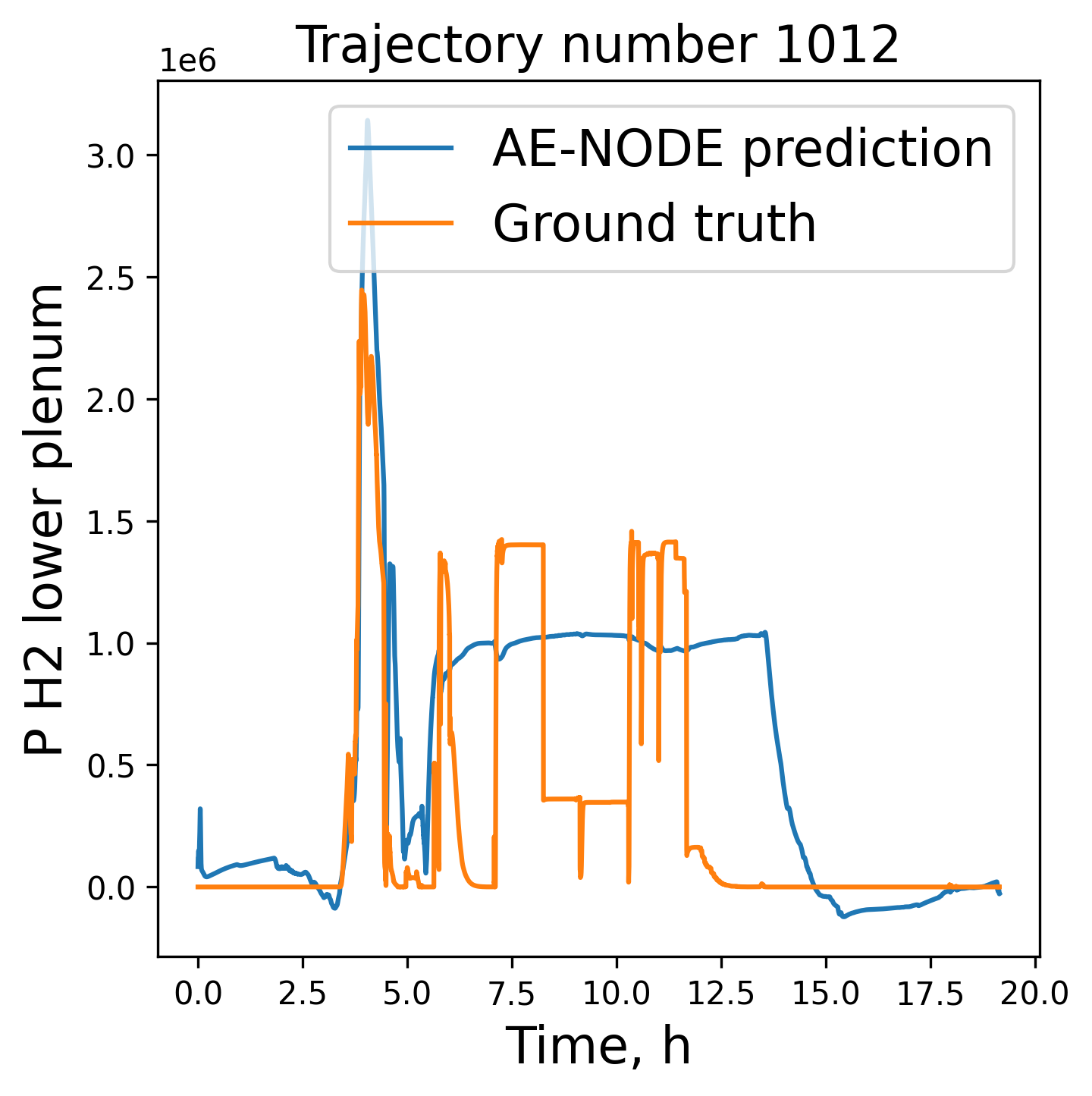}
    \end{subfigure}
    \hfill
    \begin{subfigure}[b]{0.19\textwidth}
        \includegraphics[width=\textwidth]{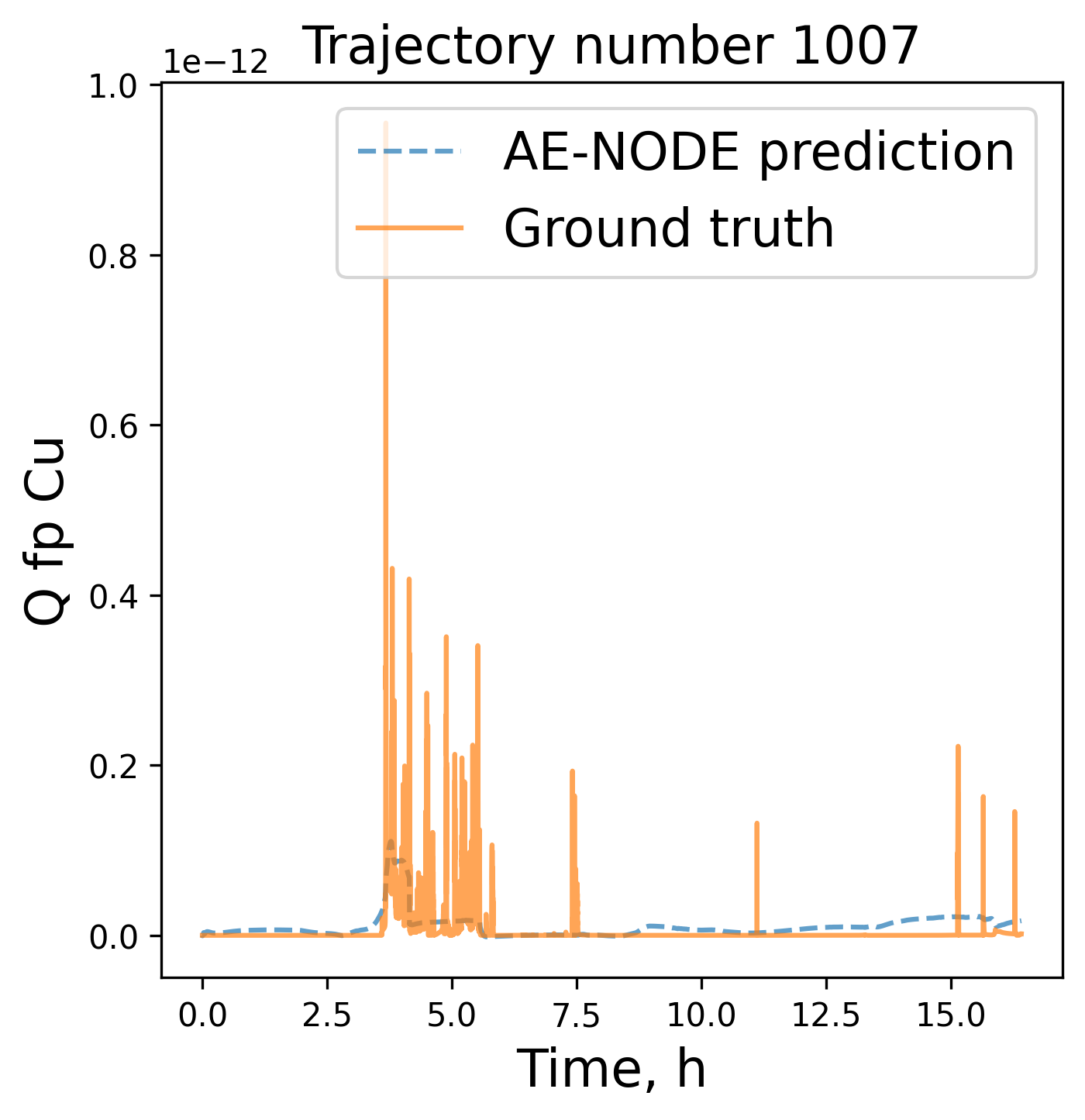}
    \end{subfigure}
    \hfill
    \begin{subfigure}[b]{0.19\textwidth}
        \includegraphics[width=\textwidth]{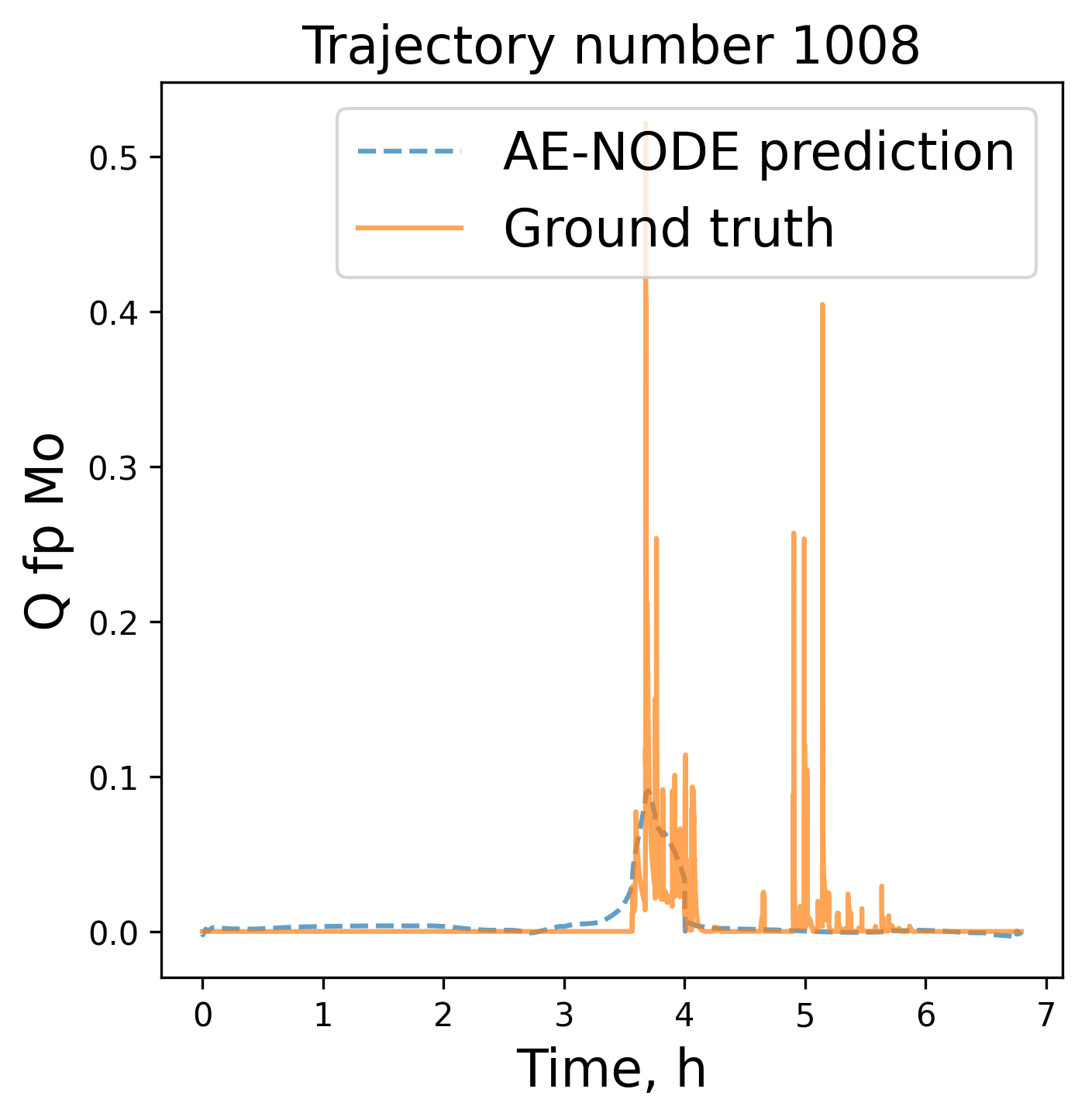}
    \end{subfigure}

    \caption{AE-NODE predictions vs.\ ground truth for scalar variables with large $\text{RMSE}_{mean}$ and $\text{RMSE}_{std}$ for SBO simulations.}
    \label{fig:near_zero_variables_SBO}
\end{figure}

\begin{figure}[htbp]
    \centering
    \begin{subfigure}[b]{0.19\textwidth}
        \includegraphics[width=\textwidth]{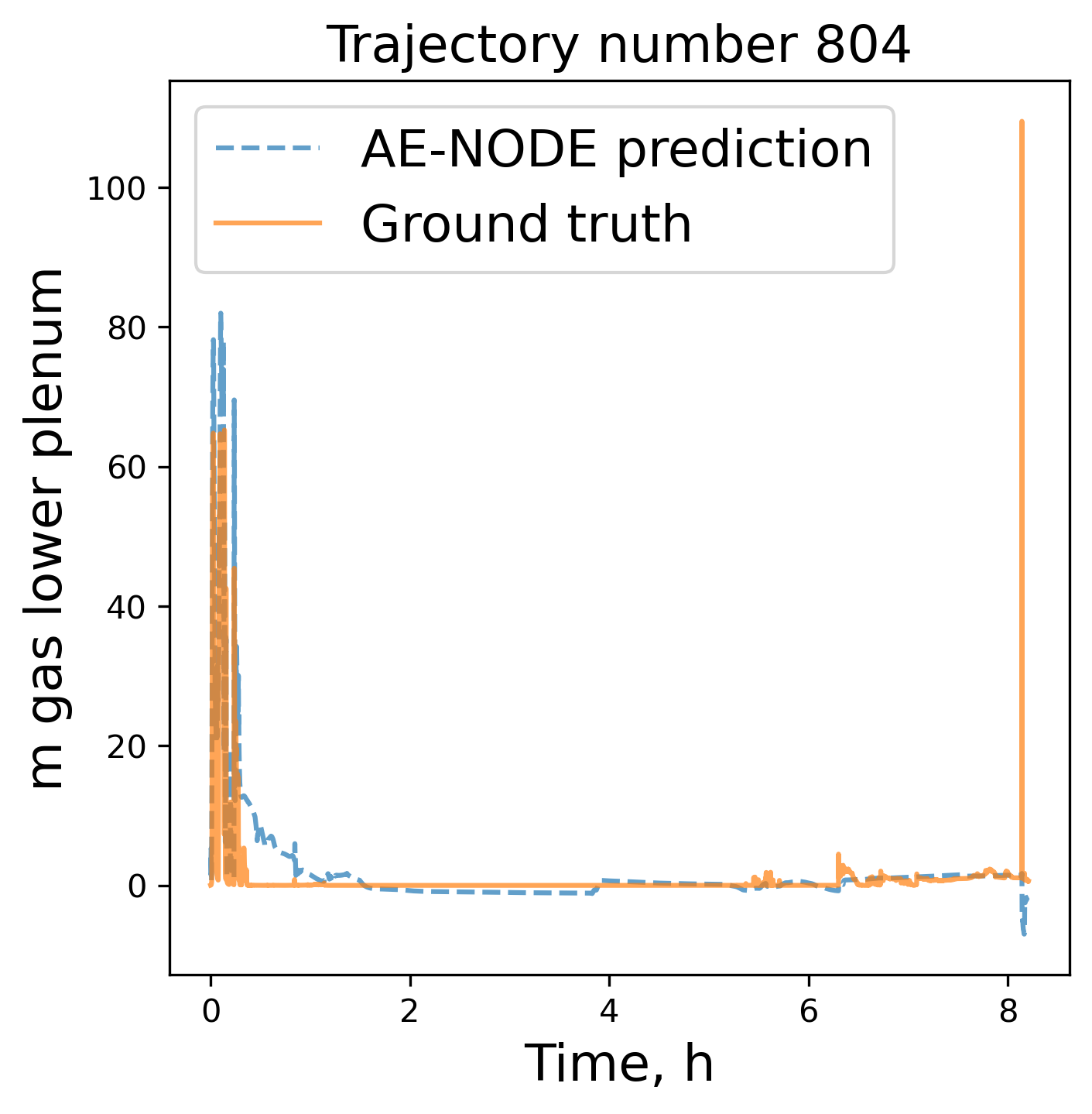}
    \end{subfigure}
    \hfill
    \begin{subfigure}[b]{0.19\textwidth}
        \includegraphics[width=\textwidth]{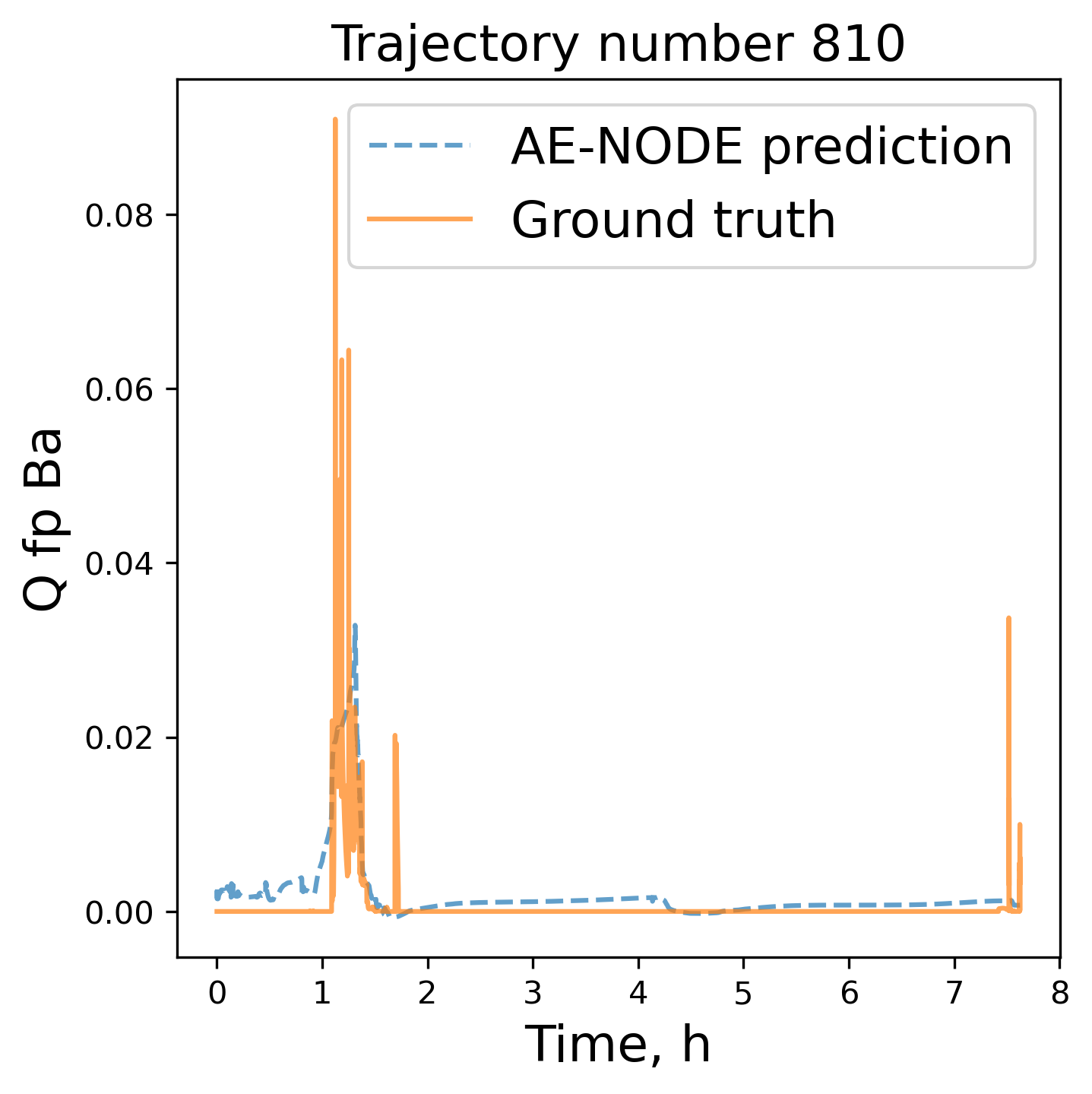}
    \end{subfigure}
    \hfill
    \begin{subfigure}[b]{0.19\textwidth}
        \includegraphics[width=\textwidth]{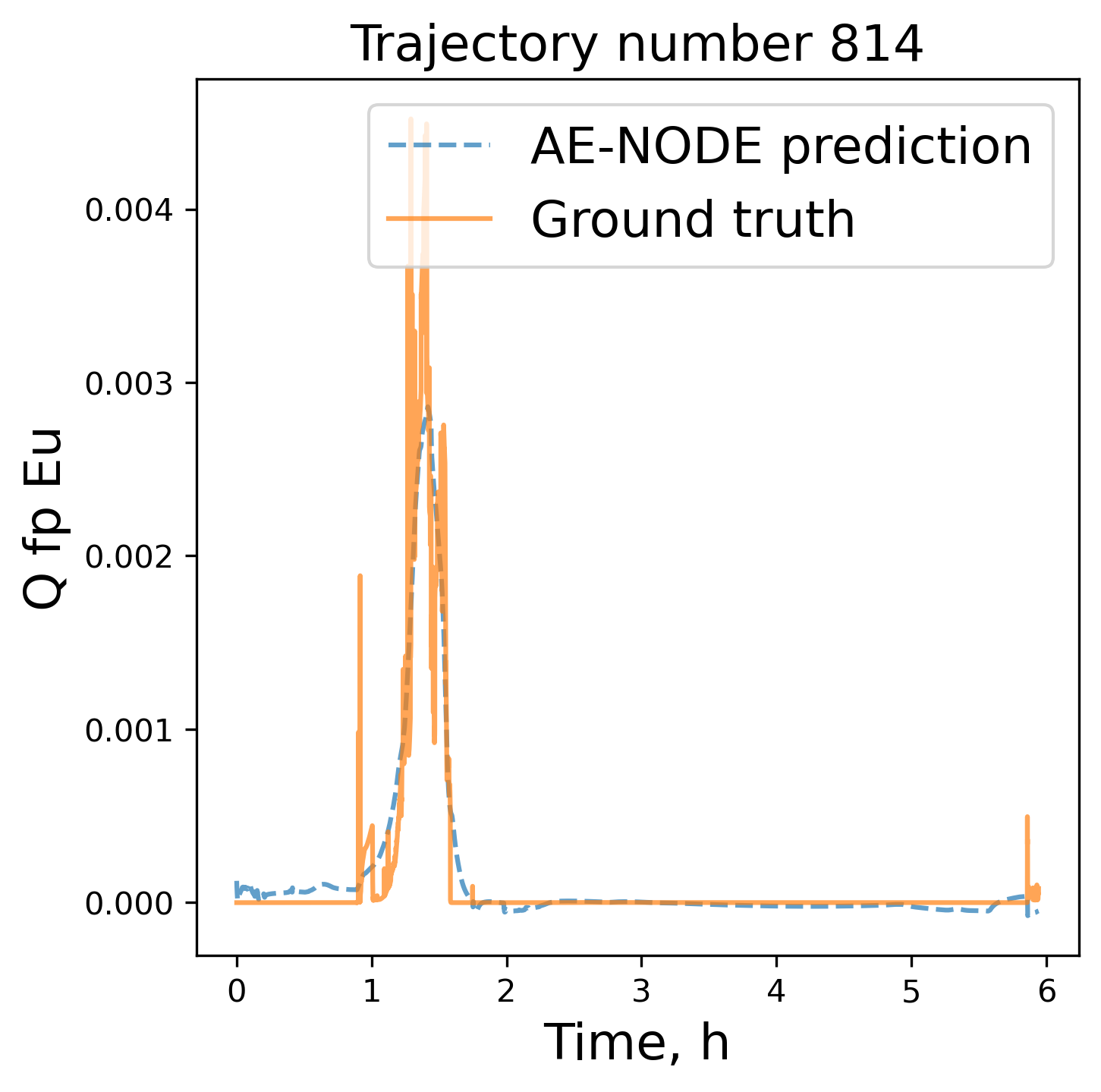}
    \end{subfigure}
    \hfill
    \begin{subfigure}[b]{0.19\textwidth}
        \includegraphics[width=\textwidth]{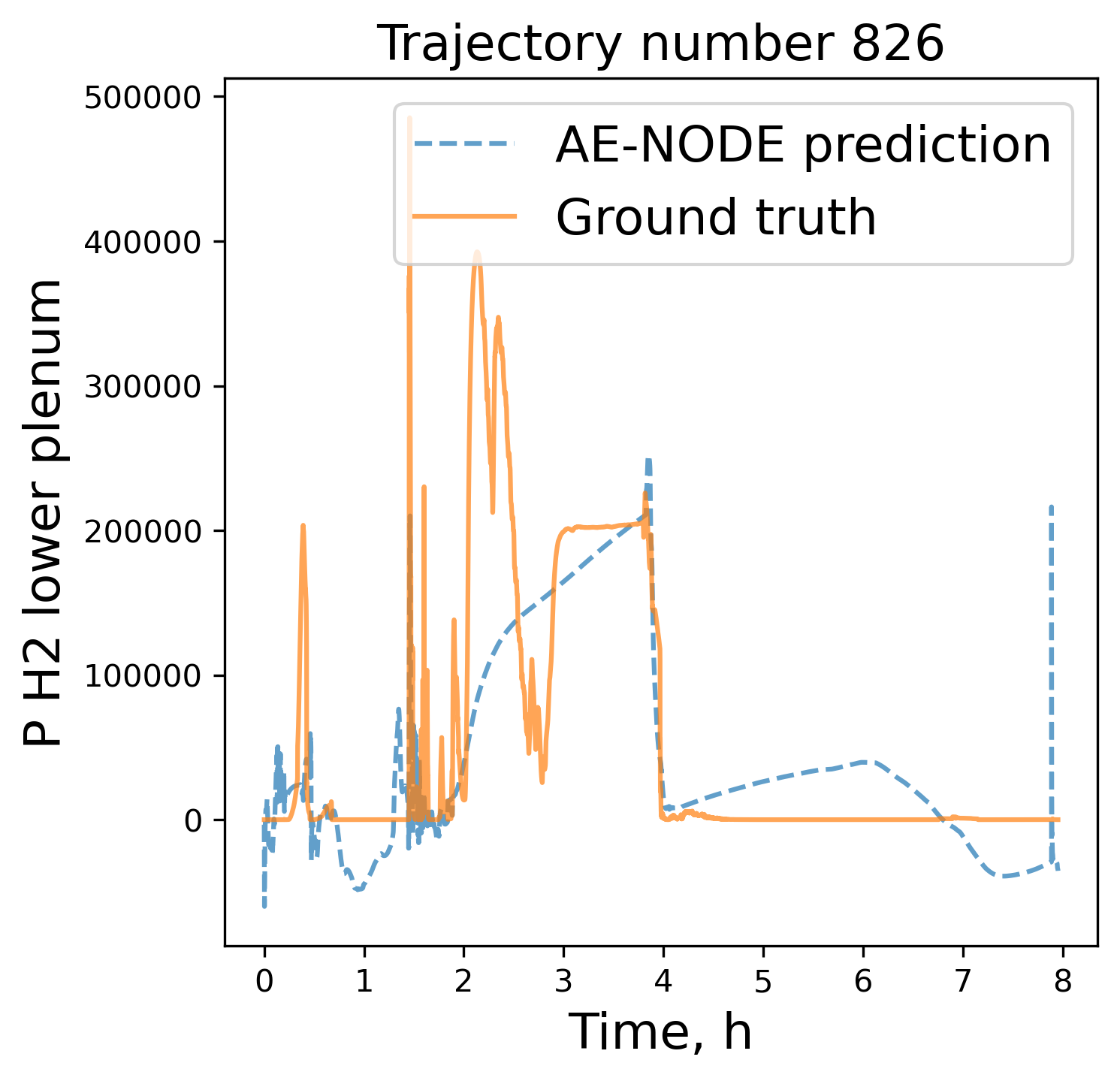}
    \end{subfigure}
    \hfill
    \begin{subfigure}[b]{0.19\textwidth}
        \includegraphics[width=\textwidth]{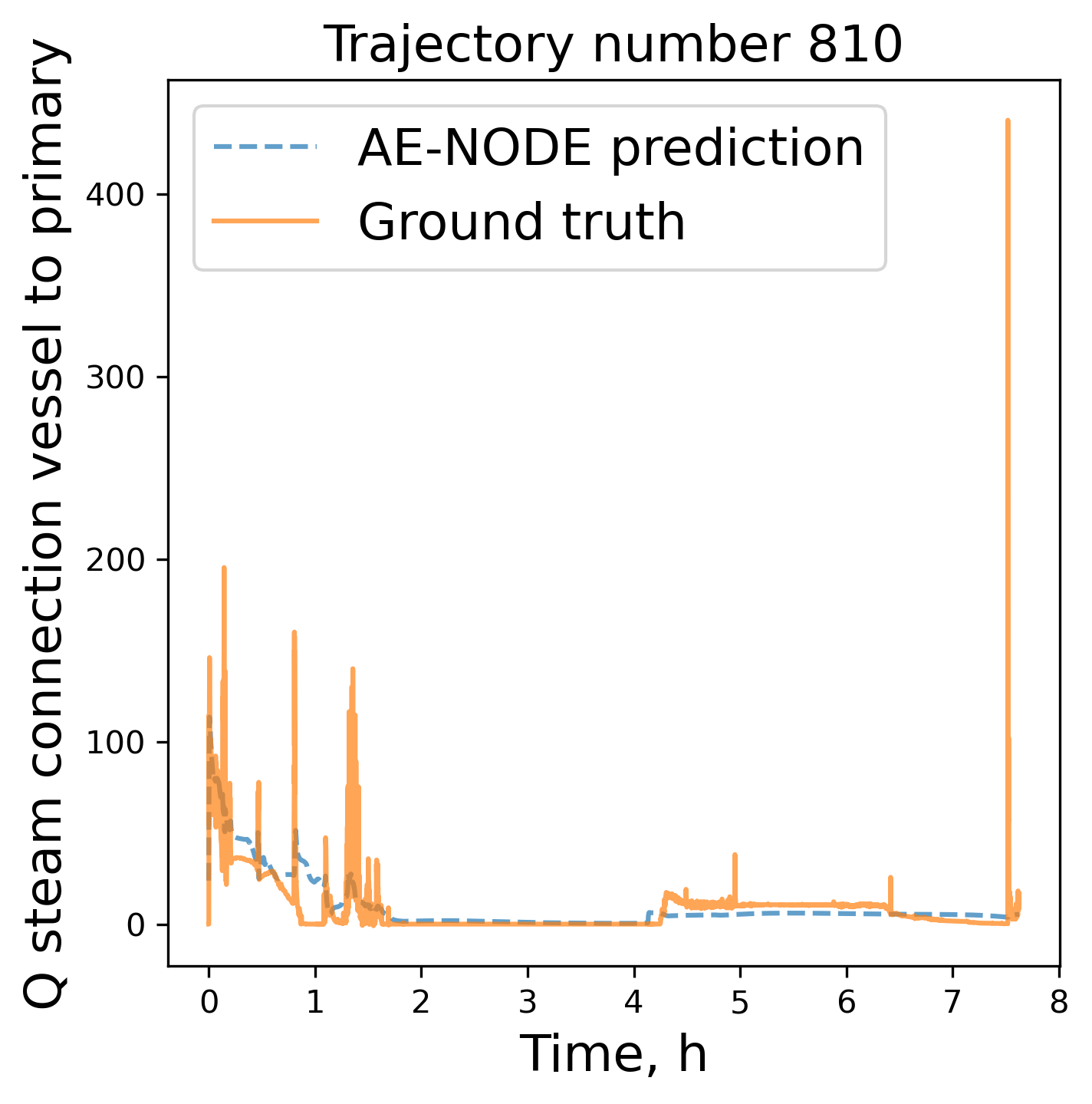}
    \end{subfigure}

    \vspace{0.5em}

    \begin{subfigure}[b]{0.19\textwidth}
        \includegraphics[width=\textwidth]{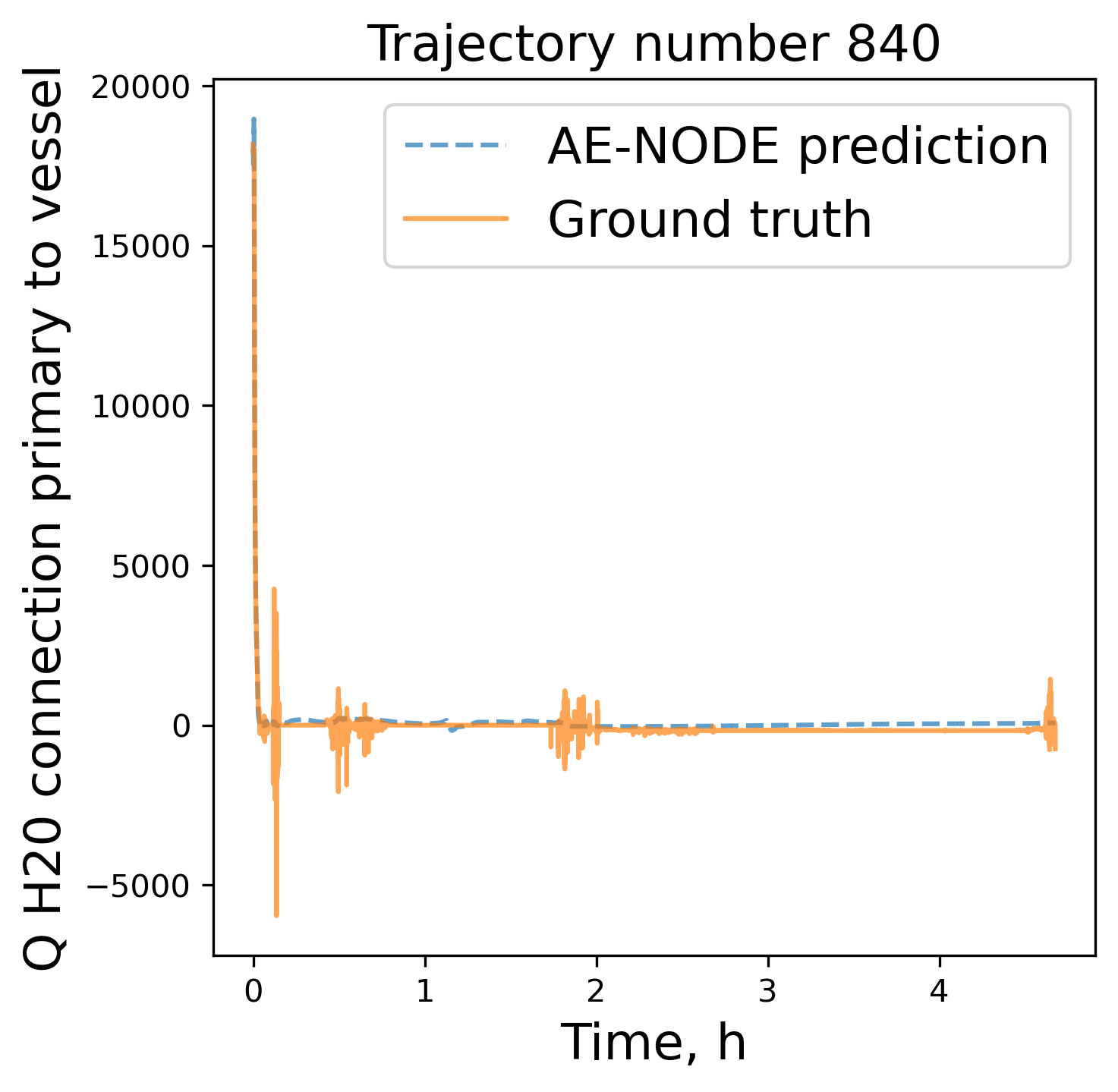}
    \end{subfigure}
    \hfill
    \begin{subfigure}[b]{0.19\textwidth}
        \includegraphics[width=\textwidth]{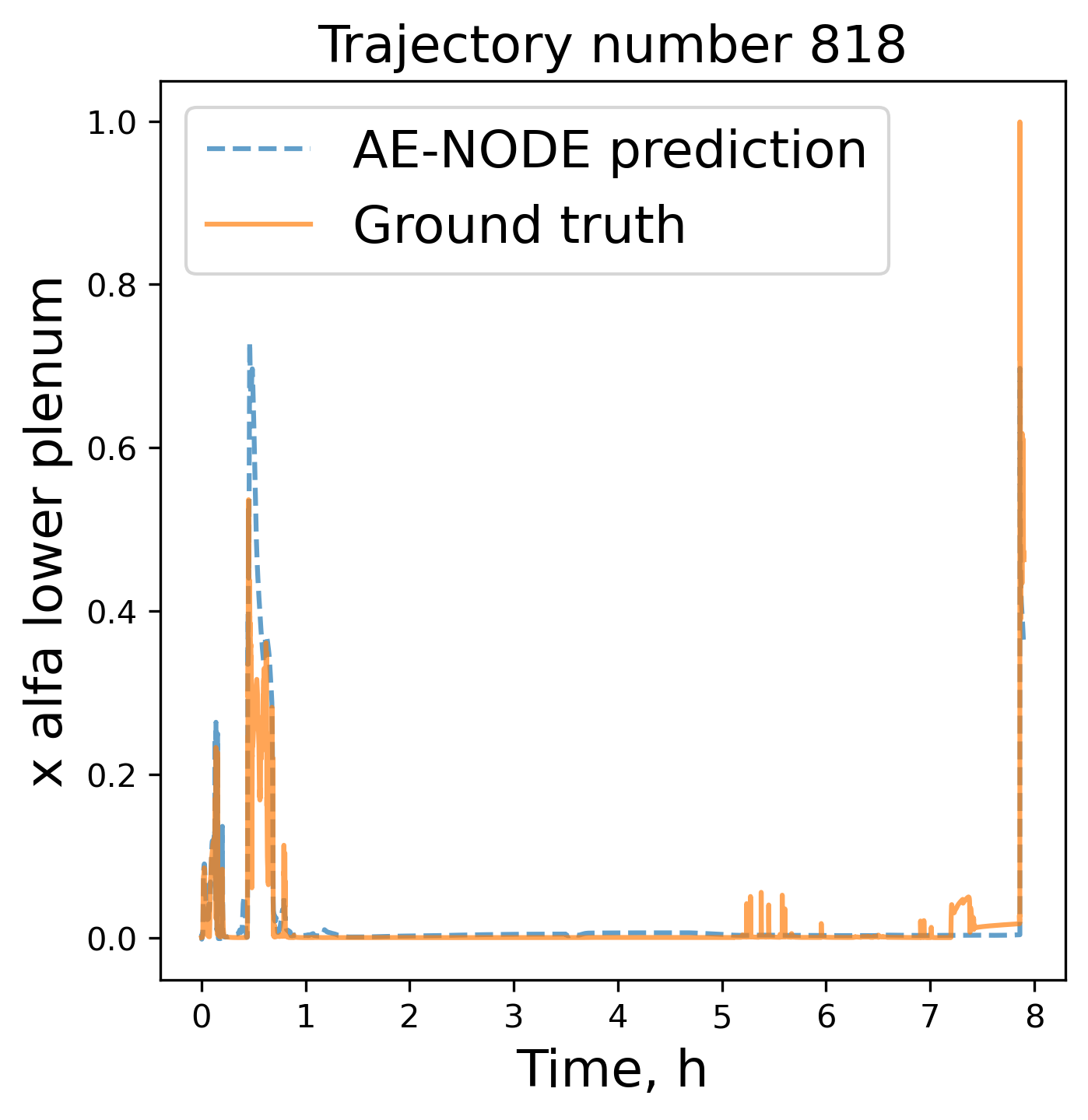}
    \end{subfigure}
    \hfill
    \begin{subfigure}[b]{0.19\textwidth}
        \includegraphics[width=\textwidth]{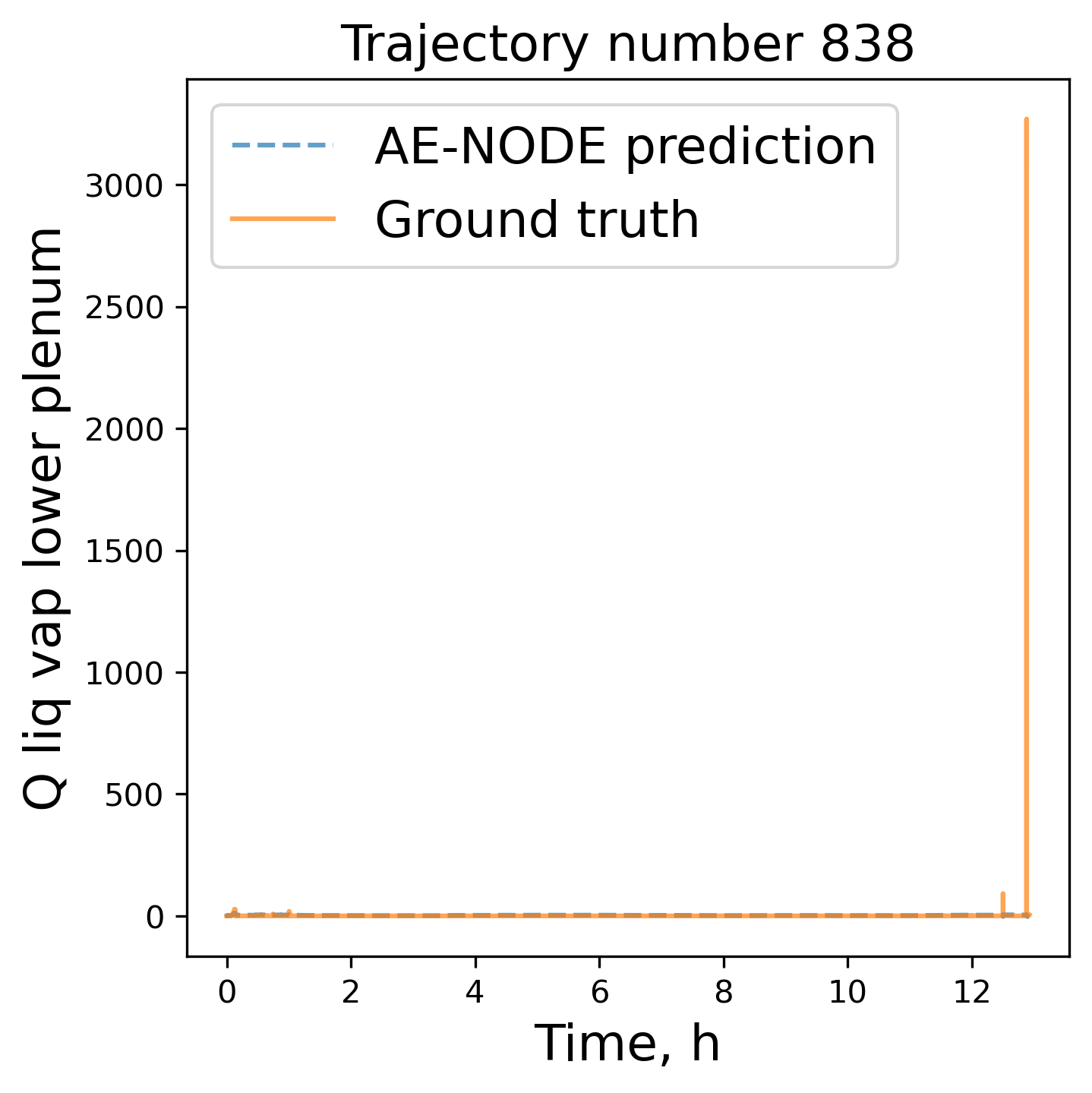}
    \end{subfigure}
    \hfill
    \begin{subfigure}[b]{0.19\textwidth}
        \includegraphics[width=\textwidth]{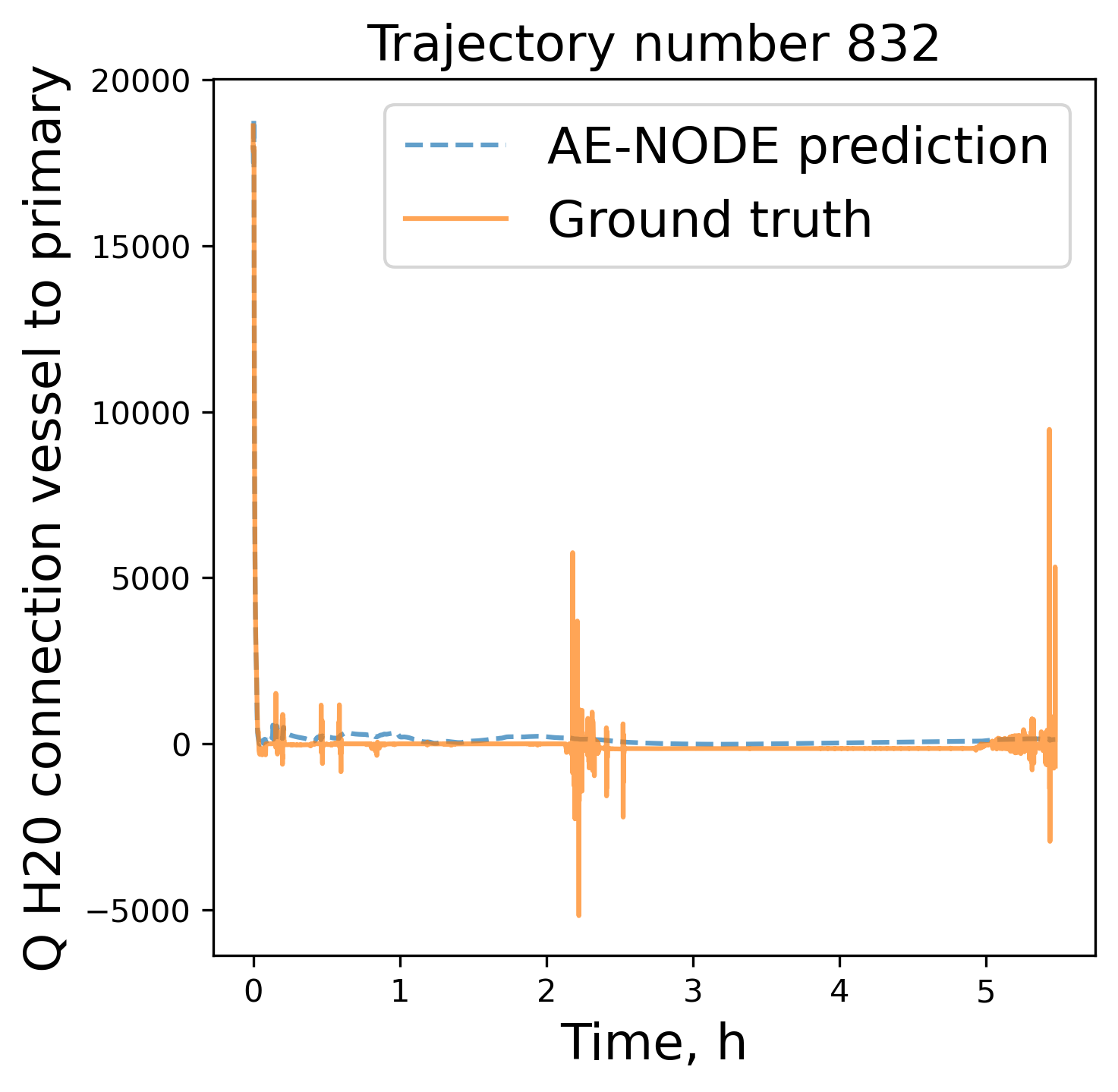}
    \end{subfigure}
    \hfill
    \begin{subfigure}[b]{0.19\textwidth}
        \includegraphics[width=\textwidth]{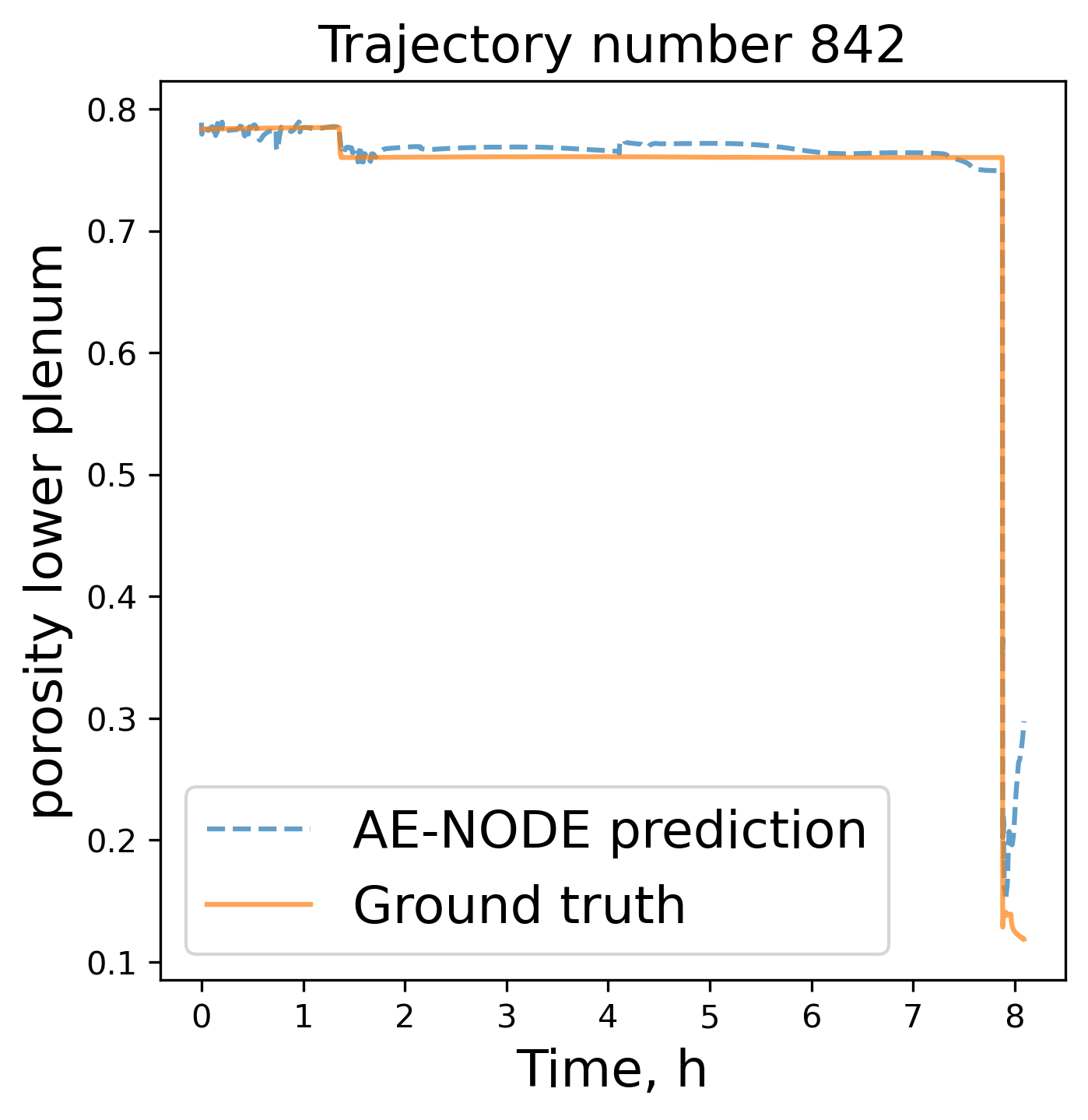}
    \end{subfigure}

    \caption{AE-NODE predictions vs.\ ground truth for scalar variables with large $\text{RMSE}_{mean}$ and $\text{RMSE}_{std}$ for LOCA simulations.}
    \label{fig:near_zero_variables_LOCA}
\end{figure}

\begin{figure}[htbp]
    \centering
    \begin{subfigure}[b]{0.48\textwidth}
        \includegraphics[width=\textwidth]{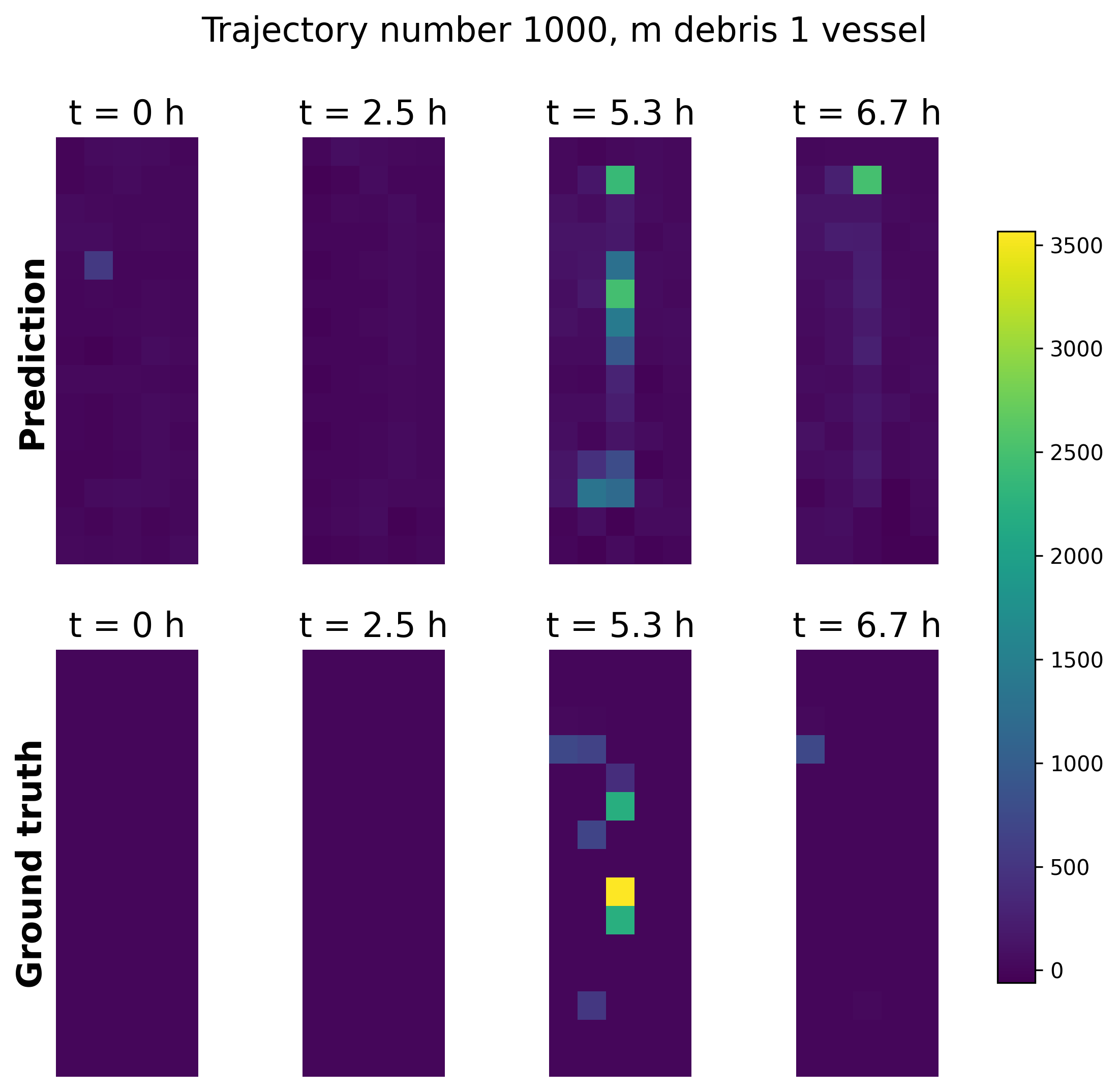}
        \caption{Debris 1 in the vessel}
        \label{fig:1000_m_debris_1_vessel}
    \end{subfigure}
    \hfill
    \begin{subfigure}[b]{0.48\textwidth}
        \includegraphics[width=\textwidth]{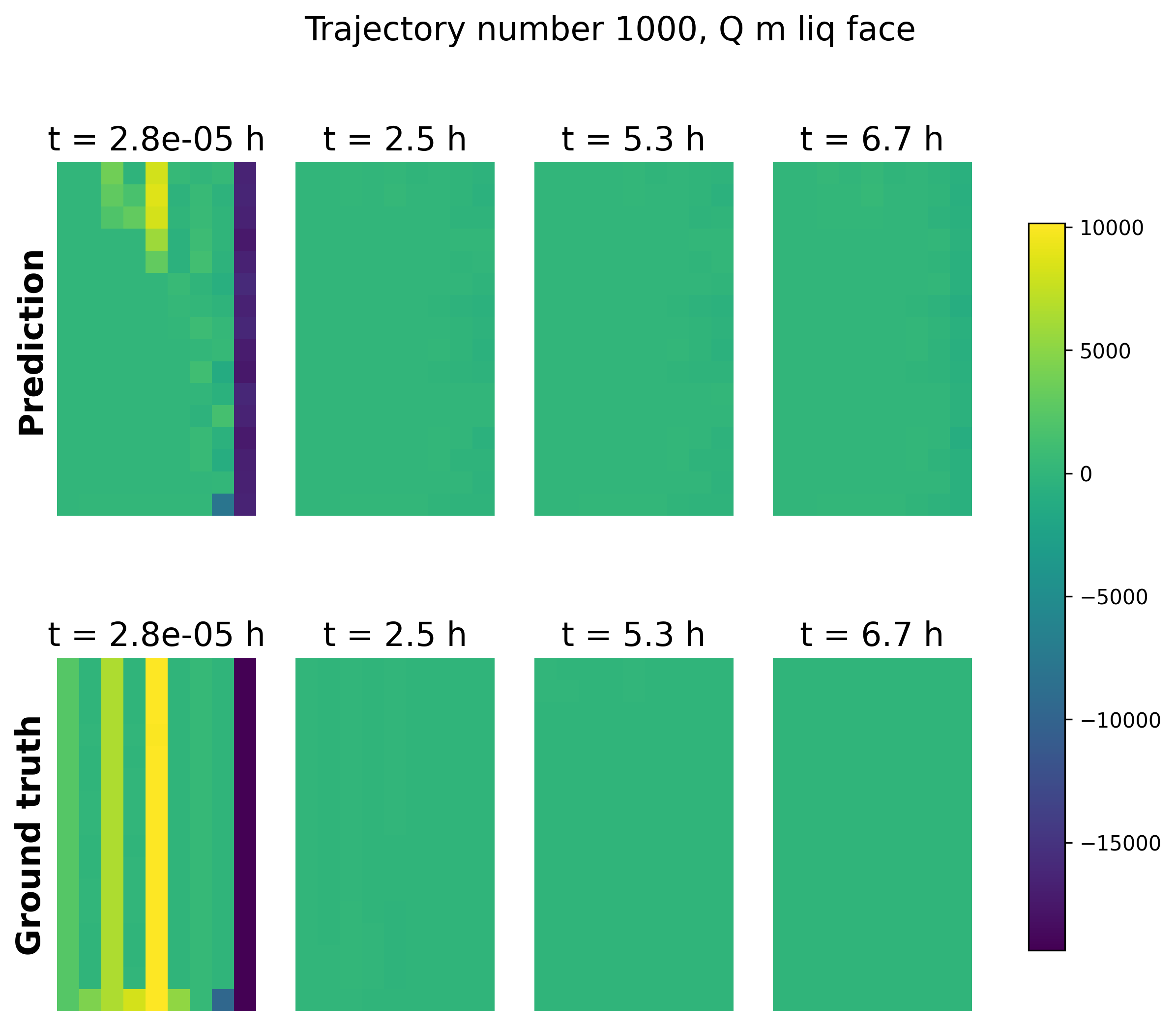}
        \caption{Q m liq face in the vessel}
        \label{fig:1000_Q_m_liq_face}
    \end{subfigure}
    \caption{AE-NODE predictions vs.\ ground truth at $4$ different time steps of debris 1 in the vessel and Q m liq face.}
    \label{fig:debris_and_q_liq}
\end{figure}

\begin{figure}[htbp]
    \centering
    \begin{subfigure}[b]{0.19\textwidth}
        \includegraphics[width=\textwidth]{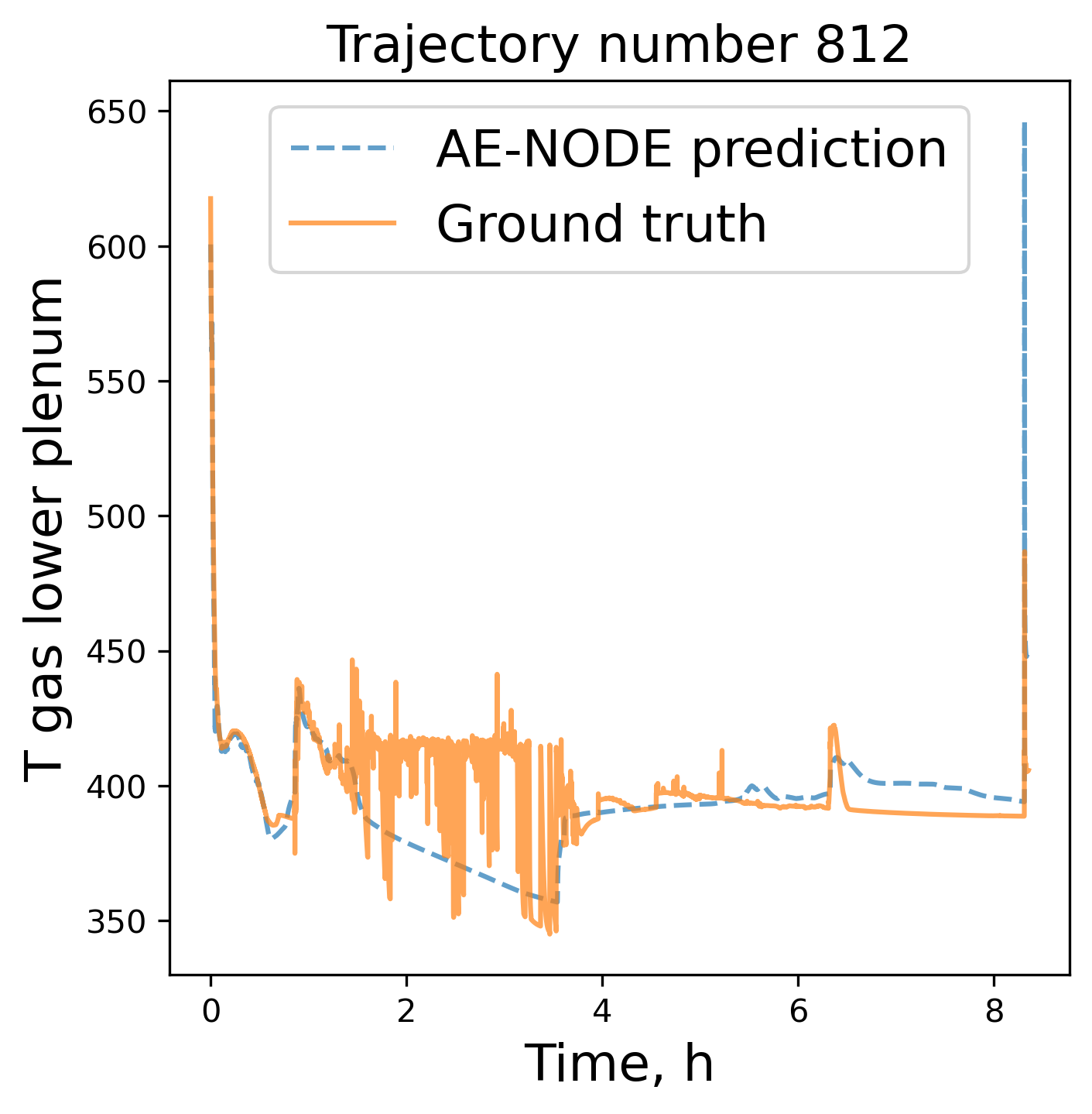}
    \end{subfigure}
    \hfill
    \begin{subfigure}[b]{0.19\textwidth}
        \includegraphics[width=\textwidth]{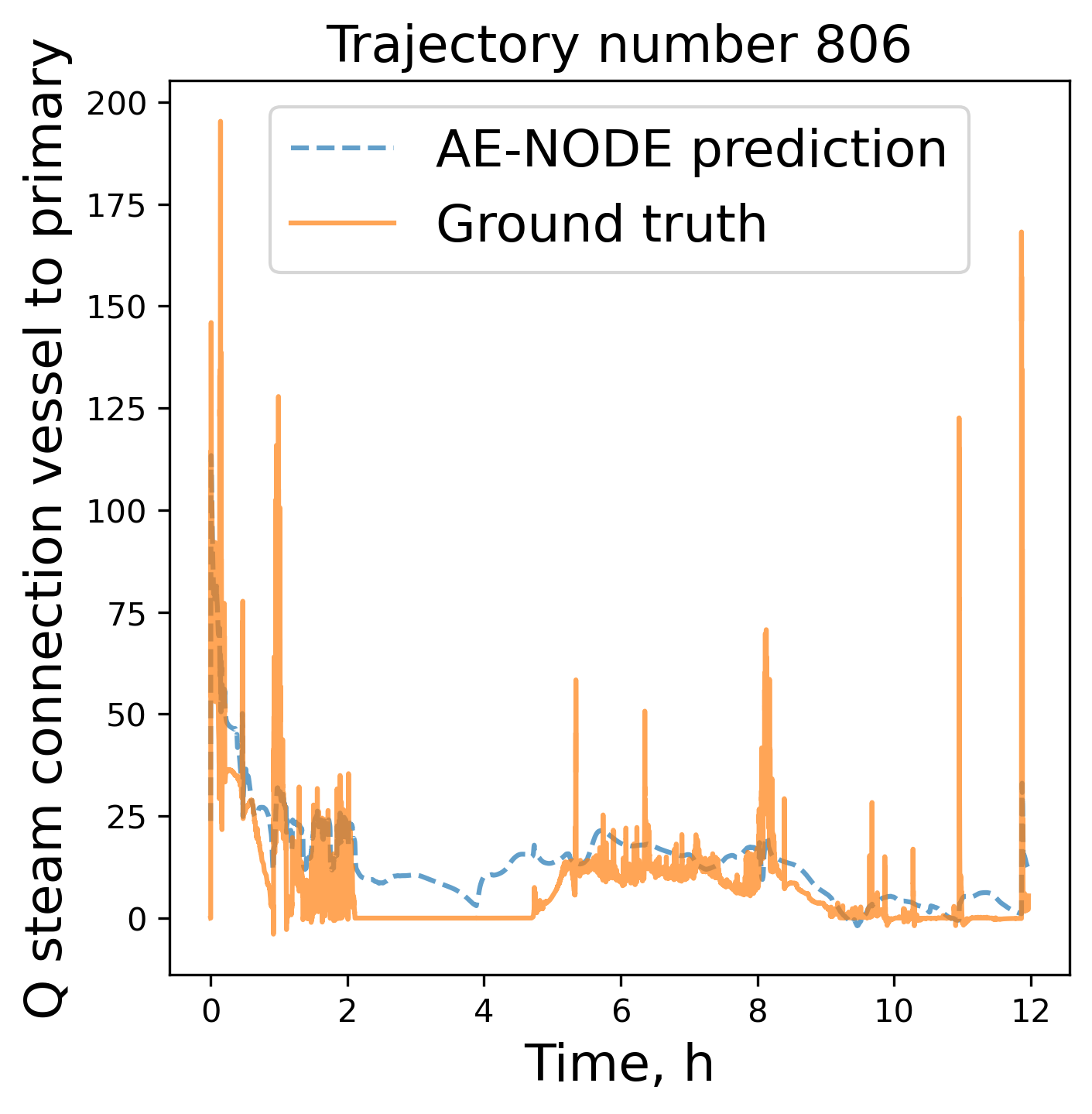}
    \end{subfigure}
    \hfill
    \begin{subfigure}[b]{0.19\textwidth}
        \includegraphics[width=\textwidth]{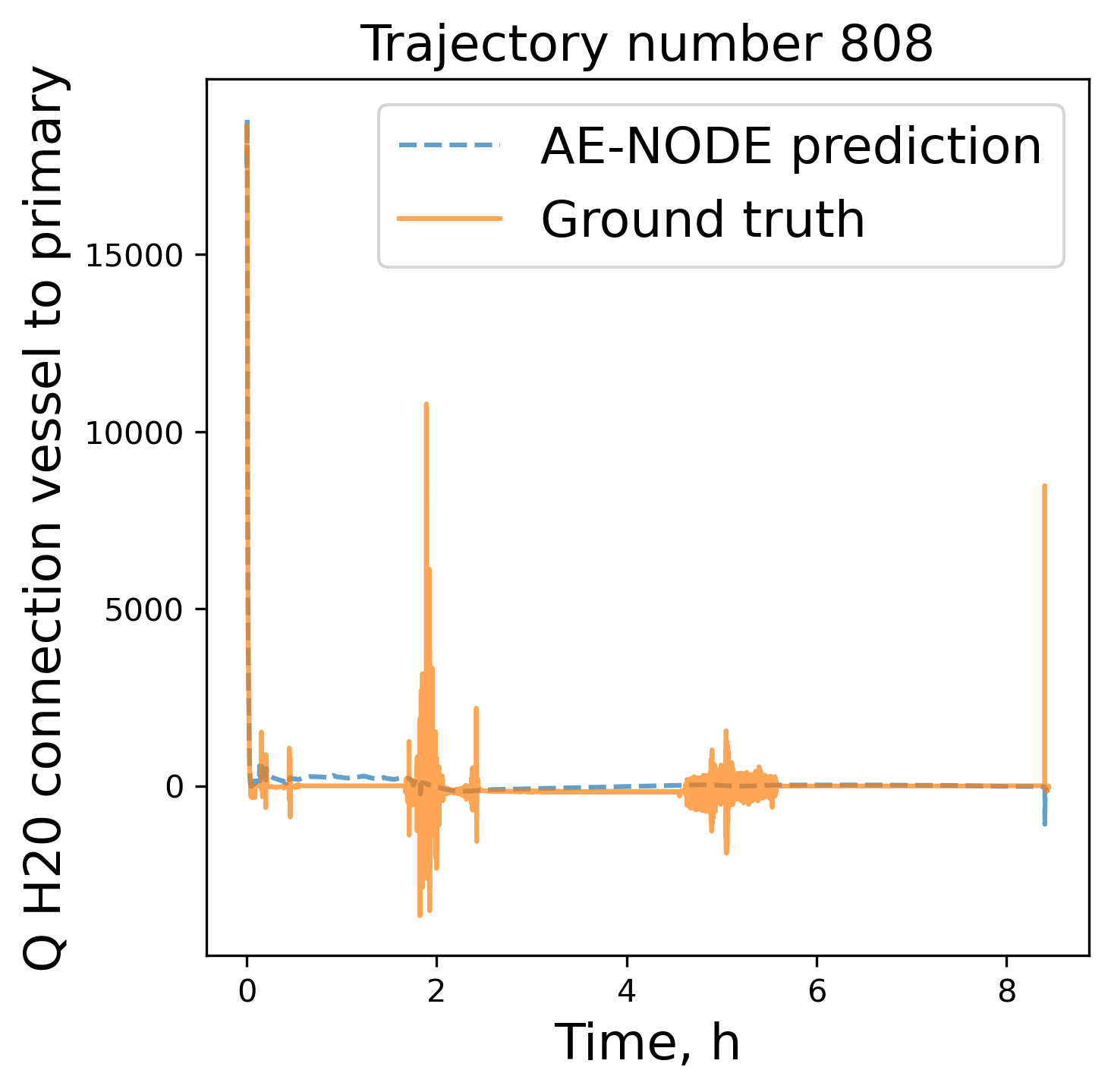}
    \end{subfigure}
    \hfill
    \begin{subfigure}[b]{0.19\textwidth}
        \includegraphics[width=\textwidth]{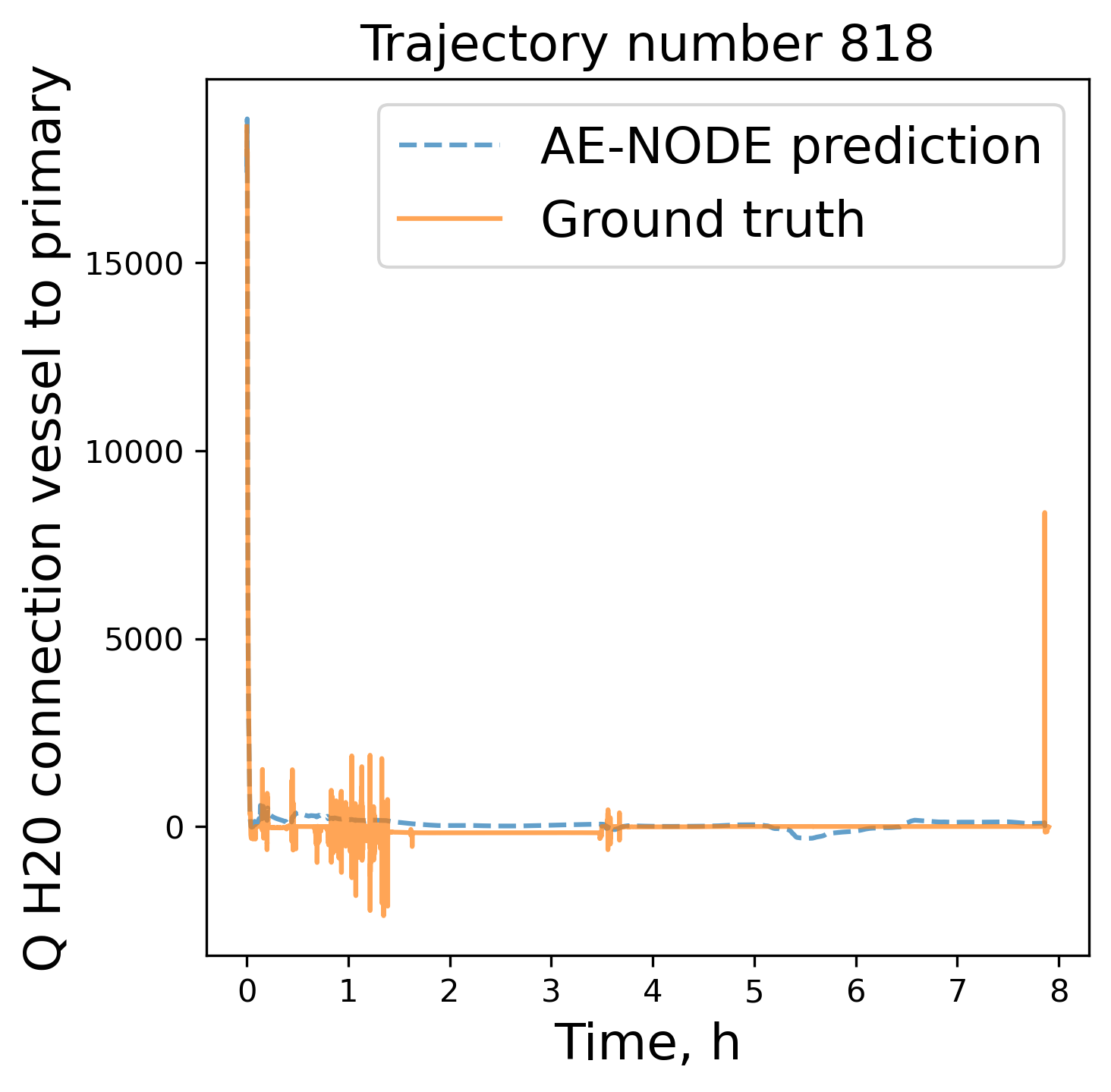}
    \end{subfigure}
    \hfill
    \begin{subfigure}[b]{0.19\textwidth}
        \includegraphics[width=\textwidth]{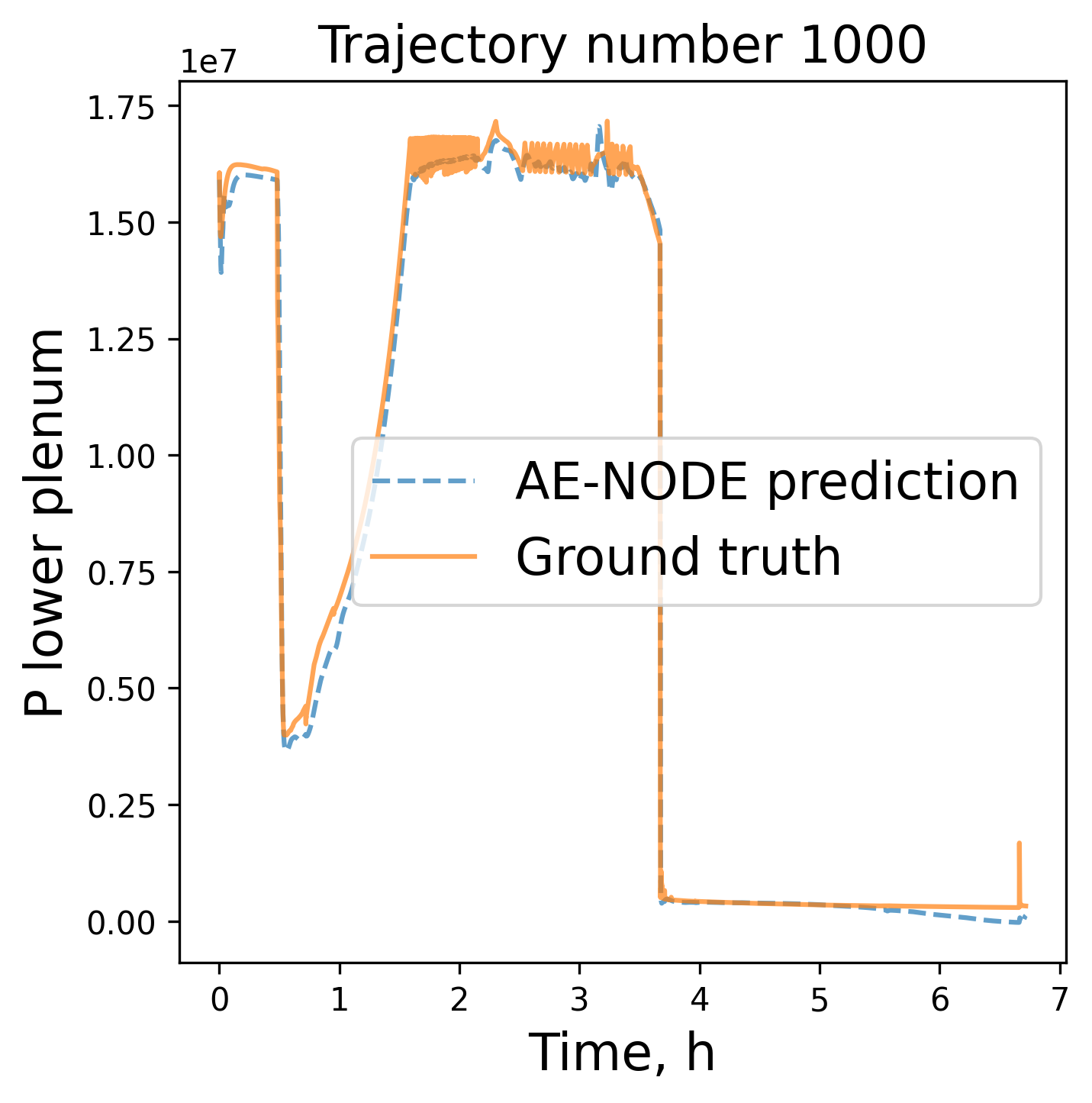}
    \end{subfigure}
    \caption{Challenging trajectory examples showing AE-NODE predictions vs.\ ground truth for trajectories 812, 806, 808, 818, and 1000. Beside the sharp jumps, there are also time intervals where the ground truth exhibits sharp and quasi-oscillating behaviour, as in time intervals $[1.5-3.8]h$, $[5.8-7.5]h$, $[4.5-5.8]h$, $[0.8-1.5]h$ and $[1.8-3]h$ from left to right.}
    \label{fig:challenging_examples}
\end{figure}

In Figure \ref{fig:near_zero_variables_SBO}, \ref{fig:near_zero_variables_LOCA} and \ref{fig:debris_and_q_liq} we show some predictions for variables that perform badly according to the metrics shown in Section \ref{sec:results}. In Figure \ref{fig:challenging_examples} we show some challenging (difficult to approximate) ASTEC variables that display sharp and quasi-periodic behaviour.


\begin{figure}[h]
  \centering
  \includegraphics[width=1.0\textwidth]{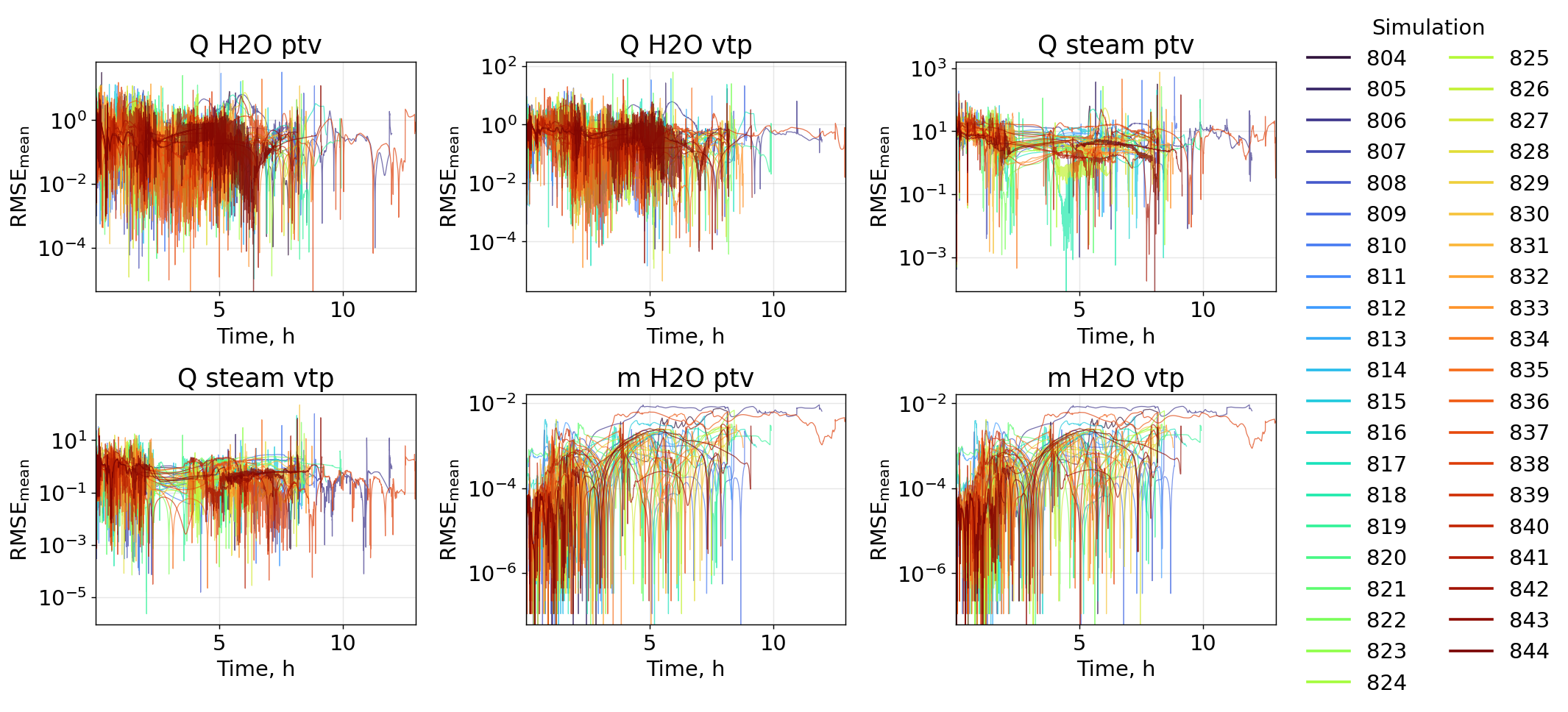}
  \caption{RMSE$_{mean}$ per time step for all the LOCA testing trajectories for the boundary variables $s_{B_1}$ and $s_{B_2}$.}
  \label{fig:boundaries_over_time_LOCA}
\end{figure}

\begin{figure}[h]
  \centering
  \includegraphics[width=1.0\textwidth]{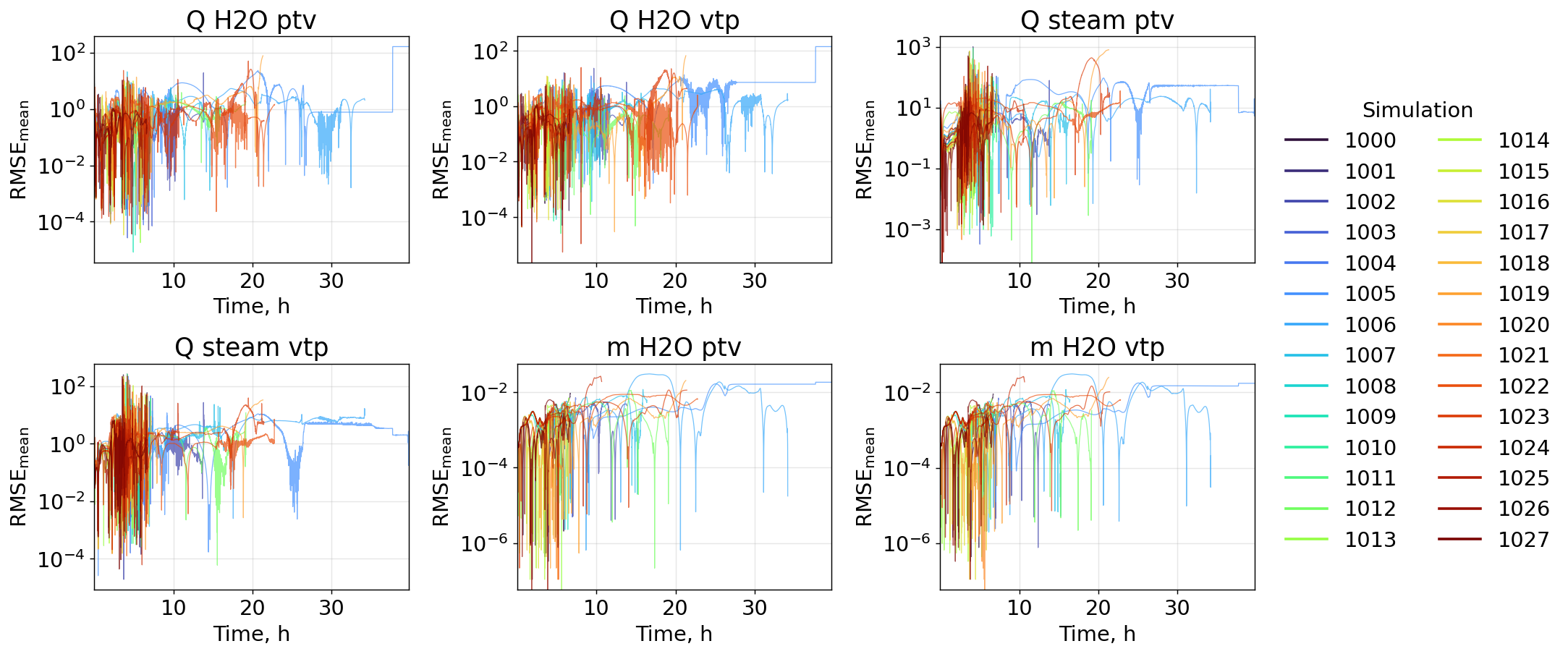}
  \caption{RMSE$_{mean}$ per time step for all the SBO testing trajectories for the boundary variables $s_{B_1}$ and $s_{B_2}$.}
  \label{fig:boundaries_over_time_SBO}
\end{figure}

In Figures \ref{fig:boundaries_over_time_LOCA} and \ref{fig:boundaries_over_time_SBO} we show the RMSE$_{mean}$ per time step for LOCA and SBO testing trajectories for the variables $s_{B_1}$ and $s_{B_2}$, which are important to couple AE-NODE to the primary circuit as shown in Figure \ref{fig:coupling_explained}.

\subsection{Latent space dynamics}
\label{appendix:latent_Space_dynamics}
\begin{figure}[htbp]
    \centering
    \begin{subfigure}[b]{0.24\textwidth}
        \includegraphics[width=\textwidth]{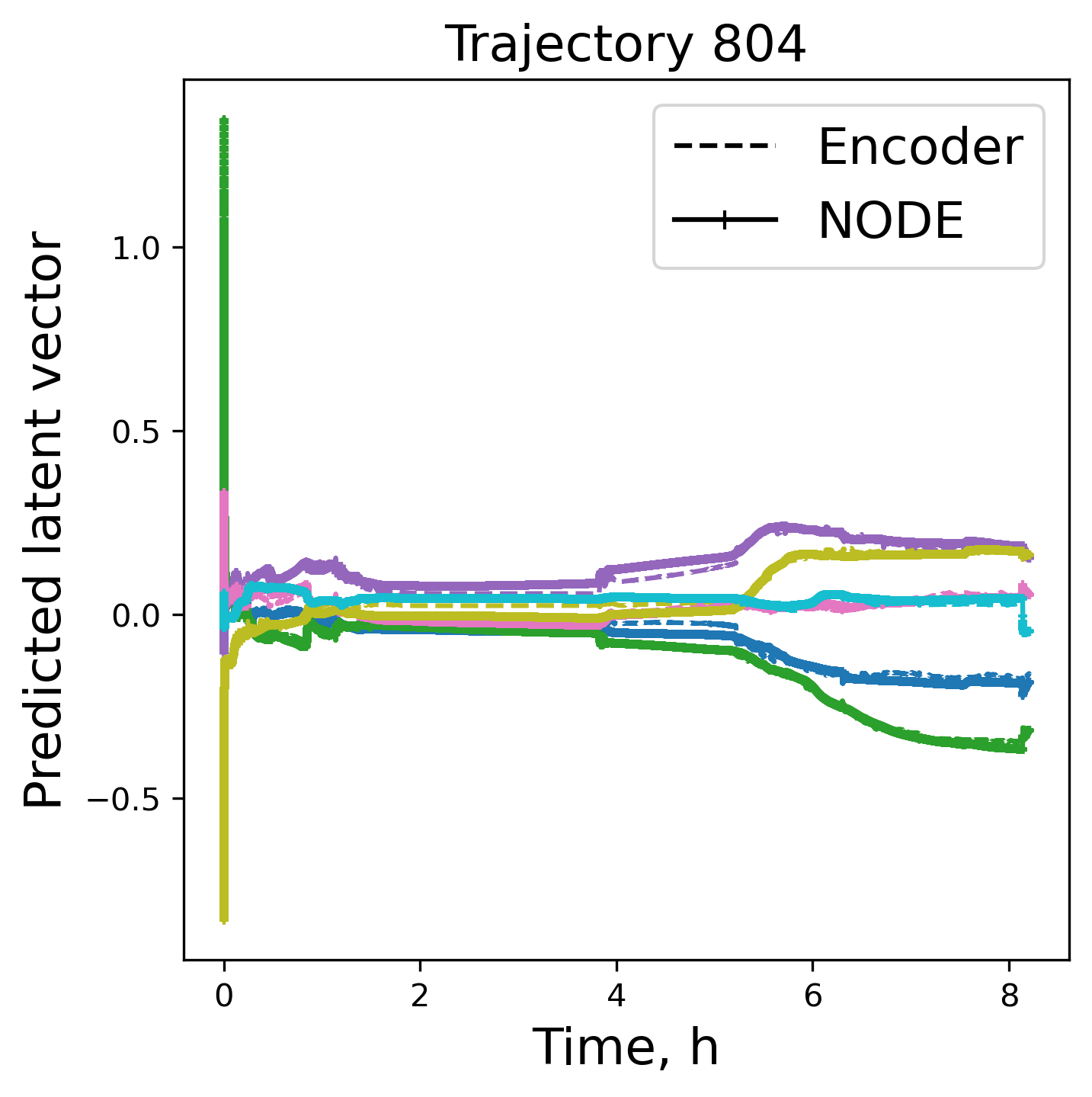}
        \caption{Trajectory 804}
        \label{fig:804_Predicted_latent_vector_LOCA}
    \end{subfigure}
    \hfill
    \begin{subfigure}[b]{0.24\textwidth}
        \includegraphics[width=\textwidth]{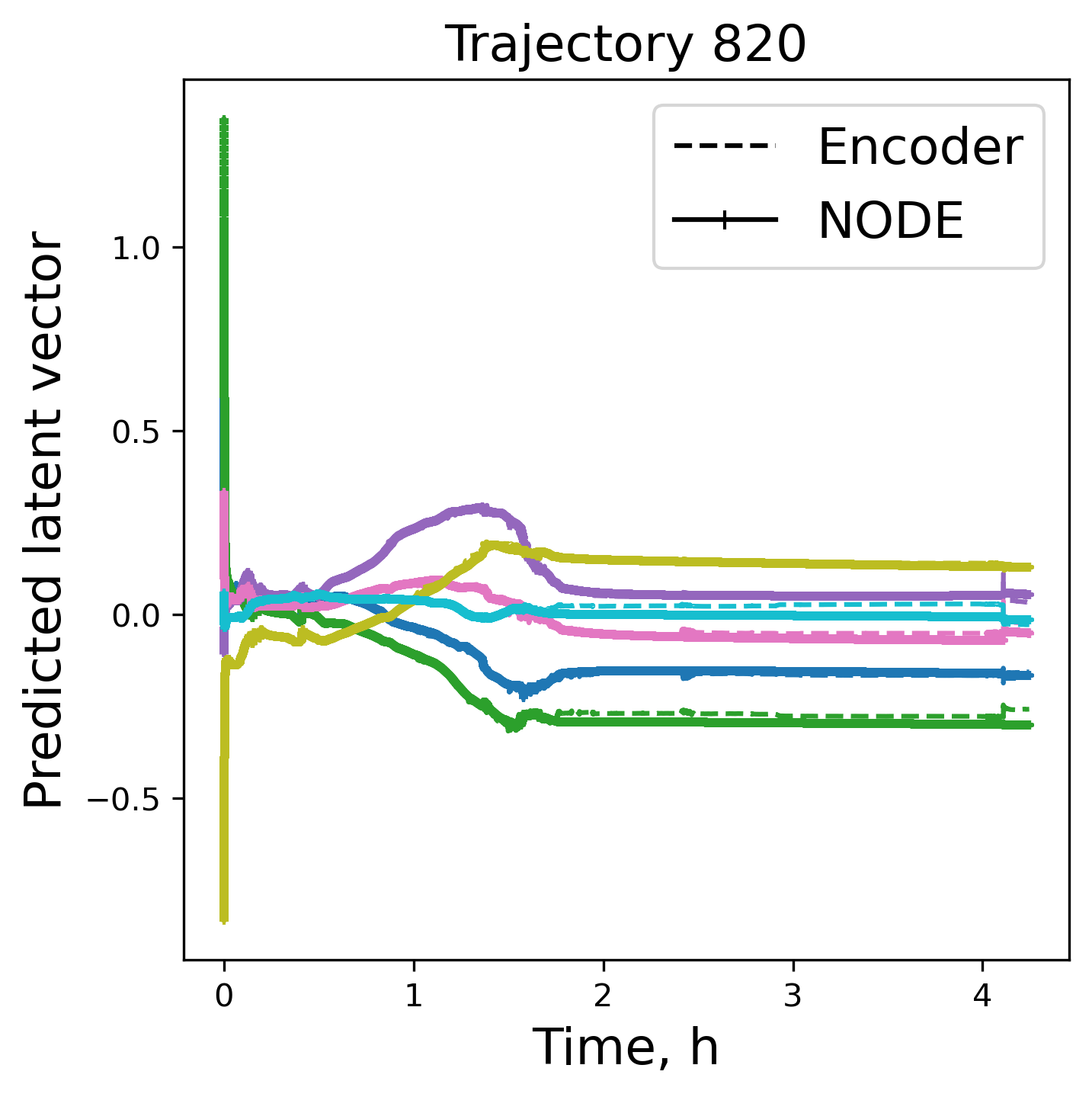}
        \caption{Trajectory 820}
        \label{fig:820_Predicted latent vector_LOCA}
    \end{subfigure}
    \hfill
    \begin{subfigure}[b]{0.24\textwidth}
        \includegraphics[width=\textwidth]{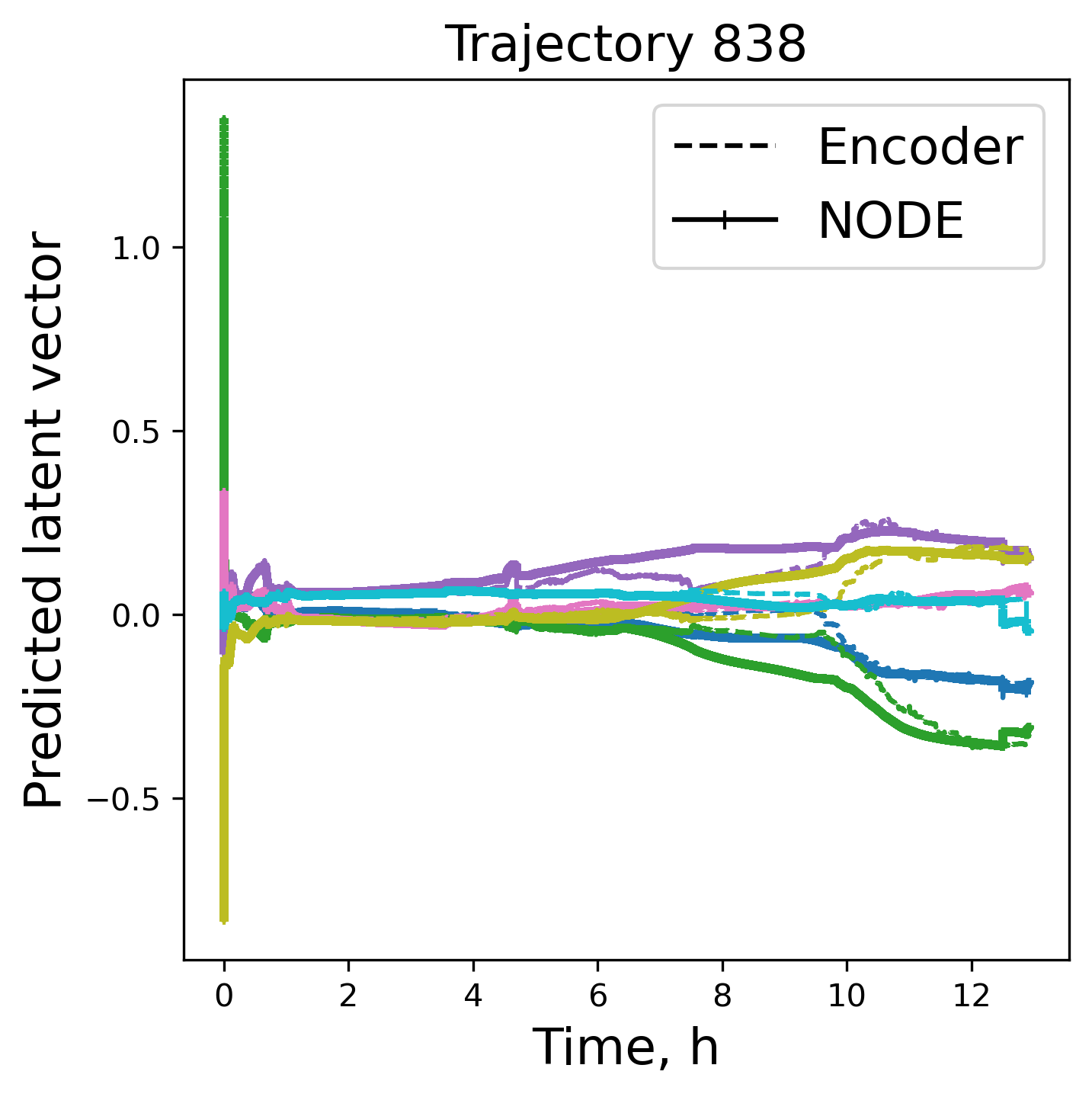}
        \caption{Trajectory 838}
        \label{fig:838_Predicted latent vector_LOCA}
    \end{subfigure}
    \hfill
    \begin{subfigure}[b]{0.24\textwidth}
        \includegraphics[width=\textwidth]{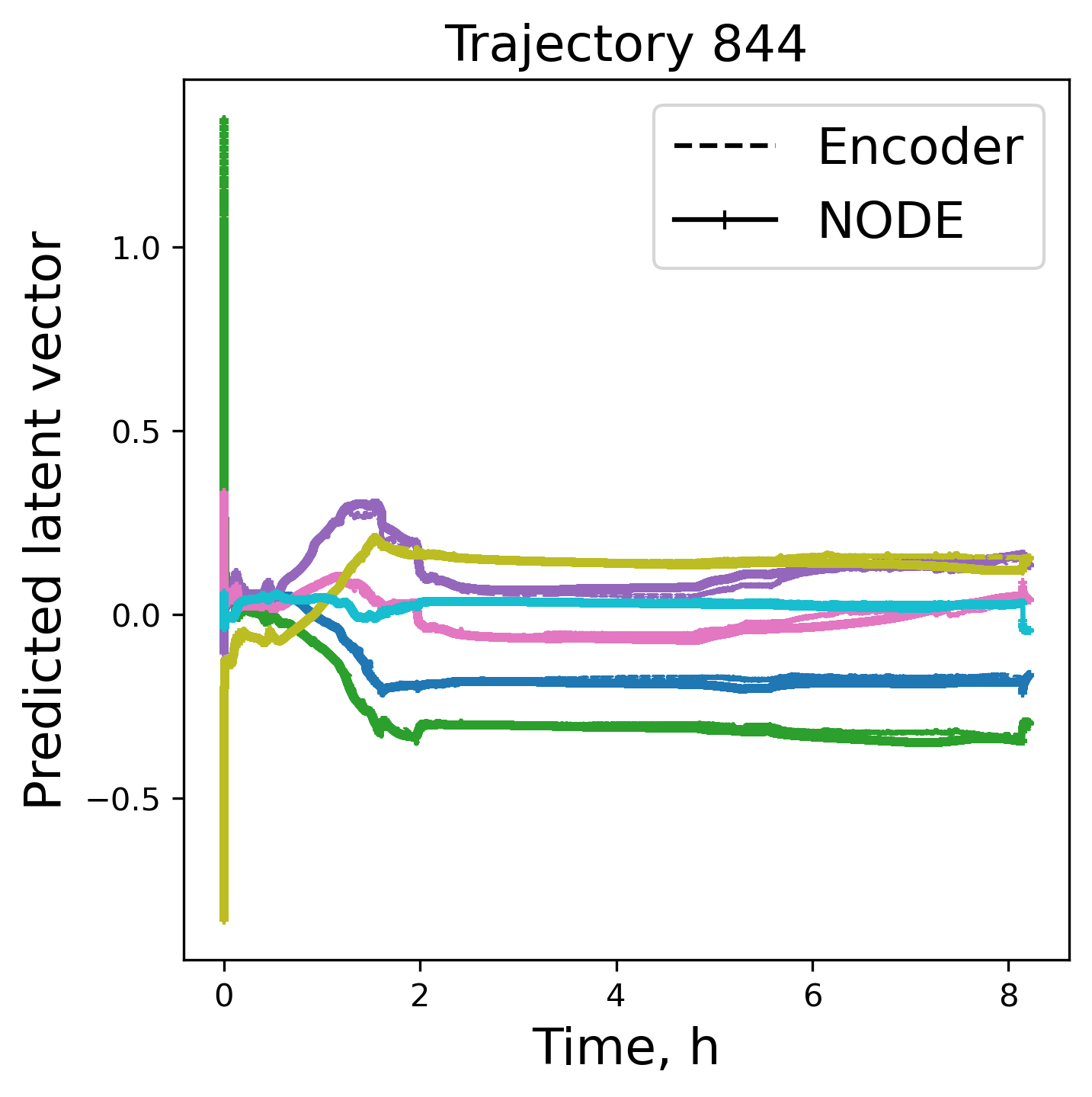}
        \caption{Trajectory 844}
        \label{fig:844_Predicted latent vector_LOCA}
    \end{subfigure}
    \caption{Auto-regressive latent vector prediction (6 dimensions) over time for some LOCA selected trajectories: dashed from Encoder, solid from NODE.}
    \label{fig:latent_space_all_LOCA}
\end{figure}

\begin{figure}[htbp]
    \centering
    \begin{subfigure}[b]{0.24\textwidth}
        \includegraphics[width=\textwidth]{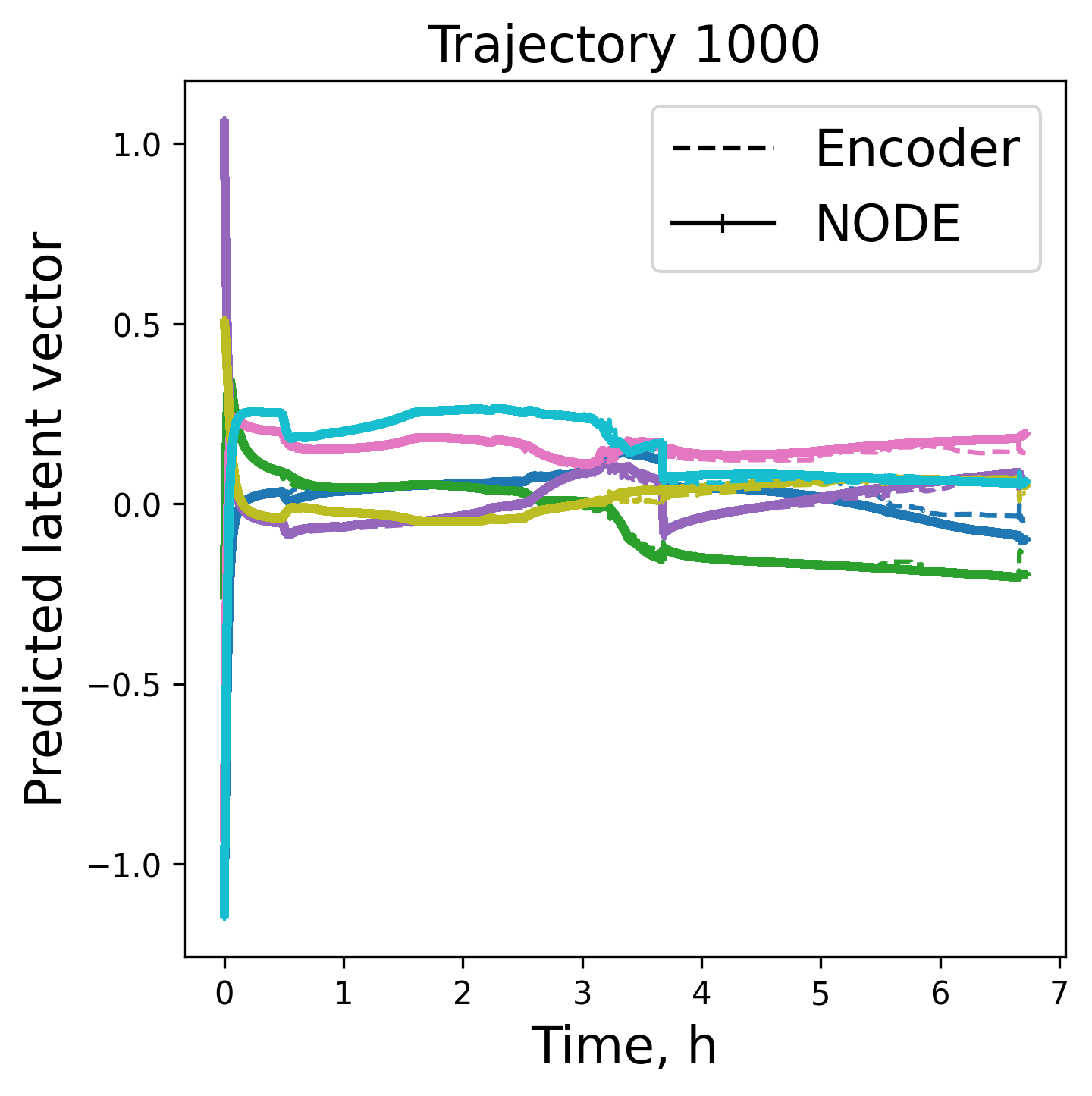}
        \caption{Trajectory 1000}
        \label{1000_Predicted latent vector_SBO.png}
    \end{subfigure}
    \hfill
    \begin{subfigure}[b]{0.24\textwidth}
        \includegraphics[width=\textwidth]{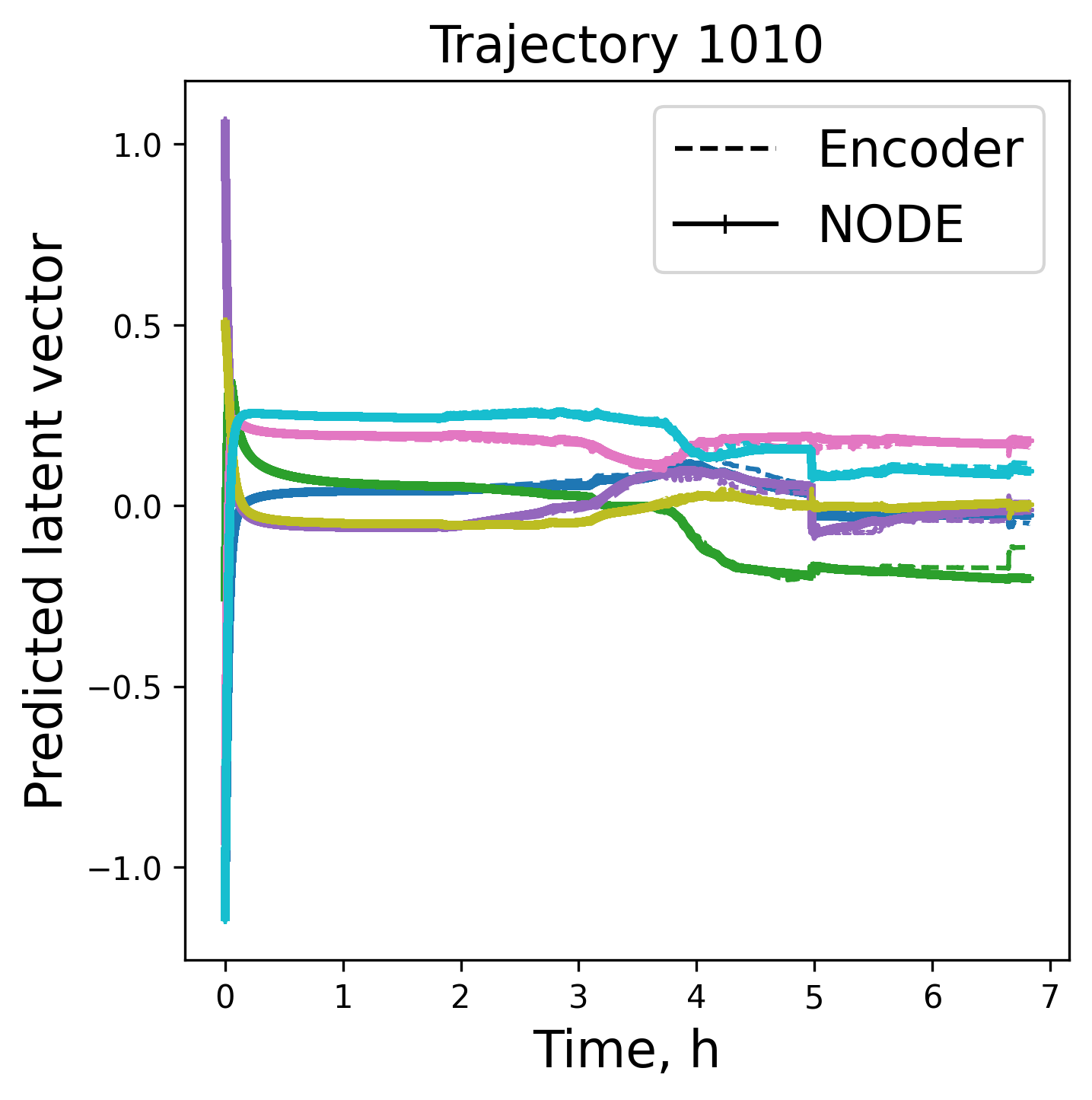}
        \caption{Trajectory 1010}
        \label{1010_Predicted latent vector_SBO.png}
    \end{subfigure}
    \hfill
    \begin{subfigure}[b]{0.24\textwidth}
        \includegraphics[width=\textwidth]{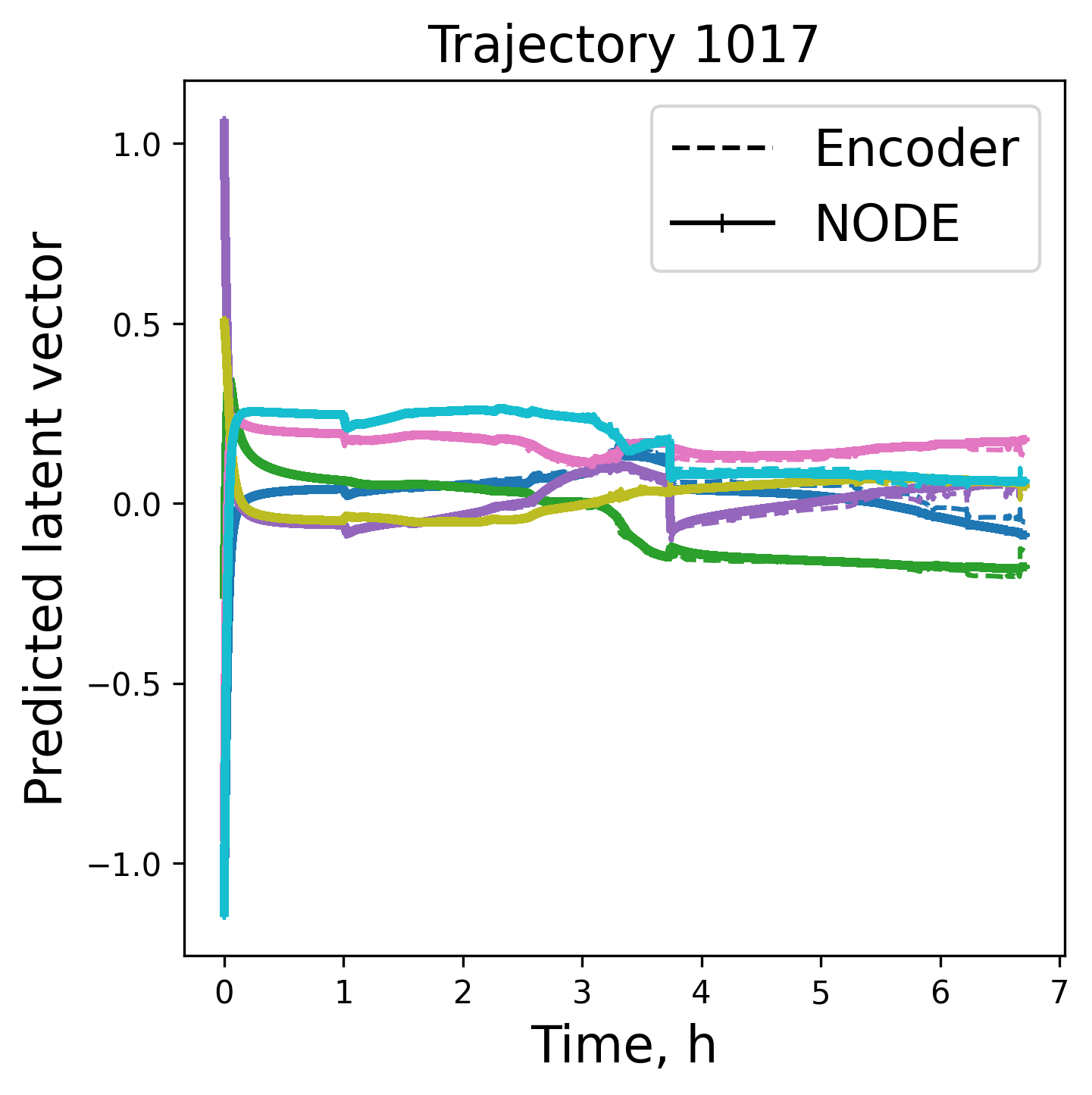}
        \caption{Trajectory 1017}
        \label{1017_Predicted latent vector_SBO.png}
    \end{subfigure}
    \hfill
    \begin{subfigure}[b]{0.24\textwidth}
        \includegraphics[width=\textwidth]{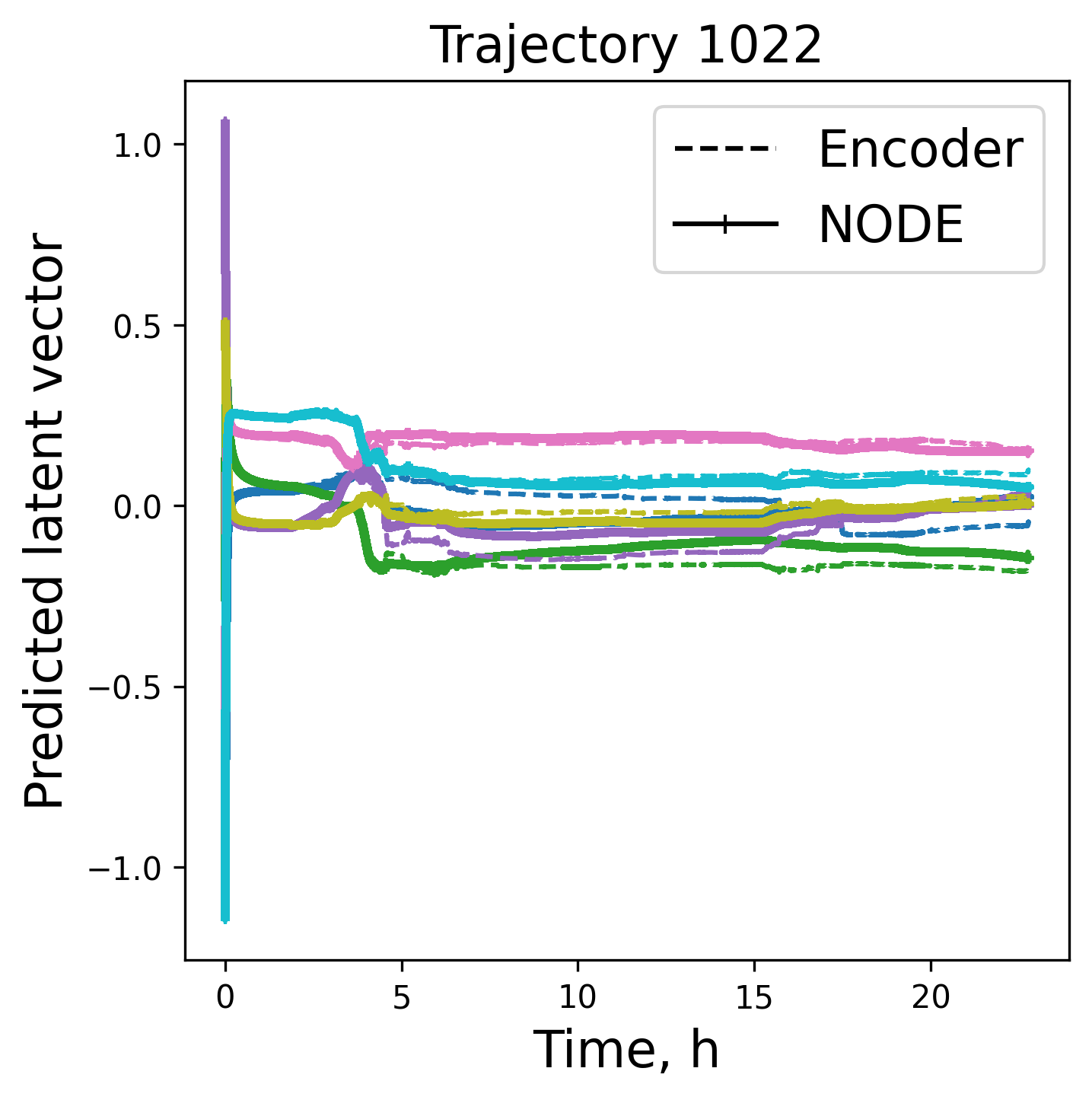}
        \caption{Trajectory 1022}
        \label{fig:1022_Predicted latent vector_SBO.png}
    \end{subfigure}
    \caption{Auto-regressive latent vector prediction (6 dimensions) over time for some SBO selected trajectories: dashed from Encoder, solid from NODE.}
    \label{fig:latent_space_all_SBO}
\end{figure}

\begin{figure}[htbp]
    \centering
    \begin{subfigure}[b]{0.24\textwidth}
        \includegraphics[width=\textwidth]{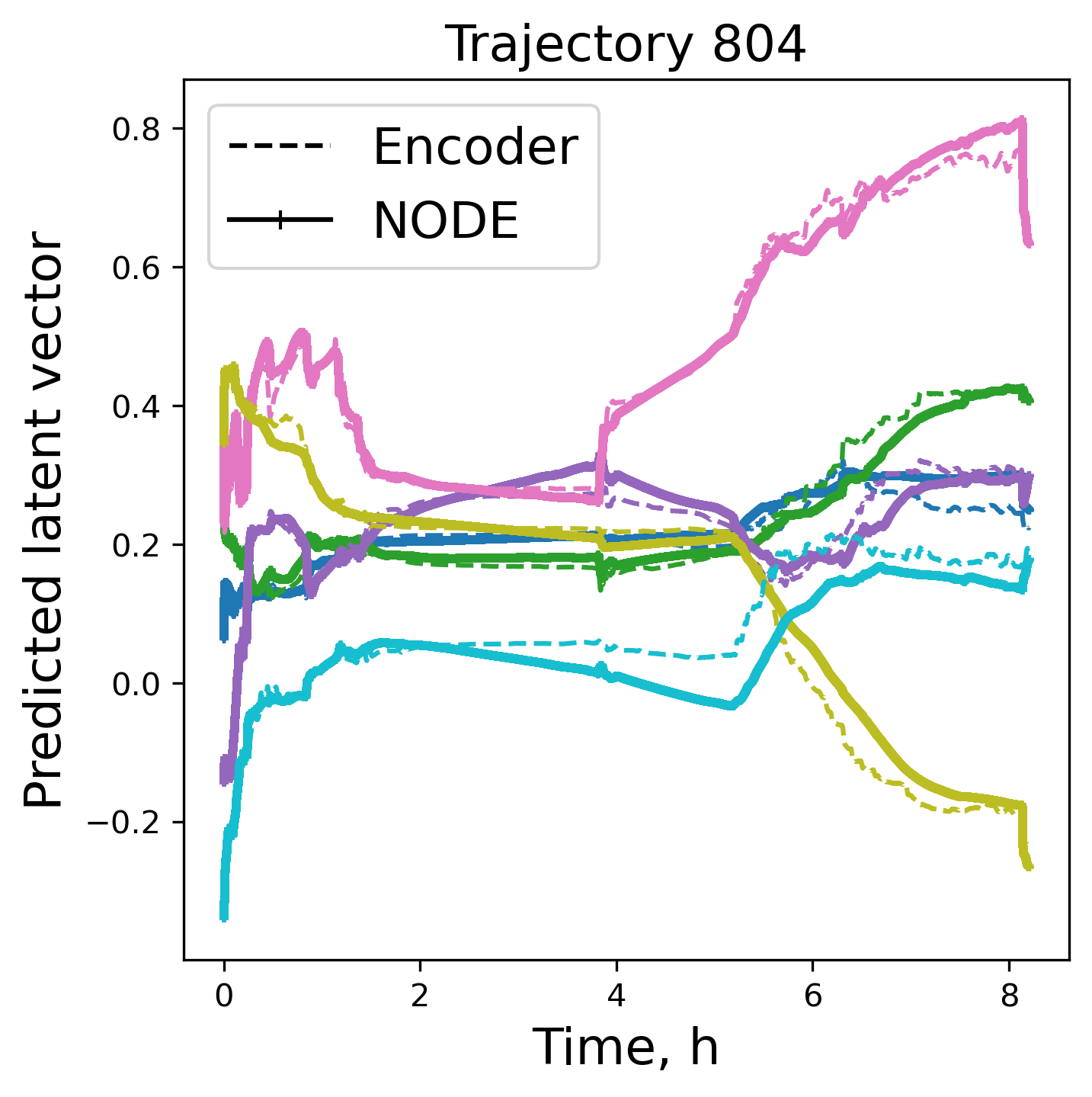}
        \caption{Trajectory 804}
        \label{fig:804_Predicted latent vector_LOCA_smoothed}
    \end{subfigure}
    \hfill
    \begin{subfigure}[b]{0.24\textwidth}
        \includegraphics[width=\textwidth]{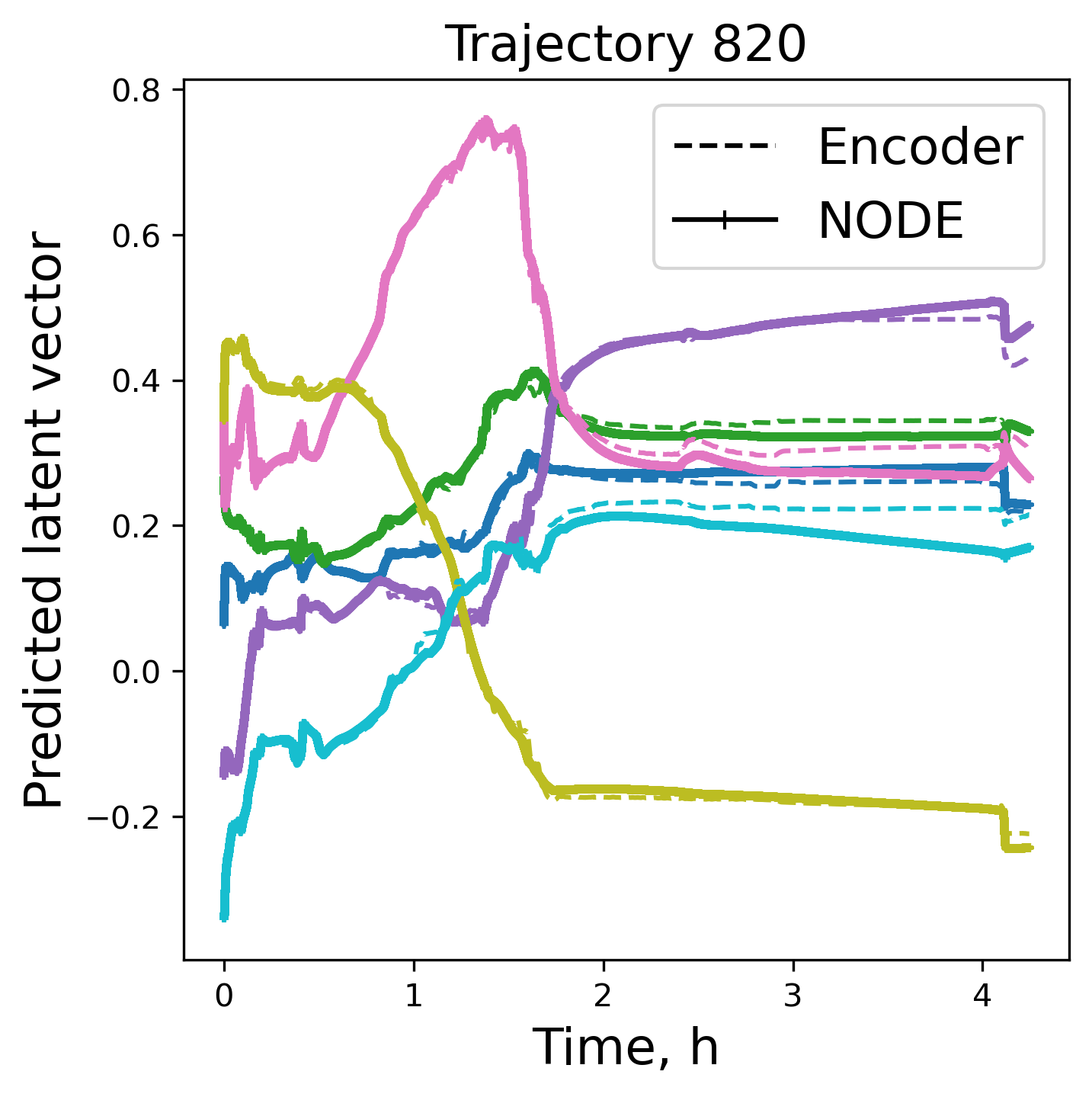}
        \caption{Trajectory 820}
        \label{fig:820_Predicted latent vector_LOCA_smoothed}
    \end{subfigure}
    \hfill
    \begin{subfigure}[b]{0.24\textwidth}
        \includegraphics[width=\textwidth]{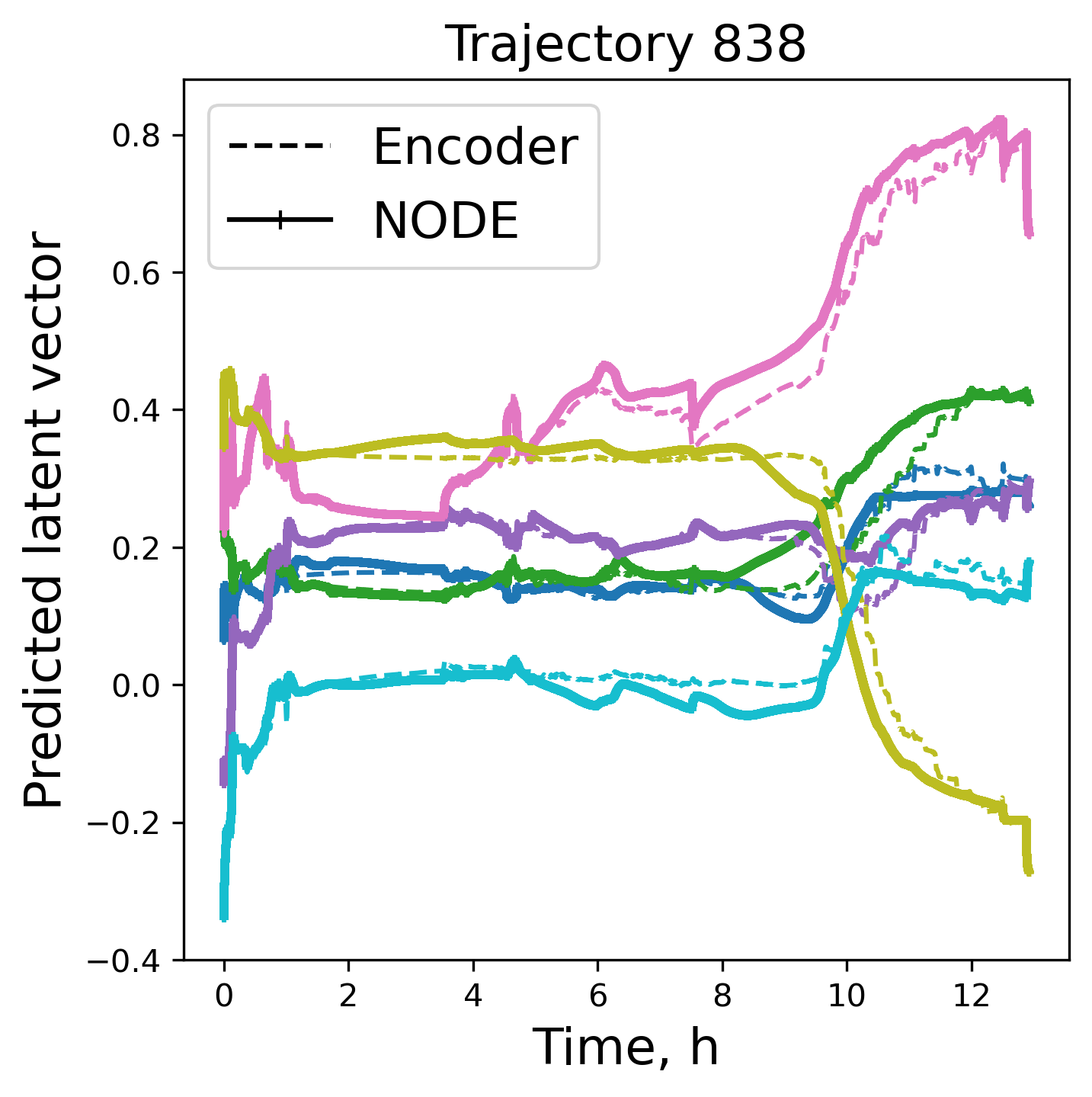}
        \caption{Trajectory 838}
        \label{fig:838_Predicted latent vector_LOCA_smoothed}
    \end{subfigure}
    \hfill
    \begin{subfigure}[b]{0.24\textwidth}
        \includegraphics[width=\textwidth]{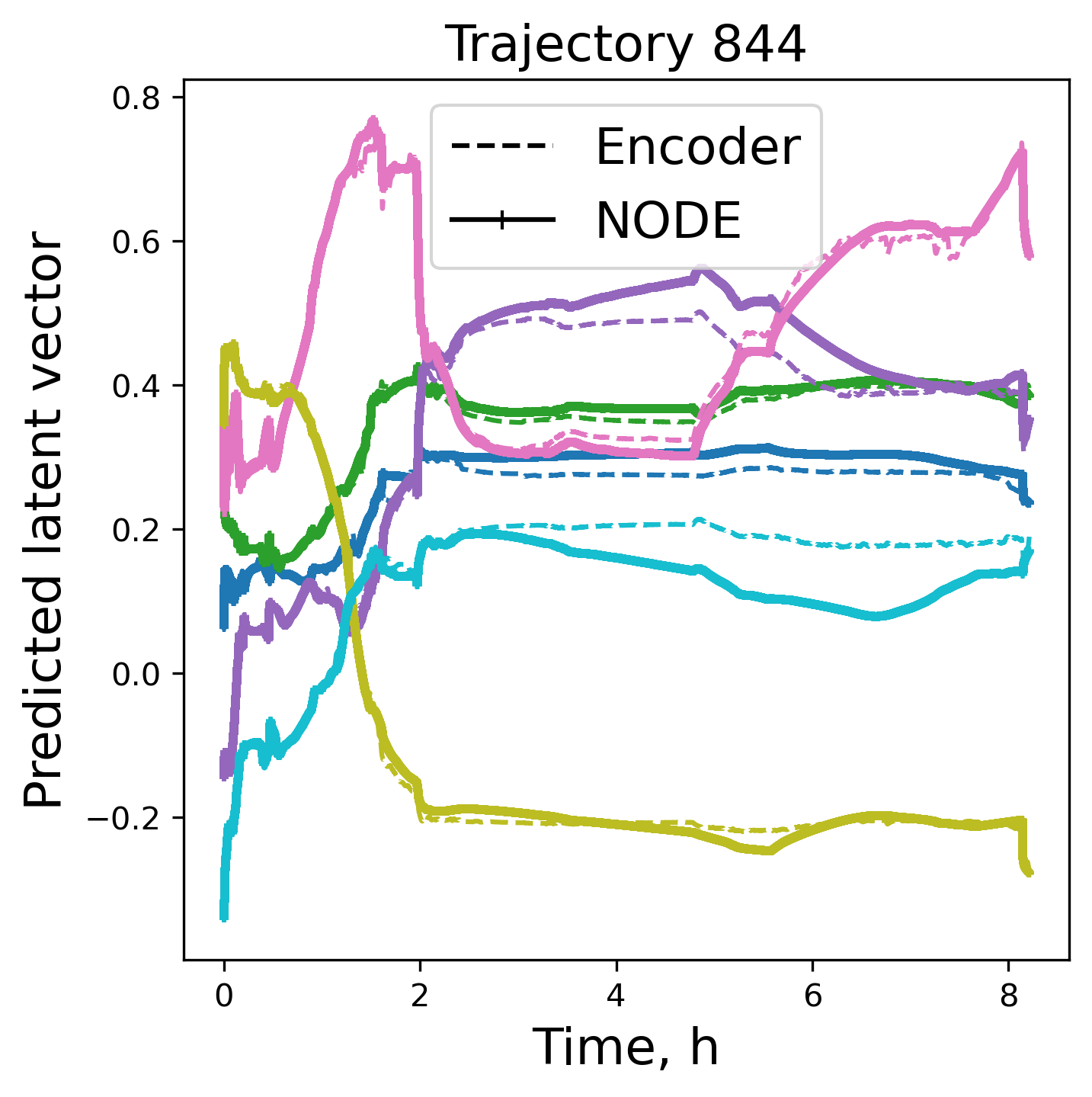}
        \caption{Trajectory 844}
        \label{fig:844_Predicted latent vector_LOCA_smoothed}
    \end{subfigure}
    \caption{Auto-regressive latent vector prediction (6 dimensions) over time for some LOCA selected trajectories when the Savitzky-Golay filter is applied: dashed from Encoder, solid from NODE.}
    \label{fig:latent_space_all_LOCA_smoothed}
\end{figure}

\begin{figure}[htbp]
    \centering
    \begin{subfigure}[b]{0.24\textwidth}
        \includegraphics[width=\textwidth]{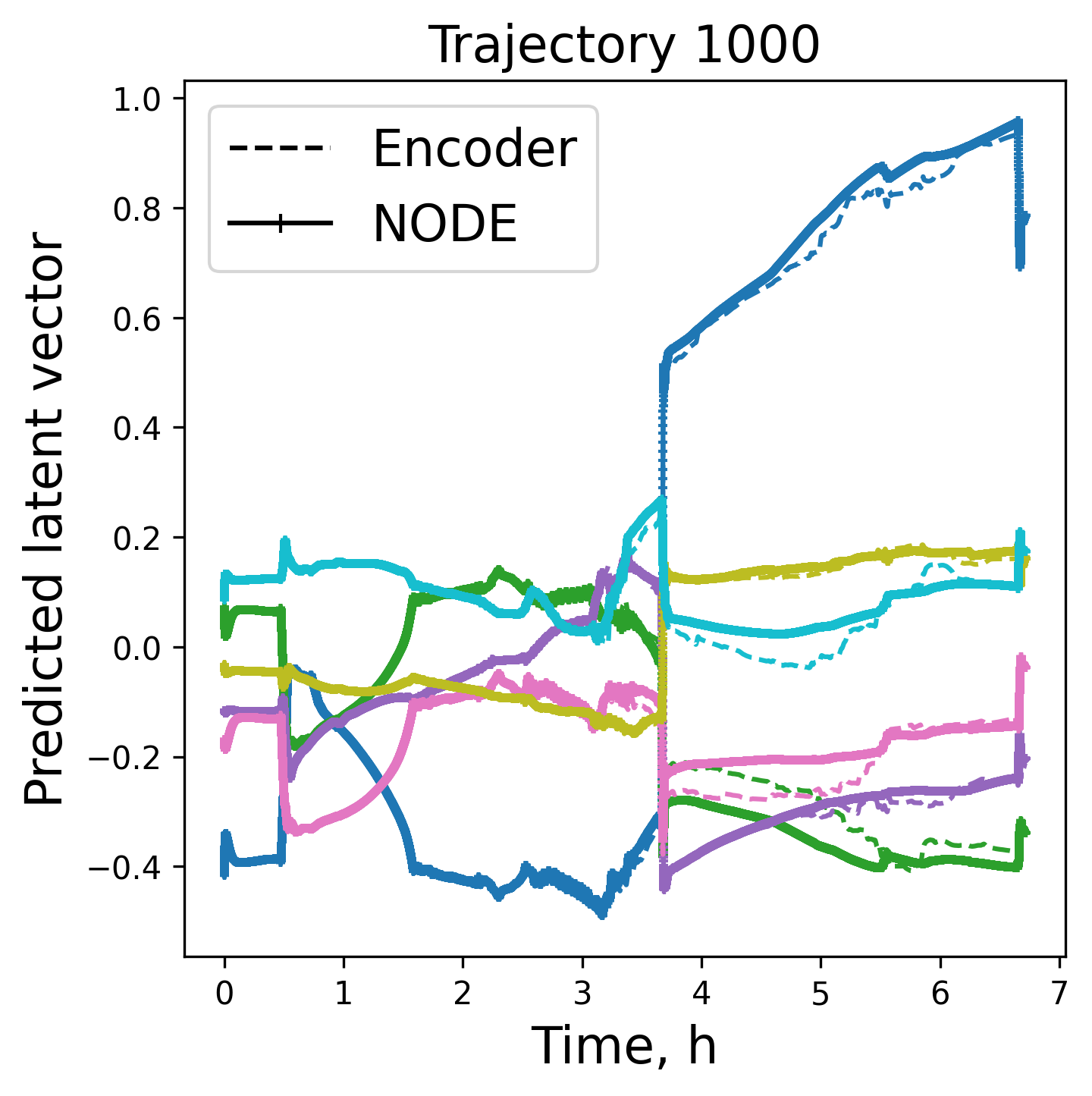}
        \caption{Trajectory 1000}
        \label{1000_Predicted latent vector_SBO_smoothed}
    \end{subfigure}
    \hfill
    \begin{subfigure}[b]{0.24\textwidth}
        \includegraphics[width=\textwidth]{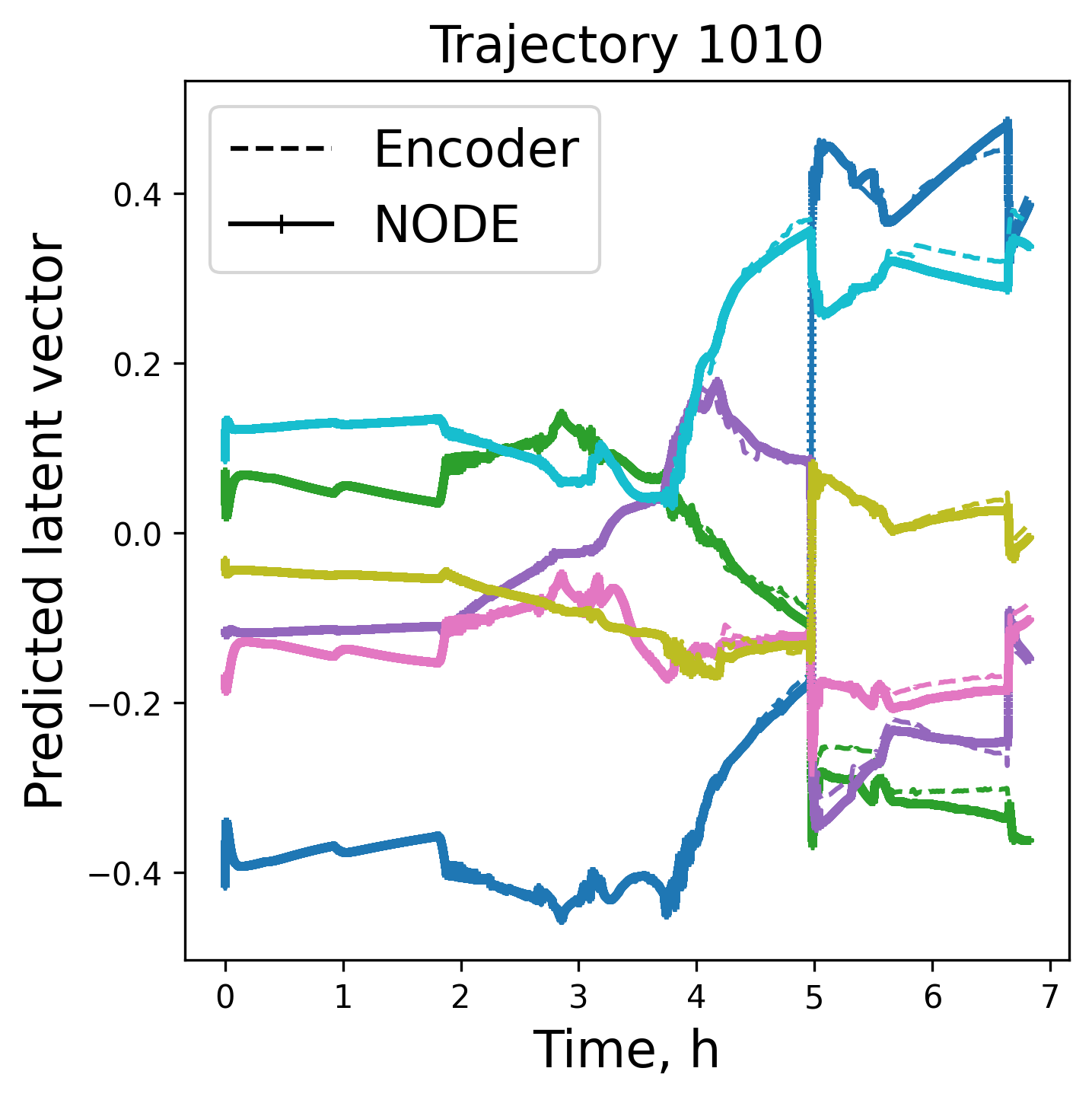}
        \caption{Trajectory 1010}
        \label{1010_Predicted latent vector_SBO_smoothed}
    \end{subfigure}
    \hfill
    \begin{subfigure}[b]{0.24\textwidth}
        \includegraphics[width=\textwidth]{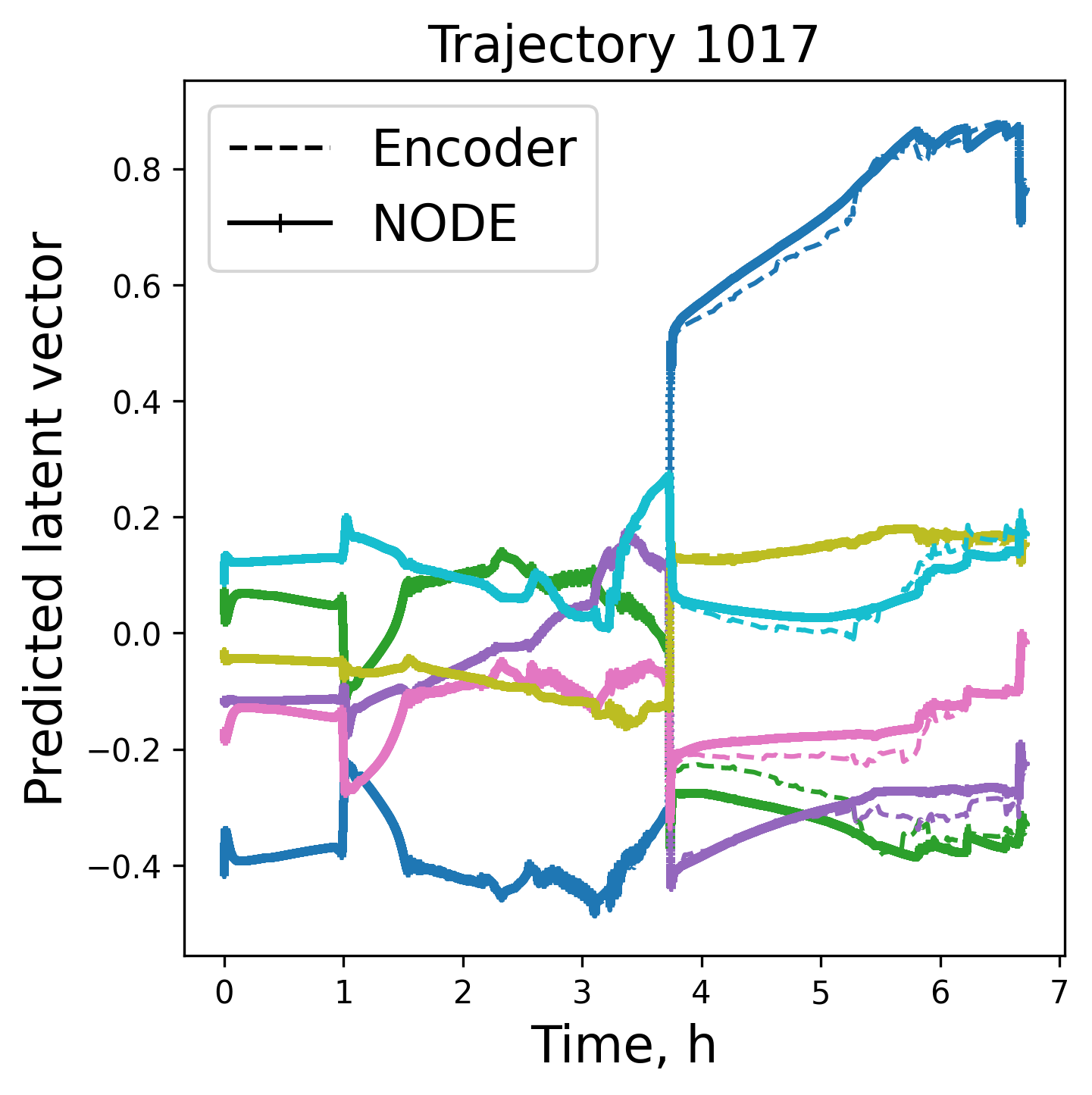}
        \caption{Trajectory 1017}
        \label{1017_Predicted latent vector_SBO_smoothed}
    \end{subfigure}
    \hfill
    \begin{subfigure}[b]{0.24\textwidth}
        \includegraphics[width=\textwidth]{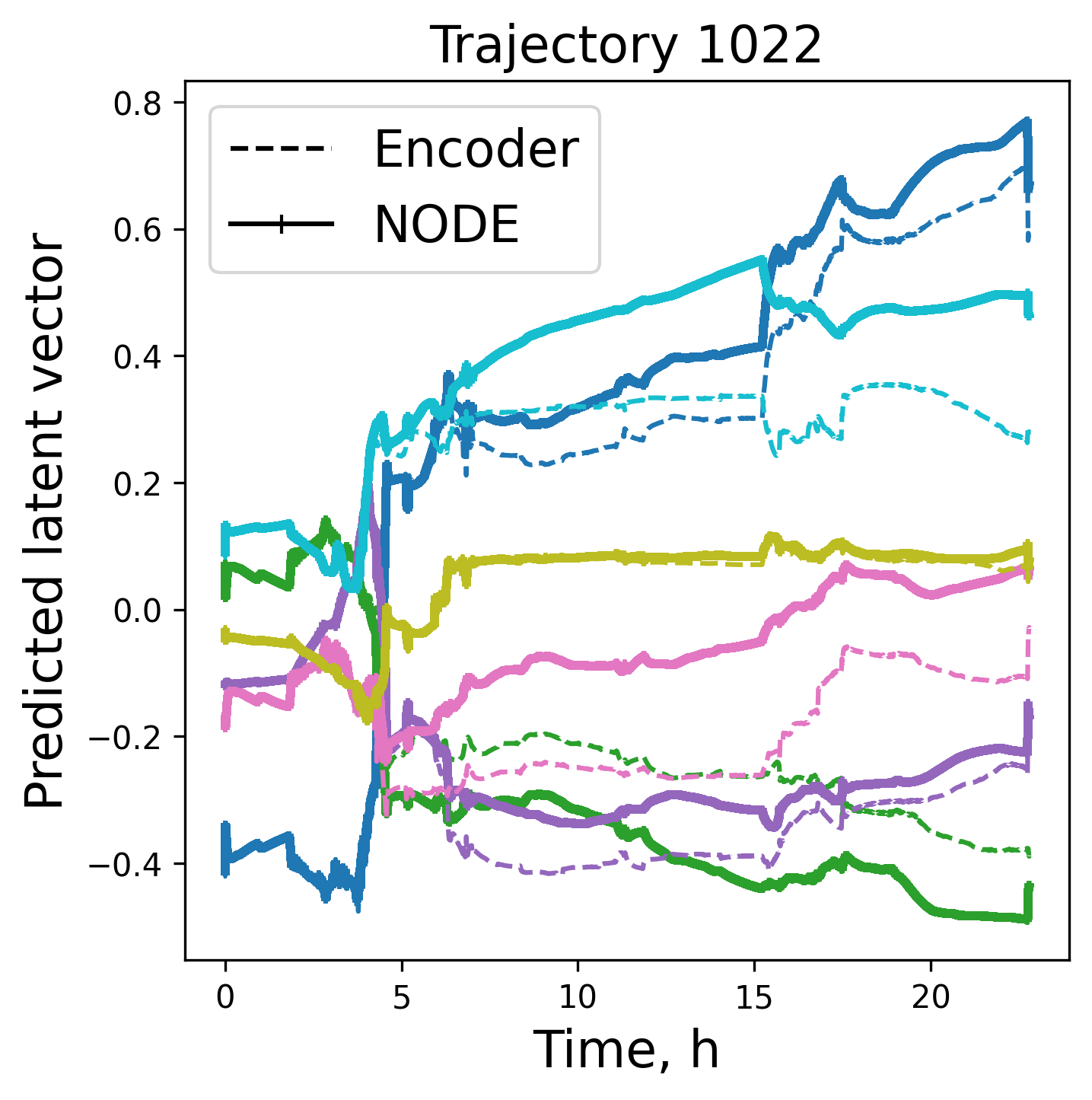}
        \caption{Trajectory 1022}
        \label{fig:1022_Predicted latent vector_SBO_smoothed}
    \end{subfigure}
    \caption{Auto-regressive latent vector prediction (6 dimensions) over time for some SBO selected trajectories when the Savitzky-Golay filter is applied: dashed from Encoder, solid from NODE.}
    \label{fig:latent_space_all_SBO_smoothed}
\end{figure}

In Figure \ref{fig:latent_space_all_LOCA} and Figure \ref{fig:latent_space_all_SBO} we show the latent evolution predicted by the AE-NODE vs the actual one identified by the Encoder; in Figures \ref{fig:latent_space_all_LOCA_smoothed} and Figure \ref{fig:latent_space_all_SBO_smoothed} we show the same quantities but when the Savitzky–Golay smoothing is applied.

\end{document}